%% file: sn-article-arxiv.tex
\documentclass[pdflatex,sn-mathphys-num]{sn-jnl}

\usepackage{tabularx}
\usepackage{float} 
\usepackage{graphicx}%
\usepackage{multirow}%
\usepackage{makecell}
\usepackage{amsmath,amssymb,amsfonts}%
\usepackage{amsthm}%
\usepackage{array}
\usepackage{mathrsfs}%
\usepackage[title]{appendix}%
\usepackage{textcomp}%
\usepackage{manyfoot}%
\usepackage{booktabs}%
\usepackage{algorithm}%
\usepackage{algorithmicx}%
\usepackage{algpseudocode}%
\usepackage{listings}%
\usepackage{geometry}  
\usepackage{lineno}
\usepackage{minitoc}
\usepackage{titletoc}
\usepackage{fancyhdr}
\usepackage[table,xcdraw]{xcolor}
\usepackage{tikz,lipsum,lmodern}
\usepackage[most]{tcolorbox}
\usepackage{pdfpages}
\usepackage{hyperref} 
\usepackage{gensymb}
\tcbuselibrary{listings}
\usepackage{listingsutf8}
\usepackage{lineno}
\usepackage{bibunits}
\usepackage{chngcntr}

\theoremstyle{thmstyleone}%
\newcommand{\model}{\mbox{EarthLink}}

\theoremstyle{thmstyletwo}%

\theoremstyle{thmstylethree}%

\begin{document}

\setlength{\bibsep}{0.5em}

\begin{bibunit}[sn-mathphys-num]

\renewcommand\linenumberfont{\normalfont\scriptsize}
\setlength\linenumbersep{25pt}


\title[Article Title]{A Self-Evolving AI Agent System for Climate Science}




\author[1,2]{\fnm{Zijie} \sur{Guo}}
\equalcont{These authors contributed equally to this work.}
\author[1,3]{\fnm{Jiong} \sur{Wang}}
\equalcont{These authors contributed equally to this work.}
\author*[1]{\fnm{Fenghua} \sur{Ling}}\email{lingfenghua@pjlab.org.cn}
\author[4]{\fnm{Wangxu} \sur{Wei}}
\author[1,5]{\fnm{Xiaoyu} \sur{Yue}}
\author[1,2]{\fnm{Zhe} \sur{Jiang}}
\author[1,6]{\fnm{Wanghan} \sur{Xu}}
\author[7]{\fnm{Jing-Jia} \sur{Luo}}
\author[4]{\fnm{Lijing} \sur{Cheng}}
\author[8]{\fnm{Yoo-Geun} \sur{Ham}}
\author[9,10]{\fnm{Fengfei} \sur{Song}}
\author[11]{\fnm{Pierre} \sur{Gentine}}
\author[12]{\fnm{Toshio} \sur{Yamagata}}
\author[1,13]{\fnm{Ben} \sur{Fei}}
\author[1]{\fnm{Wenlong} \sur{Zhang}}
\author[1]{\fnm{Xinyu} \sur{Gu}}
\author[14]{\fnm{Chao} \sur{Li}}
\author[15]{\fnm{Yaqiang} \sur{Wang}}
\author[3]{\fnm{Tao} \sur{Chen}}
\author[1,13]{\fnm{Wanli} \sur{Ouyang}}
\author[1,16]{\fnm{Bowen} \sur{Zhou}}
\author*[1]{\fnm{Lei} \sur{Bai}}\email{bailei@pjlab.org.cn}


\affil[1]{\orgname{Shanghai Artificial Intelligence Laboratory}, \orgaddress{\city{Shanghai}, \country{China}}}

\affil[2]{\orgdiv{College of Computer Science and Artificial Intelligence}, \orgname{Fudan University}, \orgaddress{\city{Shanghai}, \country{China}}}

\affil[3]{\orgdiv{College of Future Information Technology}, \orgname{Fudan University}, \orgaddress{\city{Shanghai}, \country{China}}}

\affil[4]{\orgdiv{State Key Laboratory of Earth System Numerical Modeling and Application, Institute of Atmospheric Physics}, \orgname{Chinese Academy of Sciences}, \orgaddress{\city{Beijing}, \country{China}}}

\affil[5]{\orgdiv{School of Electrical and Computer Engineering}, \orgname{The University of Sydney}, \orgaddress{\city{Sydney}, \state{New South Wales}, \country{Australia}}}

\affil[6]{\orgdiv{School of Electronic Information and Electrical Engineering}, \orgname{Shanghai Jiao Tong University}, \orgaddress{\city{Shanghai}, \country{China}}}

\affil[7]{\orgdiv{Institute for Climate and Application Research (ICAR)/CPRM/CIC-FEMD/KLME/ILCEC}, \orgname{Nanjing University of Information Science and Technology}, \orgaddress{\city{Nanjing}, \country{China}}}

\affil[8]{\orgdiv{Department of Environmental Planning, Graduate School of Environmental Studies}, \orgname{Seoul National University}, \orgaddress{\city{Seoul}, \country{South Korea}}}

\affil[9]{\orgname{Frontier Science Center for Deep Ocean Multispheres and Earth System and Physical Oceanography Laboratory, Ocean University of China}, \orgaddress{\city{Qingdao}, \country{China}}}

\affil[10]{\orgname{Laoshan Laboratory}, \orgaddress{\city{Qingdao}, \country{China}}}

\affil[11]{\orgdiv{Department of Earth and Environmental Engineering}, \orgname{Columbia University}, \orgaddress{\city{New York}, \state{NY}, \country{USA}}}

\affil[12]{\orgdiv{Application Laboratory}, \orgname{Japan Agency for Marine-Earth Science and Technology}, \orgaddress{\city{Yokohama}, \country{Japan}}}

\affil[13]{\orgdiv{Department of Information Engineering}, \orgname{The Chinese University of Hong Kong}, \orgaddress{\city{Hong Kong}, \country{China}}}

\affil[14]{\orgdiv{Key Laboratory of Geographic Information Science, Ministry of Education, School of Geographic Sciences}, \orgname{East China Normal University}, \orgaddress{\city{Shanghai}, \country{China}}}

\affil[15]{\orgname{Chinese Academy of Meteorological Sciences}, \orgaddress{\city{Beijing}, \country{China}}}

\affil[16]{\orgdiv{Department of Electronic Engineering}, \orgname{Tsinghua University}, \orgaddress{\city{Beijing}, \country{China}}}

\abstract{
Scientific progress in Earth science depends on integrating data across the planet’s interconnected spheres. However, the accelerating volume and fragmentation of multi-sphere knowledge and data have surpassed human analytical capacity. This creates a major bottleneck for discovery, especially in climate science. To handle this challenge, we introduce \model, the first self-evolving AI agent system designed as an interactive ``copilot'' for Earth scientists. Through natural language interaction, \model~automates the entire research workflow by integrating planning, code execution, data analysis, and physical reasoning into a unified process that directly addresses this limitation. Beyond efficiency, it exhibits human-like cross-disciplinary analytical ability and achieves proficiency comparable to a junior researcher in expert evaluations on core large-scale climate tasks, including model–observation comparison and climate change understanding. When tasked with an open scientific problem, specifically the discovery of precursors of the Atlantic Niño, \model~autonomously developed a research strategy, identified sources of predictability, verified its hypotheses with available data, and proposed a physically consistent mechanism. These emerging capabilities enable a new human-AI research paradigm. Scientists can focus on value and result judgments, while AI systems handle complex data analysis and knowledge integration. This accelerates the pace and breadth of discovery in Earth sciences. The system is accessible at our website \url{https://earthlink.intern-ai.org.cn}.
}

\maketitle

\section{Introduction}\label{sec1}

The pursuit of understanding Earth’s systems has long been a central endeavor of humankind. Realizing this goal requires the integration of vast datasets that capture complex interactions among the many components of the Earth. However, these data are not only vast in volume but also fragmented across sources and foundational attributes, such as physical, chemical, and biological properties, exceeding human capacity for comprehensive analysis~\cite{overpeck2011climate,reichstein2019deep,guo2017big}. This imbalance between data abundance and analytical capability has become a critical bottleneck that limits scientific discovery in Earth system research~\cite{park2023papers}.

This challenge is particularly acute in climate science, where insights must be extracted from vast, cross-disciplinary datasets to guide global warming mitigation and adaptation strategies~\cite{masson2021climate,core2023ipcc,cheng2025record,kennedy2024state,WMO2025}. Central to these efforts are Earth System Models (ESMs)~\cite{stute2001global,heinze2019esd}, which simulate the complex interactions among the atmosphere, oceans, cryosphere, and biosphere. These models form the backbone of our understanding of climate dynamics and future projections, and are systematically evaluated through the Coupled Model Intercomparison Project (CMIP)~\cite{eyring2016overview}. With each successive phase, CMIP has grown in scope, becoming the primary cross-disciplinary data archive for global climate assessment. CMIP6 alone amalgamates contributions from over 50 modeling centers, producing petabytes of data from thousands of simulations~\cite{taylor2012overview,eyring2016overview}. The sheer volume and complexity of this multi-model, multi-sphere dataset fundamentally challenge traditional methods of analysis.

Traditional research workflows, which remain manual and fragmented, struggle to keep pace with the increasing complexity of data. Even basic scientific tasks, such as model-observation comparison, require extensive custom coding and careful data handling. For example, estimating equilibrium climate sensitivity requires researchers to manually analyze hundreds of model simulations from diverse experiments, a process that is both labor-intensive and error-prone~\cite{andrews2012forcing, meehl2020context}. These workflow inefficiencies scale directly to our most critical scientific endeavors, such as the multi-year assessment cycles of the Intergovernmental Panel on Climate Change (IPCC), which relies on protracted manual analyses by thousands of scientists~\cite{calvin2023ipcc}.
 
In response to these analysis challenges, the climate science community has developed powerful diagnostic toolkits, such as ESMValTool~\cite{eyring2020earth}, CCMVal Diags~\cite{gettelman2012community}, PCMDI Metrics~\cite{lee2024systematic}, and ILAMB~\cite{collier2018international}. 
These tools have played a crucial role in standardizing workflows and promoting transparency and reproducibility across model evaluations. However, using and especially extending these frameworks often demands substantial programming expertise and deep familiarity with their architectures, posing barriers to the rapid exploration of new scientific ideas or adaptation to emerging datasets. 
Thus, while these toolkits have been instrumental in enabling consistent and transparent analyses, adapting them to the agile and exploratory workflows increasingly valued in contemporary climate research can still present challenges.

The recent rise of Large Language Models (LLMs) and intelligent agent systems offers a new path forward~\cite{wang2025history,zhang2025scientific}. These models can integrate external tools and knowledge to solve complex scientific problems, extending their capabilities far beyond language tasks~\cite{wang2024tools,fan2024survey}. This new paradigm has already automated scientific workflows in fields like biomedicine~\cite{huang2025biomni}, chemistry~\cite{boiko2023autonomous}, and materials science~\cite{kang2024chatmof}. However, in Earth sciences, progress has been limited to domain-specific question answering~\cite{bi2023oceangpt,deng2024k2,zhang2024geogpt}. The development of an agent system capable of fully automating the complex workflow of Earth system research remains an open challenge.

Here, we introduce \model, the first self-evolving AI agent system designed as an interactive ``copilot'' for Earth scientists to fill this gap. We demonstrate its capabilities in two aspects. First, expert-based assessments, conducted across a comprehensive suite of 36 large-scale tasks of varying difficulty, confirm that \model~can integrate fragmented, multi-sphere data and perform core climate tasks at a level comparable to an early-career researcher, while also possessing self-evolving capabilities for ongoing performance enhancement. Second, and more notably, we challenged \model~with a range of open scientific problems. As a prominent example, it was tasked to discover precursors for the summer Atlantic Niño at an 8-month lead time. The system autonomously designed a complete research workflow, through which it identified physically consistent precursors and, by integrating relevant knowledge, formulated a physically plausible and testable mechanism. This ability, from automated analysis to the generation of scientific insights, marks a definitive step toward AI systems that act as research partners, accelerating discovery in Earth system science.

\section{Results}\label{sec2}

\begin{figure}[H]
    \centering
    \includegraphics[width=0.99\linewidth]{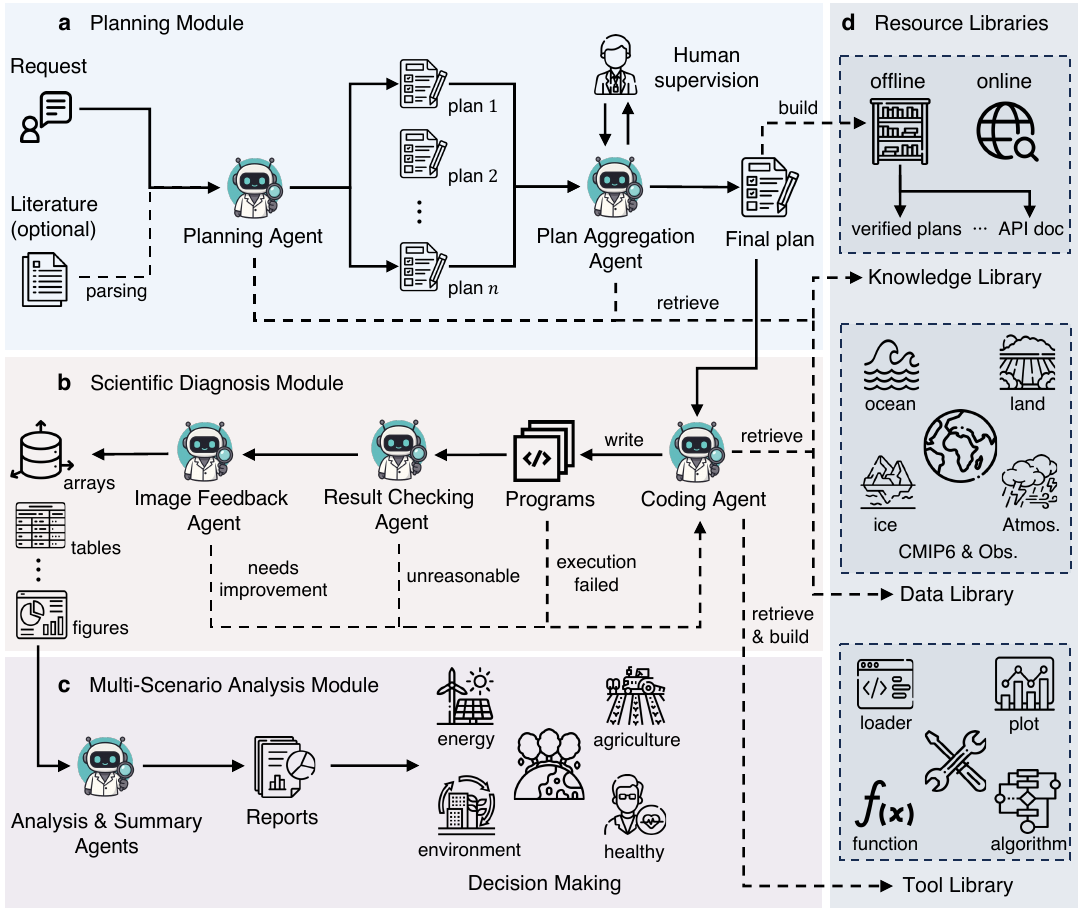}
    \caption{
    \textbf{The \model~workflow for automated climate data analysis.} 
    \textbf{a}, In the Planning Module, the system first receives a user's scientific request and optional relevant literature, from which the Planning Agent generates multiple candidate experimental plans. The Plan Aggregation Agent, with optional human supervision, then reviews and integrates these plans to form an optimal final plan.  
    \textbf{b}, The Scientific Diagnosis Module executes this plan. A Coding Agent automatically writes code, processing data and referring to tools from the Resource Libraries to perform computations and visualizations. This module features a built-in feedback loop, enabling autonomous debugging and iterative refinement through Result Checking and Image Feedback. 
    \textbf{c}, In the Multi-Scenario Analysis Module, the Analysis \& Summary Agents conduct an in-depth analysis of the scientific results from the preceding module, and synthesize these insights into well-structured and comprehensive final reports. These reports can further provide scientific interpretations across various domains, including energy, agriculture, environment, and insurance, while delivering insights relevant to policy-making. 
    \textbf{d}, The Resource Libraries serve as the foundational support for the entire workflow. They include: a Knowledge Library, which provides previously validated outputs (such as plans and code scripts), API documentation for relevant packages, and online web resources; a Data Library containing curated scientific datasets (e.g., CMIP6 data and observational records); and a Tool Library offering reusable functions, algorithms, and tools for data loading, computation, plotting, and more.
    }
    \label{fig:fig1_workflow}
\end{figure}

\model~is a multi-agent system designed to automate climate science research by seamlessly integrating vast cross-disciplinary knowledge, petabyte-scale data, and hundreds of tools for diverse diagnostic tasks. (Fig.~\ref{fig:fig1_workflow}). The workflow is organized into three synergistic modules. The Planning Module translates user queries to scientifically grounded experimental plans, supported by knowledge retrieval and optional human oversight. The Scientific Diagnosis Module then translates these plans into an executable workflow, autonomously managing data acquisition, code generation, and iterative debugging through a built-in feedback loop. Finally, the Multi-Scenario Analysis Module synthesizes experimental results into scientific reports, providing critical cross-domain insights for decision-making (see Methods section). 

\subsection{Confronting model simulations and observational data}
\begin{figure}[H]
    \centering
    \includegraphics[width=0.99\linewidth]{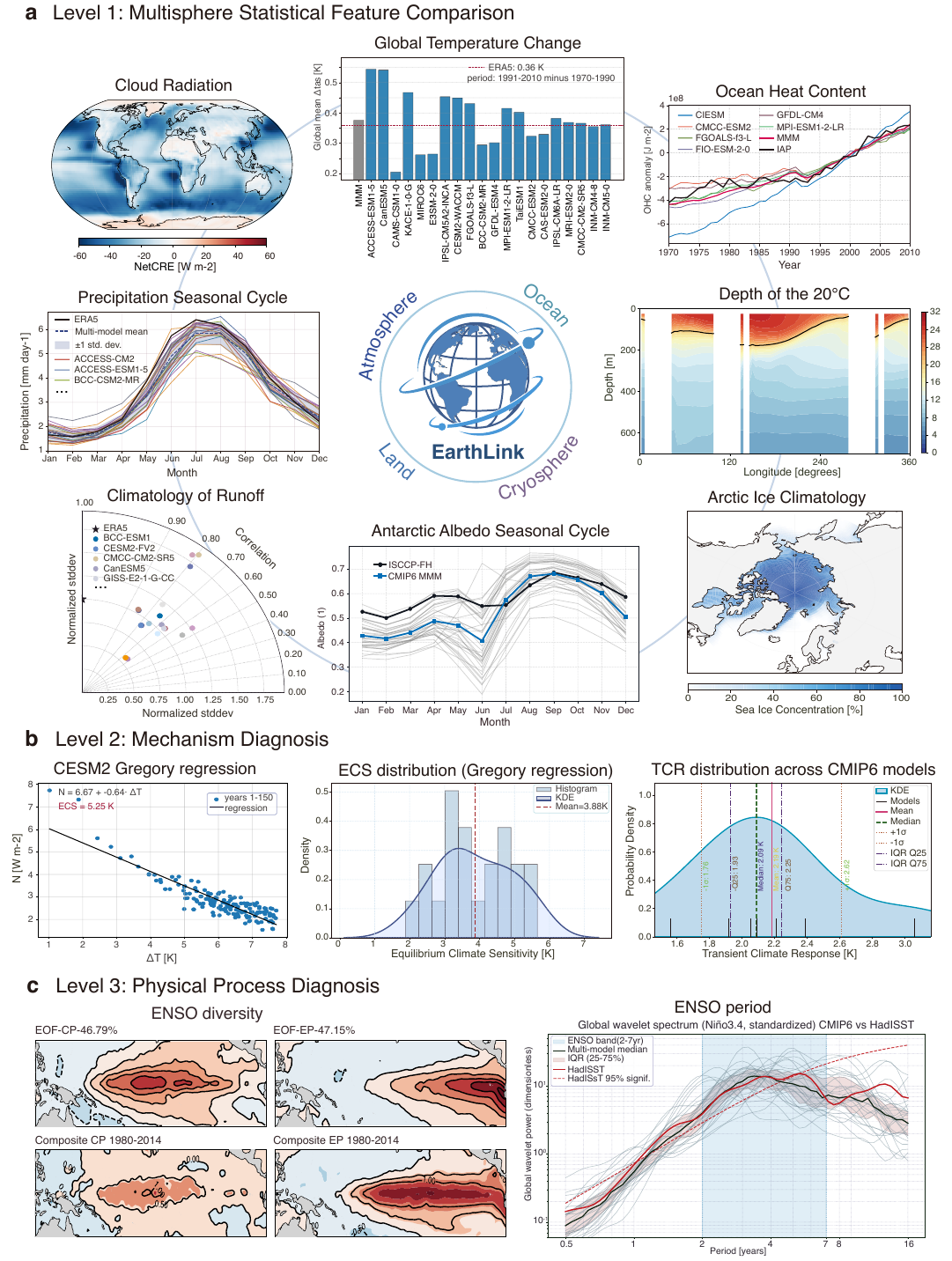}
    \caption{
    \textbf{Multi-level evaluation of \model~on a number of core climate analysis tasks.}
    \textbf{a}, Level 1: Multisphere statistical feature comparison. \model~conducts diagnostic analyses across domains by comparing the CMIP6 simulation of climatological features, such as spatial patterns and variabilities, with observations. Examples include seasonal cycles of precipitation, cloud radiative effects, global temperature change, ocean heat content (OHC) timeseries, 20°C isotherm depth, Arctic ice climatology, Antarctic surface albedo, and runoff. 
    \textbf{b}, Level 2: Mechanistic diagnosis. \model~estimates scenario-driven metrics such as equilibrium climate sensitivity (ECS) and transient climate response (TCR), demonstrating its ability to extract relevant datasets and implement standard diagnostic methods.  
    \textbf{c}, Level 3: Physical process diagnosis. The system performs advanced analyses such as ENSO diversity classification and period detection, displaying emergent capacity in physical reasoning and chain-of-thought synthesis.
    Note that most of the image elements are directly produced by \model, and the others are only slightly adjusted in layout.
    More results are shown in Supplementary Information Section C.  
    }

    \label{fig:fig2}
\end{figure}

To evaluate the scientific capabilities of \model, we designed a multi-level evaluation framework with tasks of increasing complexity foundational to climate science research (see Extended Data Tables~\ref{tab:tab1}–\ref{tab:tab3} and Method section for more details). The first three levels, described in this section, address core tasks that compare model simulations with observational data, a domain with established methodologies and verifiable outcomes (Fig.~\ref{fig:fig2}). Together, they assess how reliably \model~reproduces the reasoning chain underpinning climate model evaluation.

At the first level, we assessed \model's ability to perform a wide range of statistical analyses across different Earth system domains (Fig.~\ref{fig:fig2}a). The system conducted fundamental climatological comparisons, such as calculating spatial patterns of temperature and annual cycles (Extended Data Fig.~\ref{fig:annual_cycle}). It also performed multivariable computations, for example, deriving net cloud radiative forcing from its short-wave and long-wave components (Extended Data Fig.~\ref{fig:cre}), and integrated multidimensional information, exemplified by calculating ocean heat content across various vertical layers (Extended Data Fig.~\ref{fig:ohc}). Across all tasks, \model~produced accurate results and standard diagnostic plots that are in agreement with established climatological findings.

At the second level, we examined \model’s capacity for tasks requiring the integration of physical concepts, experimental design, and statistical analysis. A core diagnostic in this category was estimating equilibrium climate sensitivity (ECS) and transient climate response (TCR) (Fig.~\ref{fig:fig2}b). \model~correctly identified the necessary CMIP6 experiments (e.g., $\text{abrupt4xCO}_2$, $\text{1pctCO}_2$, and piControl), executed standard regression analyses or metrics calculations, and produced ECS and TCR values that fall within the ranges reported by the IPCC AR6 report. Interestingly, when explicitly instructed to estimate ECS without regression, \model~adopted a simple computational approach. It estimated ECS directly from the global temperature change during the quasi-equilibrium period by assuming a linear scaling of radiative forcing with $\text{CO}_2$ concentration. Although the resulting estimate was less accurate, this computation strategy shows that \model~can apply different physical assumptions to the same problem, just like a human.

At the third and most complex of these levels, we assessed \model's capacity for scientific reasoning through a multifaceted diagnosis of the El Niño–Southern Oscillation (ENSO) in CMIP6 models (Fig.~\ref{fig:fig2}c). When prompted with established literature methods (e.g., as in \cite{ashok2007nino,kao2009contrasting,kug2009two}) for classifying ENSO diversity, it correctly implemented the methods to distinguish between Central-Pacific (CP) and Eastern-Pacific (EP) types, reproducing their characteristic spatial patterns. When tasked with analyzing ENSO periodicity, \model~generated custom code of wavelet analysis beyond its existing tool library based on its own reasoning. This generated code correctly identified the 2–7 year periodicity of ENSO.

Taken together, these foundational levels (L1–L3) confirm that \model~can reliably replicate the standard scientific reasoning chain across multiple Earth system domains, from basic statistical comparison to complex, multi-step physical diagnosis.

\subsection{Exploring scientific scenario in climate change}

\begin{figure}[H]
    \centering
    \includegraphics[width=0.99\linewidth]{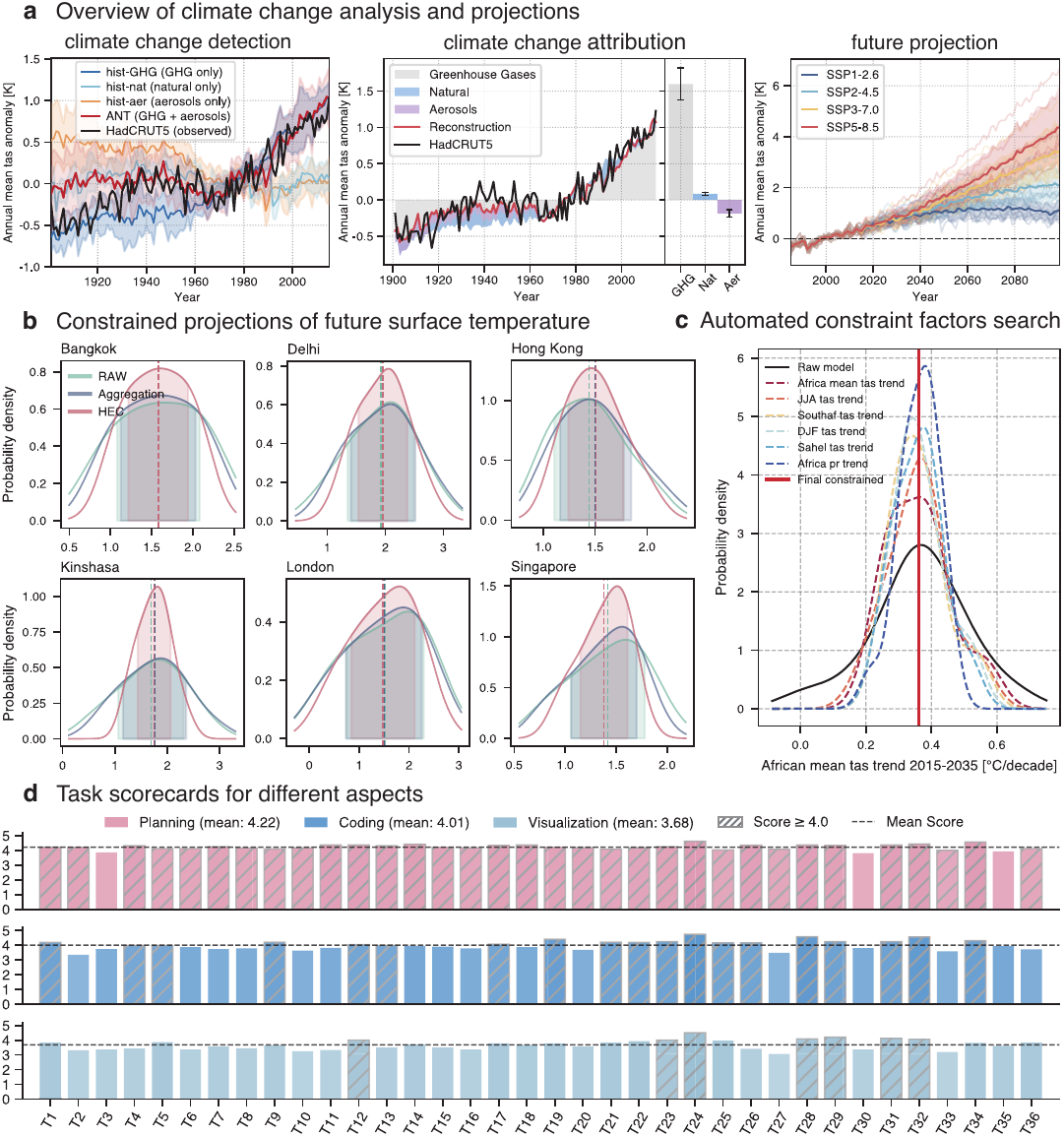}
    \caption{
    \textbf{Application of \model~to tackle future-oriented climate research challenges.}  
    \textbf{a}, Climate change detection, attribution, and future projection. \model~processes multi-model CMIP6 simulations under various experiments, accurately distinguishing between the effects of natural and anthropogenic forcings and generating global temperature anomaly timeseries.  
    \textbf{b}, Constrained projections of future surface temperature for selected regions. Using hierarchical emergent constraints (HEC) and spatial aggregation approaches, \model~reduces projection uncertainty for city-level temperatures under the SSP2-4.5 scenario (2041–2060).
    \textbf{c}, Constrained projections of future temperature changes in Africa using constraining factors automatically identified by \model. 
    Note that most of the image elements in \textbf{a}–\textbf{c} are directly produced by \model, and the others are only slightly adjusted in layout.
    \textbf{d}, Differentiated task scorecard. The system’s performance across evaluation tasks is summarized, highlighting relative strengths in planning, coding, and visualization.
    }
    \label{fig:fig3}
\end{figure}

Having validated \model~on foundational diagnostics, we next tested its ability to manage complex, future-oriented climate change scenarios (Fig.~\ref{fig:fig3}). These tasks move beyond verifiable historical analyses, requiring the system to execute the multi-stage workflows with human guidance.

First, we challenged \model~to execute the complete analysis pipeline for climate change detection, attribution, and projection under multiple DAMIP experiments \cite{gillett2016detection} (Fig.~\ref{fig:fig3}a). The system successfully replicated a workflow analogous to those used in major climate assessments. Its analysis distinguished the roles of natural forcings and anthropogenic drivers, identifying greenhouse gases as the dominant warming contributor. Furthermore, in the context of future projection, \model~correctly processed the multi-dataset inputs under different Shared Socioeconomic Pathways (SSPs) \cite{o2016scenario}, automatically identifying and visualizing key scenarios including SSP1-2.6, SSP2-4.5, SSP3-7.0, and SSP5-8.5, and generating clear, scientifically informative visualizations of the results. 

Next, we posed a more advanced research task involving the constraint of regional climate projections, a critical step for local risk assessment (Fig.~\ref{fig:fig3}b). We focus on temperature changes from 2041 to 2060 for multiple cities in different global regions under the SSP2-4.5 scenario. The system was instructed to apply two distinct methodologies to constrain projection uncertainty, including a spatial aggregation method and the hierarchical emergent constraints (HEC) approach (Supplementary Information Section A.4). It clearly distinguished the two, selected suitable analytical tools, and effectively narrowed the uncertainty range of projections, yielding more reliable regional risk estimates. 
Interestingly, the HEC script was generated autonomously by \model~based on its existing knowledge base. It derived the appropriate formula from the literature and then generated the corresponding code implementation (Extended Data Fig.~\ref{fig:code}). Verification showed that the generated code closely matched expert-developed scripts as in \cite{li2024constraining,li2025constraining}. This demonstrates \model's ability to understand geo-spatial data and its potential to deploy complex algorithms to address scientific problems.

We further evaluated its exploratory capabilities in this user-guided context by tasking \model~to constrain future temperature changes in Africa through the autonomous identification of potential constraining factors (Fig.~\ref{fig:fig3}c). In this setting, the user provided step-by-step guidance through detailed prompts, \model~autonomously handled the execution of each step, such as generating candidate factors from the data, building the regression models, and evaluating their effectiveness. The analysis identified seven effective constraints, including the summer precipitation trend. This demonstrates the system’s ability, facilitated by human-expert guidance, to explore climate-invariant relationships across multiple variables and provide insights into future investigations. See Supplementary Information Section C.4 for more details.


Finally, we conducted a formal multi-expert review to objectively quantify these capabilities. Five independent climate scientists scored \model's outputs on three core competencies: experimental planning, code correctness, and visualization quality (see Methods). We established a threshold for practical utility based on this rubric. An output scoring 4 was deemed practically useful, analogous to the work of a junior researcher, though it may lack efficiency or aesthetic polish. Across the 36 benchmark tasks, \model~achieved this level of performance in multiple cases. A breakdown of the scores revealed a clear hierarchy in its current abilities. Strategic planning emerged as its strongest attribute, followed by code generation, and then visualization quality (Fig.~\ref{fig:fig3}d). 

Beyond the evaluation framework, we further validated \model~by challenging it to reproduce several high-impact climate studies. As detailed in Supplementary Information Section D, the results show that \model~can accurately replicate the key conclusions across multiple cases, highlighting its practical utility for tackling complex, real-world research and enabling more efficient, high-impact scientific workflows.

\subsection{Investigating adaptive capabilities and self-evolving mechanisms}

To examine the adaptive nature of \model~and validate its architectural design, we conducted a series of characterization experiments (Fig.~\ref{fig:fig4}). These experiments assessed how the system adapts to different foundation models and how its performance improves through iterative self-evolution.

We first benchmarked \model’s performance when built upon five state-of-the-art large language models, spanning both open- and closed-source families. Each diagnostic task was executed three times, and all results were validated by human experts to ensure robustness. The evaluation revealed distinct patterns of adaptation and trade-offs among performance, efficiency, and cost. For example, while the open-source Llama-4-Maverick demonstrated high efficiency with minimal time and API costs, it struggled to maintain accuracy on complex tasks. Nevertheless, its reasonable performance on simpler tasks suggests considerable potential, indicating that with further domain-aligned training, open-source models could become competitive alternatives to closed-source ones, offering a faster and more affordable foundation for large-scale scientific automation. In contrast, GPT-5 exhibited the most balanced capability, achieving consistently high success rates across all levels of task complexity while maintaining stable efficiency and fewer debug iterations (Fig.~\ref{fig:fig4}a–d). 

In addition, the core architectural feature of \model~is its self-evolving mechanism, which enables knowledge accumulation and continual enhancement of system capacity (detailed in Methods section). To quantify this mechanism, we compared the system’s performance before and after one complete evolution cycle (Round~1 vs.~Round~2). In the second round, the system operated with updated internal knowledge and tool libraries derived from previous successful runs. The results show a clear evolutionary gain (Fig.~\ref{fig:fig4}e). The distribution of debug rounds shifted markedly toward lower values, particularly for higher-difficulty tasks, and the average time cost per task decreased substantially. 
These improvements show that \model~can progressively refine its reasoning and efficiency through accumulated experience, revealing the emergence of adaptive, self-evolving intelligence.

\subsection{Discovering precursors and mechanisms to understand Atlantic Niño}

To explore \model’s potential as a true scientific discovery partner, we tasked it with a fully open-ended scientific question without any human involvement. As shown in Fig.~\ref{fig:fig5}a, the system was instructed not only to identify the precursors of the summer Atlantic Niño index at an eight-month lead, aiming to improve the forecast skill to 0.5 or higher compared with the 0.39 achieved by current state-of-the-art models~\cite{bachelery2025predicting}, but also to discover the physical pathways that link those precursors.

\model~initiated a complete research workflow, identifying and screening potential precursors before systematically testing them with a suite of regression models (e.g., multiple linear regression, random forest and gradient boosting) and automated parameter tuning. It concluded that the 0.5 skill target was not reachable but identified a novel combination of predictors achieving a forecast skill (TCC) of 0.46, a notable improvement over the baseline (Fig.~\ref{fig:fig5}b). Notably, while Indo-Pacific teleconnections like ENSO (Niño 3.4) were included as potential drivers in the experimental design, \model’s analysis revealed their predictive contribution was minimal (Fig.~\ref{fig:fig5}c), suggesting that the Atlantic Niño’s development at this lead time is primarily governed by local oceanic and atmospheric preconditions.

In the final stage of its analysis, \model~further refines this mechanism through its integrative reasoning (Fig.~\ref{fig:fig5}c). It identified an unrecognized precursor of the Atlantic Niño that appears about eight months in advance, characterized by late-autumn westerly anomalies and subsurface heat content changes over the equatorial Atlantic. Building on this relationship, \model~uncovered a localized pathway in which late-autumn westerly anomalies trigger an eastward-propagating downwelling Kelvin wave. This process deepens the thermocline, allowing heat to be stored in the subsurface. The resulting subsurface heat anomaly is subsequently amplified through ocean-atmosphere coupling via the Bjerknes feedback in the following spring, leading to a peak sea surface temperature anomaly in the ATL3 region during summer (JJA). This finding enhances our understanding of the physical origins of the event. Furthermore, the analysis summarized other potential mechanisms, including the coastal Angola–Benguela influence and inter-basin teleconnections, and proposed specific directions for further model improvement (Supplementary Information Section E.1). While this AI-generated hypothesis requires further validation, it highlights \model’s potential to autonomously analyze complex climate systems and to act as an emerging research partner. This capability is further demonstrated in additional cases, as detailed in Supplementary Information Section E.

\section{Discussion}\label{sec3}

In this study, we introduce \model, an AI agent system designed to function as an interactive “copilot” for climate science. By combining a domain-specific knowledge base with dynamic planning and code generation, \model~aims to significantly streamline the scientific workflow. Multi-level evaluations demonstrate its ability to perform core scientific analyses and address complex research problems in a manner comparable to a junior human researcher. This result represents an important step toward advancing human scientific capability, shifting the scientist’s role from manual data processing to higher-level scientific judgment, substantially shortening the research cycle for complex analyses from months to days, and thereby accelerating the overall pace of scientific discovery.

In contrast to established diagnostic toolkits like ESMValTool, \model~introduces a paradigm of flexibility and interactivity. Our goal is not to replace these invaluable tools, but to unlock their full potential through intelligent orchestration. Static toolkits are limited because their components are confined within rigid frameworks. \model~addresses this by treating trusted, community-vetted functions from these tools as individual, callable tools. By doing so, it transforms a monolithic program into a flexible suite of capabilities that can be dynamically chained, combined with newly generated code, and applied to novel scientific questions. This reframes the future of scientific software as a composable ecosystem where the reliability of established tools is fused with the agility of an AI agent. In this context, \model~can further contribute to community initiatives such as the CMIP7 Rapid Evaluation Framework (REF)~\cite{hoffman2025rapid} by offering transparent workflows.

The true potential of this paradigm is realized in open-ended, human–machine collaborative discovery. In our Atlantic Niño forecast experiment, for example, the system demonstrated human-like scientific reasoning. Rather than merely improving a metric, it proposed a novel and testable hypothesis for the underlying mechanism by integrating evidence from observational data with existing mechanistic understanding. This demonstrates how an AI partner can navigate vast datasets and complex knowledge to generate creative yet plausible mechanisms, while human scientists contribute validation, interpretation, and critical judgment. As detailed in the Supplementary Information Section E, this synergy further enabled us to uncover potential teleconnections between Arctic sea ice and tropical climate signals through comprehensive literature and data reviews. Together, such collaboration empowers us to investigate scientific questions with unprecedented depth, breadth, and speed.

Despite these capabilities, it is critical to understand \model's limitations. Its reasoning is essentially interpolative, excelling at combining existing knowledge, but generating entirely new physical theories from first principles remains a challenge. Furthermore, its proficiency is directly tied to the quality of its knowledge base and the clarity of user prompts. A significant risk is the generation of ``plausibly wrong" outputs, code that runs without error but produces scientifically incorrect results. This risk underscores why our emphasis on transparent workflows is a prerequisite for trustworthy AI in science. It reaffirms that the goal is not an infallible virtual scientist, but a powerful yet imperfect tool that demands a partnership with a human expert.

Finally, our long-term vision extends beyond an open-access analysis platform. We propose that \model’s architecture provides a way to address the persistent challenge of data fragmentation in the Earth sciences.
\model~’s natural language interface functions as a universal translator, handling the tasks of locating, accessing, and harmonizing these heterogeneous data sources. As it processes more cross-domain requests, it constructs an internal semantic map of the global Earth data, becoming progressively more efficient through its self-evolving mechanism. Ultimately, we aim to develop \model~into a global, open, continuously learning resource. Such a system would accelerate discovery, lower the barriers to advanced computational analysis, and enable the entire scientific community, including resource-limited groups, to participate fully in Earth science, advancing both research progress and scientific equity.


\putbib[sn-bibliography]

\clearpage

\clearpage

\clearpage

\clearpage

\begin{figure}[H]
    \centering\includegraphics[width=0.99\linewidth]{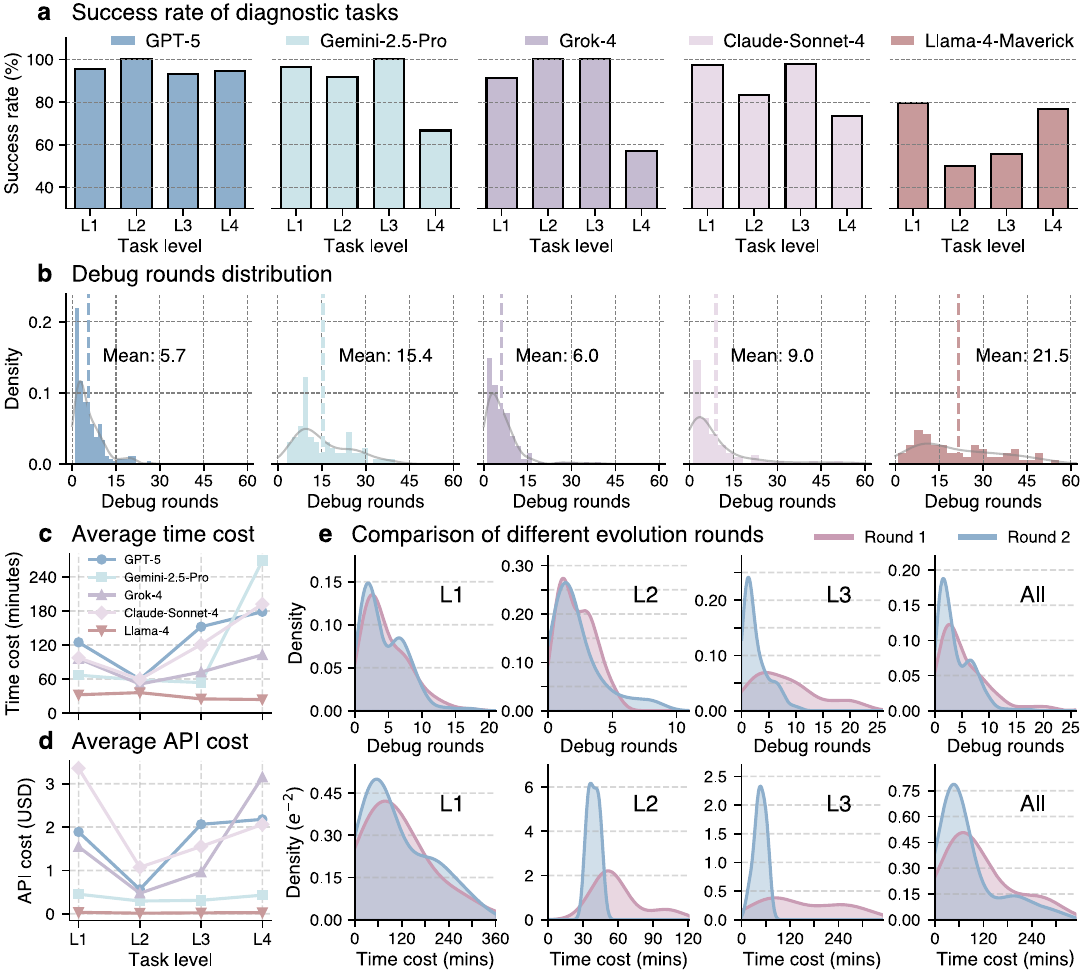}
    \caption{
    \textbf{Comparison of different foundation models and self-evolution rounds.} 
    \textbf{a}, Mean success rates of diagnostic tasks across four task levels for five foundation models (GPT-5, Gemini-2.5-Pro, Grok-4, Claude-Sonnet-4, and Llama-4-Maverick). 
    \textbf{b}, Distributions of debug rounds across all task levels for the same models, with colors corresponding to those in \textbf{a}. 
    \textbf{c},\textbf{d}, Mean time cost (\textbf{c}) and API cost (\textbf{d}) across task levels. 
    \textbf{e}, Comparison of debug rounds and time cost distributions between the first and second evolution rounds (using GPT-5 as the foundation model). Tasks at level 4 are semi-open problems and are therefore excluded from the evolution experiments.
    }    
    \label{fig:fig4}
\end{figure}

\begin{figure}[H]
    \centering
    \includegraphics[width=0.99\linewidth]{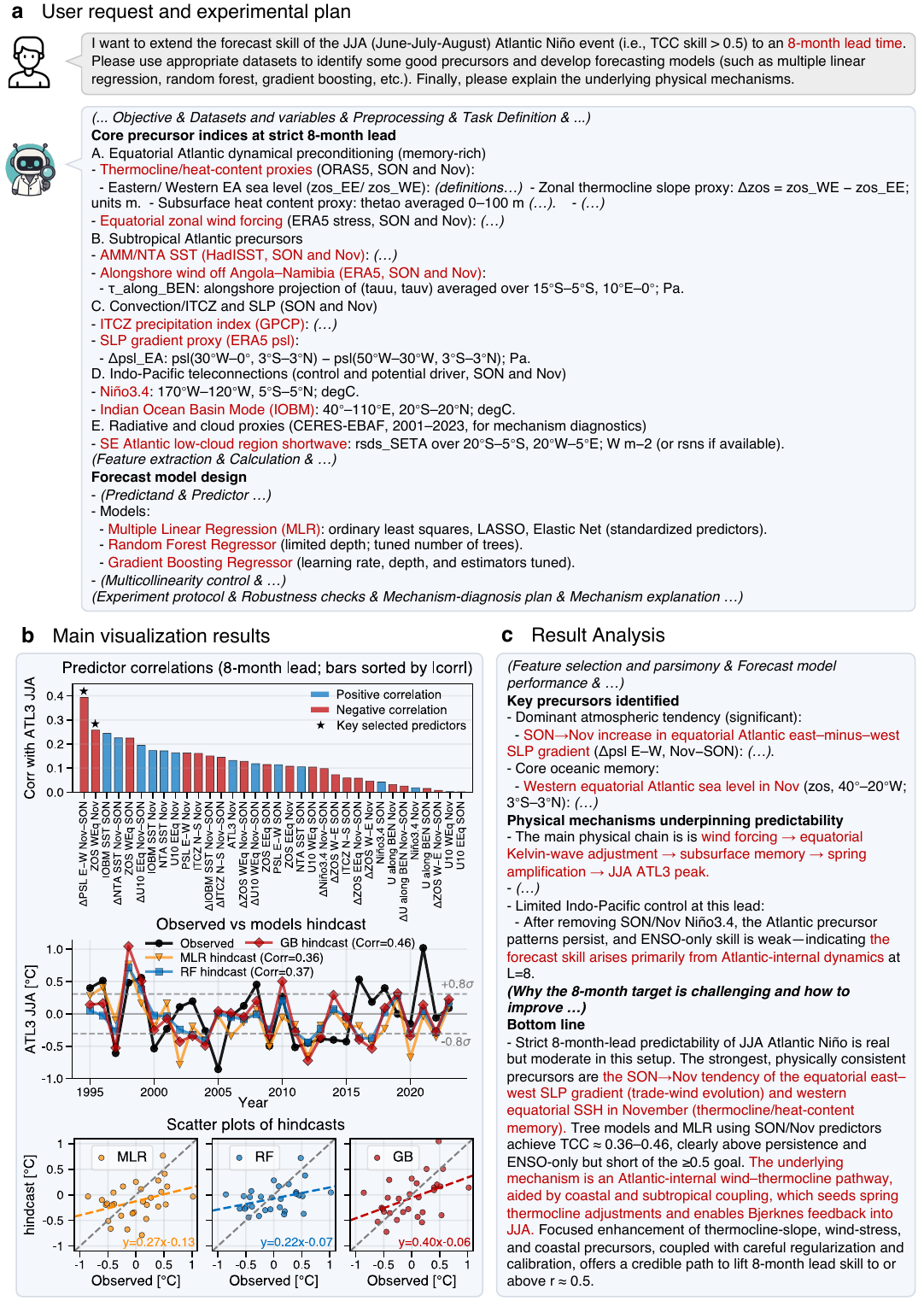}
    \caption{
    \textbf{\model’s autonomous research process for improving Atlantic Niño predictability.} 
    \textbf{a}, User request and experimental plan. The user tasked \model~to enhance the 8-month-lead forecast skill of the summer Atlantic Niño index (ATL3). The system independently designed a full experimental plan covering precursor selection, model setup, validation protocol, etc. 
    \textbf{b}, Main visualization results. 
    \model~executed the experimental plan, producing visualizations of the (top) correlations between 8-month-lead predictors and the ATL3 index, (middle) a time-series comparison of observed values against hindcasts from the multiple linear regression (MLR), random forest (RF), and gradient boosting (GB) models, and (bottom) scatter plots evaluating the hindcast skill of each model.
    Most of the image elements are directly produced by \model, and the others are only slightly adjusted in layout. 
    \textbf{c}, Result analysis. \model~integrated the diagnostic outcomes into a concise mechanistic summary, outlining the physical processes that link the identified precursors to the subsequent development of the Atlantic Niño.
    }
    \label{fig:fig5}
\end{figure}

\clearpage

\renewcommand\linenumberfont{\normalfont\scriptsize}
\setlength\linenumbersep{24pt}


\section{Methods}\label{sec4}

\subsection{Implementation of \model}

\model~is an automated, modular system designed to tackle scientific tasks in earth science through multi-agent collaboration. The system integrates external knowledge, domain-specific data, and tool libraries into a multi-stage framework that supports scientific task planning, experiment execution, result validation, and application-relevant analysis. The system is optimized for large-scale geoscientific analysis and is built to accommodate evolving tasks in Earth system science.

\model~comprises three core modules: (1) a Planning Module for experiment plan generation and refinement, (2) a Scientific Diagnosis Module for automated experiment execution and validation, and (3) a Multi-Scenario Analysis Module for comparative interpretation and synthesis. These modules interact with integrated libraries of structured knowledge, climate data, and computational tools, allowing dynamic composition and execution of scientific pipelines.

\subsubsection{Core modules}

\noindent \textbf{Planning Module.} The planning module initiates the analysis pipeline by interpreting a scientific request, typically expressed in natural language, and converting it into a formal analysis plan. Specifically, a Planning Agent parses the query $Q$ to extract key task descriptors and target variables, optionally utilizing uploaded literature $L$ (Supplementary Information Section A.2) to improve contextual understanding.
The agent consults a Knowledge Library, which contains structured representations of prior plans, domain-specific ontologies, and application programming interface (API) documentation. It generates multiple candidate plans, each represented as a sequence of analytic operations.

These candidate plans are ranked and aggregated by a Plan Aggregation Agent, which merges semantically similar pathways and prioritizes plans based on task feasibility, data availability, and methodological coherence (Supplementary Information Section B.2). Human-in-the-loop oversight is optionally included at this stage to ensure scientific interpretability and validity. The finalized optimal plan $p^\star$ is then formalized as an executable pipeline. The workflow of this module can be described as:
\begin{equation}
    p^\star = \mathcal{A}\big(\mathcal{P} \mid H\big),\ \mathcal{P} = \mathcal{G}(Q,L) = \{p_1, p_2, ..., p_m\}
\end{equation}
where $\mathcal{A}$ represents plan aggregation, $H$ refers to human supervision and $\mathcal{G}$ for candidate plan generation. $p_i$ refer to candidate plan and $p^\star$ for finalized plan.

\noindent \textbf{Scientific Diagnostic Module.} 
The formalized plan is passed to a Coding Agent, which dynamically constructs the corresponding executable codes with reference of computational functions from the Tool Library, which includes preprocessing routines, model interfaces, statistical estimators, and visualization utilities.
The plan executes on heterogeneous climate data sources, including simulations from CMIP6, reanalysis products, and observational datasets housed in the Data Library, which is organized across key Earth subsystems (e.g., ocean, land, cryosphere, atmosphere).

To ensure robustness and reliability, \model~incorporates a feedback-driven validation loop. A Result Checking Agent performs runtime diagnostics to detect statistical anomalies in outputs. Concurrently, an Image Feedback Agent inspects visualization quality and scientific consistency, identifying outputs that lack interpretability or contain outlier behavior.
When inconsistencies are detected, the system modifies parameters, recomposes functions, or reselects tools using an autonomous error-recovery routine (Supplementary Information Section B.3). This loop iterates until the convergence criteria, defined by task-specific success thresholds, are met. All successful plans and scripts are stored back into the Knowledge and Tool Libraries after being verified by experts, thereby achieving knowledge accumulation and continual improvement of system capacity. This module can be modeled as:
\begin{equation}
O^{(k+1)} = \mathcal{E}\!\left(\mathcal{M}\big(\mathcal{F}(p^\star, \mathcal{T})^{(k)}\big), \mathcal{D}\right),
\quad \text{s.t. } \mathcal{V}(O^{(k+1)}) = 1
\end{equation}
where $k$ denotes the debug round, and $\mathcal{F}$, $\mathcal{M}$, $\mathcal{E}$, $\mathcal{V}$ refer to functionality script generator, error modifier, code executor and criteria verifier, respectively. $\mathcal{V}(O^{(k+1)})$ means that convergence criteria is satisfied. $\mathcal{T}$, $\mathcal{D}$ are tool library and data library available.

\noindent\textbf{Multi-Scenario Analysis Module.} 
After successful execution and validation of the computational workflow, the resulting visualizations are transferred to the Analysis and Summary Agents. These agents synthesize the outputs into coherent scientific reports that capture the core findings of each experiment. The reports can be adapted to various application scenarios, such as climate risk assessment, energy and agricultural planning, environmental policy support, or public health evaluation (See Supplementary Information Section B.1).

By automating this critical interpretive step, the Multi-Scenario Analysis Module bridges the gap between data analysis and domain-relevant insights, enhancing both the accessibility and impact of the system’s outputs. This ensures that results are not only computationally accurate but also scientifically interpretable and practically useful across diverse decision-making contexts.

\subsubsection{Foundation LLMs}

As shown in the Fig.~\ref{fig:fig4}, GPT-5 consistently achieves the highest success rates across all four levels of task complexity and requires the fewest debug rounds to find a correct solution. Although its API cost is slightly higher than that of other models, its superior performance and efficiency make it the most robust and well-balanced choice. Therefore, we select GPT-5 as the foundation model for all modules in our framework to ensure the highest overall performance.

Furthermore, recognizing that different components within our framework have varying requirements for output diversity and reasoning depth, we assign distinct parameter configurations to the LLM for each module. This tailored approach ensures that each component is optimized for its specific task. The detailed parameter settings for each module are provided in the Supplementary Information Section A.3.1.

\subsubsection{Resource library}

Three foundational libraries underpin \model’s automated workflow, providing the knowledge, data, and computational resources necessary for reproducible and extensible climate research.

\noindent \textbf{Knowledge Library.} 
The Knowledge Library serves as a dynamic hub that supports both the Planning and Scientific Diagnosis Modules. It maintains a structured and semantically indexed knowledge base containing expert-validated diagnostic frameworks, methodological templates, and modular planning units, allowing planning agents to rapidly access prior research strategies and definitions relevant to user queries. In addition, the library provides coding agents with access to package-level documentation, including APIs from ESMValTool and other domain-specific toolkits, to facilitate the retrieval of standardized workflows and function usage. Beyond the curated offline resources, the Knowledge Library integrates online search capabilities, enabling agents to query up-to-date literature, repositories, and API documentation on the web. Through the continuous incorporation of validated outputs and user feedback, it evolves to combine static domain expertise with dynamically acquired information, ensuring both scientific rigor and adaptability in automated plan and code generation.

\noindent \textbf{Data Library.} 
The Data Library integrates climate model outputs and observational datasets into a unified, standardized repository for automated analysis. The current archive encompasses nearly all major activities of CMIP6, including CMIP, DAMIP~\cite{gillett2016detection}, GeoMIP~\cite{kravitz2011geoengineering}, and ScenarioMIP~\cite{o2016scenario}, covering over one hundred models, each with at least one ensemble member and monthly to annual data. In addition, the library incorporates observational datasets such as obs4MIPs~\cite{waliser2020observations}, HadISST~\cite{rayner2003global}, HadCRUT5~\cite{morice2021updated}, GPCP-SG~\cite{adler2003version}, and ERA5~\cite{hersbach2019era5}, with a total volume exceeding 3 PB. The library is continuously expanded to incorporate the full CMIP6 archive, with plans to include CMIP7 and subsequent datasets. By harmonizing model simulations and observations across temporal and spatial scales, the Data Library provides comprehensive coverage of the Earth system, including atmosphere, ocean, land, and cryosphere, and serves as an open resource to support global community collaboration.

\noindent \textbf{Tool Library.}
The Tool Library integrates a diverse suite of open-source tools and expert-developed algorithms for data processing, evaluation, and visualization. It includes widely used diagnostic packages such as ESMValTool~\cite{righi2020earth}, PCMDI Metrics~\cite{lee2024systematic}, and CDO, as well as Python libraries like xarray, cartopy, iris, eofs, and scikit-learn. Advanced methods, including Bayesian emergent constraints and machine-learning–based diagnostics, are also supported. To ensure interoperability, the Tool Library enables seamless translation of workflows across multiple programming languages (e.g., NCL, R, MATLAB) into standardized Python scripts via LLM-based code generation. Each tool is annotated with metadata on usage constraints and functional scope, enabling efficient task matching. Newly validated scripts and methods can be continuously incorporated, ensuring that \model~remains capable of performing both standardized community diagnostics and cutting-edge scientific analysis. In future versions, the Tool Library may also be extended to include both data-driven models~\cite{kochkov2024neural,guo2025data} and general circulation models (GCMs), encapsulated as agent-accessible modules to support predictive analysis, process attribution, and sensitivity experiments.

Together, these three libraries provide a scalable and extensible infrastructure that supports the generation, execution, and validation of automated workflows, while ensuring transparency, reproducibility, and adaptability across a wide range of climate science applications. More details about the construction and interfaces of these libraries can be found in Supplementary Information Section A.1.

\subsection{Evaluation framework}

\subsubsection{Task design} \label{sec:task_design}

To comprehensively evaluate the scientific capabilities of \model, we established a hierarchical task framework that reflects the increasing complexity of real-world climate research. This framework systematically divides climate analysis workflows into five levels, each targeting specific aspects of scientific reasoning, data processing, and analytical challenge (see Extended Data Tables~\ref{tab:tab1}–\ref{tab:tab3}). The tasks at each level were carefully designed to test \model’s ability to understand, reason, and operate across a spectrum of scientific scenarios.

\noindent\textbf{Level 1: Simple statistical analysis.}
The first level focuses on essential climatological tasks, such as data retrieval, preprocessing, calculation of annual means, spatial distributions, and interannual variability. These tasks serve as foundational exercises, testing \model’s proficiency in handling basic data structures, performing standard computations, and generating visualizations to support initial model evaluation. With 23 tasks at this level, the emphasis is on routine yet fundamental operations necessary for preliminary climate data analysis and model-observation comparison.

\noindent\textbf{Level 2: Mechanistic diagnosis.}
The second level introduces moderately complex scientific problems that require a mechanistic understanding and integration of multiple datasets. For example, tasks include estimating Equilibrium Climate Sensitivity (ECS) and Transient Climate Response (TCR), which involve understanding the physical diagnostic framework, selecting relevant experiments, and applying statistical tools such as regression analysis. Six tasks were designed at this level to assess \model’s ability to synthesize information across experiments and perform physically grounded diagnostics that move beyond simple descriptive analysis.

\noindent\textbf{Level 3: Complex scientific reasoning.}
At the third level, tasks demand advanced scientific reasoning and methodological rigor. Here, \model~must decompose complex climate phenomena, such as the diversity and periodicity of the El Ni\~no–Southern Oscillation (ENSO), into logical subtasks. This often requires integrating advanced analytical techniques (e.g., Empirical Orthogonal Function decomposition, composite analysis) and specialized domain knowledge. The six tasks in this category challenge \model~to construct and execute extended reasoning chains, demonstrating both technical proficiency and conceptual understanding of complex climate dynamics.

\noindent\textbf{Level 4: Semi-open scientific problem.}
The fourth level addresses semi-open scientific problems, which more closely simulate the open-ended questions encountered in practical climate research and policy assessment. In these cases, \model~is provided step-by-step human guidance and expected to autonomously select appropriate datasets, combine physical insight with adaptive workflows, and apply constraint methods (such as emergent constraint approaches) to refine projections and deliver decision-oriented recommendations. This level consists of a single, highly integrative task designed to probe \model’s capacity for innovation, uncertainty quantification, and workflow adaptation in addressing climate-change imposed challenges.


By structuring the evaluation in this way, we are able to systematically investigate \model’s strengths and limitations across a wide range of research scenarios. This graded approach not only benchmarks current performance but also highlights the pathways for future system enhancements and the evolution of AI agents towards more autonomous, creative scientific inquiry.

\subsubsection{Basic evaluation and scoring criteria}
To ensure rigorous and reproducible assessment of \model’s performance, we established a comprehensive scoring rubric spanning three core dimensions of scientific practice: (1) Experimental Planning and Method Design, (2) Coding Implementation, and (3) Result Synthesis and Visualization. Each task was evaluated independently by expert reviewers, who assigned scores using a 5-point Likert scale, with clear, criterion-based descriptors for each level (see Extended Data Table~\ref{tab:tab4}). This structured approach emphasizes both technical correctness and scientific reasoning, ensuring alignment with community standards and best practices.

\noindent \textbf{Experimental planning and method design.}
This dimension evaluates the clarity, scientific rigor, and feasibility of the proposed workflow. A score of 5 indicates that the plan is complete, logically sound, and scientifically executable with no evident flaws. Lower scores reflect increasing levels of methodological ambiguity, omission of key steps, or misunderstanding of core scientific concepts.

\noindent \textbf{Coding implementation.}
This dimension assesses whether the generated code is syntactically correct, functionally complete, and aligned with the proposed plan. Top scores are awarded when the system produces robust, ready-to-run scripts that require minimal or no debugging. Points are deducted for significant reliance on self-correction cycles or logical errors in tool usage.

\noindent \textbf{Result synthesis and visualization.}
This dimension focuses on the interpretability, clarity, and presentation quality of outputs—ranging from diagnostic plots to explanatory text. High-scoring outputs exhibit clean visual aesthetics, accurate labeling (units, axis titles, captions), and scientifically coherent narratives suitable for reports or publications. Lower scores reflect graphical or textual inconsistencies, poor formatting, or incomplete result delivery.

Each task was evaluated by three independent domain experts, with scores averaged across reviewers. Reviewers followed the comprehensive rubric shown in Extended Data Table~\ref{tab:tab4}, which specifies explicit standards for each score level. 
This rubric ensures fairness and objectivity in the evaluation of \model~across diverse scientific scenarios. Tasks with composite average scores above 4 were categorized as ``expert-level"; those between 2.5 and 4 were ``research-ready with minor oversight"; and scores below 2 signified either methodological or technical deficiencies that require significant expert intervention.

\section*{Data availability}
All datasets used in this study can be obtained from publicly available sources. The CMIP6 database can be obtained from \url{https://esgf-node.llnl.gov/search/cmip6/}. The full observational data source can be found in Supplementary Information Section A.1.2.

\section*{Code availability}

The code of \model~will be publicly available at \url{https://github.com/OpenEarthLab/EarthLink}. The deployed system is accessible at \url{https://earthlink.intern-ai.org.cn}.


\section*{Acknowledgements}
This work was supported by a locally commissioned task from the Shanghai Municipal Government. We acknowledge the World Climate Research Programme (WCRP), which, through its Working Group on Coupled Modelling, coordinated and promoted CMIP. We thank the climate modeling groups for producing and making available their model output, the Earth System Grid Federation (ESGF) for archiving the data and providing access and the open source scripts repository from Earth System Modelling and Observation (ESMO) group. In addition, we would like to express our sincere gratitude to the domain experts and research scientists in the climate community, whose insights helped us to conduct the evaluation. We thank the website development team (Shaowei Hou and Zheng Nie, et al.) and the data reserve team (Dong Zheng and Qihao Zheng, et.al.) of Shanghai Artificial Intelligence Laboratory for helping us quickly build the platform and database. J.-J L is supported by National Natural Science Foundation of China (Grant No. 42088101 and 42030605). W.O. is funded by The Hong Kong Jockey Club Charities Trust, the Research Grants Council of Hong Kong (Project No. CUHK14213224).

\section*{Author contribution}

F.L. and L.B. led and supervised this work. Z.G., J.W. and X.Y. contributed to the implementation of \model~and ran experiments. F.L., W.W., Z.G., J.W. and X.Y. provided analysis of the results. Z.J., W.X., B.F., W.Z. and X.G. provide technical advice and ideas. J.-J.L., L.C., Y.-G.H., F.S., P.G., T.Y., C.L., Y.W., T.C. and W.O. contributed feedback and domain insight into the project. F.L., Z.G., J.W., X.Y. and W.W. wrote the paper. All authors provided feedback and contributed to the editing of the manuscript.

\section*{Conflict of Interest}
The authors declare no competing interests.



\clearpage

\nolinenumbers

\section*{Extended Data}
\renewcommand{\figurename}{Extended Data Fig.}
\renewcommand{\tablename}{Extended Data Table}
\setcounter{figure}{0}
\setcounter{table}{0}

\begin{figure}[H]
    \centering
    \includegraphics[width=0.99\linewidth]{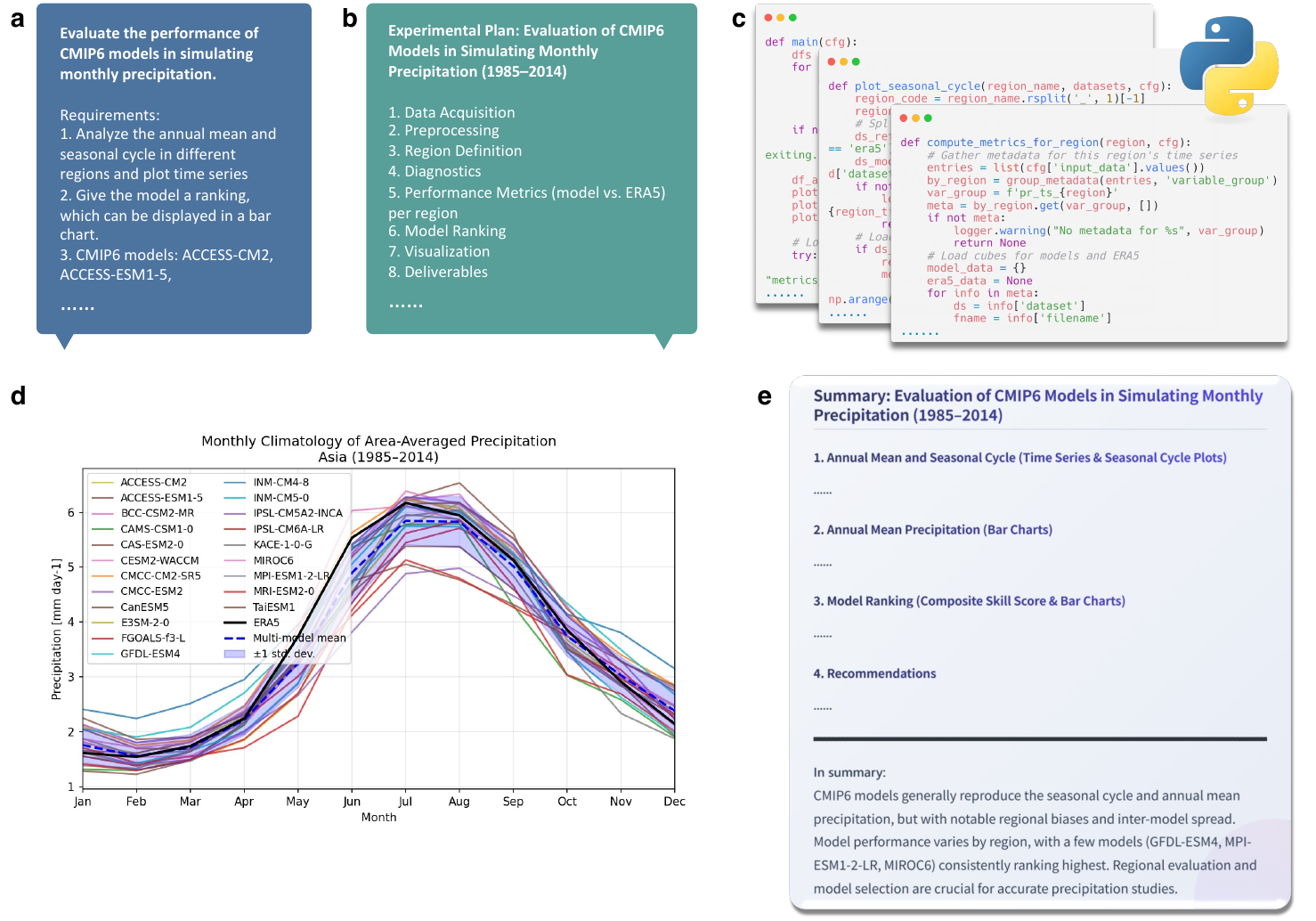}
    \caption{
    \textbf{Evaluation of CMIP6 models in simulating seasonal cycles of monthly precipitation.} 
    \textbf{a}, Task definition and diagnostic requirements, including annual mean and seasonal cycles and regional time series analyses.  
    \textbf{b}, Automated planning output from \model, detailing end-to-end workflow from data acquisition to deliverables.  
    \textbf{c}, Example code snippet generated by the system for data processing and analysis.  
    \textbf{d}, Modelled and observed precipitation seasonal cycles over a selected region. 
    \textbf{e}, Automated textual interpretation of the results, providing a plain-language summary generated by the system.
    A more complete case can be found in the Supplementary Information Section C.1.1.
    }
    \label{fig:annual_cycle}
\end{figure}
\clearpage

~
\begin{figure}[!t]
    \centering
    \includegraphics[width=0.99\linewidth]{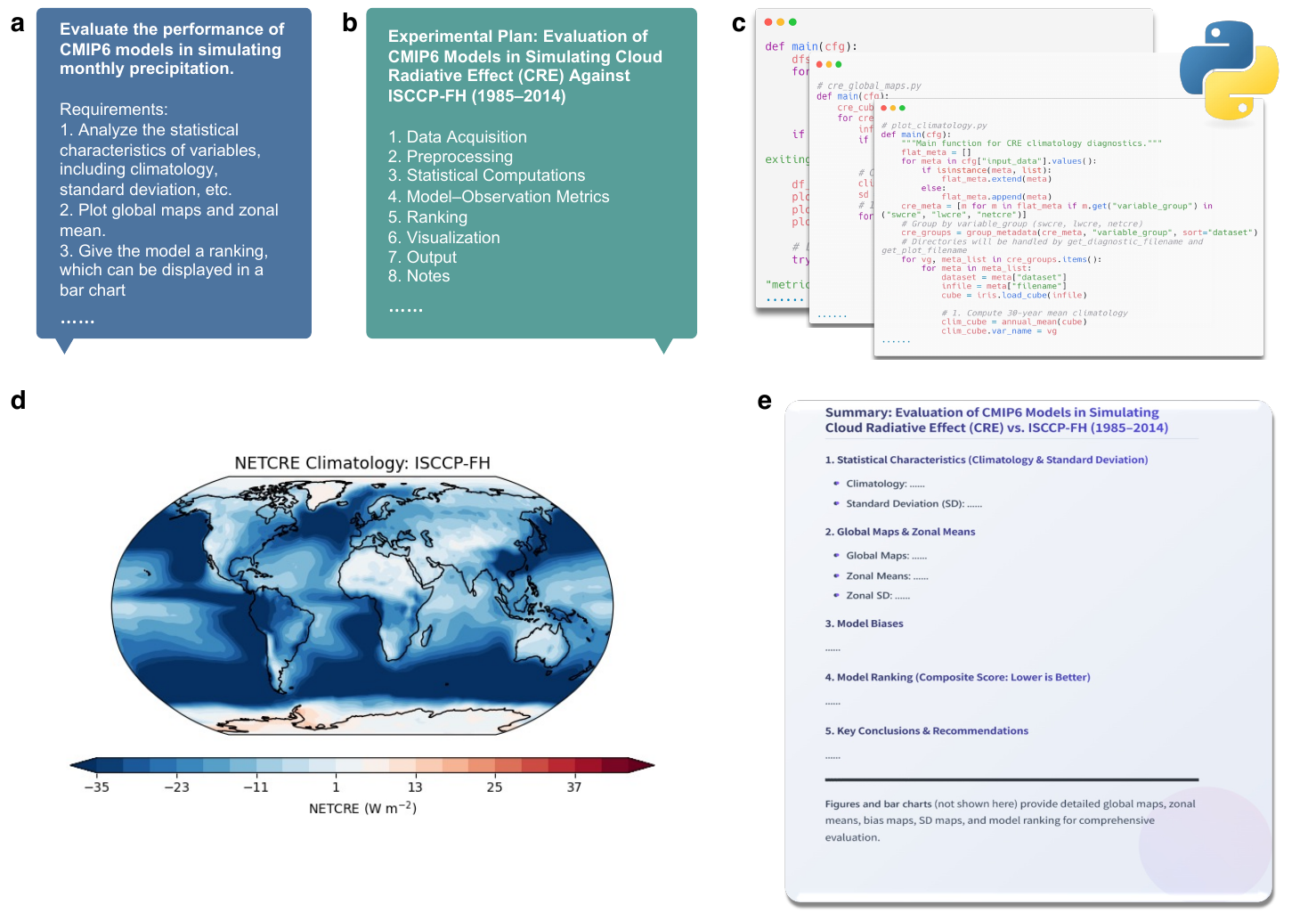}
    \caption{
    \textbf{Benchmarking CMIP6 model simulation of cloud radiative effects (CRE) against ISCCP-FH observations.}
    \textbf{a}, Task setup, requiring statistical mapping of climatology and variability and model ranking based on performance metrics.  
    \textbf{b}, End-to-end automated planning by \model, including data acquisition, computation, ranking, and visualization steps.  
    \textbf{c}, Representative code snippet for automated CRE analysis.  
    \textbf{d}, Global maps and zonal means of CRE climatology. 
    \textbf{e}, Automated textual summary of findings, demonstrating \model’s interpretative and reporting capabilities.
    A more complete case can be found in the Supplementary Information Section C.1.2.
    }
    \label{fig:cre}
\end{figure}
\clearpage

~
\begin{figure}[!t]
    \centering
    \includegraphics[width=0.99\linewidth]{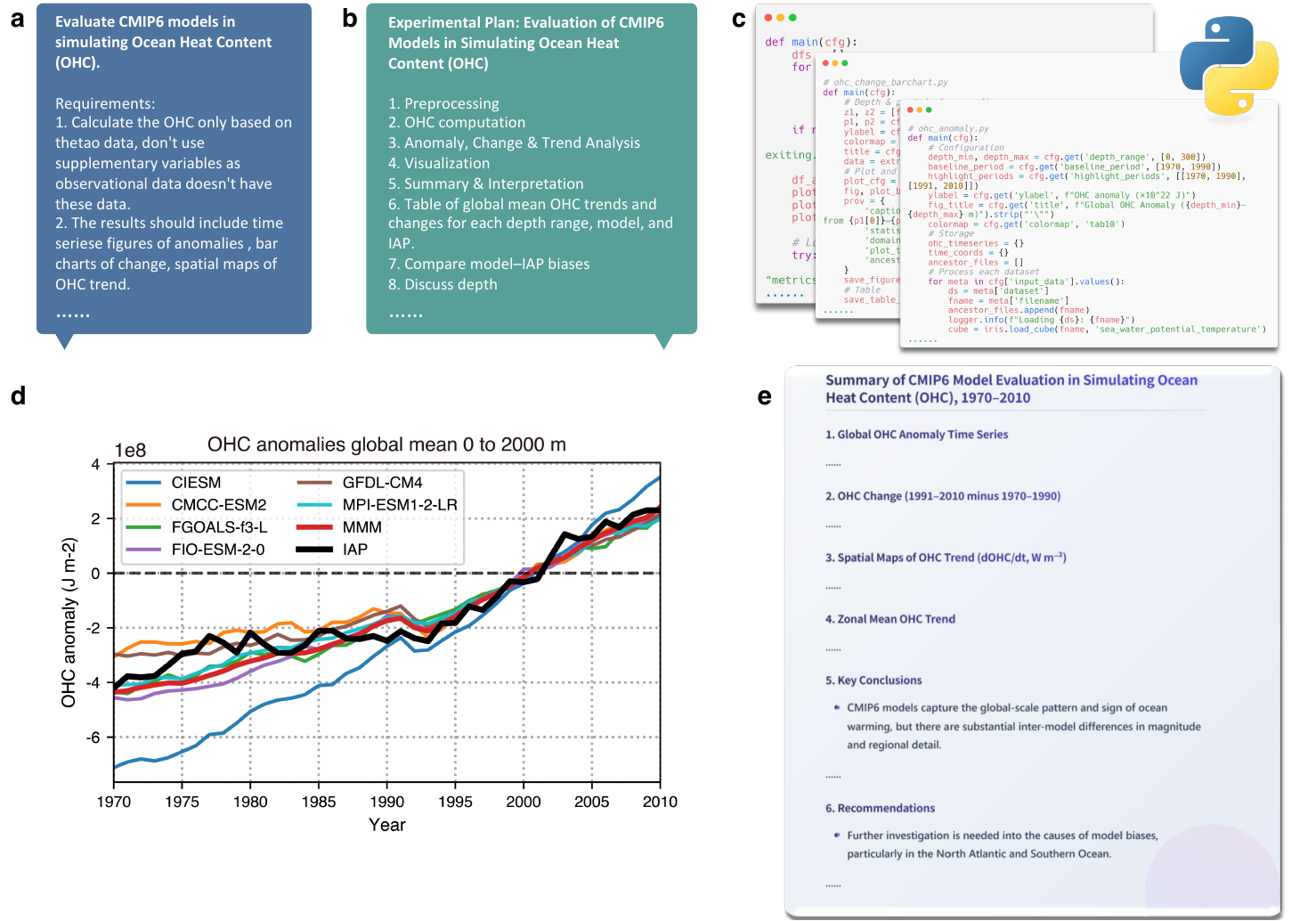}
    \caption{
    \textbf{Automated evaluation of ocean heat content (OHC) simulation in CMIP6 models.}
    \textbf{a}, Task framing for OHC evaluation using only thetao data to ensure comparability with observed datasets.  
    \textbf{b}, Full experimental plan drafted by \model~for preprocessing, calculation, anomaly/time trend analysis, and visualization.  
    \textbf{c}, Example code generated by the system for OHC calculation and plotting. 
    \textbf{d}, Time series of modeled and observed OHC anomalies. 
    \textbf{e}, Automated text interpretation and summary table, including biases and quantitative comparisons between models and observations.
    A more complete case can be found in the Supplementary Information Section C.1.3.
    }
    \label{fig:ohc}
\end{figure}
\clearpage

~
\begin{figure}[!t]
    \centering
    \includegraphics[width=0.99\linewidth]{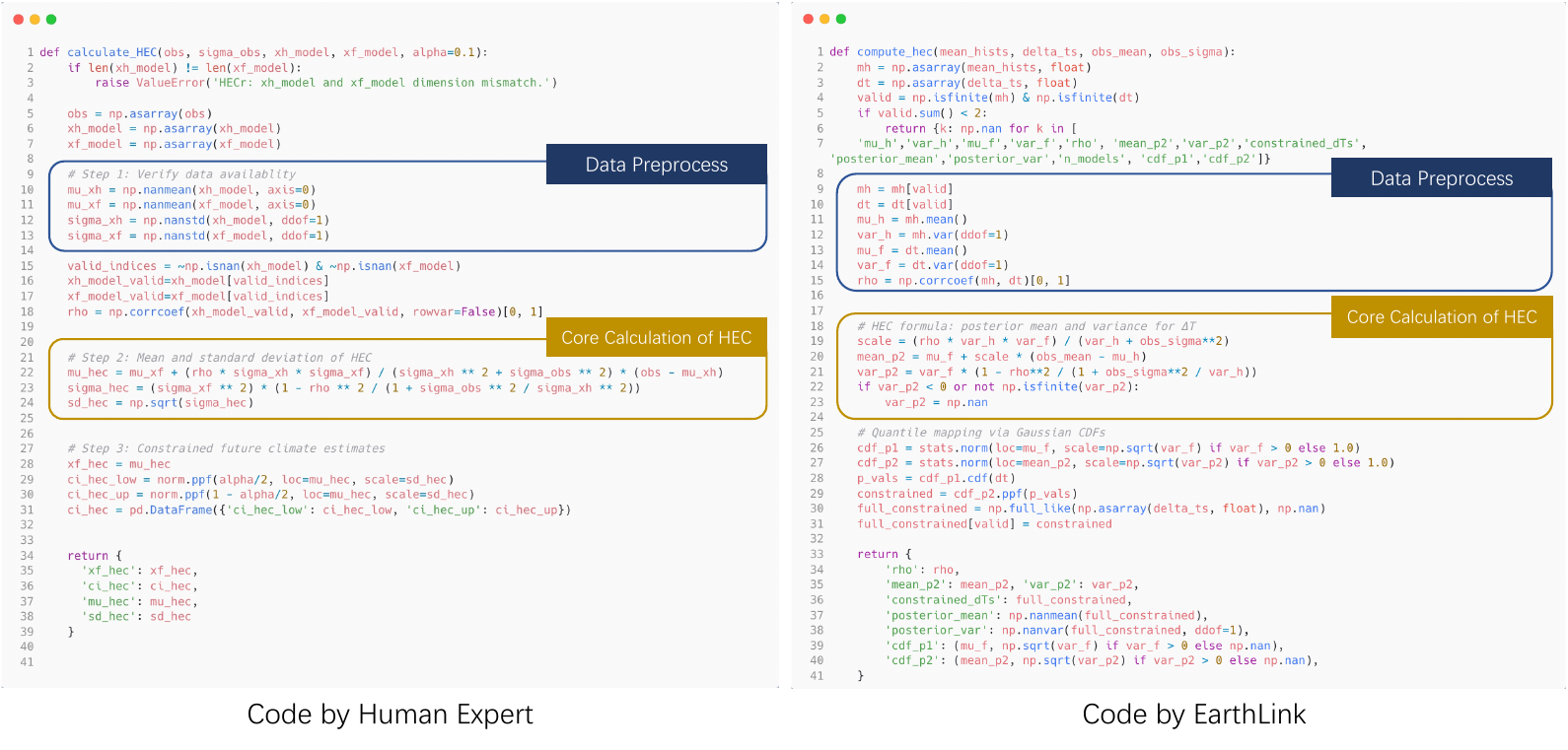}
    \caption{
    \textbf{Comparison of code generated by human expert and \model.}}

    \label{fig:code}
\end{figure}
\clearpage

~
\begin{table}[!t]
\centering
\caption{\textbf{The task complexity levels and representative capabilities of the \model~in the evaluation framework.}}
\label{tab:tab1}
    \renewcommand\arraystretch{1.5}
  \begin{tabularx}{\textwidth}{
  >{\centering\arraybackslash}m{0.2\textwidth}
  >{\centering\arraybackslash}m{0.62\textwidth}
  >{\centering\arraybackslash}m{0.1\textwidth}
  }
\toprule
\textbf{Levels} &
  \textbf{Description} &
  \textbf{Number of Tasks} \\ \midrule
  \rowcolor[HTML]{e9e9e9}
Level 1: Simple statistical analysis &
  Performs basic climatological tasks, including data retrieval, preprocessing, calculation of annual means, spatial distributions, and interannual variability, with visualizations supporting initial model evaluation. &
  23 \\
   
Level 2: Mechanistic diagnosis &
  Solve moderately complex climate problems, such as estimating Equilibrium Climate
Sensitivity (ECS) and Transient Climate
Response (TCR), by understanding the physical diagnostic framework, invoking common analyses of multiple experiment datasets and applying simple mathematical tools. &
  6 \\
\rowcolor[HTML]{e9e9e9}
Level 3: Complex scientific reasoning &
  Decomposes complex climate analyses into clear, logical subtasks. Integrates advanced analytical methods (e.g., Empirical Orthogonal Function (EOF), composite analysis) with specialized knowledge to study complex phenomena such as El Ni\~no-Southern Oscillation (ENSO) diversity, requiring rigorous methodology and extended reasoning chain. &
  6 \\
  
Level 4: Semi-open scientific problem &
  Automatically selects appropriate datasets based on detailed human problem descriptions, combining physical understanding with adaptive workflows to address open-ended climate problems. Applies constraint methods (e.g., emergent constraints) to identify the constraint factor and produce constrained forecasts and preliminary decision-oriented recommendations. &
  1 \\ \bottomrule
\end{tabularx}%
\end{table}
\clearpage

~
\begin{table}[!t]
\centering
\caption{\textbf{The evaluation L1 Tasks for \model~in our study and their corresponding results are detailed in Supplementary Information Section C.}}
\label{tab:tab2}
\renewcommand\arraystretch{1.4}
  \begin{tabularx}{\textwidth}{
  >{\centering\arraybackslash}m{0.1\textwidth}
  >{\centering\arraybackslash}m{0.15\textwidth}
  >{\centering\arraybackslash}m{0.36\textwidth}
  >{\centering\arraybackslash}m{0.28\textwidth}
  }
  \toprule
\textbf{Task ID} &
  \textbf{Task Level} &
  \textbf{Description} &
  \textbf{Variable} \\
  \midrule
\rowcolor[HTML]{e9e9e9}
Task 1 & L1 & \cellcolor[HTML]{e9e9e9} & Cloud radiative effect \\
\rowcolor[HTML]{e9e9e9}
Task 2  & L1 & \cellcolor[HTML]{e9e9e9} & Evaporation             \\
\rowcolor[HTML]{e9e9e9}
Task 3  & L1 & \cellcolor[HTML]{e9e9e9} & Precipitation           \\
\rowcolor[HTML]{e9e9e9}
Task 4  & L1 & \cellcolor[HTML]{e9e9e9} & Sea ice concentration   \\
\rowcolor[HTML]{e9e9e9}
Task 5  & L1 & \cellcolor[HTML]{e9e9e9} & Surface air temperature \\
\rowcolor[HTML]{e9e9e9}
Task 6  & L1 & \cellcolor[HTML]{e9e9e9} & Surface full wind       \\
\rowcolor[HTML]{e9e9e9}
Task 7  & L1 & \multirow{-7}{*}{\cellcolor[HTML]{e9e9e9}\makecell[c]{Comparison of global multi-model \\ simulated climatology and variability \\ for 1985–2014 with observations.}} & Thermocline             \\

Task 8 & L1 & & Evaporation \\
Task 9  & L1 &  & Precipitation           \\
Task 10  & L1 &  & Runoff                  \\
Task 11  & L1 &  & Sea ice concentration   \\
Task 12 & L1 & 
  \multirow{-5}{*}{\makecell[c]{Evaluation of global multi-model \\ simulations of annual mean and \\ seasonal cycles.}}  & Surface air temperature \\

\rowcolor[HTML]{e9e9e9}
Task 13 & L1 & \cellcolor[HTML]{e9e9e9} & Ocean heat content \\
\rowcolor[HTML]{e9e9e9}
Task 14 & L1 & \cellcolor[HTML]{e9e9e9} & Precipitation           \\
\rowcolor[HTML]{e9e9e9}
Task 15 & L1 & \cellcolor[HTML]{e9e9e9} & Runoff                  \\
\rowcolor[HTML]{e9e9e9}
Task 16 & L1 & \cellcolor[HTML]{e9e9e9} & Sea ice concentration   \\
\rowcolor[HTML]{e9e9e9}
Task 17 & L1 & \cellcolor[HTML]{e9e9e9} & Surface air temperature \\
\rowcolor[HTML]{e9e9e9}
Task 18 & L1 & \multirow{-6}{*}{\cellcolor[HTML]{e9e9e9}\makecell[c]{Comparison of global multi-model \\ simulated changes and trends from \\ 1991–2010 to 1970–1990 with \\ observations.}}  & Surface full wind       \\

Task 19 & L1 & &
  Antarctica surface albedo \\
Task 20 & L1 &  & Runoff \\
Task 21 & L1 & \multirow{-3}{*}{\makecell[c]{Comparison of CMIP simulations \\ in specific regions with \\ visualizations for selected areas.}} & Sea surface temperature   \\

\rowcolor[HTML]{e9e9e9}
Task 22 & L1 & \cellcolor[HTML]{e9e9e9} & Cloud radiative effect \\
\rowcolor[HTML]{e9e9e9}
Task 23 & L1 & \multirow{-2}{*}{\cellcolor[HTML]{e9e9e9}\makecell[c]{Comparison of differences between \\ various observations.}} & Precipitation \\
\bottomrule
\end{tabularx}%
\end{table}
\clearpage

~
\begin{table}[!t]
\centering
\caption{\textbf{The evaluation tasks for \model~in our study from L2 to L4 and their corresponding results are detailed in Supplementary Information Section C.}}
\label{tab:tab3}
\renewcommand\arraystretch{1.4}
  \begin{tabularx}{\textwidth}{
  >{\centering\arraybackslash}m{0.1\textwidth}
  >{\centering\arraybackslash}m{0.15\textwidth}
  >{\centering\arraybackslash}m{0.36\textwidth}
  >{\centering\arraybackslash}m{0.28\textwidth}
  }
  \toprule
\textbf{Task ID} & \textbf{Task Level} & \textbf{Description}                          
& \textbf{Variable}        \\
\midrule
\rowcolor[HTML]{E9E9E9} 
Task 24          & L2                  & Evaluate Equilibrium Climate Sensitivity (ECS) in different models using methods of self-choice. & ECS \\

Task 25 & L2 &  Evaluation of ECS across different models using given methods, such as Gregory regression \cite{gregory2004new}. &  ECS \\

\rowcolor[HTML]{E9E9E9} 
Task 26 & L2 & Evaluate Transient Climate Response (TCR) in different models    & TCR \\

Task 27          & L2 &  & Precipitation            \\

Task 28 & L2 &
  \multirow{-2}{*}{\makecell[c]{Comparison of climate changes \\ under different future scenarios.}} &
  Surface air temperature \\
  
\rowcolor[HTML]{E9E9E9} 
Task 29          & L2                  & Detection of global climate change using DAMIP Experiments.                                  & Surface air temperature  \\

Task 30          & L3                  & Evaluation of Atlantic Meridional Overturning Circulation (AMOC) simulation capability in CMIP6 models.                                     & AMOC                     \\

\rowcolor[HTML]{E9E9E9} 
Task 31          & L3                  & Evaluation of ENSO diversity simulation in CMIP6 using ENSO classification method in \cite{kao2009contrasting}. & ENSO diversity \\

Task 32 & L3 &
  Evaluation of ENSO diversity simulation in CMIP6 using ENSO classification method in \cite{kug2009two}. &  ENSO diversity \\
  
\rowcolor[HTML]{E9E9E9} 
Task 33 & L3 & Evaluation of ENSO period simulated in CMIP6 using wavelet analysis. & ENSO periods \\

Task 34 & L3 &
  Quantification of contributions from different forcing factors to global mean temperature warming from 1901 to 2015. &
  Climate change attribution \\
  
\rowcolor[HTML]{E9E9E9}
Task 35          & L3                  & Heat budget analysis. & Heat budge               \\

Task 36 & L4 &
  Using emergent constraints to constrain temperature trends in Africa over the next 20 years. &
  Future projection constraint \\
  
  \bottomrule
\end{tabularx}%
\end{table}

~
\begin{table}[!t]
\centering
\caption{\textbf{Comprehensive Evaluation Reference Table (Full Score 5 Points).}}
\vspace{-0.2cm}
\label{tab:tab4}
\renewcommand\arraystretch{1.4}
  \begin{tabularx}{\textwidth}{
  >{\centering\arraybackslash}m{0.1\textwidth}
  >{\centering\arraybackslash}m{0.26\textwidth}
  >{\centering\arraybackslash}m{0.26\textwidth}
  >{\centering\arraybackslash}m{0.27\textwidth}
  }
\toprule
\textbf{Score} &
  \textbf{Experimental Planning and Method Design} &
  \textbf{Code Implementation} &
  \textbf{Result Synthesis and Visualization} \\ \midrule
  \rowcolor[HTML]{E9E9E9}
5 points &
  Complete planning, scientifically rigorous, clear logic, with practical feasibility; all steps closely tied to objectives, no redundancy or errors; accurate method description; comprehensive output of results. &
  Able to accurately call data and tools, results fully conform to planning and experimental design; high code quality, no debugging or only minimal debugging is needed for successful execution. &
  High consistency between text and figures, visually pleasing and standardized charts, rich textual information and clear logic; correct axis, units, scope, and annotation in charts; overall presentation meets publication or reporting standards, effectively supporting research conclusions. \\
  
4 points &
  Overall structure is reasonable, task decomposition is clear, but some non-core redundant calculations, inefficient processes, or steps weakly related to objectives exist, without affecting overall results. &
  Able to complete all tasks, but requires relatively more debugging (e.g., debug rounds \textgreater 15), final results are correct and do not affect execution effect; non-fatal errors exist but can be corrected. &
  Text and figures are consistent, content is correct, but minor issues exist (e.g., slight color mismatches, inconsistent fonts, missing individual labels), which do not affect understanding; text expression is relatively complete but can be further optimized; overall quality is good, needing only slight polishing to reach an excellent level. \\
  
\rowcolor[HTML]{E9E9E9}
3 points &
  Able to design correct tools or implement correct experimental steps, complete main tasks, but lacks detailed explanation or sufficient interpretation of selection rationale, possibly affecting result rigor or reproducibility. &
  Able to generate intermediate data, and intermediate data basically meet expectations; core code tools also basically meet expectations, but only partial task goals are completed, or unable to fully implement the entire process. &
  Text and figures are basically consistent, but image aesthetics and textual information amount are low; obvious errors appear in images including axis labels, drawing scope, units, etc. \\
  
2 points &
  Able to find data needed for corresponding tasks, understand basic data format and rules, but cannot form effective analysis process or experimental pathway. &
  Able to load data and generate some tools or intermediate variables, but output results obviously do not meet requirements or contain logical errors, unable to support further analysis. &
  Able to provide images and reports, but reports contain obvious problems of inconsistency between text and figures. \\

\rowcolor[HTML]{E9E9E9}
1 point and below &
  Planning is incomplete or contains major errors, difficult to guide actual execution; task goals are misunderstood; data or methods seriously do not meet requirements. &
  Unable to load data, unable to generate required tools or functions; code has major errors, unable to run or continue the task. &
  Unable to provide images or comprehensive summary reports, unable to provide images meeting requirements. \\
  \bottomrule
\end{tabularx}%
\end{table}

\end{bibunit}

\clearpage

\input{sn-article-sm}



\end{document}

%% file: sn-article-sm.tex
\begin{bibunit}[sn-mathphys-num]

\lstset{
  inputencoding=utf8,
  extendedchars=true,
  basicstyle=\ttfamily\small,
  lineskip=-0.2em,
  aboveskip=-0.2em,
  belowskip=-0.2em,
  keywordstyle=\color{blue},
  commentstyle=\color{gray},
  stringstyle=\color{orange},
  breaklines=true,
  showstringspaces=false,
  escapeinside={(*@}{@*)},
  numbers=left,
  numberstyle=\bfseries\tiny\color{gray!70},
  stepnumber=1,
  numbersep=6pt,
  xleftmargin=5pt,
  xrightmargin=-10pt,
  literate=
    {“}{"}1
    {”}{"}1
    {‘}{'}1
    {’}{'}1
    {′}{'}1
    {–}{-}1
    {—}{-}1
    {−}{\textminus}1
    {–}{\textminus}1
    {‑}{\textminus}1
    {×}{{$\times$}}1
    {·}{{$\cdot$}}1
    {±}{{$\pm$}}1
    {≥}{{$\geq$}}1
    {≤}{{$\leq$}}1
    {≲}{{$\lesssim$}}1
    {≈}{{$\approx$}}1
    {…}{\ldots}1
    {ñ}{{\~n}}1
    {°}{{$\degree$}}1
    {Δ}{{$\Delta$}}1
    {σ}{{$\sigma$}}1
    {ρ}{{$\rho$}}1
    {Ω}{{$\Omega$}}1
    {ω}{{$\omega$}}1
    {φ}{{$\phi$}}1
    {θ}{{$\theta$}}1
    {α}{{$\alpha$}}1
    {β}{{$\beta$}}1
    {μ}{{$\mu$}}1
    {τ}{{$\tau$}}1
    {λ}{{$\lambda$}}1
    {ε}{{$\epsilon$}}1
    {Γ}{{$\Gamma$}}1
    {∪}{{$\cup$}}1
    {~}{{\raisebox{0.5ex}{\large\texttildelow}}}1
    {→}{{$\rightarrow$}}1
    {↔}{{$\leftrightarrow$}}1
    {ö}{{\"{o}}}1
    {⁻}{{$^{-}$}}1
    {¹}{{$^{1}$}}1
    {²}{{$^{2}$}}1
    {∈}{{$\in$}}1
    {Σ}{{$\Sigma$}}1
    {≡}{{$\equiv$}}1
    {β̂ }{{$\hat\beta$}}1
    {ᵀ}{{$\rm^T$}}1
    {√}{{$\surd$}}1
    {∇}{{$\nabla$}}1
    {∫}{{$\int$}}1
    {∂}{{$\partial$}}1
}

\newcounter{sharedbox}
\numberwithin{sharedbox}{subsection}

\newtcolorbox[use counter=sharedbox]{requestbox}[1][]{
  enhanced jigsaw,
  breakable,
  pad at break*=1mm,
  colback=white,
  fonttitle=\bfseries,
  fontupper=\sffamily,
  title=Box~\thesubsection.\thesharedbox~User Request
}

\newtcolorbox[use counter=sharedbox]{planbox}[1][]{
  enhanced jigsaw,
  breakable,
  pad at break*=1mm,
  colframe={rgb,255:red,52;green,116;blue,156},
  colback=white,
  fonttitle=\bfseries,
  fontupper=\sffamily,
  title=Box~\thesubsection.\thesharedbox~Final Experimental Plan
}

\newtcolorbox[use counter=sharedbox]{visualbox}[1][]{
  enhanced jigsaw,
  breakable,
  pad at break*=1mm,
  colframe={rgb,255:red,52;green,116;blue,156},
  top=0.5mm,
  bottom=0.5mm,
  left=0.5mm,
  right=0.5mm,
  colback=white,
  fonttitle=\bfseries,
  fontupper=\sffamily,
  title=Box~\thesubsection.\thesharedbox~Main Visualization Results (Raw Output)
}

\newtcolorbox[use counter=sharedbox]{summarybox}[1][]{
  enhanced jigsaw,
  breakable,
  pad at break*=1mm,
  colframe={rgb,255:red,52;green,116;blue,156},
  colback=white,
  fonttitle=\bfseries,
  fontupper=\sffamily,
  title=Box~\thesubsection.\thesharedbox~Analysis Summary
}

\clearpage
\setcounter{page}{1}

\nolinenumbers

\begin{center}
{\fontsize{16pt}{18pt}\selectfont \textbf{Supplementary Information}\\[4mm]}

\makeatletter
{\fontsize{15pt}{17pt}\selectfont \textbf{\@title}}
\makeatother

\end{center}

\renewcommand{\contentsname}{Table of contents}

\section*{\fontsize{13pt}{15pt}\selectfont \contentsname}
\vspace{-1em}

\startcontents[appendix]
\printcontents[appendix]{l}{1}{\setcounter{tocdepth}{3}}

\clearpage

\renewcommand{\refname}{Supplementary Information References} 

\begin{appendices}

\section{Additional details of methods}


\subsection{Construction and interfaces of resource library}

\subsubsection{Knowledge Library}

The Knowledge Library is a dual-component resource hub designed to support both the Planning and Scientific Diagnosis Modules by integrating offline and online information sources. The offline component consists of a curated database containing verified plans from previous executions and comprehensive package documentation. For the plans, we construct a repository of (user request, plan) pairs. Each user request is encoded into a high-dimensional vector using an embedding model (Qwen3-Embedding~\cite{zhang2025qwen3}) to create a vector database. When a new request is received, it is similarly vectorized and used to query this database. Several requests with the highest cosine similarity score are retrieved, and their corresponding plans are presented to the Planning Agent as references.

For API documentation, we split the documents into distinct text chunks, each of which is vectorized to build a dedicated vector database for that document. This database is exposed through a query interface that accepts a natural language query and returns the three most similar text chunks. This interface can be directly invoked by the Coding Agent to retrieve precise information on function usage and workflows.

The online component of the library is a web search tool, which can be directly called by the agents to perform real-time searches for the latest literature, repositories, and API documentation, ensuring that the system's knowledge remains current and expansive.

\subsubsection{Data Library}

The Data Library integrates vast quantities of climate model outputs and observational datasets, providing a robust data foundation for automated scientific diagnosis. The total data volume in this library exceeds 3 PB, encompassing a wide range of resources. It includes key simulation data from the CMIP6, covering major activities such as CMIP, DAMIP~\cite{gillett2016detection}, GeoMIP~\cite{kravitz2011geoengineering}, and ScenarioMIP~\cite{o2016scenario}, with data from over 70 climate models. Furthermore, it incorporates numerous authoritative observational datasets like obs4MIPs~\cite{waliser2020observations}, HadISST~\cite{rayner2003global}, HadCRUT5~\cite{morice2021updated}, GPCP-SG~\cite{adler2003version}, and ERA5~\cite{hersbach2019era5}, spanning the atmosphere, ocean, land, and cryosphere. The observational data sources in the current version are listed in Table~\ref{tab:full_data}, and the data library is constantly expanding. 

\begin{table*}[ht]
    \belowrulesep=0pt
    \aboverulesep=0pt
    \caption{\textbf{The source of observational data used in \model.} It should be noted that only a subset of some datasets has been downloaded in the current version, and the data library is constantly expanding.}
    \centering
\resizebox{\textwidth}{!}{%
    \begin{tabular}{ll}
    \toprule
Dataset & Source \\ \midrule
CERES-EBAF~\cite{loeb2018clouds}  & \url{https://ceres.larc.nasa.gov/data} \\
COBE-SST2~\cite{hirahara2014centennial} & \url{https://psl.noaa.gov/data/gridded/data.cobe2.html} \\
Duveiller2018~\cite{duveiller2018dataset} & \url{https://ndownloader.figshare.com/files/9969496} \\
ERA5~\cite{hersbach2019era5} & \url{https://cds.climate.copernicus.eu/datasets/reanalysis-era5-single-levels-monthly-means} \\
ESACCI-CLOUD~\cite{stengel2017cloud} & \url{https://public.satproj.klima.dwd.de/data/ESA_Cloud_CCI/CLD_PRODUCTS/v3.0} \\
ESACCI-SOILMOISTURE~\cite{dorigo2017esa} & \url{https://data.ceda.ac.uk/neodc/esacci/soil_moisture/data} \\
GCP2018~\cite{le2018global} & \url{https://www.icos-cp.eu/GCP/2018} \\
GCP2020~\cite{friedlingstein2020global} & \url{https://www.icos-cp.eu/GCP/2020} \\
GODAS~\cite{behringer1998improved} & \url{https://psl.noaa.gov/data/gridded/data.godas.html} \\
GPCP-SG~\cite{adler2003version} & \url{https://psl.noaa.gov/data/gridded/data.gpcp.html} \\
HadCRUT4~\cite{morice2012quantifying} & \url{https://crudata.uea.ac.uk/cru/data/temperature} \\
HadCRUT5~\cite{morice2021updated} & \url{https://crudata.uea.ac.uk/cru/data/temperature} \\
HadISST~\cite{rayner2003global} & \url{https://www.metoffice.gov.uk/hadobs/hadisst/} \\
IAP~\cite{cheng2024iapv4} & \url{http://www.ocean.iap.ac.cn/pages/dataService/dataService.html} \\
ISCCP-FH~\cite{zhang2023calculation} & \url{https://isccp.giss.nasa.gov/projects/flux} \\
NCEP-NCAR-R1~\cite{kalnay2018ncep} & \url{https://psl.noaa.gov/data/gridded/data.ncep.reanalysis.html} \\
NOAA-ERSSTv5~\cite{huang2017extended} & \url{https://www1.ncdc.noaa.gov/pub/data/cmb/ersst/v5/netcdf} \\
obs4MIPs~\cite{waliser2020observations} & \url{https://pcmdi.github.io/obs4MIPs} \\
ORAS5~\cite{copernicus2021oras5} & \url{https://cds.climate.copernicus.eu/datasets/reanalysis-oras5} \\

\bottomrule
\end{tabular}
}
    \label{tab:full_data}
\end{table*}

To manage this extensive collection, we have established a detailed metadata index for every dataset, systematically documenting critical information such as the included variables, available ensemble members, and temporal coverage. Built upon this structured metadata, we provide a suite of query interfaces (APIs). These interfaces enable agents to programmatically retrieve information, such as querying for the availability of a specific variable, listing the available ensemble members, or verifying the time period of a dataset. This functionality is crucial for supporting higher-level modules: during the planning stage, it allows agents to confirm data availability and formulate executable plans, while during the diagnosis stage, it provides the necessary parameters for data loaders to ensure accurate and efficient data access and processing.

\subsubsection{Tool Library}

The Tool Library is designed to provide robust and flexible support for the Coding Agent and is composed of two primary components: standardized software packages and a library of expert-validated code scripts.

The first component includes widely-used, domain-specific diagnostic tools and general-purpose Python scientific packages, such as ESMValTool~\cite{righi2020earth}, PCMDI Metrics~\cite{lee2024systematic}, xarray, cartopy, and scikit-learn. The Coding Agent can directly import and utilize these tools within its standard Python environment to perform a wide range of data processing, analysis, and visualization tasks.

The second component is a curated repository of modular, expert-vetted code scripts. These scripts encapsulate complete or partial workflows tailored for specific scientific analysis tasks, including custom data loading logic, complex physical computation algorithms, and specialized plotting functions. To enable intelligent retrieval of these scripts, we implement a semantic similarity-based mechanism. The process begins by using an LLM to generate a precise functional description for each script, creating a collection of (functional description, code script) pairs. Each description is then encoded into a high-dimensional vector using an embedding model to construct a vector database optimized for code retrieval, similar to that in the Knowledge Library. When a new diagnostic task is initiated, an LLM first interprets the requirements and generates a natural language description of the needed scientific computation or visualization. This description is subsequently vectorized and matched against the database to identify the most similar pairs. The code scripts from these top-matching pairs are then provided to the Coding Agent as high-quality reference examples and reusable templates, significantly enhancing the accuracy and efficiency of its code generation process.

\subsection{Literature parsing}
User-uploaded literature is first parsed and converted into a structured knowledge base that can serve as a contextual reference for subsequent scientific reasoning and experiment planning. Input documents are processed using the advanced scientific document extraction framework MinerU \cite{wang2024mineru}, which separates textual content, tables, and figures. Tabular data are converted into Markdown-formatted text to preserve structure and semantics.

For textual content, large language models (LLMs) are employed to extract metadata, including abstracts, keywords, and relevant domains, which are indexed for filtering and targeted retrieval. The full text is then segmented into semantically coherent chunks, each summarized into a concise caption by the LLM. These summaries are embedded as dense vector representations and stored as retrieval keys to support high-precision semantic search.

For extracted figures, both image data and their original captions are processed using multimodal large language models (MLLMs) to generate academic-style image summaries that highlight key visual and scientific elements. The resulting image embeddings are indexed alongside text embeddings within the same retrieval space, enabling cross-modal alignment between textual and visual evidence.

Once the knowledge base is constructed, it is used to enrich subsequent user requests. Each user query is first reformulated by the LLM to enhance technical precision and reduce ambiguity. The rewritten query is then embedded and semantically matched against the knowledge base to retrieve relevant evidence. Retrieved documents and captions are fused with the query through an MLLM-based reasoning process, producing an enriched, context-aware query representation. This enriched query is passed to the Planning Agent, which constructs executable experiment plans. To minimize information loss, the retrieved evidence is also supplied directly to the planning agent as supplementary input for reasoning.

\subsection{System configuration}

\subsubsection{Model configuration}

To ensure the highest overall performance of the framework, we have selected GPT-5 as the foundational large language model for all modules. Recognizing that the various agents within our system have distinct operational requirements, we have assigned unique parameter configurations to the LLM for each module. This tailored approach is designed to optimize the performance of each component for its specific task by carefully balancing output diversity, reasoning depth, and computational efficiency.

The temperature parameter, which controls the randomness and creativity of the model's output, is a key part of this customization. For the Coding Agent, whose primary function is to generate precise and reproducible code, the temperature is set to a low value of 0.3. This reduces output randomness, ensuring that the generated code is more stable and deterministic. Conversely, for the Planning Agent and other components where exploration of different possibilities is beneficial, we use the default value of 1.0. A higher temperature encourages the model to generate a wider and more creative array of plans, providing a richer set of options for the subsequent aggregation stage.

Similarly, we adjust the reasoning effort parameter to modulate the depth and rigor of the model's cognitive process. For the Planning Agent, this is set to low to facilitate the rapid generation of multiple candidate plans without overly deep initial deliberation. In contrast, the Plan Aggregation Agent, which is responsible for finalizing a single high-quality plan, uses the default medium setting to ensure a high degree of accuracy, logical coherence, and scientific validity. Furthermore, for the Result Checking and Image Feedback Agents, the reasoning effort is also set to low. This strategic choice is intended to prevent the model from overthinking and potentially hallucinating when analyzing multiple outputs, while also serving to reduce processing latency.

All agents not explicitly mentioned in this section operate using the model's default parameter configurations. Through this differentiated strategy, we effectively balance the competing demands for creativity, speed, and accuracy across the various specialized components of the framework.

\subsubsection{Running configuration}

To ensure robust, efficient, and stable operation of the framework, we have established a set of system-wide running configurations. During the initial planning phase, the Planning Agent is configured to generate five candidate plans in parallel, providing a diverse set of initial strategies. For all retrieval operations, whether fetching similar plans and documentation from the Knowledge Library or relevant diagnostic scripts from the Tool Library, we employ a cosine similarity metric. A retrieval is considered valid only if the similarity score exceeds a threshold of 0.5, and the system is configured to use only the top three most relevant results from any given query to maintain focus and efficiency.

To maintain control over the iterative refinement process and prevent infinite loops, we have implemented maximum iteration limits. The Result Checking loop is capped at ten rounds, while the more targeted Image Feedback loop is limited to five rounds. This principle of bounded iteration extends to the code debugging cycle. A single execution attempt is permitted a maximum of 30 debug rounds. If a solution is not found within this limit, the attempt is terminated, and a second attempt is initiated from scratch. The entire task is marked as failed only if both attempts are unsuccessful.

Beyond iterative controls, we have implemented resource management constraints to ensure system stability and timely execution. A maximum execution time of one hour is enforced for any single script. If this time limit is exceeded, the task is passed back to the Coding Agent with explicit instructions to optimize the underlying algorithm. To manage the context length of the LLM and maintain focus on the most recent issues, the debugging history is truncated to include only the error messages from the latest five rounds; older error information is discarded. Finally, to regulate the volume of generated outputs and ensure it remains within the processing capabilities of the LLM, we set a limit of 20 images per diagnostic task. If the number exceeds this limit, the Coding Agent is instructed to consolidate multiple plots into a single figure using subplots, thereby preventing analytical complexity and potential model overload.

\subsection{Emergent constraint methods}

To investigate future surface-air temperature changes across major global cities, we implement three methodological frameworks following the emergent-constraint paradigm. Projections are derived from multi-model ensembles, evaluated for both historical (1971–2020) and future (2041–2060) periods. Observational constraints are based on the spatially infilled HadCRUT5 global mean temperature record spanning 1971–2022. For each city, we quantify the distribution of future warming using: (1) RAW model projections (RAW), (2) spatially aggregated model outputs (Aggregation), and (3) a Hierarchical Emergent Constraint (HEC) approach that conditions projections on observed warming trends as shown in Fig.~\ref{fig:fig3}b. These methods progressively isolate and reduce uncertainty arising from internal climate variability and model structural differences.

\noindent\textbf{RAW.} The RAW method employs model output directly at the nearest grid point corresponding to each target urban location. For each climate model, local surface temperature anomalies are extracted, and linear warming trends are computed over the historical reference interval (1971–2020). Future warming trends for 2041–2060 are then estimated from the same local grid cell without any prior filtering or statistical constraint. The distribution of projected warming is characterized across models by computing the ensemble mean and variance. This RAW projection represents the baseline spread of model-simulated warming at native spatial resolution, preserving local variability yet retaining substantial uncertainty due to internal climate fluctuations and model divergence.

\noindent\textbf{Spatial Aggregation method.} To reduce noise arising from high-frequency climate variability, we adopt a spatial aggregation procedure consistent with previous emergent-constraint research. Each city is represented by a 3°$\times$3° latitude–longitude grid cell patch centered on the target coordinates, rather than a single native model grid cell. For each model, historical and future trends are computed from the patch-averaged temperature anomalies. Ensemble means and variances of projected warming are again calculated. This method attenuates stochastic variability and improves the signal-to-noise ratio while preserving regional climatic characteristics. Compared with RAW projections, this step yields a narrower distribution of future warming, reflecting improved robustness in the regional mean climate response.

\noindent\textbf{Hierarchical emergent constraint (HEC).} HEC imposes a probabilistic constraint linking historical warming performance to future warming projections via a hierarchical emergent relationship~\cite{li2024constraining}. Following the data-aggregation procedure, each model provides estimates of historical and future warming, along with associated variances. The observed historical warming trend from HadCRUT5 (0.20 °C per decade ± 0.03 °C) is then used to update model-simulated future warming through a Bayesian-style regression framework. The inter-model correlation between simulated historical and projected future warming is quantified, and the posterior constrained mean and variance are derived for each model. We further apply quantile mapping to adjust each model’s raw future projection to the inferred constrained distribution, thereby retaining model ranking while enforcing the observationally-conditioned probabilistic structure. The posterior means and variances form the constrained distribution, representing reduced-uncertainty warming projections conditioned on real-world climate evolution.





\section{Additional analyses} 
\subsection{Climate impact estimation across scenarios}

Benefiting from the powerful foundation model, \model~demonstrates the potential to transform quantitative climate data into policy-relevant discourse.
Despite the complexity and unpredictability of climate change's impact on policy, \model~employs a divide-and-conquer methodology to simplify the task by decomposing policy demands into domain-specific sub-problems, which are then addressed by specialized domain expert agents.

\begin{figure}[!ht]
    \centering
    \includegraphics[width=0.8\linewidth]{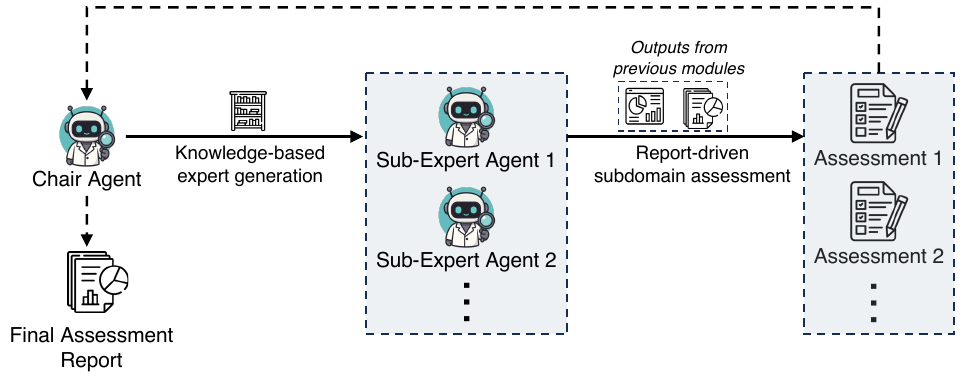}
    \caption{\textbf{The workflow of climate impact estimation.} 
    The Chair Agent retrieves relevant information from the knowledge base, synthesizes it with domain expertise, and dynamically generates domain-specific Sub-Expert Agents. Each Sub-Expert Agent performs report-driven subdomain assessments based on outputs from previous modules, producing both textual and quantitative analyses. Finally, the Chair Agent integrates all subdomain reports into a comprehensive Final Assessment Report, summarizing key findings, disagreements, and uncertainties to inform policy-relevant climate analysis.
    }
    \label{fig:climate_impact_module}
\end{figure}

As shown in Fig.~\ref{fig:climate_impact_module}, the resulting dynamic multi-agent system operates akin to a committee, where a ``chair'' agent first retrieves relevant information from the knowledge base, synthesizes it with domain expertise, determines which domain-specific agents are required, and designs appropriate prompts for each sub-expert agent.
For instance, in addressing the topic ``Agricultural Development in the Moscow Region from 2041 to 2060,'' agents such as an ``Environmental Scientist'' and an ``Agricultural Scientist'' would be deemed necessary.
In our implementation, the chair agent generates $10$ sub-expert agents for each topic.
These dynamically generated sub-expert agents take figures from the Scientific Diagnostic Module and reports from the Analysis \& Summary Agent as input, analyze the impacts of climate change within their respective domains based on retrieved domain knowledge, and produce both textual and quantitative assessments, where the former offers interpretive analysis and contextual explanation, and the latter provides numerical evaluations categorized as positive or negative.

\begin{figure}[!ht]
    \centering
    \includegraphics[width=0.75\linewidth]{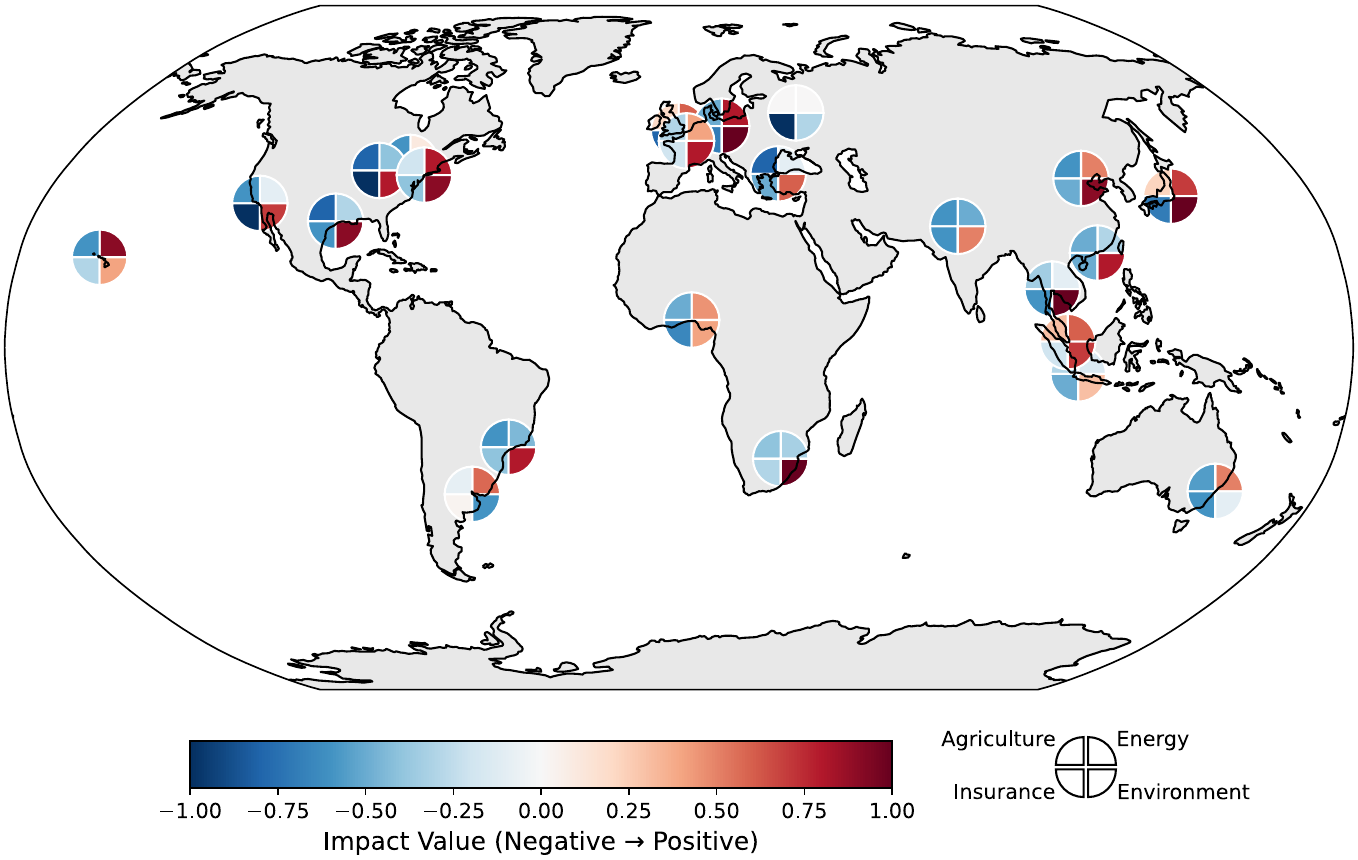}
    \caption{
Impacts of climate change over the next 20 years. The system connects quantified climate projections to sectoral impacts, providing preliminary risk assessments and qualitative narratives for agriculture, energy, insurance and environment, illustrating its ability to bridge scientific data with policy-relevant discussion.
    }
    \label{fig:climate_impact}
\end{figure}

Finally, the chair agent’s primary role is to synthesize the reports from all sub-expert agents into a structured final assessment. This output summarizes key findings and highlights areas of consensus and disagreement (for example, an agronomist might view a longer growing season as positive, while an environmental scientist may warn of increased water stress), and also identifies key uncertainties. For a high-level, heuristic overview, the system can also compute a coarse sentiment score (Fig.~\ref{fig:climate_impact}), ranging from -1 (predominantly negative assessments) to 1 (predominantly positive). However, the core scientific output remains the detailed report, which provides the rich, multi-faceted analytical landscape needed to inform nuanced policy and decision-making.

\subsection{Effect of plan aggregation}

The Planning Module is a core component responsible for generating executable experimental plans based on user requests. The initial Planning Agent generates multiple candidate plans (plan 1, ..., plan n) to explore a diverse solution space. However, these initial plans often exhibit significant variations in their approaches and quality; some may contain suboptimal methodologies, while others might overlook critical data processing steps.

To address this, the Plan Aggregation Agent is designed to critically evaluate and synthesize these candidate plans. It identifies the most effective and scientifically sound components from each candidate, discarding weaker elements. This process, often guided by human supervision, aims to construct a single, coherent Final plan that is superior to any of the individual proposals.

\begin{figure}[!ht]
    \centering
    \includegraphics[width=1.0\linewidth]{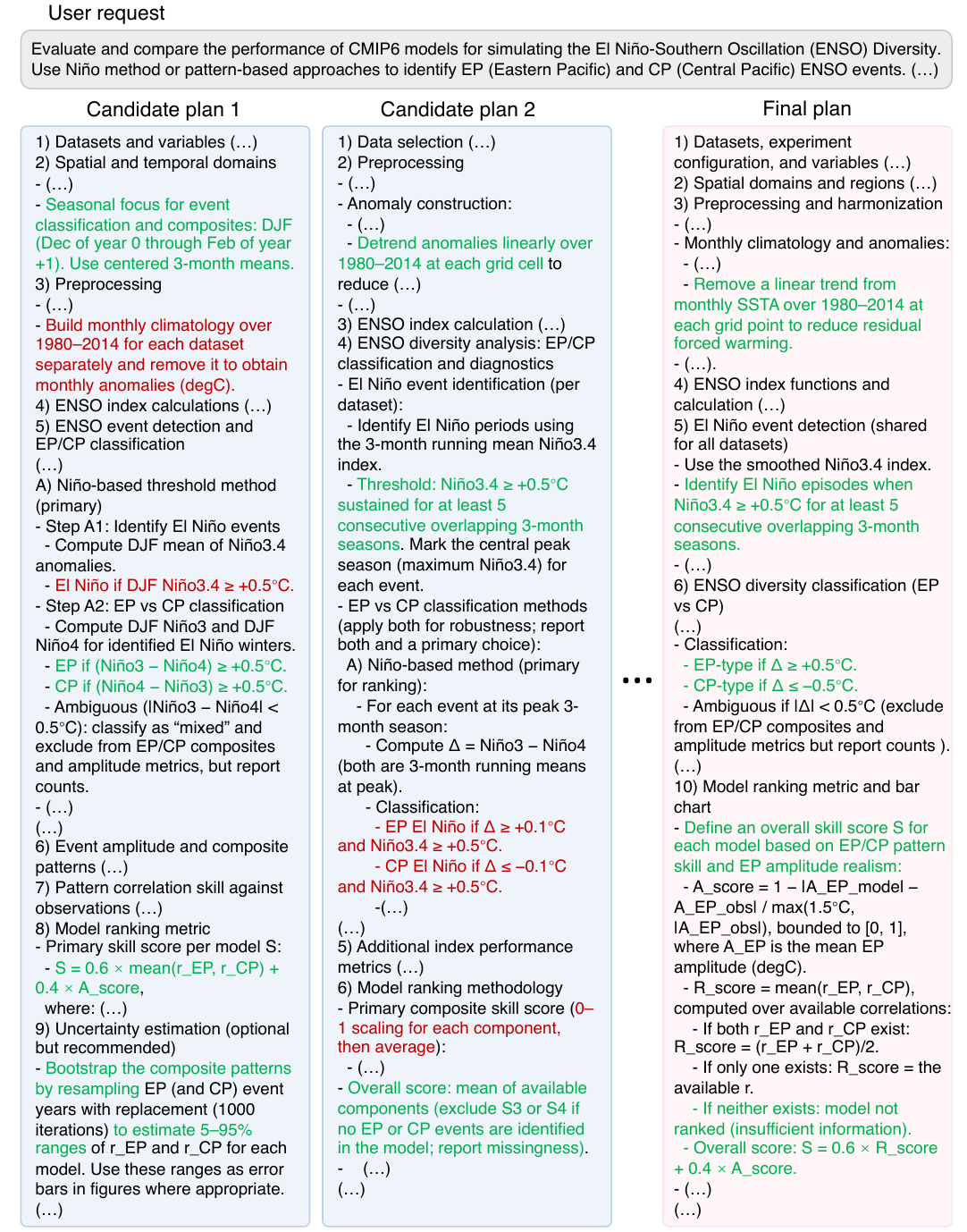}
    \caption{
An example of plan diversity generated by the Planning Module, showing two candidate plans and the final aggregated plan. Candidate plan 1 uses a simpler El Niño definition based only on the DJF season, but its EP/CP classification threshold ($\Delta \ge 0.5$°C) and weighted model ranking metric are more robust. Candidate plan 2 employs a more rigorous event definition, requiring the threshold to be exceeded for at least five consecutive seasons, and includes a critical preprocessing step of detrending the data. However, its EP/CP classification threshold is lower ($\Delta \ge 0.1$°C), and its model ranking metric is a simple multi-component average. The Final plan synthesizes the strengths of both: it adopts the more rigorous event definition and detrending process from Candidate plan 2, while incorporating the more discerning EP/CP classification threshold and the comprehensive weighted ranking metric from Candidate plan 1.
    }
    \label{fig:plan_diversity}
\end{figure}

Fig.~\ref{fig:plan_diversity} provides a concrete example of this aggregation process for a scientific task involving the evaluation of ENSO diversity. In this case, the system generated several plans, including two distinct candidates. Candidate plan 1 proposed a robust weighted ranking metric and a clear classification threshold for EP/CP events, but its event definition was overly simplistic and it critically omitted the necessary step of detrending the climate data. Candidate plan 2, in contrast, correctly included a rigorous, sustained-threshold event definition and the essential detrending preprocessing step. However, its classification threshold was less discerning, and its model ranking metric was a less sophisticated, unweighted average.

Neither plan was optimal on its own. The Plan Aggregation Agent synthesized the Final plan by selectively combining the strengths of both: it adopted the rigorous event definition and detrending method from Candidate plan 2 while integrating the more robust classification threshold and the comprehensive weighted ranking metric from Candidate plan 1. This selective integration ensures that the final experimental design is not only comprehensive but also scientifically rigorous, avoiding the pitfalls present in the initial candidate plans. The aggregation step is therefore crucial for enhancing the quality, reliability, and scientific validity of the final executable plan that guides the subsequent modules of the system.

\subsection{Effect of checking and feedback loop}

\subsubsection{Result checking loop}

The Result Checking mechanism is a critical component of the Scientific Diagnosis Module, serving as an essential safeguard to ensure the scientific validity and operational robustness of the entire automated workflow. In complex data analysis tasks, code generated by LLMs, while often syntactically correct, can produce scientifically invalid or erroneous outputs without causing an explicit program crash. Common failure modes include generating empty files, arrays filled with non-numeric values (NaNs), or data structures with incorrect dimensions or metadata. Such silent failures can corrupt the entire analysis pipeline if left undetected, leading to wasted computational resources and, more critically, incorrect scientific conclusions.

\begin{figure}[!ht]
    \centering
    \includegraphics[width=0.8\linewidth]{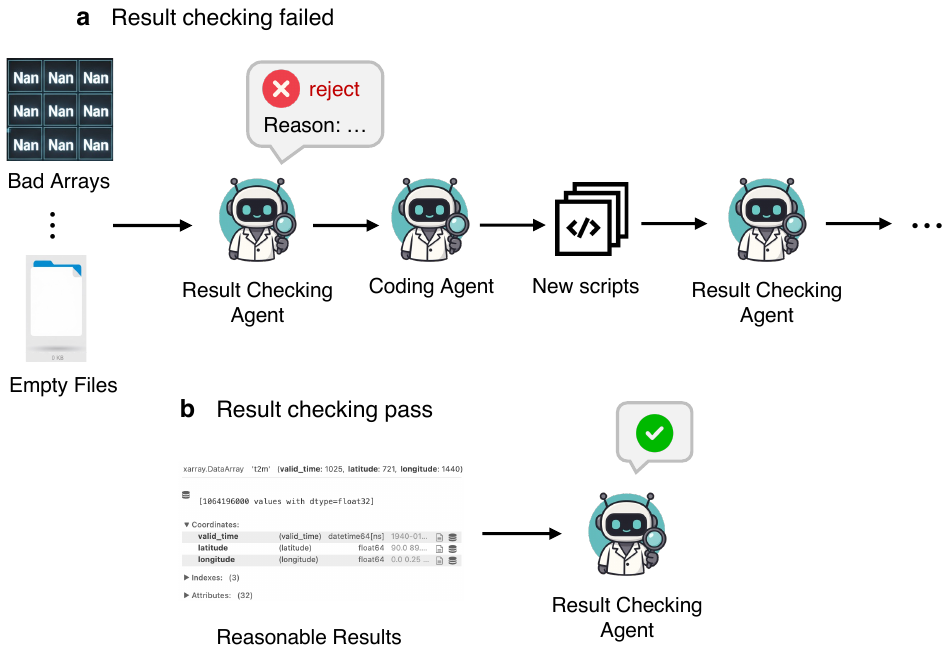}
    \caption{
The workflow of the Result Checking Agent within a self-correction loop. \textbf{a}, When the agent detects invalid outputs, such as arrays filled with NaN values or empty files, it rejects the result and provides a specific reason. This feedback triggers the Coding Agent to modify the script, initiating a debugging cycle. \textbf{b}, When the output is reasonable, for example, a correctly formatted data array with valid data, the result check passes, and the workflow proceeds.
    }
    \label{fig:result_check}
\end{figure}

The Result Checking Agent is designed to intercept these errors through a proactive verification process, as illustrated in Fig.~\ref{fig:result_check}. Following the execution of a script by the Coding Agent, the Result Checking Agent inspects the output files and data objects. If it identifies a flawed result, such as a data array entirely composed of NaNs or an empty output file, it rejects the output (Fig.~\ref{fig:result_check}a). Crucially, the rejection is accompanied by structured, descriptive feedback that specifies the nature of the error. This feedback is passed back to the Coding Agent, which uses the information to diagnose the problem and generate a revised script. This interaction forms an automated self-correction loop that iteratively refines the code until a valid output is produced.

This iterative feedback loop is fundamental to the framework's autonomy and reliability. It transforms the system from a simple code generator into a resilient agent capable of diagnosing and recovering from its own errors. When an output is deemed reasonable and passes all verification checks, such as a well-formed xarray DataArray with valid coordinates and values (Fig.~\ref{fig:result_check}b), the agent approves the result, allowing the workflow to proceed to the next stage. By systematically validating each intermediate step, the Result Checking mechanism significantly enhances the integrity of the final analysis, reduces the need for manual human oversight, and ensures that the system operates in a more robust and scientifically rigorous manner.

\subsubsection{Image feedback loop}

Beyond ensuring the numerical correctness of the data, the final presentation of scientific results through visualizations is paramount for interpretation and communication. Automated code generation can often produce plots that are technically correct but suffer from aesthetic or scientific presentation flaws. Common issues include suboptimal color maps, poor use of space, missing essential elements like color bars or labels, and awkward layouts. The Image Feedback Loop is specifically designed to address these challenges, automating the refinement process to produce publication-quality figures.

This mechanism functions as a self-correction cycle for visualizations. After a script generates a plot, the Image Feedback Agent analyzes the resulting image file, identifying specific areas for improvement. It then formulates concrete, actionable feedback, which is passed to the Coding Agent to revise the plotting script. This iterative process continues until a high-quality visualization is achieved.

\begin{figure}[!ht]
    \centering
    \includegraphics[width=1.0\linewidth]{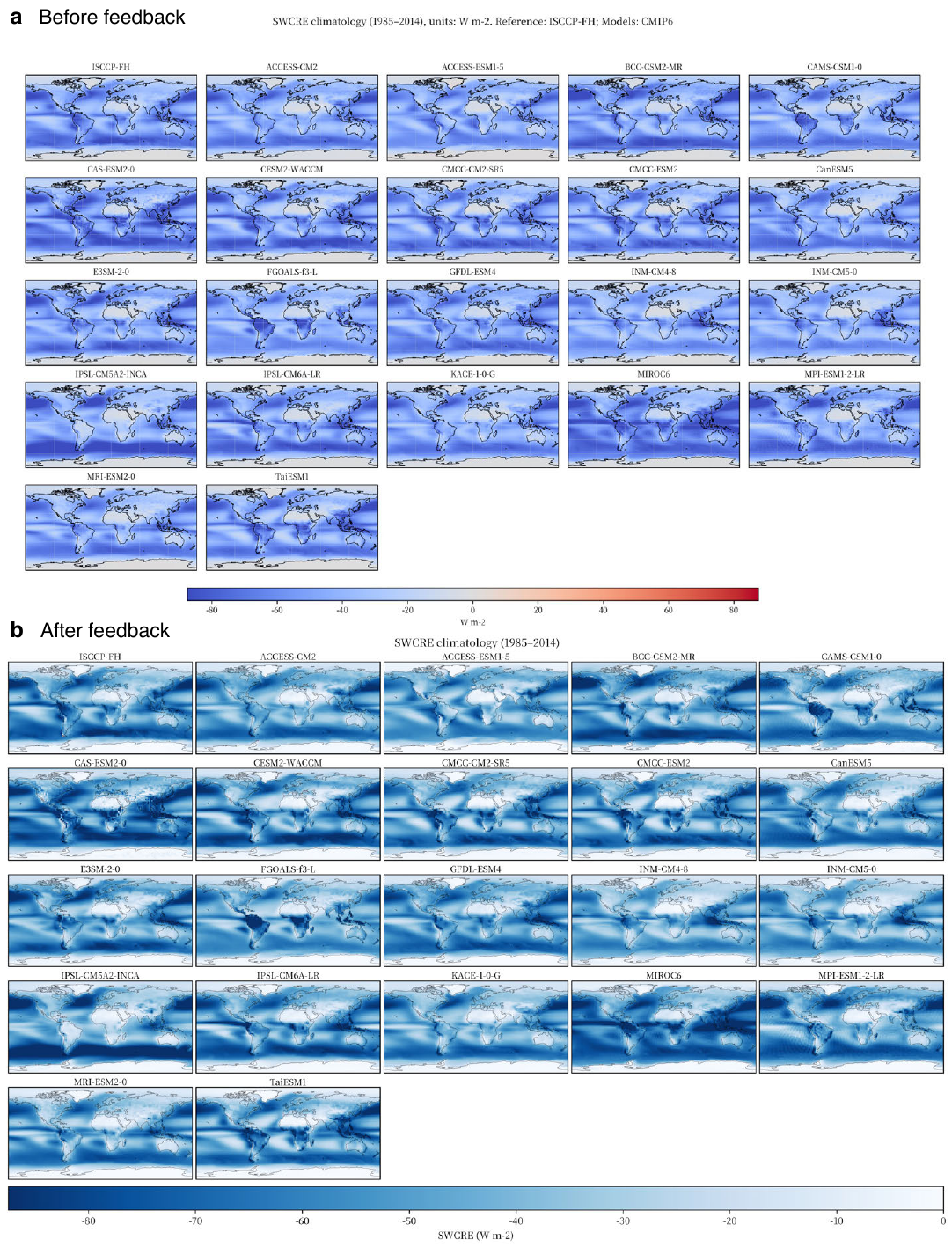}
    \caption{
Demonstration of plot refinement via the Image Feedback loop. \textbf{a}, The initial multi-panel figure generated by the code. This version exhibits several suboptimal presentation characteristics, including a symmetric color bar (-80 to 80 W m$^{-2}$) that poorly highlights the data's predominantly negative distribution and an awkward spacing between the main title and the plots. \textbf{b}, The revised figure after the Image Feedback Agent provided corrective feedback. The visualization is substantially improved: the color bar has been adjusted to a more appropriate range (-80 to 0 W m$^{-2}$) to enhance visual contrast and detail; the unnecessary white space has been cropped to allow the maps to fill their respective panels more effectively; and the title's position has been optimized for better aesthetics and readability.
    }
    \label{fig:img_feedback_1}
\end{figure}

\begin{figure}[!ht]
    \centering
    \includegraphics[width=0.9\linewidth]{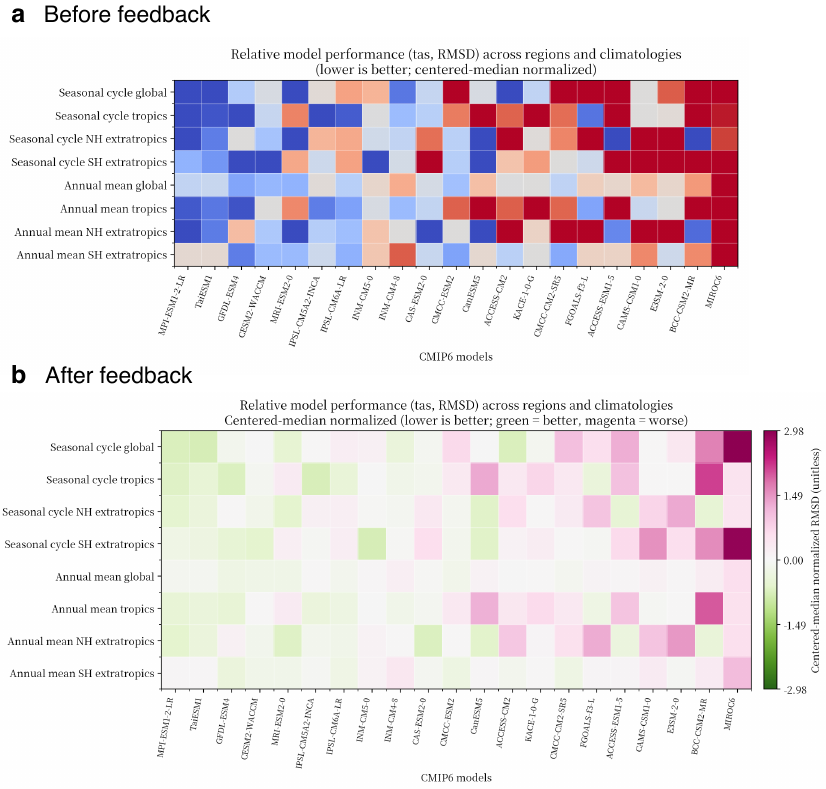}
    \caption{
A further illustration of the visualization refinement capabilities of the Image Feedback agent, applied to a model performance heatmap. \textbf{a}, The initial plot generated by the code uses a standard red-blue diverging colormap. Critically, it lacks a color bar, which renders quantitative interpretation of the model performance impossible and constitutes a significant scientific flaw. \textbf{b}, The revised figure after the agent's feedback. Two key improvements have been implemented: first, a properly labeled color bar has been added, enabling a quantitative assessment of the centered-median normalized RMSD. Second, the color scheme has been updated to a more perceptually intuitive green-magenta diverging colormap, with an improved title that explicitly states ``green = better, magenta = worse." This intervention transforms the plot from an incomplete and purely qualitative visualization into a quantitative, clear, and scientifically sound figure.
    }
    \label{fig:img_feedback_2}
\end{figure}

Fig.~\ref{fig:img_feedback_1} and~\ref{fig:img_feedback_2} demonstrate the tangible impact of this feedback loop. In the first example, the initial multi-panel plot (Fig.~\ref{fig:img_feedback_1}a) had several deficiencies: a symmetric color bar that obscured details in the predominantly negative data range, excessive white margins, and poor title spacing. After feedback, the revised figure (Fig.~\ref{fig:img_feedback_1}b) shows marked improvement. The color bar was adjusted to a more appropriate range, the white space was cropped to enhance focus on the data, and the title was repositioned, collectively improving the plot's clarity and aesthetic appeal.

The second example highlights the loop's ability to correct more critical scientific flaws. The initial heatmap (Fig.~\ref{fig:img_feedback_2}a) was scientifically incomplete as it lacked a color bar, rendering the plot uninterpretable. The feedback loop identified this crucial omission. The resulting figure (Fig.~\ref{fig:img_feedback_2}b) not only includes a properly labeled color bar but also features an improved, more intuitive green-magenta color scheme, with a descriptive title clarifying that ``green = better, magenta = worse." This intervention transformed the plot from an invalid visualization into a scientifically sound and quantitatively useful figure.

Together, these examples show that the Image Feedback Loop is essential for achieving true end-to-end automation. It elevates the framework's output from raw, unpolished plots to refined, publication-ready figures, ensuring that the final results are not only accurate but also communicated effectively. This automates a traditionally time-consuming manual step and significantly enhances the reliability and professionalism of the generated scientific products.


\section{Case study of evaluation tasks}

This section presents several cases of \model~on the evaluation tasks. Due to page limitations, only a subset of the visualization results is shown here, and the generated code scripts are omitted. The complete outputs will be made available in the GitHub repository \url{https://github.com/OpenEarthLab/EarthLink}.

\subsection{Level 1 – simple statistical analysis}

\subsubsection{Annual mean and seasonal cycle of precipitation}

Here, we tasked the \model~with a standard climate model evaluation problem: assessing the performance of CMIP6 models in simulating monthly precipitation. The agent system was instructed to use the ERA5 reanalysis dataset as the observational benchmark for the period 1985–2014. The core requirements of the task were to analyze both the annual mean precipitation and the seasonal cycle across a variety of global and regional domains, generate corresponding time series plots, and produce a final quantitative ranking of the models, summarized in a bar chart.

Upon receiving the request, the system formulated a comprehensive and methodologically sound experimental plan. This plan detailed every step of the proposed analysis, beginning with the precise identification of datasets, including 21 specific CMIP6 models and the ERA5 reference. It established a clear preprocessing workflow, including the crucial steps of unit harmonization from native model output kg m$^{-2}$ s$^{-1}$) to a more conventional metric (mm day$^{-1}$), regridding all datasets to a common 1.0° × 1.0° grid, and applying land-ocean masks derived from model land-fraction files. The plan defined a set of eleven analysis regions, spanning global, hemispheric, and key sub-continental domains such as the Amazon and Sahel. For the core analysis, \model~proposed a suite of performance metrics designed to evaluate distinct aspects of model fidelity. These included annual mean bias, as well as the Root Mean Square Error (RMSE) and Pearson correlation coefficient to assess the accuracy of the seasonal cycle's shape and the interannual variability. Crucially, the plan outlined a transparent methodology for model ranking by transforming these diverse metrics into normalized skill scores, which were then aggregated into a single composite score for each model.

Following the execution of its plan, \model~produced a detailed analysis of the results, including visualizations of the seasonal precipitation cycle across all specified regions. The summary concluded that while the multi-model mean (MMM) generally captures the primary characteristics of global precipitation, significant and systematic regional biases persist. For instance, \model~identified a prevalent wet bias across the Tropics ($+3\text{–}6\%$) and a pronounced dry bias in the Amazon region ($-13\text{–}16\%$), where the ERA5 reference consistently fell near the upper quartile of the model ensemble. The analysis of the seasonal cycle highlighted specific model deficiencies, such as an underestimated monsoon peak in the Sahel, an overly dry summer in Europe, and a slight phase lag in the Northern Hemisphere's summer precipitation maximum.

The culmination of the analysis was a final ranked list of the 21 CMIP6 models based on the composite skill score. The top-performing models were identified as E3SM-2-0, IPSL-CM6A-LR, and CMCC-ESM2, which demonstrated the most balanced performance across all regions and metrics. The agent also provided valuable interpretative insights, noting a clear "generational signal" where newer model versions generally outperformed their predecessors (e.g., IPSL-CM6A-LR over IPSL-CM5A2-INCA). The final output provided practical guidance, recommending the use of top-ranked models for applications sensitive to seasonal precipitation and advising that bias correction would be essential for studies focused on regions with large systematic errors, such as the Amazon and Sahel.

The input and output are shown below.

\begin{requestbox}
\begin{lstlisting}
Evaluate the performance of CMIP6 models in simulating monthly precipitation.
Requirements:
1. Analyze the annual mean and seasonal cycle in different regions and plot time series
2. Give the model a ranking, which can be displayed in a bar chart.
3. CMIP6 models: ACCESS-CM2, ACCESS-ESM1-5, BCC-CSM2-MR, CAMS-CSM1-0, CAS-ESM2-0, CESM2-WACCM, CMCC-CM2-SR5, CMCC-ESM2, CanESM5, E3SM-2-0, FGOALS-f3-L, GFDL-ESM4, INM-CM4-8, INM-CM5-0, IPSL-CM6A-LR, IPSL-CM5A2-INCA, KACE-1-0-G, MIROC6, MPI-ESM1-2-LR, MRI-ESM2-0, TaiESM1
4. Reference dataset: ERA5.
5. Time period: 1985-2014.
\end{lstlisting}
\end{requestbox}

\begin{planbox}
\begin{lstlisting}
Experimental plan: Evaluate CMIP6 performance in simulating monthly precipitation (1985–2014) relative to ERA5

1) Objective
- Quantitatively evaluate how well CMIP6 models simulate monthly precipitation against ERA5 over 1985–2014.
- Analyze annual mean and the seasonal cycle across multiple regions.
- Produce regional time series and a composite ranking of models, displayed in bar charts.

2) Datasets
- Reference (observational/reanalysis):
  - ERA5 monthly precipitation
  - Variable: pr (precipitation rate), units: kg m-2 s-1, frequency: monthly means (Amon-consistent)
  - Period: 1985-01 to 2014-12
- CMIP6 models (historical, activity: CMIP; MIP: Amon; frequency: mon):
  - Models: ACCESS-CM2, ACCESS-ESM1-5, BCC-CSM2-MR, CAMS-CSM1-0, CAS-ESM2-0, CESM2-WACCM, CMCC-CM2-SR5, CMCC-ESM2, CanESM5, E3SM-2-0, FGOALS-f3-L, GFDL-ESM4, INM-CM4-8, INM-CM5-0, IPSL-CM6A-LR, IPSL-CM5A2-INCA, KACE-1-0-G, MIROC6, MPI-ESM1-2-LR, MRI-ESM2-0, TaiESM1
  - Experiment: historical
  - Variable: pr (kg m-2 s-1)
  - Ensemble member: r1i1p1f1; if missing, use the lowest-numbered available realization consistently across variables
- Fixed fields (fx) from CMIP6 for weighting and masking:
  - sftlf (percentage land fraction, %; MIP: fx)
  - areacella (grid-cell area for atmospheric grid, m2; MIP: fx)
  - If a model lacks sftlf/areacella, compute weights from the target grid geometry and use a common land–sea mask based on regridded ERA5 land fraction.

3) Time period and resolution
- Period: 1985-01 through 2014-12 inclusive (30 years)
- Resolution: monthly (Amon)
- Annual means derived from monthly regional means (12-month average per year)

4) Regions
Apply longitude in degrees east from −180 to 180 (or 0–360 consistently). Define and use masks as follows:
- Global (land + ocean): 90S–90N, all longitudes
- Global land-only: 90S–90N, sftlf ≥ 50%
- Global ocean-only: 90S–90N, sftlf < 50%
- Tropics: 20S–20N
- Northern extratropics: 20N–90N
- Southern extratropics: 90S–20S
- Sub-continental boxes:
  - North America (NA): 170W–50W, 15N–70N
  - Europe (EU): 10W–40E, 35N–70N
  - East Asia (EAS): 100E–145E, 20N–50N
  - Amazon (AMZ): 80W–40W, 20S–5N
  - Sahel (SAH): 20W–30E, 10N–20N

5) Preprocessing
- Unit harmonization:
  - Convert all pr to mm day-1 using: pr_mmday = pr (kg m-2 s-1) × 86400
- Time handling:
  - Subset to 1985-01–2014-12
  - Retain native monthly means across calendars (gregorian/noleap/360-day). For annual means, average the 12 monthly means to avoid calendar-length bias
  - Ensure time series lengths are exactly 360 months per dataset
- Spatial regridding:
  - Target grid: regular 1.0° × 1.0° global grid (or ERA5’s regular grid if preferred, used consistently for all datasets)
  - Method: area-weighted bilinear interpolation for pr; use conservative or bilinear for sftlf as needed to retain fractions
- Weights and masks:
  - Area weights: use areacella on the target grid; if not available, compute grid-cell areas from the target grid bounds (cosine latitude-based areas)
  - Land/ocean masks: sftlf threshold at 50% to define land (≥50%) and ocean (<50%); derive ocean fraction as 100 − sftlf. Regrid sftlf to the target grid before forming masks
- Ensemble/member selection:
  - Use a single realization per model (r1i1p1f1 preferred); use the same member across pr and fx variables for that model

6) Core calculations
For each dataset (ERA5 and each model) and each region:
- Regional monthly means:
  - Compute area-weighted regional mean monthly time series pr_region(t) in mm day-1:
    - Weighted mean over valid cells within the region and mask: sum(pr × area × mask) / sum(area × mask)
- Annual means:
  - Compute annual means by averaging the 12 monthly regional means per year (1985–2014)
  - Compute the 1985–2014 climatological annual mean (single number per dataset and region)
- Seasonal cycle (climatology):
  - Compute the 12-month climatology by averaging each calendar month across 1985–2014
  - Compute amplitude as the standard deviation across the 12 monthly climatology values and optionally max−min
  - Determine phase as the month of maximum climatological precipitation (resolve ties by earliest month)
- Anomalies for variability:
  - Annual anomalies: subtract the 1985–2014 mean annual value from the annual mean time series
  - (Optional) Monthly anomalies: subtract the monthly climatology from each month to remove the seasonal cycle when exploring higher-frequency variability

7) Performance metrics (per model relative to ERA5, computed per region)
- Mean state (annual mean):
  - Bias_annual (mm day-1) = model climatological annual mean − ERA5 climatological annual mean
  - Relative bias (%) = 100 × Bias_annual / ERA5 climatological annual mean (report only where ERA5 ≥ 0.5 mm day-1)
- Seasonal cycle (12-month climatologies):
  - RMSE_seasonal (mm day-1): RMSE across the 12 monthly climatology values
  - Corr_seasonal (unitless): Pearson correlation across the 12 monthly climatology values
  - Amplitude ratio (unitless): std(model monthly climatology) / std(ERA5 monthly climatology)
  - Phase difference (months): circular difference between months of peak precipitation
- Interannual variability (annual means, 30 years):
  - σ_model and σ_obs (mm day-1): standard deviation of annual means
  - Variability ratio (unitless): σ_model / σ_obs
  - RMSE_annual (mm day-1): RMSE between model and ERA5 annual means
  - Corr_annual (unitless): Pearson correlation of annual mean anomalies

8) Composite skill score and ranking
- Metric-to-skill transformations (higher is better; clip to [0,1]):
  - S_bias = 1 − min(1, |Bias_annual| / B0), B0 = 1.0 mm day-1
  - S_rmse_seas = 1 − min(1, RMSE_seasonal / R0), R0 = 1.0 mm day-1
  - S_corr_seas = max(0, Corr_seasonal)  (negative correlations set to 0)
  - S_var = 1 − min(1, |σ_model − σ_obs| / V0), V0 = 0.5 mm day-1
  - S_corr_ann = max(0, Corr_annual)
- Regional composite score:
  - S_region = mean of [S_bias, S_rmse_seas, S_corr_seas, S_var, S_corr_ann]
- Overall model score and ranking:
  - Score_overall = average of S_region across all defined regions (equal weight)
  - Normalize Score_overall to 0–100
  - Rank models by Score_overall (descending). Also provide S_region (0–100) per region for context
- Multi-model mean (MMM):
  - Compute MMM (simple average of model fields after regridding) for context; calculate metrics for MMM but exclude from model ranking

9) Figures and diagnostic products
- Time series (per region):
  - Annual mean precipitation time series (1985–2014) for ERA5, each model, and MMM
- Seasonal cycle (per region):
  - 12-month climatological cycles for ERA5, each model, and MMM; annotate RMSE_seasonal and Corr_seasonal as needed
- Ranking bar charts:
  - Overall ranking: bar chart of Score_overall (0–100), sorted descending, one bar per model; include MMM as reference bar (not ranked)
  - Per-region ranking bar charts: bars of S_region (0–100) for each model
- Optional spatial context:
  - Global maps of 1985–2014 annual-mean precipitation for ERA5, MMM, and MMM bias (MMM − ERA5)
  - Seasonal cycle amplitude maps (std across months) for ERA5 and MMM

10) Quality control
- Coverage: verify each dataset provides exactly 360 monthly values for 1985–2014; if a model is missing months, exclude affected years from annual means and flag for that model/region
- Units: confirm pr conversion to mm day-1 everywhere
- Regridding consistency: ensure pr, sftlf, and areacella are on the same target grid before applying masks and weights
- Masks: verify land-only and ocean-only means use the correct sftlf-based masks; validate that regional means include sufficient valid area (>70% of intended area recommended)
- Weighting: confirm area-weighted means use appropriate weights (areacella or analytic grid-cell areas)
- Ensemble consistency: ensure a single, consistent realization per model is used throughout

11) Outputs to report
- Per model and region:
  - Annual-mean precipitation (mm day-1), Bias_annual (mm day-1), Relative bias (%)
  - Seasonal cycle metrics: RMSE_seasonal (mm day-1), Corr_seasonal (unitless), amplitude ratio (unitless), phase difference (months)
  - Interannual metrics: σ_model and σ_obs (mm day-1), variability ratio (unitless), RMSE_annual (mm day-1), Corr_annual (unitless)
  - S_region (0–100)
- Overall:
  - Score_overall (0–100) and model ranking
  - MMM metrics for context (not ranked)

12) Data access specifics
- CMIP6: activity = CMIP; experiment = historical; MIP = Amon (pr) and fx (sftlf, areacella); frequency = mon; models as listed; ensemble = r1i1p1f1 where available
- ERA5: monthly pr variable aligned to CMIP naming; period 1985–2014; convert to mm day-1

This plan uses only CMIP6 datasets and the ERA5 observation dataset requested, evaluates annual means and seasonal cycles across multiple regions, provides regional time series, and delivers a transparent model ranking suitable for presentation in bar charts.
\end{lstlisting}
\end{planbox}

\begin{visualbox}
  \begin{center}
  \includegraphics[width=1.0\linewidth]{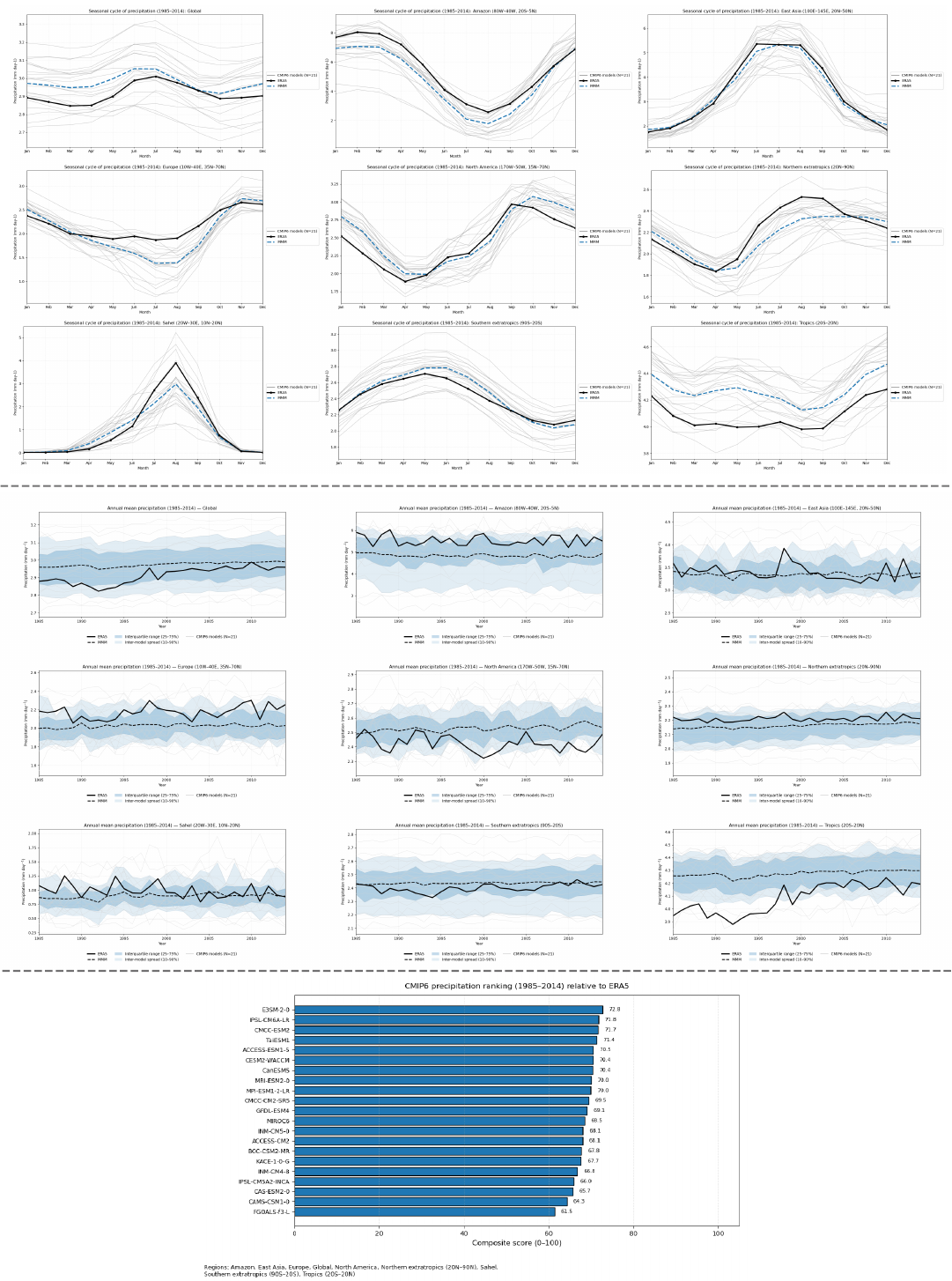}    
  \end{center}
\end{visualbox}

\begin{summarybox}
\begin{lstlisting}
# CMIP6 monthly precipitation performance vs ERA5 (1985–2014)

Scope
- Reference: ERA5 monthly precipitation
- Models (historical, r1 preferred): ACCESS-CM2, ACCESS-ESM1-5, BCC-CSM2-MR, CAMS-CSM1-0, CAS-ESM2-0, CESM2-WACCM, CMCC-CM2-SR5, CMCC-ESM2, CanESM5, E3SM-2-0, FGOALS-f3-L, GFDL-ESM4, INM-CM4-8, INM-CM5-0, IPSL-CM6A-LR, IPSL-CM5A2-INCA, KACE-1-0-G, MIROC6, MPI-ESM1-2-LR, MRI-ESM2-0, TaiESM1
- Period: 1985–2014
- Regions: Global; Tropics (20S–20N); Northern extratropics (20N–90N); Southern extratropics (90S–20S); North America; Europe; East Asia; Amazon; Sahel
- Note: The multi‑model mean (MMM) is shown for context but not ranked

Annual-mean and interannual variability (from regional annual time-series)
- Overall
  - MMM tracks ERA5 well at global and extratropical scales but with muted interannual variability (smaller year‑to‑year swings).
  - Systematic regional biases emerge: tropical wet bias; Amazon dry bias; Europe modest dry bias; Sahel/East Asia large spread and weaker modeled variability.

- Global
  - Small MMM wet bias (~+0.04 to +0.06 mm day−1, +1–2%); variability slightly damped.

- Tropics (20S–20N)
  - MMM wet bias ~+0.15 to +0.25 mm day−1 (+3–6%); captures multi‑decadal tendency but underestimates amplitude of interannual swings.

- Northern extratropics (20N–90N)
  - Slight MMM dry bias (~−0.03 to −0.05 mm day−1, −1–2%); good variability agreement.

- Southern extratropics (90S–20S)
  - Mild MMM wet bias (~+0.02 to +0.07 mm day−1, +1–3%); close tracking of ERA5.

- North America
  - MMM modest wet bias (~+0.07 to +0.12 mm day−1, +3–5%); variability muted.

- Europe
  - MMM dry bias (~−0.12 to −0.18 mm day−1, −6–9%); ERA5 often near/above the model upper quartile; variability underdone.

- East Asia
  - Small mean bias (≈0 to −0.1 mm day−1); very large inter‑model spread; MMM underestimates peaks (late 1990s, early 2010s).

- Amazon
  - Strong MMM dry bias (~−0.7 to −0.9 mm day−1, −13–16%); largest inter‑model spread; ERA5 near upper quartile.

- Sahel
  - Small mean bias on average but very large spread; MMM strongly damps interannual variability and underrepresents prominent wet years.

Seasonal cycle fidelity (from monthly climatology by region)
- Global
  - Phase and shape well captured; slight year‑round wet bias (~0.05–0.1 mm day−1); tight spread.

- Tropics
  - Correct seasonal shape; persistent wet bias (~0.15–0.3 mm day−1), largest in Nov–Dec; amplitude slightly too large.

- Northern extratropics
  - Good shape; summer peak ~1 month late and too weak by ~0.1–0.2 mm day−1.

- Southern extratropics
  - Very good phase; mild wet bias Mar–Jul (~0.05–0.1 mm day−1).

- North America
  - Reasonable cycle; peak ~1 month late with late‑summer/early‑fall wet bias (~0.1–0.2 mm day−1).

- Europe
  - Winter max/summer min captured; pronounced summer dry bias (Jun–Aug ~0.4–0.5 mm day−1 low); amplitude too large (overly dry summers).

- East Asia
  - Monsoon timing well reproduced; small negative bias at peak (Jun–Aug ~0.1–0.3 mm day−1); large model spread.

- Amazon
  - Wet- and dry-season timing captured; pervasive dry bias year‑round (≈0.5–1.0 mm day−1), sometimes exaggerated dry-season amplitude.

- Sahel
  - Onset/retreat timing broadly right (Aug peak) but monsoon intensity underestimated by ~0.8–1.1 mm day−1 at peak; very large spread.

Model ranking (composite skill score, 0–100; higher is better)
- Ranking integrates annual-mean bias, seasonal-cycle RMSE and correlation, and interannual variability across all regions.
- Bar‑chart ready list (descending):
  1) E3SM-2-0 — 72.8
  2) IPSL-CM6A-LR — 71.8
  3) CMCC-ESM2 — 71.7
  4) TaiESM1 — 71.4
  5) ACCESS-ESM1-5 — 70.5
  6) CESM2-WACCM — 70.4
  7) CanESM5 — 70.4
  8) MRI-ESM2-0 — 70.0
  9) MPI-ESM1-2-LR — 70.0
  10) CMCC-CM2-SR5 — 69.5
  11) GFDL-ESM4 — 69.1
  12) MIROC6 — 68.5
  13) INM-CM5-0 — 68.1
  14) ACCESS-CM2 — 68.1
  15) BCC-CSM2-MR — 67.8
  16) KACE-1-0-G — 67.7
  17) INM-CM4-8 — 66.8
  18) IPSL-CM5A2-INCA — 66.0
  19) CAS-ESM2-0 — 65.7
  20) CAMS-CSM1-0 — 64.3
  21) FGOALS-f3-L — 61.5

Interpretation and guidance
- Central tendency: Most models cluster near 69–70; differences are modest but systematic.
- Generational signal: Newer configurations generally outperform predecessors (e.g., IPSL-CM6A-LR > IPSL-CM5A2-INCA; INM-CM5-0 > INM-CM4-8; ACCESS-ESM1-5 > ACCESS-CM2; CMCC-ESM2 > CMCC-CM2-SR5).
- Regional priorities:
  - Strongest performance: East Asia (timing), Southern extratropics, Global.
  - Notable issues: Amazon (dry bias), Sahel (weak monsoon amplitude/variability), Europe (summer dry bias), Tropics overall (wet bias).
- Practical use: For applications sensitive to seasonal timing and amplitude, prefer top‑ranked models (e.g., E3SM-2-0, IPSL-CM6A-LR, CMCC-ESM2, TaiESM1). Consider bias correction in the Amazon, Sahel, and European summer.
\end{lstlisting}
\end{summarybox}

\subsubsection{Climatology and variability of cloud radiative effect}

Here, \model~was tasked to evaluate the performance of CMIP6 models in simulating the climatology and variability of cloud radiative effect (CRE). The agent system was instructed to perform a comprehensive statistical analysis comparing twenty-one CMIP6 models against the ISCCP-FH satellite-derived reference dataset for the period 1985–2014. The core requirements included diagnosing the climatology and variability of shortwave (SW), longwave (LW), and net CRE, producing global and zonal-mean visualizations, and ultimately generating a quantitative performance ranking of the models. 

\model~initiated the task by first generating a detailed and methodologically rigorous experimental plan. This plan precisely defined the objective, listed all necessary datasets and variables, and outlined the derivation of CRE components from native top-of-atmosphere radiative fluxes. Crucially, it specified meticulous data preparation steps, including temporal subsetting, calendar harmonization, and spatial regridding of all model and observational data to a common grid using bilinear interpolation. The plan also detailed the calculation of key statistical characteristics, such as annual-mean climatology and the standard deviation of deseasonalized monthly anomalies, ensuring that all spatial metrics were appropriately weighted by the cosine of latitude. This was followed by a sophisticated, multi-metric ranking methodology designed to synthesize model performance across spatial patterns, mean biases, and interannual variability into a single composite skill score.

\model~then executed this plan, producing a final analysis summary and a set of visualizations, exemplified by the figure showing the zonal mean of all three CRE components across the models and observations. The summary systematically diagnosed key model biases, such as the pervasive overestimation of SWCRE over subtropical stratocumulus decks and the underestimation of LWCRE in deep convective regions. It connected these statistical biases to underlying physical interpretations, noting that SWCRE errors point to deficiencies in low-cloud representation while LWCRE weaknesses suggest issues with simulating high, cold clouds. The analysis culminated in a definitive ranking, identifying GFDL-ESM4 as the top-performing model and highlighting how the final net CRE performance in many models results from a complex compensation of errors between the shortwave and longwave components.

The input and output are shown below.

\begin{requestbox}
\begin{lstlisting}
Evaluate the performance of CMIP6 models in simulating the cloud radiative effect.
Requirements:
1. Analyze the statistical characteristics of variables, including climatology, standard deviation, etc.
2. Plot global maps and zonal mean.
3. Give the model a ranking, which can be displayed in a bar chart.
4. CMIP6 models: ACCESS-CM2, ACCESS-ESM1-5, BCC-CSM2-MR, CAMS-CSM1-0, CAS-ESM2-0, CESM2-WACCM, CMCC-CM2-SR5, CMCC-ESM2, CanESM5, E3SM-2-0, FGOALS-f3-L, GFDL-ESM4, INM-CM4-8, INM-CM5-0, IPSL-CM6A-LR, IPSL-CM5A2-INCA, KACE-1-0-G, MIROC6, MPI-ESM1-2-LR, MRI-ESM2-0, TaiESM1
5. Reference dataset: ISCCP-FH.
6. Time period: 1985-2014.
\end{lstlisting}
\end{requestbox}

\begin{planbox}
\begin{lstlisting}
Title: Evaluation of CMIP6 model performance in simulating Cloud Radiative Effect (CRE) against ISCCP-FH, 1985–2014

1) Objective
- Quantify how well selected CMIP6 models reproduce top-of-atmosphere (TOA) cloud radiative effect in shortwave (SW), longwave (LW), and net components relative to ISCCP-FH over 1985–2014.
- Diagnose statistical characteristics (climatology and standard deviation).
- Produce global maps and zonal-mean diagnostics.
- Rank models using a composite skill score and display results in a bar chart.

2) Datasets and scope
- Activity/experiment/frequency/MIP: CMIP / historical / mon / Amon.
- CMIP6 models (one realization per model; prefer r1i1p1f1, else use the lowest-numbered available member consistently across variables and period): ACCESS-CM2, ACCESS-ESM1-5, BCC-CSM2-MR, CAMS-CSM1-0, CAS-ESM2-0, CESM2-WACCM, CMCC-CM2-SR5, CMCC-ESM2, CanESM5, E3SM-2-0, FGOALS-f3-L, GFDL-ESM4, INM-CM4-8, INM-CM5-0, IPSL-CM6A-LR, IPSL-CM5A2-INCA, KACE-1-0-G, MIROC6, MPI-ESM1-2-LR, MRI-ESM2-0, TaiESM1.
- Reference observations: ISCCP-FH (satellite).
- Time period: 1985-01 to 2014-12 (inclusive).
- Required variables (Amon; units W m-2; positive up):
  - rsut: TOA outgoing shortwave radiation
  - rsutcs: TOA outgoing shortwave radiation, clear-sky
  - rlut: TOA outgoing longwave radiation
  - rlutcs: TOA outgoing longwave radiation, clear-sky

3) Derived variables (CRE definitions; consistent with positive-up fluxes)
- swcre_up = rsut − rsutcs [W m-2]
- lwcre_up = rlut − rlutcs [W m-2]
- netcre_up = swcre_up + lwcre_up [W m-2]
- Compute swcre_up, lwcre_up, netcre_up for both models and ISCCP-FH using the same formulas.

4) Data preparation
- Temporal handling:
  - Subset all datasets to 1985-01 through 2014-12.
  - Harmonize calendars (gregorian/standard, noleap, 360_day) by aligning to year-month indices.
  - For annual-mean climatologies, weight monthly means by days-in-month according to each dataset’s calendar.
- Spatial handling:
  - Regrid all fields (models and ISCCP-FH) to a common regular latitude–longitude grid (e.g., 2.5° × 2.5°) using bilinear interpolation (appropriate for TOA flux intensities).
  - Construct and apply a common valid-data mask based on the intersection of valid ISCCP-FH and model data after regridding.
- Units and QC:
  - Confirm all fluxes are in W m-2 with positive-up sign convention; no unit conversion expected.
  - Require all four native flux variables (rsut, rsutcs, rlut, rlutcs) for 1985–2014 to compute CRE; if any are missing for a model, exclude that model from metrics requiring the missing component and from the composite ranking.
- Area weights:
  - Use cosine-of-latitude weights for all spatial means, RMSEs, and correlations on the common grid.

5) Calculations and statistical characteristics
- Monthly climatology (1985–2014):
  - For swcre_up, lwcre_up, netcre_up, compute 12 monthly mean maps (Jan–Dec) at each grid cell.
- Annual-mean climatology:
  - Compute time-weighted annual-mean climatology maps from monthly data for each CRE component.
- Deseasonalized monthly anomalies and interannual variability:
  - At each grid cell, subtract the corresponding monthly climatology (Jan–Dec) from each monthly value to form anomalies.
  - Compute standard deviation (STD) of monthly anomalies over 1985–2014 at each grid cell for each CRE component (map of interannual variability).
- Zonal means:
  - For each component, compute zonal-mean annual climatology (longitude-average of the annual-mean map using area weights).
  - For each component, compute zonal-mean STD (longitude-average of the STD map).

6) Model–observation comparison metrics (per model; computed for each CRE component)
All spatial metrics are computed on the common grid with area weighting and the common mask, using ISCCP-FH as reference.
- Mean-state (annual climatology):
  - Spatial bias map: Model − ISCCP-FH [W m-2].
  - Global-mean bias: Area-weighted mean of the bias map [W m-2].
  - Spatial RMSE (map): Area-weighted RMSE between model and ISCCP-FH annual-mean maps [W m-2].
  - Pattern correlation (map): Area-weighted Pearson correlation between model and ISCCP-FH annual-mean maps [unitless].
  - Zonal-mean RMSE: RMSE between model and ISCCP-FH zonal-mean annual climatology [W m-2].
- Variability (interannual; monthly anomalies):
  - Variability RMSE (map): Area-weighted RMSE between model and ISCCP-FH STD maps [W m-2].

7) Ranking methodology
- Metrics included per CRE component: spatial RMSE (annual climatology), pattern correlation (annual climatology), zonal-mean RMSE (annual climatology), global-mean absolute bias (annual climatology), variability RMSE (STD map).
- Normalization to common “higher is better” scores per component:
  - For error-type metrics E ∈ {RMSE_map, RMSE_zonal, |bias_global|, RMSE_STDmap}, define S_E = 1 − (E − E_min)/(E_max − E_min), bounded to [0,1], computed across models for that component.
  - For correlation r_map, define S_r = (r − r_min)/(r_max − r_min), bounded to [0,1], computed across models for that component.
- Per-component score:
  - S_component = mean of {S_RMSE_map, S_r, S_RMSE_zonal, S_|bias|, S_RMSE_STDmap}.
- Overall model score and ranking:
  - S_overall = mean of the three component scores (SW, LW, Net) with equal weights.
  - Rank models by descending S_overall.
  - Tie-breakers: lower mean of component RMSE_map, then higher mean of component correlation.

8) Figures and visualizations
- Global maps (for each component: swcre_up, lwcre_up, netcre_up):
  - ISCCP-FH annual-mean climatology maps (1985–2014).
  - Per-model annual-mean bias maps (Model − ISCCP-FH).
  - ISCCP-FH STD-of-anomalies maps and per-model variability bias maps (Model STD − ISCCP-FH STD).
- Zonal-mean plots:
  - Annual-mean zonal climatology curves: ISCCP-FH and all models for each component.
  - Zonal-mean STD curves: ISCCP-FH and all models for each component.
- Ranking bar chart:
  - Bars for S_overall per model (sorted). Optionally include separate bars for S_component (SW, LW, Net) alongside the overall chart.

9) Statistical and computational details
- Weight spatial metrics by cosine of latitude on the common grid.
- Use the same regridding, mask, and weights for all components and metrics to ensure fair comparisons.
- Form anomalies by removing the 12-month climatology at each grid cell; compute STD over 1985–2014 monthly anomalies.
- Annual means are time-weighted by days per month within each dataset’s calendar.

10) Outputs and deliverables
- Tables (per model and per component):
  - Global-mean annual climatology [W m-2], global-mean bias [W m-2], spatial RMSE [W m-2], pattern correlation [unitless], zonal-mean RMSE [W m-2], variability RMSE [W m-2].
  - Per-component scores and S_overall, with final rank ordering.
- Figures:
  - Global maps: ISCCP-FH climatologies, per-model climatology biases, ISCCP-FH STD maps, per-model STD biases.
  - Zonal means: annual climatology and STD comparisons.
  - Bar chart of model ranking (S_overall), with model names.

11) Data availability and constraints
- ISCCP-FH variables rsut, rsutcs, rlut, rlutcs are available over 1984-01 to 2016-12; use 1985-01 to 2014-12.
- CMIP6 variables rsut, rsutcs, rlut, rlutcs are available in Amon for the listed models in the historical experiment; restrict to 1985–2014.
- If any required variable is unavailable for a model over the analysis period, exclude that model from composite ranking and report which metrics were not computed.

12) Summary of data selection for implementation
- CMIP6: CMIP/historical/Amon/mon; models as listed; variables rsut, rsutcs, rlut, rlutcs; period 1985–2014.
- Observation: ISCCP-FH rsut, rsutcs, rlut, rlutcs; period 1985–2014.
- Derived evaluation variables: swcre_up, lwcre_up, netcre_up [W m-2].
\end{lstlisting}
\end{planbox}

\begin{visualbox}
  \begin{center}
  \includegraphics[width=1.0\linewidth]{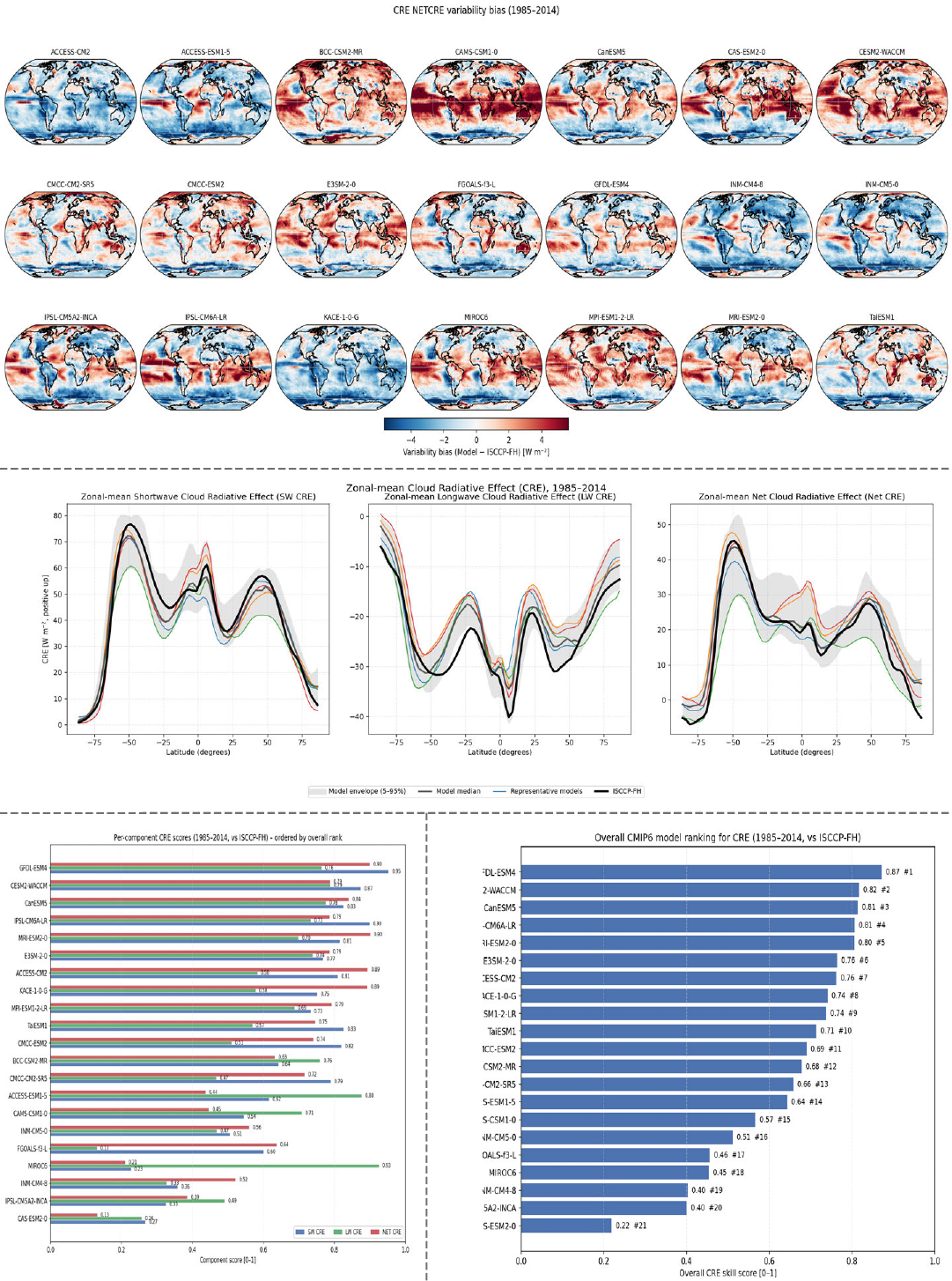}    
  \end{center}
\end{visualbox}

\begin{summarybox}
\begin{lstlisting}
# CMIP6 Cloud Radiative Effect (CRE) evaluation against ISCCP-FH (1985–2014)

Overview
- Reference: ISCCP-FH (positive-up flux convention). Period: 1985–2014.
- Models: 21 CMIP6 models listed by the user.
- CRE derived from TOA fluxes: SWCRE = rsut − rsutcs; LWCRE = rlut − rlutcs; NETCRE = SWCRE + LWCRE.
- Metrics synthesized from annual-mean climatology and interannual variability (STD of deseasonalized monthly anomalies), using area weighting.

Observed (ISCCP-FH) baseline
- SWCRE: negative (cooling), strongest over subtropical marine stratocumulus decks (SE Pacific/Atlantic/Indian) and storm tracks; weak over deserts and poles.
- LWCRE: positive (warming), strongest in deep convective tropics (ITCZ, warm pools) and storm tracks; weaker in subtropical subsidence/polar regions.
- NETCRE: small magnitude with sign varying by regime; slightly positive in subsidence subtropics, slightly negative where SW cooling dominates (storm tracks, parts of deep tropics).

Mean-state climatology: global maps (Model − ISCCP-FH)
- SWCRE biases (largest contributor to NET errors)
  - Pervasive positive biases over subtropical stratocumulus decks (+5 to +10 W m−2): models reflect too little SW (low-cloud amount/optical depth too weak).
  - Negative biases over the Southern Ocean storm belt (≈40–60°S; −5 to −10 W m−2): too-strong SW cooling.
  - Mixed but often positive biases over warm pools and many land regions.
- LWCRE biases
  - Widespread negative anomalies (−2 to −6 W m−2) in convective tropics and storm tracks: underestimation of high, cold cloud greenhouse effect.
  - Regional positive LWCRE biases in a few models/regions (+3 to +8 W m−2).
  - Intermodel spread is smaller than for SWCRE.
- NETCRE biases (SW + LW)
  - Tropics/subtropics: net positive bias common (SW cooling too weak outweighs LW).
  - Southern Ocean: net negative bias (SW cooling too strong; LW does not compensate).
  - Dipoles near ITCZ/eastern ocean basins indicate regime/ITCZ placement errors.
- Magnitudes and spread
  - SWCRE biases frequently reach ±10 W m−2; LWCRE typically within ±6 W m−2; NETCRE commonly within ±8 W m−2.
  - Greatest intermodel spread over stratocumulus regions, the equatorial Pacific, and the Southern Ocean.

Zonal-mean climatology (annual means)
- SWCRE
  - SH storm track (~50–60°S): ensemble underestimates by ~3–8 W m−2; large spread (≈±10 W m−2).
  - 0–25°N: several models too strong by ~5–10 W m−2 (overly bright subtropical maximum).
  - NH storm track (~45–55°N): underestimation by ~3–6 W m−2 is common.
- LWCRE
  - Tropics (10°S–10°N): too weak (less negative) by ~5–8 W m−2 vs ISCCP-FH.
  - Subtropics/midlatitudes: 2–5 W m−2 less negative than observed.
  - High latitudes (≥70°): too weak LWCRE by ~4–7 W m−2.
- NETCRE
  - SH storm track peaks are reasonably captured due to partial SW–LW compensation.
  - 0–25°N: net too positive by ~5–10 W m−2.
  - Arctic: observed net turns negative (~−5 W m−2) but most models remain positive (+6 to +10 W m−2).

Interannual variability: STD of monthly anomalies (maps and zonal implications)
- Observed patterns
  - SWCRE: strongest variability (≈12–19 W m−2) in subtropical cloud belts and storm tracks.
  - LWCRE: maxima (≈10–13 W m−2) in convective tropics and storm tracks.
  - NETCRE: a blend; peaks in storm tracks and convective regions.
- Systematic model behaviors
  - SWCRE: underestimation in extratropical storm tracks/Southern Ocean; overestimation in convective tropics and monsoon regions (typical biases ±3–6 W m−2).
  - LWCRE: many models underestimate tropical variability; some overestimate across the tropics; biases typically ±1–3 W m−2.
  - NETCRE: mirrors SW dominance over oceans; negative in storm tracks, positive in tropics.
- Zonal implications
  - SW: negative variability biases at 40–60° (both hemispheres) and positive near the equator.
  - LW: negative variability bias centered on the equator for many models; weaker positive biases in midlatitudes.
  - NET: negative in storm tracks, positive in tropics.

Physical interpretation
- Too-weak tropical/high-latitude LWCRE implies cloud tops too warm/low or insufficient high/ice cloud (weak cloud greenhouse).
- SWCRE issues reflect low-cloud regime deficiencies (amount/optical depth) in stratocumulus and storm tracks.
- NETCRE errors follow latitude-dependent SW–LW imbalance: positive in tropics/subtropics, negative over the Southern Ocean; Arctic net sign error is common.

Composite skill ranking (bar-chart figure; higher is better [0–1])
- Method: per-component scores (SW, LW, NET) combine normalized global-mean absolute bias, spatial RMSE (map), pattern correlation, zonal-mean RMSE, and variability RMSE; overall score is the mean of the three components.
- Overall ranking (score)
  1) GFDL-ESM4 (0.87)
  2) CESM2-WACCM (0.82)
  3) CanESM5 (0.81)
  4) IPSL-CM6A-LR (0.81)
  5) MRI-ESM2-0 (0.80)
  6) E3SM-2-0 (0.76)
  7) ACCESS-CM2 (0.76)
  8) KACE-1-0-G (0.74)
  9) MPI-ESM1-2-LR (0.74)
  10) TaiESM1 (0.71)
  11) CMCC-ESM2 (0.69)
  12) BCC-CSM2-MR (0.68)
  13) CMCC-CM2-SR5 (0.66)
  14) ACCESS-ESM1-5 (0.64)
  15) CAMS-CSM1-0 (0.57)
  16) INM-CM5-0 (0.51)
  17) FGOALS-f3-L (0.46)
  18) MIROC6 (0.45)
  19) INM-CM4-8 (0.40)
  20) IPSL-CM5A2-INCA (0.40)
  21) CAS-ESM2-0 (0.22)
- Component-wise behavior
  - SWCRE shows the widest skill spread and dominates performance contrasts (top near ~0.9; bottom near ~0.13).
  - LWCRE skill is generally higher and less dispersed (many ≥0.7).
  - NETCRE often benefits from SW–LW compensation; several models score higher in NET than in either component individually.
- Notable model characteristics from scores
  - High, balanced skill: GFDL-ESM4; also strong: CESM2-WACCM, CanESM5, IPSL-CM6A-LR.
  - NET strong via compensation: MRI-ESM2-0, ACCESS-CM2, E3SM-2-0.
  - Imbalance cases: MIROC6 exhibits excellent LW but weak SW and modest NET; FGOALS-f3-L and CAS-ESM2-0 limited by very weak SW performance.
  - Family contrasts: IPSL-CM6A-LR markedly improves over IPSL-CM5A2-INCA; INM-CM5-0 improves over INM-CM4-8 but remains below median; ACCESS-CM2 outperforms ACCESS-ESM1-5 primarily via better NET.

Key takeaways
- SWCRE errors (low-cloud regimes and Southern Ocean) are the dominant source of CRE biases and intermodel spread; improving these would yield the largest gains in NETCRE.
- LWCRE is generally better captured but remains too weak in the deep tropics and high latitudes in many models, indicating deficiencies in high, cold cloud representation.
- NETCRE realism often arises from compensation between SW and LW errors; highest fidelity comes from balanced improvements in both components.
- Best overall performers reproduce both the spatial patterns and the amplitude of mean and variability (especially in NET), while the lowest-ranked models show pervasive SWCRE deficiencies across regimes.
\end{lstlisting}
\end{summarybox}

\subsubsection{Change and trend of ocean heat content}

Here, we tasked \model~to evaluate the performance of six CMIP6 models in simulating global Ocean Heat Content (OHC) for the period 1970–2010. Key constraints were imposed to test \model's ability to work with incomplete data and adhere to specific physical definitions. \model~was instructed to calculate OHC using only the potential temperature field (thetao), emulating scenarios where supplementary grid information is unavailable in observational datasets. The evaluation required a multi-faceted analysis across three distinct depth layers (0–300 m, 0–700 m, and 0–2000 m) and involved generating time series of OHC anomalies, bar charts of decadal change, and spatial maps of long-term trends, all benchmarked against the IAP observational dataset.

In response to this request, \model~formulated a comprehensive and methodologically sound experimental plan. The plan correctly identified the required datasets, including the specific models and ensemble members, and outlined a robust workflow for data harmonization and analysis. It delineated precise steps for preprocessing, including vertical integration across specified depth ranges while accounting for bathymetry, and the calculation of annual means. It further specified the analytical procedures for deriving the required outputs: OHC anomalies relative to a 1991–2010 baseline, the change between the 1970–1990 and 1991–2010 periods, and the spatial trend in watts per square meter (W m$^{-2}$). For robust comparison, \model~planned to regrid all spatial data to a common 1°x1° grid and compute a multi-model mean (MMM) to assess ensemble performance.

Following the execution of this plan, \model~produced a detailed analysis summarizing the results. The agent's synthesis concluded that the CMIP6 models, particularly the MMM, successfully captured the overall sign and decadal phasing of OHC increase across all depth layers, including transient cooling events like the Mount Pinatubo eruption. However, the analysis also quantified systematic biases. For the 0–300 m layer, \model~found that most models and the MMM overestimated the decadal OHC change, with the MMM showing a +40.0\% bias relative to the IAP. Spatially, \model's analysis of the trend maps revealed common regional biases across depths, such as an underestimation of warming in the subpolar North Atlantic and an overestimation in the Southern Ocean and along western boundary currents. The agent provided a nuanced, model-by-model performance breakdown, identifying high-end warmers (e.g., CIESM, FIO-ESM-2-0) and models that aligned more closely with observations in specific metrics (e.g., GFDL-CM4 for 0–700 m change). Significantly, \model~demonstrated a capacity for self-critique by flagging the generated bar chart for the 0–2000 m depth range as "not physically credible" due to internal inconsistencies, concluding that it likely reflected a processing error and should not be used for inference—a crucial step in ensuring scientific validity.

The input and output are shown below.

\begin{requestbox}
\begin{lstlisting}
Evaluate CMIP6 models in simulating Ocean Heat Content (OHC).
Requirements:
1. Calculate the OHC only based on thetao data, don't use supplementary_variables as observational data doesn't have these data. 
2. The results should include time seriese figures of anomalies , bar charts of change, spatial maps of OHC trend, et.al.
3. Analyze at different depth ranges: 0-300m, 0-700m, 0-2000m.
4. CMIP6 models: CIESM, CMCC-ESM2, FGOALS-f3-L, FIO-ESM-2-0, GFDL-CM4, MPI-ESM1-2-LR
5. Reference thetao Dataset: IAP. Note that the original variable unit of IAP may not be thetao. If necessary, convert unit to thetao according to TEOS-10.
6. Time period: 1970-2010 (use 1991-2010 minus 1970-1990 as change; calculate trend in whole period).
7. The defination of OHC trend (or alternatively, OHC rate, or tendency): The time derivative of OHC (dOHC/dt), given in units of J yr −1 or W m −2.
8. For all figures, choose appropriate colormap, bars or contour levels to highlight the key features.
\end{lstlisting}
\end{requestbox}

\begin{planbox}
\begin{lstlisting}
Title
Evaluation of CMIP6 models in simulating Ocean Heat Content (OHC) using thetao only, relative to IAP, 1970–2010

Objective
- Quantitatively evaluate OHC anomalies, decadal change, and trends (dOHC/dt) in selected CMIP6 models relative to IAP for three depth ranges: 0–300 m, 0–700 m, 0–2000 m.
- Compute OHC strictly from sea water potential temperature (thetao), without using supplementary variables.

Datasets
- Observational reference
  - IAP (Institute of Atmospheric Physics) ocean temperature, project: OBS6
  - Variable: thetao (Sea Water Potential Temperature), mip: Omon, units: degC, temporal coverage: 1940–present; use 1970-01 to 2010-12.
  - Verification and unit handling:
    - Confirm that IAP provides thetao [degC]. If instead the field is in situ temperature or Conservative Temperature, convert to potential temperature at 0 dbar following TEOS-10 using pressure inferred from depth and nominal salinity (kept constant for conversion only). Proceed using thetao thereafter. No other observational variables are used downstream.

- CMIP6 simulations
  - Activity: CMIP; Experiment: historical; Frequency: mon; MIP table: Omon; Variable: thetao [degC]; Grid: native (gn).
  - Models and ensemble members:
    - CIESM: r1i1p1f1
    - CMCC-ESM2: r1i1p1f1
    - FGOALS-f3-L: r1-2i1p1f1
    - FIO-ESM-2-0: r1i1p1f1
    - GFDL-CM4: r1i1p1f1
    - MPI-ESM1-2-LR: r1i1p1f1
  - Period analyzed: 1970-01 to 2010-12.

Spatial/temporal scope
- Domain: global ocean (land masked by thetao missing values).
- Depth ranges: 0–300 m; 0–700 m; 0–2000 m.
- Temporal resolution: monthly input; compute monthly OHC and then annual means for anomaly time series, inter-period change, and trend estimation.

Key constraints and constants (thetao-only)
- Only thetao is used for OHC; do not use areacello, volcello, thkcello, salinity, density, or cp fields from models or observations.
- Seawater property constants (fixed across all datasets):
  - ρ0 = 1026 kg m−3
  - cp0 = 3990 J kg−1 K−1
  - Combined volumetric heat capacity: ρ0·cp0 = 4.09 × 10^6 J m−3 K−1
- Earth geometry constant:
  - Earth radius R = 6,371,000 m
- Seconds per year = 31,557,600 s

Preprocessing and harmonization
1) Time handling
- Subset all datasets to 1970-01 through 2010-12.
- Harmonize calendars by forming annual means from monthly data per dataset using their native calendar.

2) Vertical coordinates and thickness
- Use depth coordinate (lev/olevel; units: m). Prefer depth bounds (lev_bnds/olevel_bnds) when available.
- If bounds are absent, construct layer interfaces from mid-levels; top interface = 0 m, bottom interface extrapolated from last mid-level spacing.
- For each target depth H ∈ {300, 700, 2000} m and each grid cell:
  - Compute in-range layer thickness Δzk(H) = max[0, min(zbot,k, H) − max(ztop,k, 0)].
  - If local bathymetry is shallower than H, integrate only to the bottom (thetao is typically masked below seabed).

3) Horizontal areas (no areacello)
- Compute grid cell area A(i,j) from latitude/longitude bounds when provided. Otherwise, compute from central lat/lon spacing via spherical geometry:
  - A ≈ R^2 × Δλ (radians) × [sin(φ+Δφ/2) − sin(φ−Δφ/2)].

4) Ocean masks and coverage
- Define a time-invariant ocean mask per dataset from where thetao is valid at any time in 1970–2010.
- Require sufficient temporal coverage for trend estimation (e.g., at least 80% of annual means available in 1970–2010 at a grid cell).

OHC computation (thetao-only)
A) Monthly depth-integrated OHC per unit area
- For each dataset, grid cell (i,j), month t, and depth limit H:
  - OHC_H(i,j,t) = ρ0 × cp0 × Σk [ θ(i,j,k,t) × Δzk(H) ]  with θ in K (degC + 273.15; equivalently, use degC because differences are identical).
  - Units: J m−2.

B) Annual means
- Form annual mean OHC_H_ann(i,j,year) from monthly OHC_H.

C) Global integrals and global mean per-area
- Ocean area for a dataset and H: Ao(H) = Σ A(i,j) over valid ocean cells contributing to [0, H].
- Global total OHC: OHC_H_tot(year) = Σ [ OHC_H_ann(i,j,year) × A(i,j) ] [J].
- Global-mean per-area OHC: OHC_H_gm(year) = OHC_H_tot(year) / Ao(H) [J m−2].

D) Anomalies and inter-period change
- Anomaly baseline: 1991–2010.
- Global anomaly time series: OHC′_H_gm(year) = OHC_H_gm(year) − mean[OHC_H_gm(1991–2010)] [J m−2].
- Period change (for bars): ΔOHC_H_tot = mean[OHC_H_tot(1991–2010)] − mean[OHC_H_tot(1970–1990)] [J]; also compute per-area ΔOHC_H_gm [J m−2].

E) Trends (OHC tendency; dOHC/dt)
- Global trend:
  - Fit an ordinary least squares (OLS) linear trend to OHC_H_tot(year), 1970–2010.
  - Report slope in J yr−1 and convert to W by dividing by seconds per year; also provide W m−2 by dividing W by Ao(H).
- Spatial trend maps:
  - For each grid cell, fit OLS trend to OHC_H_ann(i,j,year), 1970–2010.
  - Report slope as J m−2 yr−1; convert to W m−2 by dividing by seconds per year.

Regridding for spatial comparison (per-unit-area fields only)
- After vertical integration, regrid annual OHC_H_ann and derived trend maps from each dataset’s native grid to a common 1°×1° regular lat–lon grid using an area-conservative method appropriate for extensive per-unit-area quantities.
- Use a common ocean mask on the target grid formed by the intersection of valid ocean cells across IAP and all models for spatial comparisons and the multi-model mean (MMM).

Model–observation comparisons and summary diagnostics
- Time series (global-mean per-unit-area anomalies, annual, 1970–2010):
  - Plot IAP and each model; compute correlation and RMSE versus IAP per depth range.
- Period change:
  - Compare ΔOHC_H_tot and ΔOHC_H_gm for each model vs IAP.
- Spatial OHC trend maps (W m−2):
  - Compare each model’s map to IAP; compute pattern correlation and area-weighted RMSE on the common grid.
- Multi-model mean:
  - Compute MMM from the six models’ regridded fields for time series and maps; compare MMM vs IAP using the same metrics.

Figure specifications
1) Time series of anomalies (global)
- Three panels: 0–300 m, 0–700 m, 0–2000 m.
- Annual OHC′_H_gm (J m−2) for 1970–2010 for IAP and each model; include MMM.
- Baseline explicitly indicated (1991–2010 mean removed).
- Visual guidance: lines that clearly distinguish models; highlight IAP. Include a zero line. Use axis limits that avoid clipping.

2) Bar charts of change
- For each depth range, bars for IAP, each model, and MMM showing ΔOHC_H_tot (J, preferably annotated in ZJ; 1 ZJ = 1e21 J).
- A companion set using ΔOHC_H_gm (J m−2) may be included.
- Visual guidance: consistent ordering across depths; include uncertainty bars based on interannual variability within each epoch if computed.

3) Spatial maps of OHC trend (1970–2010)
- For each depth range, provide:
  - IAP trend (W m−2),
  - MMM trend (W m−2),
  - MMM minus IAP difference (W m−2).
- Visual guidance: diverging colormap centered at 0 for trend and difference to emphasize warming vs cooling; contour/level spacing chosen to resolve key features without saturation; mask areas with insufficient data.

Computation and unit details
- θ: degC (treated as K for integration).
- OHC per unit area: J m−2.
- Global totals: J (report also in ZJ).
- Trends: J yr−1 (global total), J m−2 yr−1 and W m−2 (local and global-mean).
- Change between periods: J (global totals) and J m−2 (per-area).

Quality control
- Ensure vertical integration respects partial layers at 0, 300, 700, and 2000 m and truncation at the local seabed.
- Use consistent Ao(H) (static in time) for each dataset/depth range to avoid area changes aliasing into trends.
- Exclude grid cells from spatial trend maps with insufficient valid annual means.
- Check for unrealistic spikes; if present, exclude those years/cells from trend estimation.

Deliverables
- Annual, depth-integrated OHC per unit area fields (J m−2) for 0–300, 0–700, 0–2000 m on native and common 1° grids for IAP and all six models.
- Global OHC anomaly time series figures (annual, 1970–2010) for each depth range.
- Bar charts of ΔOHC (1991–2010 minus 1970–1990) as global totals (J, with ZJ labels) and per-area (J m−2).
- Spatial maps of OHC trend (W m−2) for IAP, MMM, and MMM–IAP for each depth range.
- Summary statistics: time series correlation and RMSE vs IAP; trend and change biases; spatial pattern correlation and RMSE for trend maps.

Compliance with user requirements
- OHC computed only from thetao using constant ρ0 and cp; no supplementary variables are used.
- Depth ranges, period, trend definition, and requested figures are implemented exactly as specified.
- IAP used as the sole observational reference; potential temperature conversion according to TEOS-10 is included as a verification step but is not needed if IAP thetao is confirmed.
\end{lstlisting}
\end{planbox}

\begin{visualbox}
  \begin{center}
  \includegraphics[width=1.0\linewidth]{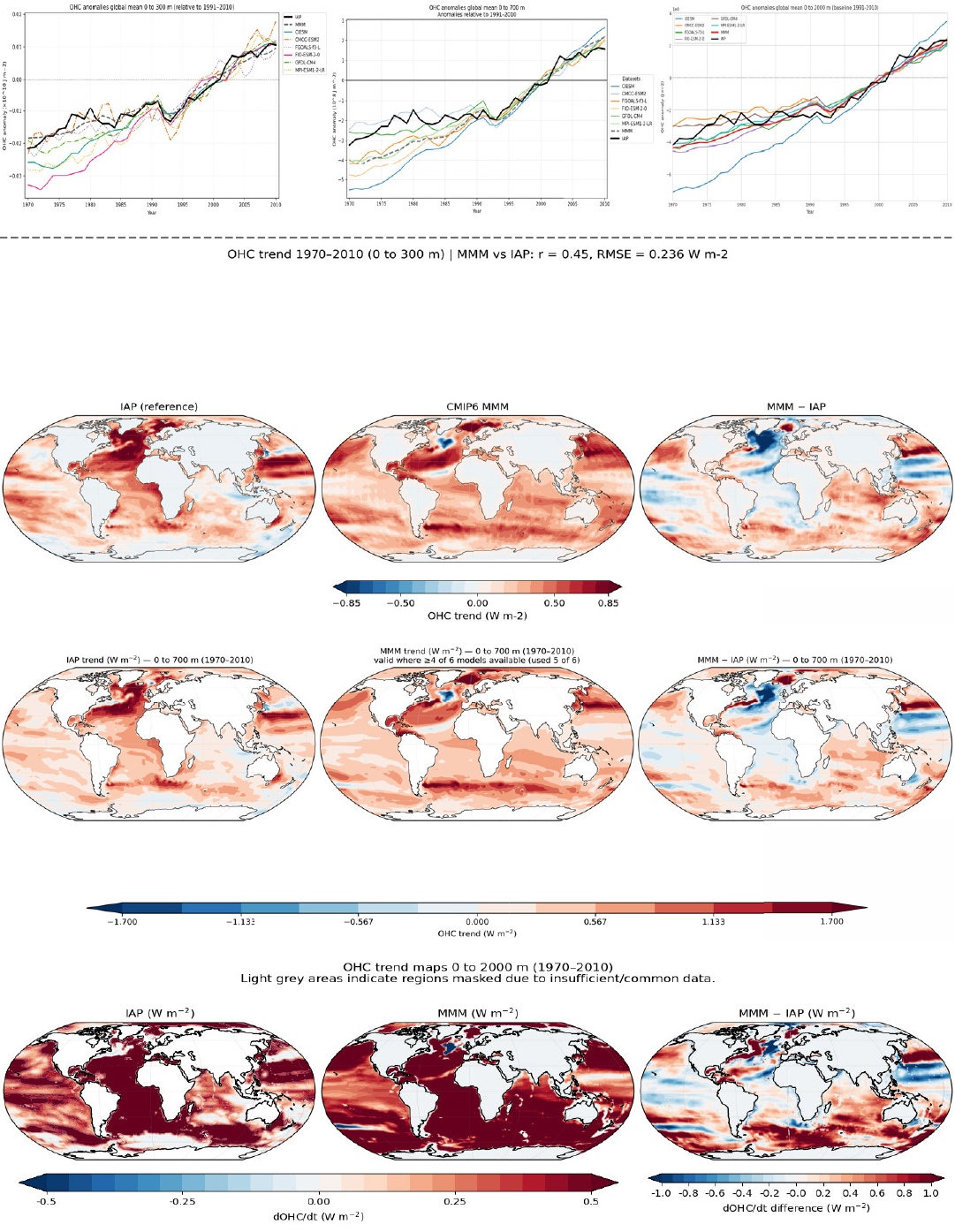}    
  \end{center}
\end{visualbox}

\begin{summarybox}
\begin{lstlisting}
# Evaluation summary: CMIP6 OHC (thetao-only) vs IAP, 1970–2010

Scope
- Depth ranges: 0–300 m, 0–700 m, 0–2000 m.
- Metrics/figures: global OHC anomaly time series (baseline 1991–2010), global ΔOHC bars (1991–2010 minus 1970–1990), spatial maps of OHC trend (dOHC/dt; W m−2).
- Reference: IAP; Models: CIESM, CMCC-ESM2, FGOALS-f3-L, FIO-ESM-2-0, GFDL-CM4, MPI-ESM1-2-LR; MMM = multi-model mean.

0–300 m

- Global anomaly time series
  - All datasets show sustained upper-ocean warming with a clear multidecadal rise and a distinct Pinatubo-related dip (~1991–1993) captured by most models.
  - MMM closely tracks IAP in phase with muted interannual variance (expected from averaging). Models differ mainly in warming amplitude.
  - Model notes: CMCC-ESM2 and MPI-ESM1-2-LR warm fastest late in the record; FGOALS-f3-L under-warms after ~2003; FIO-ESM-2-0 starts very cold and converges toward IAP in the 2000s.
  - Approximate global-mean uptake from slopes: ~0.2–0.25 W m−2 (IAP scale).

- Global ΔOHC bars (totals, ZJ; 1991–2010 minus 1970–1990)
  - IAP: 49.63 ZJ.
  - Models (bias vs IAP): CMCC-ESM2 54.04 (+8.9%), MPI-ESM1-2-LR 41.78 (−15.8%), GFDL-CM4 61.26 (+23.4%), FGOALS-f3-L 66.76 (+34.5%), CIESM 69.61 (+40.2%), FIO-ESM-2-0 76.31 (+53.8%).
  - MMM: 69.51 ZJ (+40.0%). Intermodel spread is large (41.78–76.31 ZJ).
  - Performance (closest to IAP): CMCC-ESM2 (best) < MPI-ESM1-2-LR < GFDL-CM4 < FGOALS-f3-L < CIESM ≈ MMM < FIO-ESM-2-0.

- Spatial OHC trend maps (W m−2; 1970–2010)
  - Pattern skill is moderate (MMM vs IAP r ≈ 0.45; RMSE ≈ 0.236 W m−2). Both show pervasive upper-300 m warming.
  - Systematic MMM biases: too weak in the subpolar North Atlantic and equatorial Pacific; too strong along western boundary current extensions and across broad Southern Hemisphere mid-latitudes. MMM is too smooth, missing observed regional heterogeneity.

- Takeaways (0–300 m)
  - Sign and timing of warming are well captured; amplitude biases are common. Most models (and MMM) overestimate the 1991–2010 minus 1970–1990 change, with CMCC-ESM2 closest and FIO-ESM-2-0 highest.

0–700 m

- Global anomaly time series
  - Strong, quasi-monotonic warming with late-1990s acceleration reproduced by all models and MMM.
  - Models tend to be more negative than IAP in the 1970s–mid-1980s (relative to the 1991–2010 baseline), implying stronger multi-decadal trends than IAP.
  - Approximate global-mean uptake: IAP ~0.37–0.38 W m−2; MMM ~0.49–0.50 W m−2 (MMM high by ~0.1 W m−2).
  - Model notes: CMCC-ESM2 and MPI-ESM1-2-LR align well post-1995 but end slightly high; CIESM warms fastest by 2010; GFDL-CM4 shows strong early-1990s dip and weaker end-period increase.

- Global ΔOHC bars (totals, ZJ)
  - IAP: 78.43 ZJ. MMM: 86.77 ZJ (+10.6%).
  - Models (bias vs IAP): GFDL-CM4 87.93 (+12.1%), MPI-ESM1-2-LR 66.48 (−15.2%), CMCC-ESM2 63.79 (−18.7%), FGOALS-f3-L 58.72 (−25.1%), FIO-ESM-2-0 112.13 (+43%), CIESM 131.56 (+68%).
  - Ranking (closest to IAP): MMM ≈ GFDL-CM4 < MPI-ESM1-2-LR < CMCC-ESM2 < FGOALS-f3-L << FIO-ESM-2-0 << CIESM.

- Spatial OHC trend maps (W m−2)
  - Broad agreement on widespread warming. MMM underestimates intense warming in western boundary currents and subpolar North Atlantic; overestimates across much of the Southern Ocean and parts of the subtropical Atlantic.
  - Ensemble smoothing plus common circulation/mixing biases likely contribute to too-weak hotspots and too-strong diffuse warming elsewhere.

- Takeaways (0–700 m)
  - MMM captures timing and sign but is modestly high in global uptake and shows compensating regional biases. GFDL-CM4 is closest on the decadal-change metric; CIESM and FIO-ESM-2-0 strongly overestimate.

0–2000 m

- Global anomaly time series
  - Strong, near-monotonic deep-ocean warming in all datasets; major dips (early 1980s/early 1990s) are captured.
  - MMM is very close to IAP after the mid-1990s. Model notes: CIESM exhibits the steepest trend (high bias); FGOALS-f3-L, FIO-ESM-2-0, MPI-ESM1-2-LR track IAP well.
  - Approximate global-mean uptake: IAP ~0.50–0.57 W m−2; MMM ~0.48–0.53 W m−2; CIESM higher (~0.8–0.9 W m−2).

- Global ΔOHC bars (totals, ZJ)
  - The 0–2000 m bar chart is internally inconsistent (e.g., zeros for IAP and several models; sign reversals; outlier magnitudes). These results are not physically credible and likely reflect processing issues (mask/area/thickness/sign handling). Do not use this figure for inference.

- Spatial OHC trend maps (W m−2)
  - IAP and MMM show ubiquitous deep warming, strongest in the Southern Ocean and western boundary current regions.
  - MMM − IAP differences indicate a positive amplitude bias across much of the Southern Hemisphere and boundary current regions, with underestimation along the equator and in parts of the subpolar North Atlantic. Regional biases can reach ~±1 W m−2.

- Takeaways (0–2000 m)
  - Temporal evolution is well captured by MMM and several individual models; CIESM warms too strongly. Spatially, MMM has a broad positive bias (notably in the Southern Ocean) and misses some tropical/subpolar features seen in IAP.
  - The ΔOHC bar figure for this depth is unreliable and should be recomputed before drawing conclusions on decadal change magnitudes.

Cross-depth synthesis

- Consistent strengths
  - The CMIP6 MMM reproduces the sign and decadal phasing of OHC anomalies across all depth ranges and captures major transient cooling events. Inter-model spread narrows after ~1995.

- Systematic biases
  - Amplitude: MMM tends to overestimate decadal OHC increase and/or trend in 0–300 m and 0–700 m; for 0–2000 m, MMM is close in time series but shows regional positive biases in maps.
  - Regional patterns: Common across depths—too weak warming in the subpolar North Atlantic and equatorial Pacific; too strong in western boundary current extensions and broad Southern Hemisphere mid-latitudes/Southern Ocean; patterns are overly smooth.

- Model ranking highlights
  - Closer to IAP: GFDL-CM4 (0–700 m), CMCC-ESM2 (0–300 m), and the MMM overall for time series.
  - High-end warmers: CIESM (notably at 0–700 m and 0–2000 m) and FIO-ESM-2-0 (0–300 m and 0–700 m).
  - Low-end: FGOALS-f3-L tends to under-warm at 0–700 m; MPI-ESM1-2-LR is low for 0–300 m change but near MMM thereafter.

- Notes on figure fidelity
  - The 0–2000 m ΔOHC bar chart contains evident processing errors and should be recalculated; the time series and trend maps at 0–2000 m are internally consistent and physically plausible.
\end{lstlisting}
\end{summarybox}

\subsection{Level 2 – mechanistic diagnosis}

\subsubsection{Equilibrium climate sensitivity}

\model~was tasked to estimate Equilibrium Climate Sensitivity (ECS) from a suite of 22 Coupled Model Intercomparison Project Phase 6 (CMIP6) models. The request required \model~to employ the Gregory regression method on data from abrupt-4xCO2 experiments. Key deliverables included visualizing the regression for each model, ranking the models by their calculated ECS values, and providing an ensemble analysis, including the multi-model mean and a probability density function (PDF) of the results. This task tests \model's ability to formulate a multi-step plan, handle complex climate model data, perform standard scientific analysis, and synthesize the results into coherent visualizations and summaries.

In response, \model~generated a detailed and scientifically rigorous experimental plan. The protocol outlined the precise methodology for calculating ECS by fitting a linear model, $\Delta N = a + b \Delta T$, where $\Delta N$ is the annual anomaly in the global mean top-of-atmosphere (TOA) net radiation and $\Delta T$ is the annual anomaly in global mean surface air temperature. The plan correctly identified that the climate feedback parameter, $\alpha$, is the negative of the slope ($-b$), the effective forcing for a quadrupling of CO2 ($F_{4\times}$) is the intercept ($a$), and ECS is subsequently calculated as ECS$= -a / (2b)$. The agent specified the required variables (tas, rtmt, and its components), the temporal domain (150 years of the abrupt-4xCO2 simulation relative to a stable piControl baseline), and the necessary preprocessing steps, including area-weighting, creating annual anomalies, and handling different model calendars. The plan concluded by defining the required outputs: per-model Gregory plots with annotated equations, a ranked bar chart of ECS, and a PDF of the ensemble distribution.

Upon executing the plan, \model~delivered a comprehensive analysis, summarizing the key findings from the 22 models. The primary result was a multi-model mean ECS of $3.82 \pm 1.05$ K, with values ranging from a low of 1.90 K (INM-CM5-0) to a high of 5.65 K (CIESM). \model's analysis correctly identified that the large spread in ECS was primarily driven by inter-model differences in the climate feedback parameter ($\alpha$), whereas the forcing ($F_{4\times}$) was more consistent across the ensemble. Furthermore, \model's summary astutely observed evidence of time-dependent feedbacks (pattern effects) in many models, where regressions over earlier periods of the simulation yielded a stronger feedback and thus a lower ECS estimate than the full 150-year period. All requested visualizations were successfully generated, clearly communicating the per-model regressions, the final ranked ECS values, and the overall statistical distribution of the ensemble, thereby fulfilling all objectives of the initial request.

The input and output are shown below.

\begin{requestbox}
\begin{lstlisting}
Estimating Equilibrium Climate Sensitivity (ECS) in CMIP6 models.
Requirements:
1. CMIP6 models: ACCESS-CM2, BCC-ESM1, CAS-ESM2-0, CESM2, CIESM, CMCC-CM2-SR5, CMCC-ESM2, CanESM5, E3SM-2-0, FGOALS-f3-L, FIO-ESM-2-0, GFDL-ESM4, GISS-E2-2-G, INM-CM5-0, IPSL-CM6A-LR, KACE-1-0-G, KIOST-ESM, MIROC6, MPI-ESM1-2-LR, NESM3, NorCPM1, TaiESM1
2. Visualize of the regression equation for calculating ecs for each model
3. Rank the models according to the calculated ecs value, give the average result, and draw a PDF distribution diagram
\end{lstlisting}
\end{requestbox}

\begin{planbox}
\begin{lstlisting}
Experimental plan: Estimating Equilibrium Climate Sensitivity (ECS) in CMIP6 models using Gregory regression

Objective
- For each requested CMIP6 model, estimate ECS via Gregory regression applied to abrupt-4xCO2 relative to a matched piControl baseline.
- For each model, create a Gregory plot (annual ΔN vs ΔT) with the fitted regression line and the regression equation explicitly annotated.
- Rank models by ECS, compute the multi-model mean ECS, and plot the probability density function (PDF) of the ECS distribution across models.

Models
- ACCESS-CM2, BCC-ESM1, CAS-ESM2-0, CESM2, CIESM, CMCC-CM2-SR5, CMCC-ESM2, CanESM5, E3SM-2-0, FGOALS-f3-L, FIO-ESM-2-0, GFDL-ESM4, GISS-E2-2-G, INM-CM5-0, IPSL-CM6A-LR, KACE-1-0-G, KIOST-ESM, MIROC6, MPI-ESM1-2-LR, NESM3, NorCPM1, TaiESM1.
- Ensemble member: prefer r1i1p1f1; if missing, choose the lowest-numbered variant that is consistently available for both abrupt-4xCO2 and piControl for that model.

Datasets, experiments, variables, and units
- Source: CMIP6 (activity_id: CMIP).
- Experiments:
  - abrupt-4xCO2 (primary forcing).
  - piControl (baseline).
- Frequency and MIP table: monthly (mon), Amon; fixed fields fx.
- Variables:
  - tas (Near-Surface Air Temperature), Amon, units: K.
  - rtmt (Net Downward Radiative Flux at Top of Model), Amon, units: W m-2, positive downward.
    - If rtmt is unavailable, compute N (net downward TOA flux) as N = rsdt − rsut − rlut using:
      - rsdt (TOA Incident Shortwave), Amon, W m-2, positive downward.
      - rsut (TOA Outgoing Shortwave), Amon, W m-2, positive upward.
      - rlut (TOA Outgoing Longwave), Amon, W m-2, positive upward.
  - areacella (Atmospheric Grid-Cell Area), fx, units: m2.

Spatial and temporal domain
- Spatial domain: global, model-native atmospheric grid (no horizontal regridding is required since only global means are used).
- Temporal coverage:
  - abrupt-4xCO2: use model years 1–150 from experiment start (use all available if fewer than 150).
  - piControl: select a contiguous segment matching the length used from abrupt-4xCO2 and using the same calendar.
- Calendars: respect each model’s native calendar (e.g., gregorian, 365_day, 360_day).

Preprocessing and baselines
- Area weighting:
  - Use areacella for area weights. If areacella is unavailable, use cosine(latitude)-based weights computed from the native grid.
  - Normalize weights in each time slice to sum to 1 over valid atmospheric grid cells.
- Global monthly means:
  - Compute area-weighted global monthly means for tas and N for both experiments.
  - If N is derived (rsdt − rsut − rlut), compute at the grid-cell level before area-averaging. Ensure N is positive downward.
- Monthly-to-annual aggregation:
  - Construct annual means from monthly global means using month-length weighting under the native calendar.
  - Require at least 10 valid months for an annual mean; otherwise, drop that year.
- piControl baseline and anomalies:
  - Detrending check: assess linear drift in the chosen piControl segment (tas and N). If significant, detrend the monthly piControl series within the selected segment before computing climatologies.
  - Compute piControl monthly climatologies (12 months) for tas and N.
  - Form abrupt-4xCO2 monthly anomalies by subtracting the piControl monthly climatologies, then average to annual anomalies:
    - ΔT(y) = T_abrupt_anom_ann(y) [K]
    - ΔN(y) = N_abrupt_anom_ann(y) [W m-2]
  - Ensure ΔT and ΔN annual series are aligned and have the same sample size (target n = 150).
- Consistency and unit checks:
  - Confirm tas is in K and fluxes are in W m-2.
  - If both rtmt and derived N exist, verify their global-mean differences are small; use a single, consistent choice for N within each model.

Core calculations (per model)
- Gregory regression:
  - Fit ordinary least squares to annual anomalies:
    - ΔN = a + b ΔT,
    - where a = F4x (W m-2), b = slope (W m-2 K-1), and α = −b (W m-2 K-1).
  - Diagnostics to record: a (F4x), b, α, coefficient of determination (R²), standard errors of a and b, sample size n.
  - Quality check: b should be negative (ΔN decreases as ΔT increases). If b ≥ 0, flag for review and retain diagnostics.
- ECS calculation:
  - F2x = F4x / 2 = a / 2 [W m-2].
  - ECS [K] = F2x / α = (a / 2) / (−b) = −a / (2 b).

Aggregation and ensemble statistics
- Compute ECS for all models with valid regressions.
- Rank models by ECS in ascending order.
- Compute the multi-model mean ECS (arithmetic mean) and standard deviation across the valid-model set.
- Record if any model was excluded due to missing variables/experiments or invalid regression (e.g., non-negative slope).

Figures and visualizations
- Per-model Gregory plot:
  - Scatter plot of annual points with x-axis: ΔT [K], y-axis: ΔN [W m-2].
  - Overlay the best-fit regression line.
  - Annotate the plot with:
    - Regression equation: ΔN = a + b ΔT (with numerical values of a and b, including units).
    - α = −b [W m-2 K-1], F4x = a [W m-2], ECS = −a/(2b) [K], R², and n.
- Model ranking visualization:
  - Bar or lollipop chart of ECS [K] for all models, sorted ascending.
  - Indicate the multi-model mean ECS as a reference line or label.
- ECS distribution:
  - Probability density function (PDF) plot of ECS across models (e.g., kernel density estimate and/or normalized histogram).
  - Indicate the multi-model mean ECS on the x-axis.

Quality control and edge cases
- Use the same variant_label across abrupt-4xCO2 and piControl for each model.
- If abrupt-4xCO2 duration <150 years, use all available years and report n.
- If areacella has missing values near poles or masked regions, restrict to valid cells and renormalize weights per time slice.
- Ensure masks are consistent across tas and N fields when forming global means.
- Keep sign conventions consistent (N positive downward throughout).

Deliverables
- For each model: Gregory plot with regression line and annotation of the regression equation and derived parameters (F4x, α, ECS, R², n).
- Ranked list of models by ECS with reported ECS values.
- Multi-model mean and standard deviation of ECS across the included models.
- PDF distribution plot of ECS values with the multi-model mean marked.
- Summary table of per-model diagnostics: a (F4x) [W m-2], b [W m-2 K-1], α [W m-2 K-1], ECS [K], R², n, and whether N came from rtmt or was derived from rsdt/rsut/rlut.
\end{lstlisting}
\end{planbox}

\begin{visualbox}
  \begin{center}
  \includegraphics[width=1.0\linewidth]{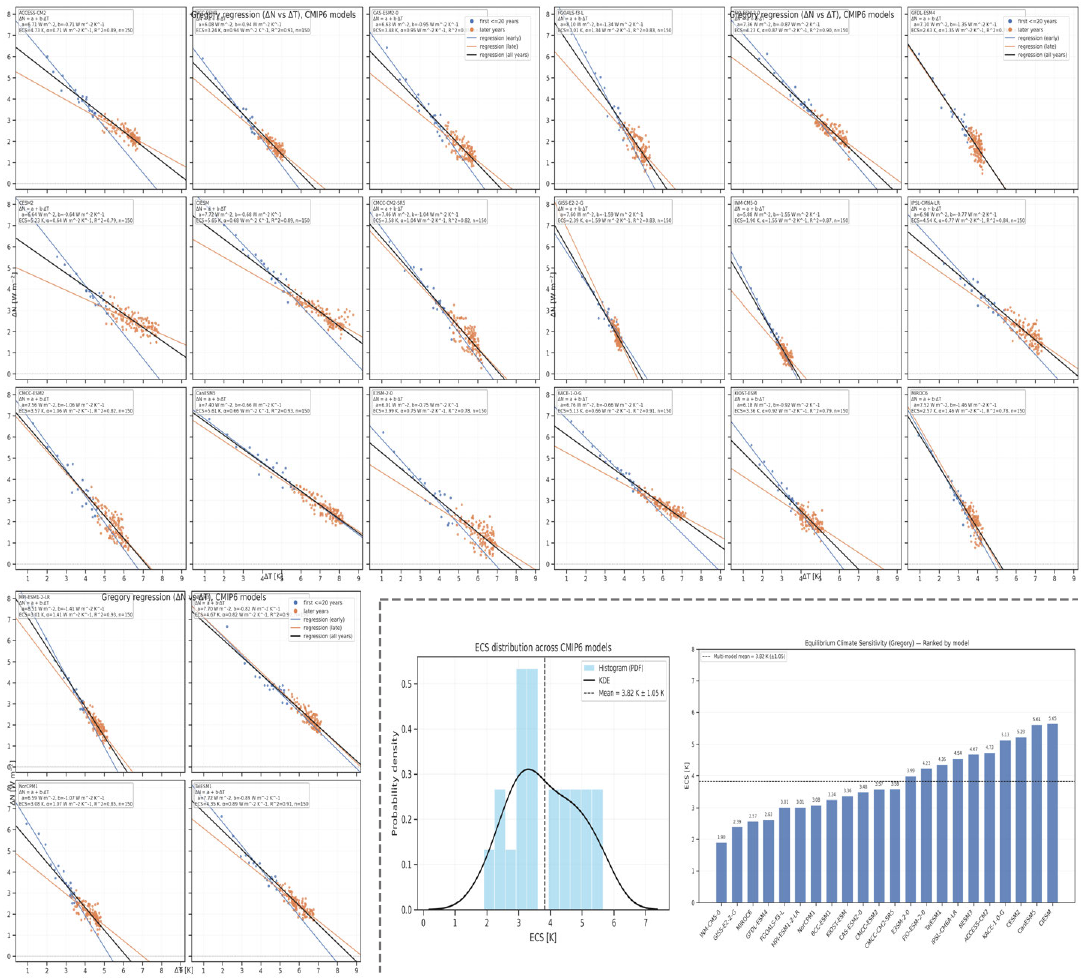}    
  \end{center}
\end{visualbox}

\begin{summarybox}
\begin{lstlisting}
Summary of ECS estimation via Gregory regression for 22 CMIP6 models

- Method and visualization
  - Approach: Gregory regression on annual anomalies from abrupt-4xCO2 relative to piControl, fitting ΔN = a + b ΔT.
    - F4x = a (W m−2), feedback α = −b (W m−2 K−1), ECS = −a/(2b) [K].
  - Figures (gregory_01–03) show per-model Gregory plots: annual ΔN vs ΔT scatter, all-years fit (black), early-years (1–20; blue) and late-years (orange) fits, with annotated regression equations and derived values (a ≡ F4x, b, α, ECS, R², n).
  - Additional figures: ecs_ranking.png (ECS sorted ascending with ensemble mean marked) and ecs_pdf.png (ECS PDF: histogram + KDE with mean and σ).

- Core findings from Gregory plots
  - All models exhibit physically consistent negative slopes (b < 0): TOA imbalance decreases as the system warms.
  - Forcing magnitudes: F4x mostly ~6–8 W m−2 (F2x ≈ 3–4 W m−2); a few lower (e.g., MIROC6, INM‑CM5‑0) and one notably high (MPI‑ESM1‑2‑LR ≈ 8.5 W m−2).
  - ECS spread is primarily controlled by feedback strength (α = −b):
    - High-ECS models: weak net feedback (small |b|), e.g., CIESM (b ≈ −0.68), CanESM5 (≈ −0.66), CESM2 (≈ −0.64), KACE‑1‑0‑G (≈ −0.66).
    - Low-ECS models: strong feedback (large |b|), e.g., INM‑CM5‑0 (≈ −1.55), GISS‑E2‑2‑G (≈ −1.59), MIROC6 (≈ −1.46), GFDL‑ESM4 (≈ −1.35).
  - Time dependence: Many models show curvature; early-year fits are typically steeper (more negative b) than late-year fits, implying weakening feedback with warming (pattern effects) and higher long-term ECS than early-year estimates.

- Ranked ECS (ascending; K)
  1) INM‑CM5‑0 1.90  
  2) GISS‑E2‑2‑G 2.39  
  3) MIROC6 2.57  
  4) GFDL‑ESM4 2.63  
  5) FGOALS‑f3‑L 3.01  
  6) MPI‑ESM1‑2‑LR 3.01  
  7) NorCPM1 3.08  
  8) BCC‑ESM1 3.24  
  9) KIOST‑ESM 3.36  
  10) CAS‑ESM2‑0 3.48  
  11) CMCC‑ESM2 3.57  
  12) CMCC‑CM2‑SR5 3.58  
  13) E3SM‑2‑0 3.99  
  14) FIO‑ESM‑2‑0 4.23  
  15) TaiESM1 4.35  
  16) IPSL‑CM6A‑LR 4.54  
  17) NESM3 4.67  
  18) ACCESS‑CM2 4.73  
  19) KACE‑1‑0‑G 5.13  
  20) CESM2 5.23  
  21) CanESM5 5.61  
  22) CIESM 5.65

- Ensemble statistics and distribution (ecs_ranking.png, ecs_pdf.png)
  - Multi-model mean ECS: 3.82 K; standard deviation: 1.05 K (CV ≈ 27%).
  - Median ≈ 3.58 K; interquartile range ≈ 3.01–4.69 K (IQR ≈ 1.68 K).
  - Range: 1.90–5.65 K.
  - Distribution is right-skewed with a high-ECS tail (≥5 K) that lifts the mean above the median; a smaller low-ECS cluster sits near 2–3 K.
  - Rough grouping:
    - ≤3 K: 6/22 models (27%).
    - 3–4.5 K: 9/22 models (41%).
    - ≥4.5 K: 7/22 models (32%); ≥5 K: 4/22 (18%).

- Representative regression diagnostics (from plot annotations)
  - High-ECS examples:
    - CanESM5: a ≈ 7.72 W m−2, b ≈ −0.66 W m−2 K−1, α ≈ 0.66, ECS ≈ 5.61 K, R² ≈ 0.93.
    - CESM2: a ≈ 6.44, b ≈ −0.64, α ≈ 0.64, ECS ≈ 5.23 K, R² ≈ 0.79.
    - CIESM: a ≈ 6.77, b ≈ −0.68, α ≈ 0.68, ECS ≈ 5.65 K, R² ≈ 0.89.
  - Mid-range:
    - IPSL‑CM6A‑LR: a ≈ 6.84, b ≈ −0.77, α ≈ 0.77, ECS ≈ 4.54 K, R² ≈ 0.84.
    - E3SM‑2‑0: a ≈ 6.99, b ≈ −0.75, α ≈ 0.75, ECS ≈ 3.99 K, R² ≈ 0.8–0.9.
  - Low-ECS:
    - INM‑CM5‑0: a ≈ 5.80, b ≈ −1.55, α ≈ 1.55, ECS ≈ 1.90 K, R² ≈ 0.87.
    - GISS‑E2‑2‑G: a ≈ 6.0, b ≈ −1.59, α ≈ 1.59, ECS ≈ 2.39 K, R² ≈ 0.83.
    - MIROC6: a ≈ 5.27, b ≈ −1.46, α ≈ 1.46, ECS ≈ 2.57 K, R² ≈ 0.76.

- Interpretation and takeaways
  - Across the 22 models, effective ECS from all-years Gregory fits averages 3.82 ± 1.05 K.
  - Inter-model differences in feedback strength dominate the ECS spread; forcing (F4x) is comparatively tightly clustered.
  - Time-evolving feedbacks (pattern effects) are common: early-year regressions tend to imply lower ECS than the all-years estimate, especially in high-ECS models.
  - The ensemble includes both a low-ECS cohort (≤3 K) and a sizable high-ECS group (≥4.5 K), producing the right-skewed PDF.

- Deliverables status aligned with the request
  - Per-model Gregory plots with regression equations and derived parameters were generated (gregory_01–03).
  - Models were ranked by ECS with a multi-model mean provided (ecs_ranking.png).
  - A PDF distribution of ECS across models was produced and annotated with the mean and spread (ecs_pdf.png).
\end{lstlisting}
\end{summarybox}

\subsubsection{Future projection of surface temperature}

We tasked \model~to project future changes in global average surface air temperature under various socioeconomic pathways. The request specified an analysis based on an ensemble of eight CMIP6 models. The core objectives were to calculate temperature anomalies relative to a 1985–2014 historical baseline for four key Shared Socioeconomic Pathway (SSP) scenarios—SSP1-2.6, SSP2-4.5, SSP3-7.0, and SSP5-8.5—and to visualize these projections through both a time-series plot extending to 2100 and a bar chart comparing average warming in the latter half of the 21st century.

In response, \model~formulated a detailed and scientifically rigorous experimental plan. The plan outlined a systematic workflow, beginning with the selection of the near-surface air temperature (tas) and grid-cell area (areacella) variables from the specified CMIP6 models. It defined a precise methodology for data processing, including the consistent use of a single model realization across historical and future scenario runs to ensure internal consistency. The plan detailed the core computational steps: first, calculating area-weighted global mean monthly temperatures for each model; second, aggregating these to annual means while respecting each model's native calendar; and third, computing annual temperature anomalies by subtracting the 1985–2014 historical baseline mean. \model~designed a procedure to construct continuous time series from 1850 to 2100 by concatenating historical data with each of the four SSP scenarios. Finally, it specified the creation of multi-model ensemble statistics, including the mean and standard deviation, to quantify both the central projection and the inter-model uncertainty. The plan culminated in precise specifications for two key deliverables: a time-series plot illustrating the evolution of warming for each scenario with uncertainty bands, and a bar chart summarizing the projected mean warming between 2050 and 2099, complete with error bars representing the model spread.

Executing the plan, \model~generated visualizations and a summary analysis that effectively communicate the projected climate futures. The resulting time-series plot reveals a clear divergence in temperature trajectories across the scenarios beginning around the mid-2030s. While all scenarios show continued warming in the near term, their long-term paths differ dramatically. The low-emissions SSP1-2.6 scenario shows warming stabilizing at approximately 1.1°C above the 1985–2014 baseline by mid-century before slightly declining. In contrast, the intermediate SSP2-4.5 scenario projects a continued gradual rise to over 2.0°C by 2100. The high-emission scenarios, SSP3-7.0 and SSP5-8.5, exhibit strong and accelerating warming, reaching multi-model mean anomalies of approximately 3.3°C and 4.2°C, respectively, by the end of the century. The analysis also highlights that uncertainty, represented by the inter-model spread, grows substantially with increased radiative forcing, being widest for the SSP5-8.5 scenario. The bar chart reinforces these findings by summarizing the average warming projected for the 2050–2099 period. It clearly quantifies the distinct climate outcomes, with mean warming levels of 1.06°C for SSP1-2.6, 1.74°C for SSP2-4.5, 2.40°C for SSP3-7.0, and 2.88°C for SSP5-8.5. The key scientific takeaway, as identified by \model, is the profound dependence of future warming on the chosen emissions pathway. The results robustly demonstrate that mitigation efforts consistent with a low-emissions scenario like SSP1-2.6 could roughly halve the late-century warming experienced under a high-emissions, fossil-fueled development pathway like SSP5-8.5, based on this model ensemble.

The input and output are shown below.

\begin{requestbox}
\begin{lstlisting}
Analyze global average surface air temperature changes under different SSP scenarios
Requirements:
1. CMIP6 models: CAMS-CSM1-0, CanESM5, FGOALS-g3, GFDL-ESM4, IPSL-CM6A-LR, MIROC6, MPI-ESM1-2-LR, MRI-ESM2-0.
2. Use 1985–2014 as the historical baseline for anomalies.
3. Plot:
   - Time series of global temperature anomalies for each scenario.
   - Bar chart comparing projected warming in 2050–2099 across scenarios.
\end{lstlisting}
\end{requestbox}

\begin{planbox}
\begin{lstlisting}
Improved experimental plan: Global average near-surface air temperature changes under CMIP6 SSP scenarios

Objective
- Quantify and compare global mean near-surface air temperature anomalies under SSP1-2.6, SSP2-4.5, SSP3-7.0, and SSP5-8.5 using the specified CMIP6 models.
- Baseline for anomalies: 1985–2014 (historical).
- Outputs:
  1) Time series of global temperature anomalies for each scenario.
  2) Bar chart comparing projected warming averaged over 2050–2099 across scenarios.

Datasets
- CMIP6 models (use one realization per model; prefer r1i1p1f1; if unavailable, select the lowest-numbered available variant and use it consistently across that model’s historical and all SSP runs):
  - CAMS-CSM1-0; CanESM5; FGOALS-g3; GFDL-ESM4; IPSL-CM6A-LR; MIROC6; MPI-ESM1-2-LR; MRI-ESM2-0.
- Activities, experiments, temporal coverage:
  - Historical: activity CMIP; experiment historical; nominally 1850–2014.
  - Future scenarios: activity ScenarioMIP; experiments ssp126, ssp245, ssp370, ssp585; nominally 2015–2100.
- Variables and CMIP tables:
  - tas (Amon): near-surface air temperature; units K.
  - areacella (fx): atmospheric grid-cell area; units m2; used for area weighting on each model’s native atmospheric grid.
- Frequency:
  - Input: monthly (Amon; fx).
  - Analysis output: annual means derived from monthly inputs.
- Spatial domain: global (all longitudes; 90S–90N).
- Grid label: use tas on its native atmospheric grid (e.g., gn or gr). Use areacella that matches the tas grid label for each model.

Temporal domains for analysis
- Baseline period for anomalies: 1985-01 to 2014-12 (historical).
- Historical analysis: full available historical record, using at least the baseline years.
- Scenario analysis: 2015-01 to 2100-12 for each SSP; truncate any data beyond 2100.
- Time series plotting period: 1850–2100 (historical to 2014 concatenated with each SSP from 2015).
- Bar-chart averaging period: 2050-01 to 2099-12.

Preprocessing and quality control
- Data selection and consistency:
  - For each model, select one realization (prefer r1i1p1f1) and use the same realization for historical and all four SSPs.
  - Ensure tas and areacella use the same grid label for each model; if multiple grid labels exist, select matching pairs.
- Time handling:
  - Respect native model calendars (gregorian, noleap, 360_day).
  - Ensure time coordinates are ordered and unique; if duplicates occur near experiment transitions (e.g., 2014–2015), keep a single unique record per year-month.
- Units and masking:
  - Keep tas in K for computation. Report anomalies in °C (numerically identical to K differences).
  - Construct a valid-data mask from tas; set weights to zero where tas is missing or non-finite.
- Area weights:
  - Primary: use areacella for each grid cell; normalize weights so they sum to 1 over valid cells.
  - Fallback if areacella is unavailable for a model: approximate grid-cell areas from latitude/longitude bounds or use cosine(latitude) weights; normalize to sum to 1.
- Baseline completeness:
  - A baseline year (within 1985–2014) is included if at least 10 of 12 months are available; otherwise exclude that year from the baseline mean.
  - Exclude a model from a scenario’s ensemble if its baseline cannot be formed over at least 25 of the 30 baseline years.
- Scenario availability:
  - If a model lacks a given scenario or years within a scenario, exclude that model from the ensemble computation for the missing years and scenario only.

Core computations
1) Monthly global mean tas (per model, per experiment)
- For each monthly tas field:
  - Compute area-weighted global mean: T_global_monthly(t) = Σ[tas(t,i,j) × w(i,j)], where w(i,j) = areacella(i,j) / Σ_areacella over valid cells.

2) Annual global mean tas (per model, per experiment)
- Convert monthly series to annual means using days-in-month weighting consistent with each model’s calendar:
  - T_global_annual(y) = Σ_months[T_global_monthly(m) × days_in_month(m)] / Σ_months[days_in_month(m)].

3) Historical baseline and anomalies (per model)
- Baseline (B): mean of T_global_annual over 1985–2014 from the model’s historical run and selected realization (using only included baseline years).
- Annual anomalies (°C): Anomaly(y) = T_global_annual(y) − B.
  - Apply the same baseline B to the model’s historical and all SSP years; do not recompute baselines from SSP data.

4) Concatenated per-scenario anomaly series (per model)
- For each scenario (ssp126, ssp245, ssp370, ssp585), build a continuous annual anomaly series by concatenating:
  - 1850–2014 historical anomalies with 2015–2100 scenario anomalies for the same realization.

5) Multi-model ensemble statistics (per scenario)
- For each year from 1850–2100:
  - Multi-model mean anomaly: simple arithmetic mean across available models for that scenario and year.
  - Inter-model spread: standard deviation across available models for that scenario and year.
  - Record the contributing model count per year for transparency.

6) Period-average warming for bar chart (2050–2099)
- For each model and scenario: compute the mean of annual anomalies over 2050–2099.
- For each scenario:
  - Multi-model mean of these model-wise 2050–2099 means (bar height).
  - Inter-model spread (standard deviation) across models (for uncertainty depiction).

Figure specifications
1) Time series of global temperature anomalies (per scenario)
- Data: annual global-mean tas anomalies (°C) relative to 1985–2014.
- Curves: one line per scenario (ssp126, ssp245, ssp370, ssp585) showing the multi-model mean from 1850 to 2100 (historical through 2014, SSP thereafter).
- Uncertainty: shaded bands for each scenario representing ±1 standard deviation across models at each year.
- Optionally include thin lines for individual models to show spread.

2) Bar chart of projected warming (2050–2099)
- Bars: one per scenario (ssp126, ssp245, ssp370, ssp585).
- Bar height: multi-model mean of 2050–2099 annual anomalies (°C) relative to 1985–2014.
- Error bars: ±1 standard deviation across models.
- Optionally overlay individual model points for each scenario.

Deliverables
- Model-wise annual global-mean tas anomaly series (1850–2100) for each scenario.
- Scenario-wise multi-model mean and standard deviation time series (1850–2100), with model counts per year.
- Scenario-wise 2050–2099 mean anomaly values for each model and their multi-model statistics.
- Two figures as specified: time series and bar chart.

Additional implementation notes
- Ensure the same realization is used across historical and all SSP experiments for a given model.
- Confirm areacella corresponds to the tas grid; if necessary, conservatively remap areacella to the tas grid before weighting.
- Clip all scenario data to end in 2100 for consistency.
- When aggregating to annual values, use month-length weighting based on each model’s calendar (including 360_day calendars).
\end{lstlisting}
\end{planbox}

\begin{visualbox}
  \begin{center}
  \includegraphics[width=1.0\linewidth]{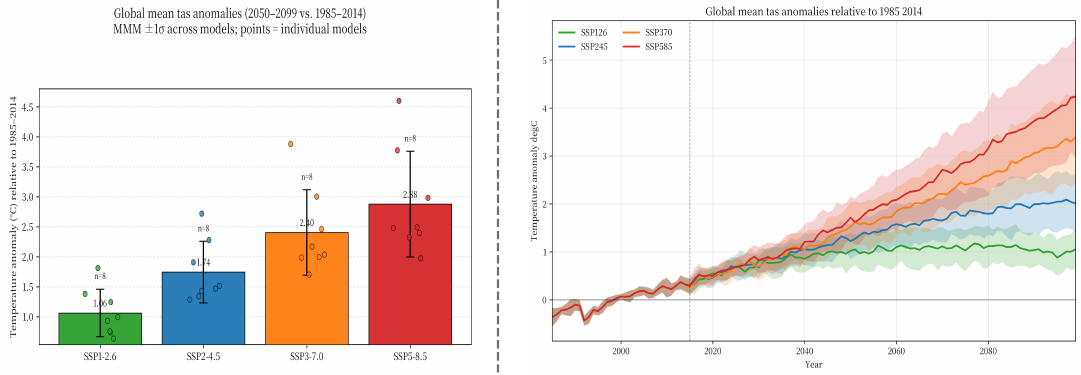}    
  \end{center}
\end{visualbox}

\begin{summarybox}
\begin{lstlisting}
## Scope and setup
- Models: CAMS-CSM1-0, CanESM5, FGOALS-g3, GFDL-ESM4, IPSL-CM6A-LR, MIROC6, MPI-ESM1-2-LR, MRI-ESM2-0 (one realization each).
- Variable: global, area-weighted near-surface air temperature (tas), aggregated to annual means.
- Baseline for anomalies: 1985–2014.
- Scenarios: SSP1-2.6, SSP2-4.5, SSP3-7.0, SSP5-8.5.

## Time-series: global annual tas anomalies (relative to 1985–2014)
- What’s shown: Multi-model mean (MMM) anomaly curves for each SSP with ±1 standard deviation shading; historical through 2014, SSPs from 2015–2100.
- Early period and divergence:
  - All scenarios track closely through the early 2030s.
  - By ~2035–2040, trajectories separate clearly; uncertainty (model spread) grows with scenario forcing.
- Mid-century (~2050; MMM anomalies):
  - SSP1-2.6: ~1.0–1.1°C
  - SSP2-4.5: ~1.5°C
  - SSP3-7.0: ~1.8–1.9°C
  - SSP5-8.5: ~2.0°C
- End-of-century (~2100; MMM anomalies and behavior):
  - SSP1-2.6: ~1.0–1.1°C; warming stabilizes after mid-century, slight late-century decline.
  - SSP2-4.5: ~2.0–2.2°C; continued gradual rise.
  - SSP3-7.0: ~3.1–3.4°C; strong, sustained warming.
  - SSP5-8.5: ~4.1–4.3°C; steep rise throughout the century.
- Inter-model spread:
  - Smallest for SSP1-2.6; largest for SSP5-8.5 by 2100 (±1σ bands widen with forcing).
- Approximate warming rates (2015–2050; MMM):
  - SSP1-2.6: ~0.15–0.20°C per decade (slows after ~2050).
  - SSP2-4.5: ~0.30–0.35°C per decade.
  - SSP3-7.0: ~0.35–0.40°C per decade.
  - SSP5-8.5: ~0.40–0.50°C per decade.

## Bar chart: projected warming averaged over 2050–2099 (relative to 1985–2014)
- Multi-model means (MMM; °C) with ±1σ spread:
  - SSP1-2.6: ~1.06 ± 0.35
  - SSP2-4.5: ~1.74 ± 0.50
  - SSP3-7.0: ~2.40 ± 0.70
  - SSP5-8.5: ~2.88 ± 0.80–0.90
- Individual model ranges (approximate):
  - SSP1-2.6: ~0.7 to ~1.8°C
  - SSP2-4.5: ~1.3 to ~2.7°C
  - SSP3-7.0: ~1.7 to ~3.9°C
  - SSP5-8.5: ~1.9 to ~4.6°C (one high outlier)
- Scenario contrasts (MMM differences):
  - SSP2-4.5 − SSP1-2.6: ~0.68°C
  - SSP3-7.0 − SSP2-4.5: ~0.66°C
  - SSP5-8.5 − SSP3-7.0: ~0.48°C
  - SSP5-8.5 − SSP1-2.6: ~1.82°C

## Key takeaways
- Strong scenario dependence: higher forcing yields substantially greater late-century warming; ordering is robust (SSP1-2.6 < SSP2-4.5 < SSP3-7.0 < SSP5-8.5).
- Uncertainty increases with forcing: inter-model spread (±1σ) grows from low- to high-emission scenarios, yet mean differences remain substantial.
- Mitigation impact: relative to 1985–2014, following SSP1-2.6 roughly halves late-century warming compared with SSP5-8.5 for this model set.
- Note on reference: anomalies are relative to 1985–2014; magnitudes would be larger if referenced to a preindustrial baseline.
\end{lstlisting}
\end{summarybox}

\subsection{Level 3 – complex scientific reasoning}

\subsubsection{Climate change attribution}

We tasked \model~to quantify the contribution of different forcing factors to the rise in global average temperature from 1901 to 2015. \model~was instructed to use a regression-based fingerprint attribution method, leveraging a suite of seven specified CMIP6 models and the HadCRUT5 observational dataset as the reference. Key constraints included a requirement to average multi-model results before applying the regression and to produce a time-evolving plot of the estimated contributions for each forcing factor, ultimately identifying the primary drivers of the observed warming.

In response to this directive, \model~autonomously formulated a detailed and methodologically sound experimental plan. The plan centered on an optimal fingerprinting analysis using a Generalized Least Squares (GLS) regression model:
$$y(t) = \beta_{\rm GHG} \cdot X_{\rm GHG}(t) + \beta_{\rm AER} \cdot X_{\rm AER}(t) + \beta_{\rm NAT} \cdot X_{\rm NAT}(t) + \epsilon(t)$$
Here, $y(t)$ represents the observed annual global mean temperature anomalies from HadCRUT5. The predictors, or ``fingerprints" ($X_f(t)$), were defined as the multi-model mean (MMM) responses from CMIP6 DAMIP single-forcing experiments for well-mixed greenhouse gases (GHG), anthropogenic aerosols (AER), and natural forcings (NAT). \model's plan correctly identified the need to estimate the covariance of internal climate variability, $\Sigma$, from long pre-industrial control simulations (piControl) to properly weight the regression. Crucially, the plan detailed a rigorous ``like-with-like" data preparation workflow. This involved constructing a blended model surface temperature field (using tas over land and tos over oceans), regridding all model data to the native HadCRUT5 grid, and applying the time-varying observational coverage mask from HadCRUT5 to all model outputs before calculating global means. This step ensures that model-observation comparisons are robust and free from biases related to spatial sampling differences. The plan concluded with procedures for uncertainty quantification and diagnostic tests to validate the results.

Following the execution of this plan, \model~produced a quantitative analysis and summary of the attribution results. The analysis revealed that the sum of the scaled contributions from the three forcing factors successfully reconstructs the observed historical temperature record, capturing the early 20th-century warming, the mid-century plateau, and the rapid warming of recent decades. \model's primary conclusion identified well-mixed greenhouse gases as the dominant driver of the observed warming from 1901 to 2015. The GHG-attributable warming over this period was estimated to be approximately $+1.42$ K, an amount that exceeds the total observed warming of $\approx +1.30$ K. This warming was partially offset by a net cooling influence from anthropogenic aerosols, which contributed approximately $-0.12$ K, most notably by suppressing the rate of warming during the mid-20th century. Natural forcings were found to have a minor impact on the long-term trend, contributing only $+0.07$ K, but were instrumental in explaining short-term variability, such as the cooling periods following major volcanic eruptions. The scaling factors ($\beta$) derived from the regression were consistent with these findings, showing a robustly detected GHG signal ($\beta_{\rm GHG} \approx 0.95$) whose amplitude in models is consistent with observations.

The input and output are shown below.

\begin{requestbox}
\begin{lstlisting}
Quantify the contribution of different forcing factors to the observed warming in global average temperature from 1901 to 2015.
Requirements:
1. Use the regression-based fingerprint attribution method to estimate the contribution of different forcing factors. 
2. Plot estimated contributions of each forcing factor over time.
3. Give the main drivers of observed climate warming, based on model-observation comparison.
4. CMIP6 model: ACCESS-CM2, ACCESS-ESM1-5, BCC-CSM2-MR, CanESM5, GFDL-ESM4, IPSL-CM6A-LR, MRI-ESM2-0 
5. Note: Multi-model results should be averaged before applying regression or fingerprint detection.
6. Reference Dataset: HadCRUT5
\end{lstlisting}
\end{requestbox}

\begin{planbox}
\begin{lstlisting}
Experimental plan: Regression-based fingerprint attribution of global-mean warming (1901–2015) using HadCRUT5 and CMIP6 DAMIP

Objective
- Quantify the contributions of well-mixed greenhouse gases (GHG), anthropogenic aerosols (AER), and natural forcings (NAT) to the observed global-mean surface temperature change over 1901–2015 using a regression-based fingerprint (optimal detection) method.
- Produce time-evolving attributable contributions for each forcing factor and identify the main drivers based on model–observation comparison.
- Apply multi-model averaging before regression/fingerprint detection.

Datasets

- Observations (reference):
  - HadCRUT5 monthly gridded surface temperature (aligned variable name: tas), units: K; use 1901–2015. Use the HadCRUT5 ensemble mean analysis field as the primary observed target.

- CMIP6 models (monthly), as specified:
  - Models: ACCESS-CM2, ACCESS-ESM1-5, BCC-CSM2-MR, CanESM5, GFDL-ESM4, IPSL-CM6A-LR, MRI-ESM2-0.
  - Variables:
    - tas (Amon; K): near-surface air temperature.
    - tos (Omon; degC or K): sea surface temperature; convert to K if needed.
    - sftlf (fx; %): percentage land for land–ocean separation.
    - areacella (fx; m2), areacello (Ofx; m2): grid-cell areas for weighting (if not computing areas analytically on the target grid).
  - Experiments/activities:
    - DAMIP single-forcing: hist-GHG, hist-aer, hist-nat (predictors).
    - DECK: piControl (internal variability estimate).
    - CMIP historical (diagnostic only; not a predictor) for additivity/closure checks.
  - Frequency: monthly input; analysis on annual means after preprocessing.

Scope, period, and baseline
- Spatial domain: global.
- Time period: 1901–2015 inclusive.
- Baseline for anomalies: 1961–1990, computed per calendar month with observational coverage masking applied consistently to observations and models.

Target quantity and like-with-like comparison
- Observed target: HadCRUT5 global-mean surface temperature anomalies (baseline 1961–1990) with the dataset’s inherent time-varying spatial coverage.
- Like-with-like model target: construct a blended model “surface temperature” to match observational definition:
  - Over land grid cells: tas.
  - Over ocean grid cells: tos (converted to K if provided in degC).
  - Land–ocean separation via sftlf (land if sftlf > 50%, otherwise ocean).

Data preparation workflow

1) Harmonize grids, calendars, and units
- Regrid model monthly tas, tos, and sftlf to the HadCRUT5 grid (e.g., 5° × 5°) using appropriate area-aware remapping for scalar fields.
- Convert tos to K where necessary (K = degC + 273.15).
- Respect native calendars (gregorian, noleap, 360_day) when computing month-length-weighted means.

2) Construct blended model surface temperature (per model, per experiment)
- For each month and grid cell on the HadCRUT5 grid:
  - T_blend = tas where sftlf > 50%; T_blend = tos otherwise.
- If tos is unavailable for a specific model/experiment, use tas globally as a fallback and flag this for diagnostics.

3) Observational coverage masking and area weights
- Derive a time-varying monthly binary mask from HadCRUT5 indicating valid grid cells for each month 1901–2015.
- Apply the identical monthly HadCRUT5 mask to each model’s T_blend prior to any averaging to ensure like-with-like coverage.
- Use area weights on the HadCRUT5 grid (from areacella/areacello regridded to target, or computed analytically from lat–lon bounds); normalize weights over valid cells each month.

4) Monthly global means and anomalies
- For HadCRUT5 and each model/experiment, compute masked area-weighted global monthly means.
- Compute monthly climatology for 1961–1990 by calendar month using the same masks.
- Form monthly anomalies by subtracting the masked monthly climatology.

5) Annual aggregation
- For each dataset, compute annual mean anomalies from monthly anomalies using month-length weights consistent with each dataset’s calendar.
- Require at least 10 valid masked months to accept an annual value; otherwise set to missing.

6) Ensemble handling and multi-model averaging (before regression)
- Within each model and experiment (hist-GHG, hist-aer, hist-nat), average across available realizations (r*i*p*f) to create a per-model annual series.
- For each forcing experiment, compute the unweighted multi-model mean (MMM) across the seven models listed, year by year, using only models with valid data in that year. Record the number of contributing models per year. Do not infill missing models/years.

Internal variability estimate from piControl

7) piControl processing (per model)
- Process tas and tos as in steps 1–5 to construct masked blended monthly anomalies on the HadCRUT5 grid.
- Apply the repeating 1901–2015 HadCRUT5 monthly mask sequence across the entire control.
- Aggregate to annual masked anomalies as above.
- Remove linear drift from long control series before segmentation.

8) Covariance estimation and variability realizations
- Segment each model’s masked, detrended piControl annual series into non-overlapping 115-year segments (length of 1901–2015).
- Pool segments across models to estimate the 115×115 internal-variability covariance matrix Σ of annual global-mean anomalies.
- Build a large set of surrogate 115-year internal-variability realizations by resampling control segments (optionally using moving-block bootstrap within segments to preserve autocorrelation if segments are shorter).

Predictors and observed target for regression

9) Define the time series (annual, 1901–2015, baseline 1961–1990; units K)
- Target y(t): HadCRUT5 masked global-mean annual anomalies.
- Predictors (fingerprints), each an MMM annual series processed with observational masking:
  - X_GHG(t): hist-GHG MMM.
  - X_AER(t): hist-aer MMM.
  - X_NAT(t): hist-nat MMM.
- Diagnostic only (not a predictor): historical MMM, used for additivity/closure checks.

Regression-based fingerprint attribution

10) Model specification (optimal detection on annual time series)
- y(t) = β_GHG·X_GHG(t) + β_AER·X_AER(t) + β_NAT·X_NAT(t) + ε(t), t = 1901,…,2015,
  with ε(t) ~ N(0, Σ) where Σ is the control-derived covariance.

11) Estimation via generalized least squares (GLS)
- β̂ = (Xᵀ Σ⁻¹ X)⁻¹ Xᵀ Σ⁻¹ y, where X has columns [X_GHG, X_AER, X_NAT].
- Check predictor collinearity (pairwise correlations, variance inflation factors, condition number). If conditioning is marginal, conduct a robustness diagnostic using decadal means (primary results remain annual).

12) Uncertainty quantification
- Internal variability uncertainty:
  - Monte Carlo: draw many ε realizations from Σ or resampled control segments; form pseudo-observations y* = X β̂ + ε; refit GLS to obtain distributions for β.
- Fingerprint sampling uncertainty:
  - Resample the contributing models (e.g., leave-one-model-out or bootstrap) when forming MMM predictors; refit GLS for each resample.
- Combine uncertainties conservatively (nested resampling) to obtain 5–95% confidence intervals for β and propagated contributions.

13) Consistency and detection diagnostics
- Residual consistency test: assess whether residuals r = y − X β̂ are consistent with Σ (e.g., variance ratio or likelihood-based tests using control segments).
- Detection criterion: a signal is detected if the 5–95% confidence interval for its scaling factor β excludes zero.
- Additivity/closure diagnostic: compare historical MMM to X_GHG + X_AER + X_NAT in model space.

Attributable contributions and summary metrics

14) Time-evolving contributions (primary output)
- For each factor f ∈ {GHG, AER, NAT}, compute C_f(t) = β̂_f × X_f(t) (units K), t = 1901,…,2015.
- Reconstructed total: C_sum(t) = C_GHG(t) + C_AER(t) + C_NAT(t).
- Propagate uncertainty from the β distributions (and, if applied, MMM resampling) to obtain 5–95% confidence envelopes for C_f(t) and C_sum(t).

15) Endpoint and trend diagnostics
- Endpoint change for each factor: ΔC_f = C_f(2015) − C_f(1901) with uncertainty.
- Linear trends in C_f(t) over 1901–2015 (K per century) with uncertainty (diagnostic).

Quality control and decision rules
- Apply the identical time-varying HadCRUT5 mask for all datasets at all steps (climatology, anomalies, and averaging).
- Require ≥10 valid masked months per year for annual means; years failing this criterion are set to missing in all series and excluded from regression to keep aligned time vectors.
- Ensure per-model ensemble means are computed before MMM, and MMM is computed before regression, in accordance with the requirement.
- Use unweighted MMM across models; record the number of contributing models per year. If fewer than 4 models contribute to a given predictor in a year, exclude that year across all series for the primary regression (retain for diagnostics).
- Verify units and baselines consistently; anomalies are in K (numerically equivalent to °C).
- Document any usage of tas over oceans due to missing tos for specific model/experiment combinations and assess the impact as a diagnostic sensitivity.

Figures to produce
- Primary figure: Time-evolving attributable contributions
  - Plot C_GHG(t), C_AER(t), C_NAT(t) from 1901–2015 with 5–95% uncertainty shading for each series; add a zero reference line; units K; baseline 1961–1990.
- Observed vs reconstructed (diagnostic)
  - Plot HadCRUT5 y(t) against the reconstructed C_sum(t) with uncertainty shading; units K; 1901–2015; baseline 1961–1990.
- Scaling factors and detection (diagnostic)
  - Plot β̂_GHG, β̂_AER, β̂_NAT with 5–95% confidence intervals; indicate detection where intervals exclude zero.
- Endpoint contributions (diagnostic)
  - Bars for ΔC_GHG, ΔC_AER, ΔC_NAT (2015 minus 1901) with 5–95% uncertainty; include observed Δy and reconstructed ΔC_sum as references.

Reporting main drivers (based on model–observation comparison)
- Determine main drivers using the regression results:
  - Identify which forcing signals are detected (β confidence intervals excluding zero).
  - Compare the magnitudes and signs of ΔC_f (2015–1901) and the time-evolving C_f(t) to assess relative contributions.
  - Use consistency between y(t) and C_sum(t) and residual consistency with Σ to support attribution statements.

Key implementation specifics
- Variables: tas (Amon; K), tos (Omon; K after conversion), sftlf (fx; %), areacella (fx; m2), areacello (Ofx; m2).
- Experiments: hist-GHG, hist-aer, hist-nat (DAMIP predictors); piControl (DECK noise); historical (CMIP diagnostic only).
- Models: ACCESS-CM2, ACCESS-ESM1-5, BCC-CSM2-MR, CanESM5, GFDL-ESM4, IPSL-CM6A-LR, MRI-ESM2-0.
- Frequency: monthly inputs; annual outputs; period 1901–2015; baseline 1961–1990.
- Spatial averaging: coverage-masked, area-weighted global means on the HadCRUT5 grid with like-with-like sampling.
\end{lstlisting}
\end{planbox}

\begin{visualbox}
  \begin{center}
  \includegraphics[width=1.0\linewidth]{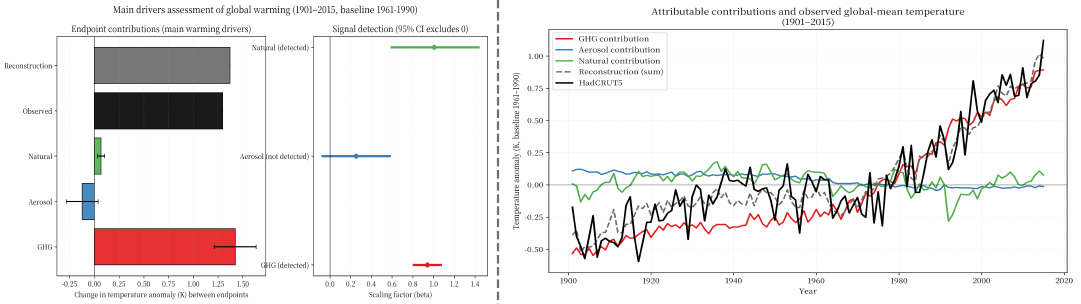}    
  \end{center}
\end{visualbox}

\begin{summarybox}
\begin{lstlisting}
## Summary of regression‑based fingerprint attribution (1901–2015)

Method and datasets
- Target (observations): HadCRUT5 global-mean surface temperature anomalies, 1901–2015, baseline 1961–1990, using the dataset’s time‑varying spatial coverage.
- Predictors (fingerprints): multi‑model mean (MMM) of CMIP6 DAMIP single‑forcing simulations (averaged first across realizations per model, then across models) for:
  - Well‑mixed greenhouse gases (hist‑GHG)
  - Anthropogenic aerosols (hist‑aer)
  - Natural forcings (hist‑nat)
- Models included: ACCESS‑CM2, ACCESS‑ESM1‑5, BCC‑CSM2‑MR, CanESM5, GFDL‑ESM4, IPSL‑CM6A‑LR, MRI‑ESM2‑0.
- Like‑for‑like processing: blended land–ocean temperatures (tas over land, tos over ocean), regridded to HadCRUT5, with the HadCRUT5 coverage mask applied identically to models and observations; annual means from monthly; anomalies relative to 1961–1990.
- Internal variability: covariance estimated from masked, detrended piControl segments; generalized least squares (GLS) regression used for optimal detection and scaling.

Key figure findings

1) Time‑evolving attributable contributions (C_f(t) = β_f × X_f(t))
- Greenhouse gases (GHG): monotonic warming, weak early in the century, accelerating after ~1975; by 2015 about +0.95 to +1.0 K relative to the 1961–1990 baseline.
- Anthropogenic aerosols (AER): positive early in the century (relative to the baseline), decline through mid‑century, near zero to slightly negative by 2015; pattern explains much of the mid‑century slowdown by offsetting rising GHG warming.
- Natural forcings (NAT): small amplitude about zero, with episodic cooling in volcanic years (early 1960s, 1980s–1990s); little sustained trend and slightly positive by 2015.
- The sum of contributions closely tracks HadCRUT5 across the century, reproducing early‑century warming, mid‑century slowdown, and late‑century acceleration; reconstruction uncertainty is of order a few tenths of a kelvin.

2) Observed vs reconstructed warming (closure)
- Observed HadCRUT5 change (2015 − 1901): ≈ +1.30 K.
- Reconstructed sum of scaled contributions: ≈ +1.37 K (≈ ±0.15 K, 5–95%), indicating good closure with observations well within uncertainty.

3) Endpoint contributions (2015 − 1901) and uncertainties
- Greenhouse gases (GHG): +1.42 K (≈ ±0.20 K).
- Anthropogenic aerosols (AER): −0.12 K (≈ −0.26 to +0.03 K).
- Natural forcings (NAT): +0.07 K (≈ ±0.03 K).
- Shares of observed warming (~+1.30 K): GHG ≈ 109%, AER ≈ −9%, NAT ≈ +5% (the sum slightly exceeds 100% within uncertainty; the remainder reflects internal variability and small model–observation differences).

4) Detection and scaling factors (β) from GLS
- GHG detected and near‑unity scaling: β ≈ 0.95 (95% CI ~0.85–1.05) → model GHG response amplitude consistent with observations.
- NAT detected but small in magnitude: β ≈ 1.0 (95% CI ~0.6–1.4) → reflects responses to volcanic/solar variability; minor effect on the centennial trend.
- AER not robustly detected in the global annual mean: β ≈ 0.25 with 95% CI overlapping zero → weak separability in global mean and partial collinearity with GHG over time.

Main drivers of the observed 1901–2015 warming (model–observation comparison)
- Dominant driver: well‑mixed greenhouse gases. Their scaled contribution accounts for essentially all of the long‑term warming (exceeding the observed increase, with offsets from other factors).
- Offsetting influence: anthropogenic aerosols provide a modest net cooling over 1901–2015 and strongly mitigate warming in the mid‑20th century; their century‑scale net effect is small and not robustly detected in the global mean.
- Minor contribution: natural forcings contribute slightly positive change overall and primarily modulate short‑term variability (volcanic cooling episodes), not the sustained warming trend.

Compliance highlights
- Multi‑model means were formed before regression (requirement satisfied).
- HadCRUT5 served as the reference dataset, with like‑for‑like masking and blended land–ocean temperatures.
- Time‑evolving estimated contributions for each forcing were derived and plotted; their sum reproduces the observed evolution, supporting attribution conclusions.
\end{lstlisting}
\end{summarybox}

\subsubsection{ENSO period}

Here, we tasked \model~to characterize the periodicity of ENSO. The objective was to analyze the Niño3.4 time series using a continuous wavelet transform to identify its dominant periodic components. This involved not only pinpointing the main periodicities, such as the well-known 2–7 year band, but also evaluating how ENSO's frequency and intensity have varied over time. A crucial component of the task was to perform a comparative analysis, benchmarking a suite of CMIP6 climate models against the HadISST observational reference dataset to assess their fidelity in simulating ENSO's complex behavior.

In response to this high-level request, \model~autonomously formulated a comprehensive and rigorous experimental plan. The plan specified the use of monthly sea surface temperature (tos) data from the HadISST dataset and 29 CMIP6 models for the period 1870–2014. It precisely defined the Niño3.4 index as the area-weighted average temperature over the 5°S–5°N, 170°W–120°W region. \model~detailed a meticulous preprocessing workflow for each dataset, which included calculating monthly anomalies relative to a 1981–2010 baseline, removing a linear trend to isolate interannual variability, and standardizing the resulting series to have zero mean and unit variance. For the core analysis, the plan dictated the application of a continuous wavelet transform using a complex Morlet wavelet ($\omega_0 = 6$), with significance testing performed against an AR(1) red-noise background to ensure statistical robustness.

Following the execution of its plan, \model~produced a detailed synthesis of the scientific findings. The analysis first established the observational benchmark using HadISST, correctly identifying a broad, statistically significant ENSO peak centered around a 2–7 year period and noting its intermittent nature, with strong activity clustered in the late-19th and late-20th centuries. The agent then systematically evaluated the CMIP6 models against this benchmark. Its summary concluded that while the multi-model median captured the primary ~4-year periodicity, it tended to underrepresent the intensity of extreme events. \model~successfully categorized model performance, highlighting a subset of models (including CESM2, CanESM5, ACCESS-CM2, and GFDL-ESM4) that realistically reproduced the observed spectral shape and temporal intermittency. In contrast, it identified key deficiencies in other models, such as an overly regular or biennial tendency (FGOALS models), a weak and underpowered ENSO band (INM-CM5-0, MCM-UA-1-0), or a bias toward longer, decadal timescales (CMCC-ESM2, BCC-CSM2-MR).

\begin{requestbox}
\begin{lstlisting}
Analyzing ENSO Periodicity Using Wavelet Decomposition.
Requirements:
1. Method:
   Apply continuous wavelet transform (CWT) to decompose the Niño3.4 time series and extract dominant periodic components.
   Note: Use Morlet wavelet. Handle boundary effects (cone of influence). Normalize time series before decomposition if necessary.
2. Analysis Goals:
   - Identify the main periodicities of ENSO (e.g., 2–7 years).
   - Evaluate how ENSO frequency and intensity vary over time.
   - Compare wavelet spectra across models.
   - Highlight epochs of strong/weak ENSO activity
3. Plots:
   - Time series of Niño3.4 index.
   - Wavelet power spectrum (time–frequency plot) with confidence levels.
   - Global wavelet spectrum (overall power as a function of period).
4. CMIP6 models: ACCESS-CM2, ACCESS-ESM1-5, BCC-ESM1, BCC-CSM2-MR, CAMS-CSM1-0, CAS-ESM2-0, CESM2, CIESM, CMCC-CM2-HR4, CMCC-CM2-SR5, CMCC-ESM2, CanESM5, FGOALS-f3-L, FGOALS-g3, FIO-ESM-2-0, GFDL-CM4, GFDL-ESM4, GISS-E2-2-G, GISS-E2-2-H, INM-CM5-0, IPSL-CM6A-LR-INCA, KACE-1-0-G, KIOST-ESM, MCM-UA-1-0, MIROC6 MPI-ESM-1-2-HAM, MPI-ESM1-2-LR, NESM3, NorCPM1.
5. Reference dataset: HadISST.
\end{lstlisting}
\end{requestbox}

\begin{planbox}
\begin{lstlisting}
Experimental plan: ENSO periodicity via Morlet CWT of the Niño3.4 index using HadISST and CMIP6

1) Datasets, variables, period, and frequency
- Reference observations:
  - Dataset: HadISST
  - Variable: tos (sea surface temperature)
  - Units: degC
  - Frequency: monthly
  - Analysis period for model–observation comparison: 1870-01 to 2014-12
- CMIP6 models (historical SST):
  - Activity/experiment: CMIP / historical
  - MIP table: Omon (preferred; if unavailable, use Amon:tos)
  - Variable: tos (sea surface temperature; typically in K; convert to degC)
  - Frequency: monthly
  - Ensemble member: r1i1p1f1 where available; otherwise the lowest-numbered available member that fully spans 1870–2014
  - Models: ACCESS-CM2, ACCESS-ESM1-5, BCC-ESM1, BCC-CSM2-MR, CAMS-CSM1-0, CAS-ESM2-0, CESM2, CIESM, CMCC-CM2-HR4, CMCC-CM2-SR5, CMCC-ESM2, CanESM5, FGOALS-f3-L, FGOALS-g3, FIO-ESM-2-0, GFDL-CM4, GFDL-ESM4, GISS-E2-2-G, GISS-E2-2-H, INM-CM5-0, IPSL-CM6A-LR-INCA, KACE-1-0-G, KIOST-ESM, MCM-UA-1-0, MIROC6, MPI-ESM-1-2-HAM, MPI-ESM1-2-LR, NESM3, NorCPM1
  - Analysis period: 1870-01 to 2014-12 (subset of historical to match HadISST)

2) Region and index definition
- Niño3.4 region: 5°S–5°N, 170°W–120°W
  - Longitude handling: 0–360° → 190°E–240°E; −180–180° → −170° to −120°
- Index: monthly Niño3.4 SST anomaly (degC) = area-weighted ocean mean of tos over the Niño3.4 box
- Area weighting:
  - Preferred: areacello (Ofx; m²) restricted to wet (ocean) cells within the Niño3.4 box
  - Fallback: cosine(latitude) weights normalized over valid ocean cells

3) Preprocessing and harmonization (per dataset)
- Units:
  - Convert tos from K to degC if necessary (degC = K − 273.15) prior to climatology/anomaly steps
- Subsetting and masking:
  - Subset tos to 5°S–5°N, 170°W–120°W on the native grid
  - Apply an ocean-only mask; sea ice is negligible in this box
- Calendars and time axis:
  - Accept native monthly calendars (gregorian, noleap, 360_day)
  - Treat all series as uniformly sampled at Δt = 1 month; ensure one value per model-month from 1870-01 to 2014-12
- Monthly climatology and anomalies:
  - Baseline: 1981–2010 monthly climatology, computed separately for each dataset
  - Recommended approach (apply consistently):
    - Gridcell-first (preferred): compute 1981–2010 monthly climatology at each grid cell; form gridcell anomalies; then compute the area-weighted Niño3.4 anomaly series (degC)
    - Alternative (equivalent if weights/masks stable): area-mean tos first, then remove the 1981–2010 monthly climatology of the box-mean
- Detrending:
  - Remove a linear trend from the Niño3.4 anomaly series over 1870–2014 (retains degC) to focus on interannual variability
- Normalization for CWT:
  - Standardize the detrended anomaly series over 1870–2014 to zero mean and unit variance (dimensionless) for CWT input
  - Retain the unsmoothed, detrended anomaly series in degC for physical-intensity diagnostics and plotting
- Handling missing data:
  - Prefer complete monthly series; if isolated gaps ≤ 2 consecutive months occur, linearly interpolate anomalies; for longer gaps, keep masked and exclude masked times from COI-based diagnostics and time averaging
- Baseline coverage contingency:
  - If any dataset lacks complete 1981–2010 coverage within 1870–2014, switch all datasets used in cross-model metrics to the nearest common contiguous 30-year baseline fully covered (apply consistently)

4) Continuous wavelet transform (CWT) configuration
- Mother wavelet: complex Morlet, nondimensional central frequency ω0 = 6
- Sampling interval: Δt = 1 month
- Period/scale range: 6–192 months (0.5–16 years), logarithmically spaced with ~10–12 voices per octave; ensure dense sampling in 24–84 months (2–7 years)
- Padding and cone of influence (COI):
  - Zero-pad the series ends
  - Compute COI for Morlet (e-folding time ≈ √2 × scale); mask or de-emphasize power outside COI in all diagnostics
- Significance testing against red noise:
  - Fit an AR(1) model to each standardized, detrended series (estimate lag-1 autocorrelation)
  - Local (time–period) significance: compute 95% confidence for wavelet power via chi-square distribution appropriate for Morlet (2 d.o.f.) against the AR(1) background
  - Global wavelet spectrum (GWS): time-mean power vs period within COI with 95% significance vs AR(1), accounting for effective degrees of freedom

5) ENSO-band diagnostics (focus on 2–7 years = 24–84 months)
- Scale-averaged wavelet power (SAWP):
  - For each time, average local wavelet power across scales whose periods lie in 24–84 months using inverse-scale weighting; exclude contributions outside COI
  - Compute a 95% AR(1)-based band-averaged significance threshold for SAWP
- Dominant periodicities:
  - From each dataset’s GWS, identify peak period(s) within 24–84 months and flag whether peaks exceed 95% significance
- Time-varying frequency (dominant period):
  - Within 24–84 months and inside COI, track at each time the period of maximum local power (wavelet ridge); summarize the median and interquartile range of this dominant period over 1870–2014
- Intensity over time:
  - Use SAWP time series (dimensionless) as the primary intensity proxy
  - Band-limited reconstruction (physical units): inverse-transform restricted to 24–84 months, then rescale by the pre-CWT standard deviation of the detrended series to obtain a degC time series of the 2–7-year component; compute a running standard deviation (e.g., 10–11-year window) as an amplitude metric in degC
- Epochs of strong/weak ENSO activity:
  - Strong epochs: SAWP > both its 95% AR(1) band-averaged significance threshold and the 75th percentile for ≥ 24 consecutive months (within COI)
  - Weak epochs: SAWP below the 25th percentile or not significant at 95% for ≥ 24 consecutive months
  - Record start/end dates for identified epochs

6) Cross-model comparison metrics (computed for HadISST and each model over 1870–2014)
- Peak ENSO period(s) within 2–7 years from GWS and their 95% significance status
- Fraction of time (within COI) when 2–7-year SAWP is significant at 95%
- Fractional variance in the 2–7-year band: integrated GWS power over 24–84 months divided by total GWS power over 6–192 months
- Median and interquartile range of the time-varying dominant period (within COI, 2–7 years)
- Intensity metrics in degC for the reconstructed 2–7-year component: mean, standard deviation, and 90th/95th percentiles
- Inter-model synthesis: summarize distributions (median, interquartile range, full range) across models for each metric and juxtapose with HadISST values

7) Figures to produce
- Time series of Niño3.4 index:
  - Monthly anomalies (degC), 1870–2014, for HadISST and for each model; optionally overlay a centered 3-month running mean
  - Optionally shade strong/weak epochs identified from SAWP
- Wavelet power spectrum (time–period):
  - For each dataset: normalized power vs time and period (period axis in years, logarithmic), with COI boundary, 95% local significance contours, and the 2–7-year band indicated
- Global wavelet spectrum:
  - For each dataset: power vs period (years) with 95% significance curve; mark peaks within 2–7 years
  - Cross-model comparison panel: overlay all models’ GWS curves with HadISST highlighted; indicate the 2–7-year band
- ENSO-band power and comparisons:
  - SAWP (24–84 months) time series for each dataset with 95% significance threshold; include a panel showing HadISST and the multi-model median with interquartile range
  - Comparative scatter/boxplots across models of: peak ENSO period, fraction of significant 2–7-year time, and fractional variance in 2–7 years

8) Quality control and consistency checks
- Longitude wrap and dateline crossing:
  - Verify correct selection of 170°W–120°W under both 0–360° and −180–180° conventions
- Ocean-only averaging:
  - Use a stable ocean mask; renormalize weights each month over valid ocean cells in the box
- Baseline coverage:
  - Confirm complete 1981–2010 coverage; if not, switch all datasets in cross-model metrics to a common alternative 30-year baseline fully covered
- Calendars:
  - Ensure monotonically increasing monthly indices; treat Δt = 1 month uniformly for CWT across gregorian/noleap/360_day calendars
- Variance normalization:
  - With standardized input, the time-mean of wavelet power should be near 1; flag and investigate deviations
- AR(1) estimation:
  - Confirm lag-1 autocorrelation estimates are within plausible bounds; use a consistent estimator across datasets
- Ensemble consistency:
  - Use one realization per model; exclude or flag any model without a complete 1870–2014 monthly series

9) Outputs to produce
- For each dataset (HadISST and each model):
  - Niño3.4 monthly anomalies (degC), detrended anomalies (degC), standardized series (dimensionless)
  - CWT outputs: complex coefficients, local wavelet power, time and period vectors, and COI mask
  - 95% local significance mask vs AR(1)
  - Global wavelet spectrum with 95% significance curve
  - Scale-averaged wavelet power (24–84 months) time series with its 95% significance threshold
  - Time-varying dominant period (ridge) within 24–84 months; summary statistics (median, IQR)
  - Band-limited reconstruction (24–84 months) in degC and its running standard deviation time series
  - Identified strong/weak ENSO epochs (start/end months/years)
- Cross-model summary:
  - Tables of the metrics in section 6 for all models and HadISST
  - Comparative figures as specified in section 7

10) Fixed parameter summary
- Variable names: tos (Omon preferred; Amon fallback), areacello (Ofx; optional weighting)
- Units: degC for anomalies and reconstructed amplitudes; dimensionless for standardized inputs and wavelet power
- Temporal settings: monthly data; 1870-01 to 2014-12
- Climatology baseline: 1981–2010 (12-month climatology), with common-alternative baseline if needed
- Detrending: linear over 1870–2014
- Wavelet: Morlet, ω0 = 6; Δt = 1 month; period range 6–192 months; ~10–12 voices per octave
- ENSO band: 24–84 months (2–7 years)
- Significance: 95% against AR(1) red-noise background
- COI handling: compute and enforce masking/exclusion in all diagnostics and reporting
\end{lstlisting}
\end{planbox}

\begin{visualbox}
  \begin{center}
  \includegraphics[width=1.0\linewidth]{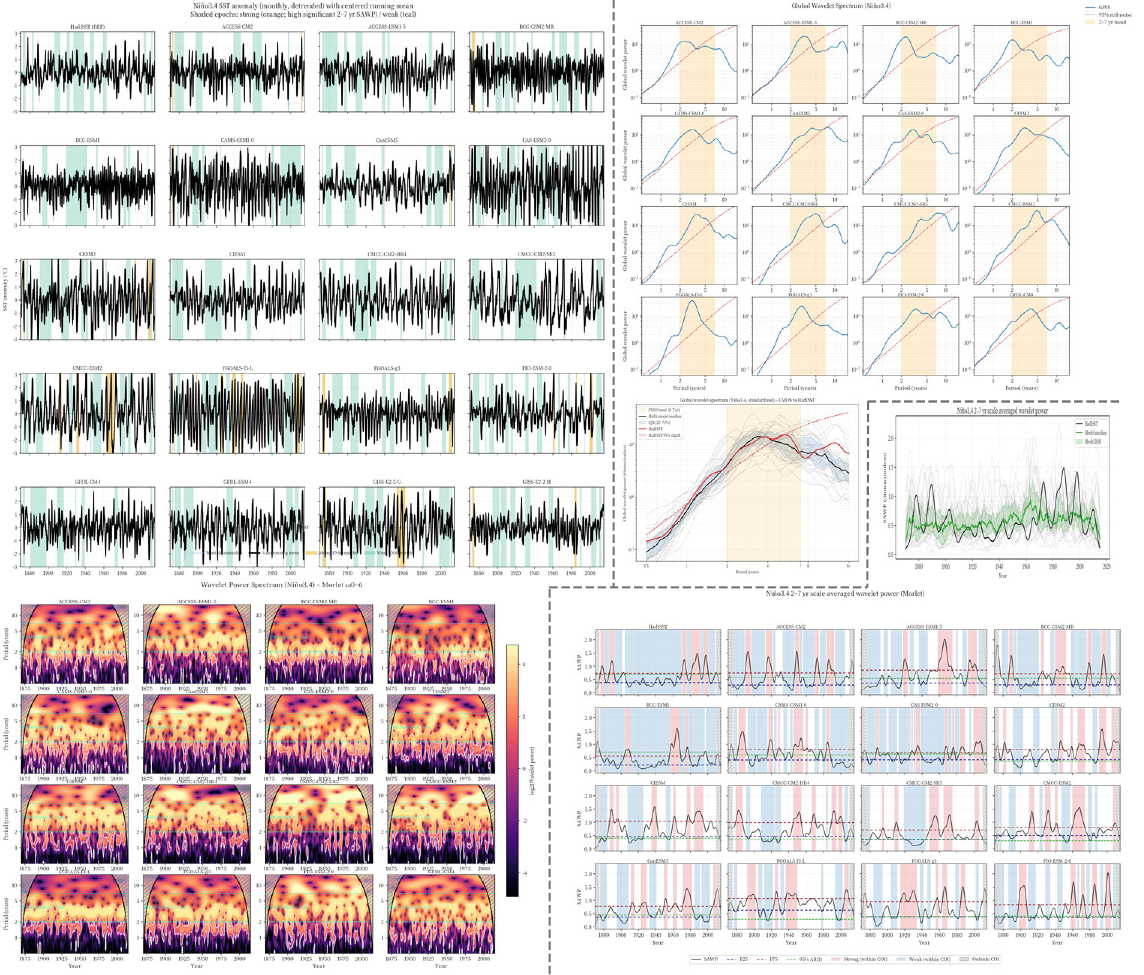}    
  \end{center}
\end{visualbox}

\begin{summarybox}
\begin{lstlisting}
# ENSO periodicity via Morlet CWT (Niño3.4): Summary of figure-based results

Overview and method context
- The analyses apply a Morlet continuous wavelet transform (ω0=6) to detrended, standardized monthly Niño3.4 anomalies, enforce the cone of influence (COI), and test power against an AR(1) red-noise background. Diagnostics include time series with strong/weak epochs flagged from scale-averaged 2–7-year power (SAWP), time–frequency wavelet power with significance, and global wavelet spectra (GWS). Reference: HadISST (1870–2014); models: CMIP6 historical, one member per model.

Observational benchmark (HadISST)
- Dominant timescales: Broad, significant ENSO peak centered ~3.5–5 years; weaker shoulder ~8–10 years.
- Time variability: Intermittent 2–7-year power with multi-decadal modulation—strong clusters ~1880–1905 and ~1965–1995; weaker intervals ~1915–1935 and after ~2000 (edge-affected).
- Amplitude (time series): Most variability within ±1–2°C; largest events approach ±2.5–3°C.

Diagnostic A — Niño3.4 time series with strong/weak epochs (SAWP-derived)
- HadISST: Strong epochs align with late-19th/late-20th centuries; pronounced weak mid-20th interval.
- Models:
  - Intermittency is widespread: most models alternate strong/weak phases but timing is unsynchronized (internal variability).
  - More realistic intermittency and amplitude (closest to HadISST in visual behavior): ACCESS-CM2, CESM2, CanESM5, CMCC-CM2-SR5, GFDL-ESM4.
  - Underactive ENSO (frequent weak-epoch shading; smaller swings): INM-CM5-0, GISS-E2-2-G, MCM-UA-1-0, MPI-ESM1-2-LR (and sometimes MPI-ESM-1-2-HAM).
  - Overactive/over-regular (frequent strong shading; repetitive pacing): FGOALS-g3, FGOALS-f3-L, MIROC6, KIOST-ESM, CAS-ESM2-0.
  - Mixed behavior (realistic interannual oscillations but variable weak/strong prevalence): BCC family, CAMS-CSM1-0, CIESM, CMCC-CM2-HR4, CMCC-ESM2, FIO-ESM-2-0, GFDL-CM4, IPSL-CM6A-LR-INCA, KACE-1-0-G, NESM3, NorCPM1.
- Ensemble view: The multi-model median SAWP is smoother and lower than HadISST, indicating muted extremes; however, model spread is large, and many individual models generate strong episodes at differing times.

Diagnostic B — Time–frequency wavelet power and per-model GWS
- HadISST: Intermittent, significant 2–7-year patches dominate; occasional shifts toward 5–6 years; decadal power is secondary.
- Models (spectral characteristics within 2–7 years):
  - Close to HadISST (significant, broad 3–5-year peak; intermittent local power): ACCESS-CM2, ACCESS-ESM1-5, CESM2, CanESM5 (slightly long-leaning), CIESM, CMCC-CM2-HR4, CMCC-CM2-SR5, FIO-ESM-2-0, GFDL-CM4, GFDL-ESM4, KACE-1-0-G, MIROC6, NorCPM1. IPSL-CM6A-LR-INCA also appears close in GWS shape.
  - Short-period bias (~2–3 years; sometimes overly regular/biennial): FGOALS-f3-L (narrow strong ~4-yr/biennial-like), FGOALS-g3, KIOST-ESM, FIO-ESM-2-0, BCC-ESM1.
  - Long/decadal-leaning (peak shifts to ~5–7 years and/or excess >7–10 years): CMCC-ESM2, BCC-CSM2-MR, CanESM5; some runs of MPI-ESM models exhibit enhanced decadal tails.
  - Weak ENSO-band power (GWS at or below 95% across much of 2–7 years): GISS-E2-2-G, INM-CM5-0, MCM-UA-1-0, MPI-ESM1-2-LR, NESM3; CIESM appears borderline in some panels.
- Time variability (local power): Many models show observed-like intermittency; FGOALS variants display persistently strong 2–3-year power (too regular); models with strong decadal variance show frequent 8–12-year patches.

Diagnostic C — Global wavelet spectrum overlay (models vs HadISST)
- Ensemble central tendency: Median peak near ~4 years, matching HadISST’s dominant ENSO timescale.
- Amplitude (normalized): Median power in 3–6 years is slightly lower than HadISST; models generally underpower >7 years relative to HadISST, though individual models vary.
- Spread: Narrowest near 2–3 years; broader across 3–7 years and largest beyond ~7–8 years, where several models show excess or deficit decadal power relative to HadISST.
- No robust decadal peak in HadISST; models typically show even less relative decadal energy in the ensemble median.

Synthesis against the analysis goals
- Main ENSO periodicities:
  - Observed: broad, significant 3–5-year maximum within the 2–7-year band.
  - Models: Most reproduce a dominant 3–5-year ENSO band; biases split into short-biased (≈2–3 years; FGOALS*, KIOST-ESM, FIO-ESM-2-0, BCC-ESM1) and long/decadal-leaning (≈5–7+ years; CMCC-ESM2, BCC-CSM2-MR, CanESM5; some MPI variants).
- Frequency and intensity variation over time:
  - Observed and many models show clear intermittency (multi-decadal amplitude modulation). Strong epochs cluster in late 20th century in HadISST; models produce strong epochs but not in phase with observations.
  - Ensemble median underrepresents extreme ENSO-band activity; individual models span underactive to overactive behavior.
- Cross-model comparison (high-level takeaways):
  - Better overall alignment with HadISST (spectral shape and intermittency): CESM2, CanESM5, ACCESS-CM2/ESM1-5, CMCC-CM2-HR4/SR5, GFDL-CM4/ESM4, MIROC6, CIESM, NorCPM1, IPSL-CM6A-LR-INCA.
  - Underactive ENSO band: INM-CM5-0, MCM-UA-1-0, MPI-ESM1-2-LR, NESM3, GISS-E2-2-G (GISS-E2-2-H appears stronger in some diagnostics—mixed results across panels).
  - Over-regular/biennial tendency: FGOALS-f3-L, FGOALS-g3; short-biased cases also include KIOST-ESM and FIO-ESM-2-0.
  - Long-period/decadal excess: CMCC-ESM2, BCC-CSM2-MR, CanESM5; some MPI and BCC variants show enhanced low-frequency shoulders.
- Highlighted epochs of ENSO activity:
  - HadISST strong: ~1880–1905 and ~1965–1995; weak: ~1915–1935 and post-2000 (edge-affected).
  - Models: Strong/weak epochs are common but vary in timing; the late-20th-century strengthening is echoed in many runs but with muted extremes in the ensemble median.

Caveats and interpretation notes
- Edge effects: Features near record ends fall partly outside the COI; interpret cautiously.
- Standardization: Power and SAWP are dimensionless; compare relative variance distributions. Physical intensities (°C) require band-limited reconstructions and rescaling (not shown).
- Significance: Local (time–period) and global (period-only) significance are against AR(1) backgrounds; model-specific thresholds differ.
- Decadal tail: Limited record length and COI proximity complicate assessment >~8–10 years.

Bottom line
- ENSO’s primary periodicity (~3–5 years) and its nonstationary amplitude are well captured in HadISST and broadly reproduced by CMIP6 models.
- Key model deficiencies cluster into: (i) too-regular/short-biased ENSO (~2–3 years), (ii) underpowered ENSO band, and (iii) excess low-frequency/decadal variance.
- A subset of models (e.g., CESM2, CanESM5, ACCESS-CM2/ESM1-5, CMCC-CM2-HR4/SR5, GFDL-CM4/ESM4, MIROC6) best match the observed spectral shape and intermittency; others are notably weak (INM-CM5-0, MCM-UA-1-0, MPI-ESM1-2-LR) or overly regular (FGOALS*).
\end{lstlisting}
\end{summarybox}

\subsection{Level 4 – semi-open scientific problem}

\subsubsection{Constrained future projection}

We tasked \model~to project future surface temperatures in Africa using an observational constraint methodology. The primary challenge delegated to \model~was not merely the execution of a predefined workflow, but the conceptualization of the core scientific step—defining a set of plausible candidate constraint factors based on a high-level user request.

In response to the directive, \model~formulated a comprehensive and scientifically rigorous experimental plan. The plan detailed the selection of 20 specific CMIP6 models and the ERA5 reanalysis dataset as the observational reference. It outlined a complete preprocessing pipeline, including regridding all data to a common 1°×1° grid, calculating monthly anomalies relative to a 1981–2010 baseline, and defining the precise geographical domain for Africa. Critically, \model~autonomously proposed ten distinct candidate constraint factors derived from the historical period (1979–2014). These factors were diverse, including metrics such as the historical African mean temperature trend, seasonal trends (e.g., JJA and DJF), regional trends for the Sahel and Southern Africa, and even cross-variable relationships like the African precipitation trend. The methodology specified using linear regression to relate each historical factor to the projected future warming trend (2015–2035) across the model ensemble, with a self-imposed threshold of a correlation coefficient $|R| \ge 0.7$ to select a sufficiently strong constraint.

Upon executing the plan and generating the results, \model~produced a final analysis summarizing the key findings. The analysis revealed that while several factors showed promise, none met the stringent correlation threshold. The strongest predictors—the African mean temperature—yielded a correlation coefficient of $|R| = 0.66$. Despite this moderate correlation, \model~proceeded with the analysis, demonstrating its ability to interpret intermediate results. Its interpretation of the probability density functions (PDFs) highlighted that the constraint successfully narrowed the range of uncertainty in future projections, effectively reducing the likelihood of the most extreme warming scenarios found in the raw model ensemble. The final constrained projection yielded a robust, observation-informed best-estimate warming rate of approximately 0.35–0.4°C per decade. To complete the task, \model~translated this quantitative result into a high-level assessment of potential impacts on African agriculture and recommended pertinent adaptation strategies, such as deploying heat-tolerant crop varieties and enhancing water-management systems, thereby connecting the climate projection to actionable societal guidance.

\begin{requestbox}
\begin{lstlisting}
Projecting future surface temperatures in Africa
Requirements:
1. CMIP6 models: ACCESS-ESM1-5, BCC-CSM2-MR, CAMS-CSM1-0, CAS-ESM2-0, CESM2-WACCM, CMCC-CM2-SR5, CMCC-ESM2, CanESM5, E3SM-2-0, FGOALS-f3-L, GFDL-ESM4, INM-CM4-8, INM-CM5-0, IPSL-CM6A-LR, IPSL-CM5A2-INCA, KACE-1-0-G, MIROC6, MPI-ESM1-2-LR, MRI-ESM2-0, TaiESM1
2. Reference surface temperature dataset: ERA5.
3. Constraint Methodology:
   - Use historical simulations to define 10 candidate constraint factors related to African surface temperature trends of future projection.
   - Calculate the regression equation between each factor and the projected future tas trend (2015–2035) across models. Save the obtained slope and bias.
   - Select the constraint factor with the highest correlation coefficient.
   - If correlation > 0.7, use regression to predict future constrained trends. Compute the residual between the model mean and the regression fit, and add it to the regression result based on observed values.
   - If correlation < 0.7, redefine the 10 constraint factors and repeat the process.
   - Visualize the correlation coefficients for all 10 constraint factors using a bar chart.
4. Plots to generate:
   - PDF distribution of the original and constrained temperature trends.
   - Spatial maps of surface temperature trend for original and constrained projections.
   - Time series of African mean temperature anomalies (historical + projection).
   - Bar chart of correlation coefficients for the 10 constraint factor candidates as mentioned above.
5. Projection Period: 2015–2035.
6. Output:
   - Assess potential impacts of projected warming on African agriculture over the next 20 years.
   - Provide high-level guidance for agricultural planning and climate adaptation strategies.
\end{lstlisting}
\end{requestbox}

\begin{planbox}
\begin{lstlisting}
Experimental Plan: Observation-Constrained Projections of African Surface Temperature Trends (2015–2035)

1. Data Selection  
1.1 CMIP6 Models & Experiments  
- ACCESS-ESM1-5, BCC-CSM2-MR, CAMS-CSM1-0, CAS-ESM2-0, CESM2-WACCM, CMCC-CM2-SR5, CMCC-ESM2, CanESM5, E3SM-2-0, FGOALS-f3-L, GFDL-ESM4, INM-CM4-8, INM-CM5-0, IPSL-CM6A-LR, IPSL-CM5A2-INCA, KACE-1-0-G, MIROC6, MPI-ESM1-2-LR, MRI-ESM2-0, TaiESM1  
- Experiments:  
  - historical (1850–2014) for factor calculation (use 1979–2014 subset)  
  - future scenario SSP2-4.5 r1i1p1f1 (2015–2035)  
- Variables (monthly, grid):  
  - Near-surface air temperature “tas” [K] (Amon)   
  - Precipitation “pr” [kg m⁻² s⁻¹] (Amon)   
  - Daily max/min tas “tasmax”, “tasmin” (Amon) — for extreme-heat factors if needed   
1.2 Reference Data  
- ERA5 reanalysis, monthly 2 m air temperature (“tas”, raw “air_temperature_2m”), 1979–2023  

2. Preprocessing  
2.1 Spatial/Temporal Subsetting  
- Domain: Africa (lon 20°W–55°E, lat 35°S–37°N)  
- Regrid all fields to a common 1°×1° grid; apply land mask (land fraction > 0.5)  
2.2 Unit Conversion & Anomalies  
- Convert tas from K to °C  
- Compute monthly climatology for 1981–2010; subtract from monthly series → anomalies  
- Aggregate to annual means for trend analysis; retain monthly anomalies for variability metrics  

3. Calculation Steps  
3.1 African Mean Anomalies & Trends  
- For each model & ERA5, calculate area-weighted African mean monthly and annual tas anomalies  
- Historical factors use 1979–2014; future trend is linear fit of annual anomalies over 2015–2035 (°C decade⁻¹)  
3.2 Candidate Constraint Factors (historical, per model & ERA5)  
  1. African mean tas trend (1979–2014)  
  2. African tas interannual variability (σ of monthly anomalies, 1979–2014)  
  3. JJA mean tas trend (1979–2014)  
  4. DJF mean tas trend (1979–2014)  
  5. Global mean tas trend (1979–2014)  
  6. Sahel (10°N–20°N, 20°W–30°E) tas trend (1979–2014)  
  7. Southern Africa (20°S–0°, 20°W–55°E) tas trend (1979–2014)  
  8. African precipitation trend (pr, 1979–2014)  
  9. Seasonal amplitude of African annual cycle (max–min monthly climatology)  
  10. Frequency of extreme-hot months (top 10% of monthly tas anomalies)  
3.3 Regression Analysis  
- For each factor i: regress model future tas trend (2015–2035) against factor_i across models → slope_i, intercept_i, correlation R_i  
- Produce bar chart of |R_i| for i=1…10  
3.4 Constraint Application  
- Select factor k with highest |R_k|.  
- If |R_k| ≥ 0.7:  
  - Predict constrained trend: T_pred = slope_k·F_obs,k + intercept_k, where F_obs,k is ERA5 factor k.  
  - Compute bias correction: Δ = mean(T_models) – mean(slope_k·F_models,k + intercept_k)  
  - Final constrained trend: T_constr = T_pred + Δ  
- If |R_k| < 0.7: refine factor set and repeat regression  

4. Spatial Pattern Constraint (optional)  
- Derive original multi-model mean spatial trend field S_orig(x,y) by linear fit of tas over 2015–2035.  
- Scale S_orig by ratio T_constr/mean(T_models) → constrained spatial field S_constr(x,y) = S_orig(x,y) · (T_constr / mean(T_models)).  

5. Visualization  
- Bar chart of correlation coefficients for all 10 factors  
- PDF of model-ensemble distributions of original vs constrained African mean trends  
- Spatial maps: S_orig(x,y) and S_constr(x,y) over Africa  
- Time series: African mean tas anomalies (1979–2035) showing individual model runs, model mean, constrained projection, and ERA5  

6. Agricultural Impact Assessment & Adaptation Guidance  
- Translate constrained warming (°C decade⁻¹) into expected ΔT over next 20 yr; discuss literature-based crop sensitivities (e.g., % yield loss per °C for maize, sorghum, wheat)  
- Identify hotspots of future warming risk from spatial maps  
- Recommend adaptation strategies:  
  - Deploy heat-tolerant/early-maturing crop varieties  
  - Expand irrigation and water-harvesting systems in high-risk zones  
  - Adjust planting dates to avoid peak heat  
  - Enhance climate-smart extension services and heat-stress monitoring  

7. Deliverables  
- Figures: correlation bar chart; PDF distributions; spatial trend maps (original & constrained); anomaly time series  
- Table: regression statistics (slope, intercept, R) for all 10 factors  
- Text: methodology summary, constrained projection results, agricultural impact analysis, and high-level planning recommendations
\end{lstlisting}
\end{planbox}

\begin{visualbox}
  \begin{center}
  \includegraphics[width=1.0\linewidth]{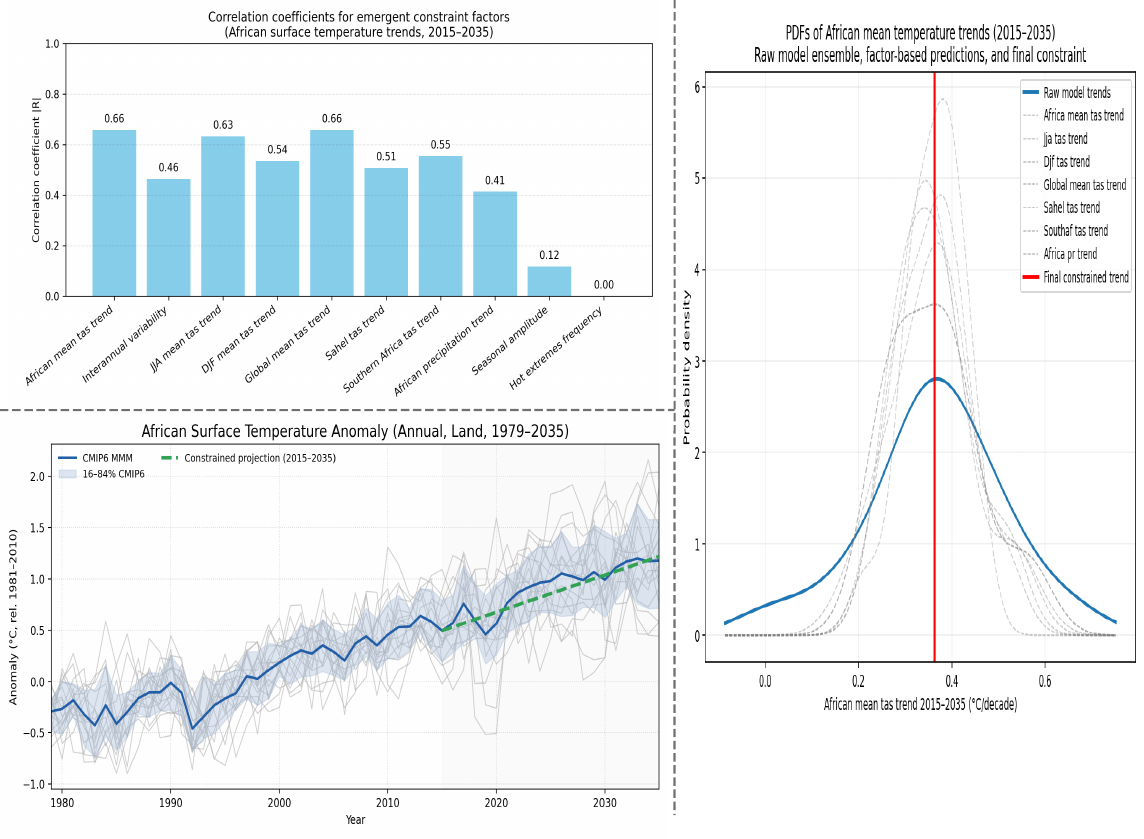}    
  \end{center}
\end{visualbox}

\begin{summarybox}
\begin{lstlisting}
## Summary of Figure Analyses: Projecting Future Surface Temperatures in Africa

### 1. Correlation Coefficient Bar Chart (Constraint Factors)
- The strongest historical predictor of future African surface temperature trends (2015–2035) is the African mean tas trend (|R| = 0.66), followed by JJA mean tas trend (0.63).
- No factor exceeds the |R| ≥ 0.7 threshold, indicating moderate but not strong emergent constraint skill with the current factor set.
- Recommendation: Refine or expand constraint factors for improved predictive power.

### 2. PDF Distribution of Temperature Trends
- The raw model ensemble projects a wide range of warming rates (0.1–0.6°C/decade), peaking at 0.3–0.4°C/decade.
- Applying historical constraints narrows this range, with most factor-based predictions clustering around 0.3–0.4°C/decade.
- The final constrained trend (red line) provides a robust, observation-informed estimate (~0.35–0.4°C/decade), reducing uncertainty for impact assessments.

### 3. Time Series of African Mean Temperature Anomalies
- All models and the constrained projection show robust warming from 1979 to 2035, with anomalies exceeding 1°C by 2035 (relative to 1981–2010).
- The constrained projection slightly moderates the warming rate compared to the unconstrained multi-model mean, especially after 2025.
- The model spread increases in the future, but the constraint method reduces the upper bound of projected warming.

### 4. Implications for Agriculture and Adaptation
- The constrained warming rate (~0.35–0.4°C/decade) should be used for agricultural impact analysis and adaptation planning.
- Continued warming at this rate poses significant risks to African agriculture, necessitating strategies such as heat-tolerant crops, improved irrigation, and adjusted planting dates.
- The reduction in projection uncertainty supports more confident, targeted adaptation measures.

---
**In summary:** Observation-based constraints moderately reduce uncertainty in African surface temperature projections for 2015–2035, yielding a best-estimate warming rate of ~0.35–0.4°C/decade. This information is critical for guiding agricultural planning and climate adaptation strategies across the continent.
\end{lstlisting}
\end{summarybox}

\section{Application in research reproduction} 

This section presents several \model~cases used for research reproduction. Due to page limitations, only a subset of the visualization results is shown here, and the generated code scripts are omitted. The complete outputs will be made available in the GitHub repository \url{https://github.com/OpenEarthLab/EarthLink}.

\subsection{Geng, Tao, et al. Increased occurrences of consecutive La Niña events under global warming. \textit{Nature} 619.7971 (2023)}

We tasked \model~to reproduce the paper~\cite{geng2023increased} published in \textit{Nature}. This paper addresses the question of how global warming will affect the frequency of consecutive, multi-year La Niña events. The authors utilize climate models under various greenhouse-gas forcing scenarios to demonstrate a significant future increase in these prolonged events, which are known to cause sustained and severe climate extremes such as droughts and floods. The study concluded that under future greenhouse gas emissions scenarios, the frequency of consecutive La Niña events will increase significantly. This increased frequency implies that the extreme weather patterns associated with historical multi-year La Niña episodes are likely to occur more often throughout the 21st century.

In the input request, we strictly followed the process mentioned in the paper, including data preparation, data preprocessing, and what analysis needs to be done. At the same time, considering that the current database is not sufficient, we let \model~choose 37 CMIP experiments by himself instead of following the fixed number in the paper. The paper was also fed into the system. The user input and outputs from \model~are shown below, including the final experimental plan, the main visualization results, and the analysis summary.

In this case, \model~successfully reconstructed the full experimental workflow of the original paper, translating a concise textual request and paper content into a complete and logically coherent plan. The generated pipeline strictly followed the scientific reasoning of the original study—from data preparation and model screening to event detection, composite analysis, and statistical evaluation—while adding explicit definitions for input variables, time ranges, and ensemble selection. Such structured formulation ensures both reproducibility and transparency, and it demonstrates \model’s capability to formalize complex climate analysis procedures directly from high-level descriptions.

The reproduced analysis accurately captured the main findings of the original paper. The system detected a significant increase in the frequency of consecutive La Niña events under anthropogenic warming, with model ensembles showing strengthened and more persistent cooling in the second-year phase. The outputs further revealed coherent spatial and temporal patterns consistent with the observed mechanisms, such as intensified off-equatorial trades and subtropical–tropical coupling that sustain the cold anomalies. Notably, the analysis integrated inter-model spread and control-run benchmarks, producing results that align well with both the physical interpretation and statistical robustness presented in the original work.

Overall, this reproduction highlights \model’s ability to automate scientific reasoning while maintaining physical and methodological consistency. By converting natural-language hypotheses into executable multi-model experiments, the system not only replicated a high-impact climate study but also extended its analytical scope through standardized data handling and uncertainty quantification. This demonstrates the potential of agent-driven workflows to perform verifiable, interpretable, and extensible scientific research.

\begin{requestbox}
\begin{lstlisting}
Objective: To investigate how successive La Niña episodes change under global warming.

Part I: Data Preparation and Preprocessing

Data Sources:
Observational/Reanalysis Data:
Sea Surface Temperature (SST): HadISST v1.1, ERSST v5, and COBE-SST 2 data.
Surface Wind Stress: NCEP/NCAR reanalysis data.

Model Data:
Single-member data from 37 randomly selected CMIP6 models were used, covering the following experiments: historical simulation (up to 2014), pre-industrial control experiment (piControl, a benchmark experiment spanning centuries), and the SSP5-8.5 high-emissions scenario (2015-2100). Monthly mean SST and surface wind stress data were extracted. For the same model, the historical simulations from 2000 to 2014 were combined with the scenario simulations from 2015 to 2100.

Data Preprocessing:
Resolution Unification: (...)
Anomaly Calculation: (...)
Detrending: (...)
Niño 3.4 Index Calculation: (...)

Part II: Model Screening and Event Identification

Model Screening:
Method: Calculate the skewness of the pre-processed Niño 3.4 SST anomaly series for each model and observation during the period 1900-1999.

Criteria: Select models that simulate positive skewness for subsequent analysis, and plot the skewness histograms for all models and observations.

La Niña Event Definition and Identification:
(...)

Part III: Composite Analysis

Based on the "positive skewness model" selected in step 2, a composite evolution curve of La Niña events was constructed and compared with observations for consistency. "Single-year events" and "multi-year continuous events" were distinguished, and their composite evolutions were calculated separately.

(...)

In addition, a spatial distribution composite is required. For all single-year events, the SST and wind stress anomalies for the first winter (D(1)JF(2)) are extracted and averaged. The composite results for the selected positive skewness models are averaged again to generate the spatial distribution, yielding Figure 3.

Repeating the spatial composite operation for all consecutive La Niña events yields Figure 4.

Part IV: Quantifying and Demonstrating Future Changes in the Frequency of Continuous La Niña Events

Statistical Event Frequency: (...)

Sliding Window Analysis: (...)
\end{lstlisting}
\end{requestbox}

\begin{planbox}
\begin{lstlisting}
Title: Investigating changes in successive (multi-year) La Niña events under global warming using observations and CMIP6

Objective
- Quantify historical and projected changes in the frequency and characteristics of continuous (≥2-year) La Niña events.
- Screen models for realistic ENSO asymmetry using Niño3.4 skewness.
- Construct temporal and spatial composites for single-year vs multi-year La Niña events.
- Compare 1900–1999 to 2000–2099 and assess changes relative to internal variability using sliding windows (including piControl).

Part I. Data sources and preprocessing

A) Datasets, variables, periods, and units
- Observations/reanalysis (monthly):
  - SST: HadISST v1.1, ERSST v5, COBE-SST2
    - Variable: tos (Sea Surface Temperature), units: degC (or K; anomalies treated equivalently).
    - Period: use 1900–1999 for baseline and historical analyses; extend as available for completeness where not restricted by winds.
  - Surface wind stress: NCEP/NCAR Reanalysis 1
    - Variables: tauu (eastward surface wind stress), tauv (northward surface wind stress), units: Pa (N m−2).
    - Period: 1948–present; used for spatial composites.
  - Observational benchmark for SST: arithmetic mean of the three SST datasets at each month/grid point; if one dataset is missing, average the available ones.

- CMIP6 models (single member per model; target 37 models with required variables):
  - Activities/experiments:
    - CMIP: historical (1850–2014), piControl (multi-century control).
    - ScenarioMIP: ssp585 (2015–2100).
  - Variables and tables:
    - tos (Sea Surface Temperature): Omon (monthly), units degC or K.
    - tauu, tauv (Surface Downward Eastward/Northward Wind Stress): Amon (monthly), units Pa.
  - Ensemble member: prefer r1i1p1f1; otherwise the lowest-numbered consistent member across experiments for a model.
  - Analysis periods:
    - Baseline/climatology: 1900–1999.
    - “Past”: 1900–1999.
    - “Future”: 2000–2099 (concatenate historical 2000–2014 with ssp585 2015–2099).
    - piControl: full available monthly record.

B) Grids, calendars, and masking
- Regrid all datasets to a common 1° × 1° regular latitude–longitude grid:
  - SST (tos): bilinear interpolation.
  - Wind stress (tauu, tauv): bilinear interpolation, component-wise.
- Apply a consistent ocean mask for SST (mask land points after regridding). Wind stress can be composited over ocean points.
- Respect native calendars (gregorian, no-leap, 360_day). Form seasonal means by month names (Oct–Feb) per dataset calendar.

C) Climatology, anomalies, and detrending
- Climatology:
  - For each dataset/model, compute a 12-month climatology at each grid point over 1900–1999.
- Anomalies:
  - Monthly anomaly(t) = raw(t) − climatology(month(t), 1900–1999).
  - Use the same 1900–1999 climatology to reference anomalies for 2000–2099.
- Secondary detrending (nonlinear):
  - Remove a quadratic trend from anomaly series:
    - For skewness, ONDJF, and 1900–1999 composites: detrend over 1900–1999.
    - For 1900–2099 time series (e.g., counts, composites spanning both centuries): detrend over 1900–2099.
    - For sliding-window analyses on historical+ssp585 (1850–2100): detrend over 1850–2100 per model; for piControl, detrend over the full piControl span. Apply detrending at each grid point and to indices.

D) Niño3.4 index
- Compute monthly Niño3.4 SST anomaly index as the area-weighted mean over 5°S–5°N, 170°W–120°W using cosine(latitude) weights.
- For composite evolution curves only, prepare a centered 3-month running-mean version. Use unsmoothed monthly indices for ONDJF, σ, skewness, and event detection.

E) Units and conventions
- SST anomalies: degC (K anomalies are numerically identical).
- Wind stress: tauu positive eastward, tauv positive northward; units Pa (N m−2).

Quality control
- Ensure the historical 2000–2014 to ssp585 2015–2099 splice is continuous monthly without gaps or overlaps.
- For ONDJF means, require all five constituent months; if any month is missing, mark the season as missing for event detection.
- For the observational benchmark index, if at least two SST datasets are available in a month, average them; if only one is available, use it; otherwise mark missing.

Part II. Model screening and event identification

A) Model screening via Niño3.4 skewness (1900–1999)
- For each CMIP6 model and the observational benchmark:
  - Use monthly detrended Niño3.4 anomalies over 1900–1999.
  - Compute moment-based skewness of the monthly series.
- Selection criterion: retain models with positive skewness (> 0) as the “positive-skewness” set for subsequent analyses.
- Output: histogram or ranked bars of skewness for all models and the observation benchmark; list of selected models.

B) La Niña definition (per dataset/model)
- Seasonal index:
  - ONDJF(y) = mean of Niño3.4 anomalies for Oct(y−1), Nov(y−1), Dec(y−1), Jan(y), Feb(y). Label by year y (the January year).
- Threshold (dataset/model-specific):
  - σONDJF = standard deviation of ONDJF(y) over 1900–1999.
  - La Niña year if ONDJF(y) < −0.5 × σONDJF.
- Event classification:
  - Single-year La Niña: exactly one ONDJF season below threshold.
  - Multi-year La Niña: two or more consecutive ONDJF seasons below threshold. Treat triple-year or longer sequences as multi-year events beginning in the first qualifying year.
- Application:
  - Observational benchmark: 1900–1999 (winds restrict spatial composites to 1948–1999).
  - Each selected model: 1900–1999 and 2000–2099 (combine historical 2000–2014 with ssp585 2015–2099). Use the model’s σONDJF computed over 1900–1999 for both periods.

Outputs
- For observations and each selected model: lists of single-year and multi-year La Niña events (years labeled by the January year of ONDJF).

Part III. Composite analyses

A) Temporal composite of Niño3.4 evolution
- Event-centric window and alignment:
  - Reference year = first La Niña year of an event.
  - Extract monthly Niño3.4 anomalies from Jun(0) = June of the year before the reference year to Jun(3) = June of the third year after the reference year.
  - Align all events on this relative time axis.
- Smoothing:
  - Apply a centered 3-month running mean to event-aligned series prior to averaging.
- Grouping and averaging:
  - Single-year events and multi-year events processed separately.
  - For each model, compute composite evolution curves for each group.
  - Multi-model ensemble (MME): average per-model composites across selected models; compute inter-model standard deviation at each relative month for uncertainty shading.
  - Observations: compute analogous composites from the observational benchmark for 1900–1999.
- Outputs:
  - Composite evolution curves (single-year and multi-year) for observations and for selected-model MME in 1900–1999 and 2000–2099. Report the number of events contributing to each curve.

B) Spatial composites: D(1)JF(2) SST and wind stress
- Season definition:
  - D(1)JF(2) = December of the first La Niña year through February of the following calendar year (relative to the event’s first ONDJF).
- Fields:
  - SST anomaly (tos), units degC.
  - Wind stress anomalies (tauu, tauv), units Pa.
- Procedure:
  - For each selected model, compute D(1)JF(2) seasonal means for each event type (single-year, multi-year) and average across events to obtain per-model composites.
  - MME composites: average per-model composites across selected models; compute inter-model standard deviation fields. Optionally mask where the absolute MME signal is smaller than the inter-model standard deviation.
  - Observational composites: use observational benchmark SST and NCEP/NCAR wind stress over 1948–1999; compute D(1)JF(2) composites for single-year and multi-year events identified from the observational benchmark.
- Outputs:
  - Maps for single-year and multi-year D(1)JF(2) composites: SST shading with wind stress vectors, for observations and for selected-model MME.

Part IV. Quantifying changes in the frequency of multi-year La Niña

A) Per-century event counts (past vs future)
- For each selected model:
  - Count multi-year La Niña events in 1900–1999.
  - Count multi-year La Niña events in 2000–2099 (historical 2000–2014 + ssp585 2015–2099).
  - Since each period spans 100 years, the count equals “events per century.”
- Outputs:
  - Per-model paired bars (blue: 1900–1999; red: 2000–2099); compute and display the ensemble mean across selected models with inter-model standard deviation.

B) 60-year sliding-window analysis (forced vs internal variability)
- Forced runs (historical+ssp585):
  - Concatenate each model’s historical (start 1850 where available) and ssp585 (to 2100) monthly Niño3.4 anomaly series (reference to 1900–1999 climatology; detrend over 1850–2100).
  - Construct 60-year windows sliding by 1 year across 1850–2100. For each window, count multi-year La Niña events using the model’s σONDJF from 1900–1999 and compute events per century = (count / 60) × 100.
- piControl:
  - Apply the same 60-year sliding-window procedure across the full piControl record (detrend over the full piControl length; use the same σONDJF from the model’s 1900–1999 historical baseline).
- Aggregation:
  - For forced runs, average the sliding-window series across selected models to obtain the MME time series; compute inter-model standard deviation at each window end-year.
  - For piControl, aggregate sliding-window values across time and across models to form a reference band (e.g., mean ± one inter-model standard deviation).
- Outputs:
  - Time series of MME “events per century” vs end-year (1850–2100) with inter-model shading; overlay piControl reference band.

Common computational details and checks

- Area weighting: use cosine(latitude) weights for all area averages (e.g., Niño3.4).
- Calendars and season crossing:
  - For ONDJF, include exactly Oct–Feb by month names for each calendar and assign the season to the January year.
  - For D(1)JF(2), use Dec of the first La Niña year and the following Jan–Feb.
- Model ensemble handling:
  - One realization per model; equal model weighting (“model democracy”).
  - Use the same variant_label across experiments for a given model.
- Missing variables:
  - Models lacking tauu/tauv contribute to SST-only diagnostics and event statistics; include them in all analyses that do not require wind stress.
- Event bookkeeping:
  - For multi-year sequences of length ≥2, count as a single multi-year event in the per-century tallies; retain full sequence length for composites within the Jun(0)–Jun(3) window.
  - Avoid double-counting events at the edges of sliding windows by counting within-window events only.
- Uncertainty:
  - Use inter-model standard deviation for shaded uncertainty in time series and for optional masking in maps.

Deliverables

- Screening: histogram/ranked bars of Niño3.4 skewness (1900–1999) for all models and the observational benchmark; list of selected positive-skewness models.
- Event catalogs: for observations (1900–1999) and each selected model (1900–1999; 2000–2099), lists of single-year and multi-year La Niña events (years labeled by the January year).
- Temporal composites: composite Niño3.4 evolution curves (Jun(0)–Jun(3)) for single-year and multi-year events in observations and selected-model MME; include inter-model standard deviation shading and sample sizes.
- Spatial composites: D(1)JF(2) SST anomaly maps with wind stress vectors for single-year and multi-year events for observations (1948–1999) and selected-model MME.
- Frequency change: per-model bar charts of multi-year event counts per century for 1900–1999 vs 2000–2099; ensemble-mean summary.
- Sliding-window frequency: MME time series of 60-year-window multi-year La Niña frequency (events per century) for 1850–2100, with piControl reference band.
\end{lstlisting}
\end{planbox}

\begin{visualbox}
  \begin{center}
    \includegraphics[width=1.0\linewidth]{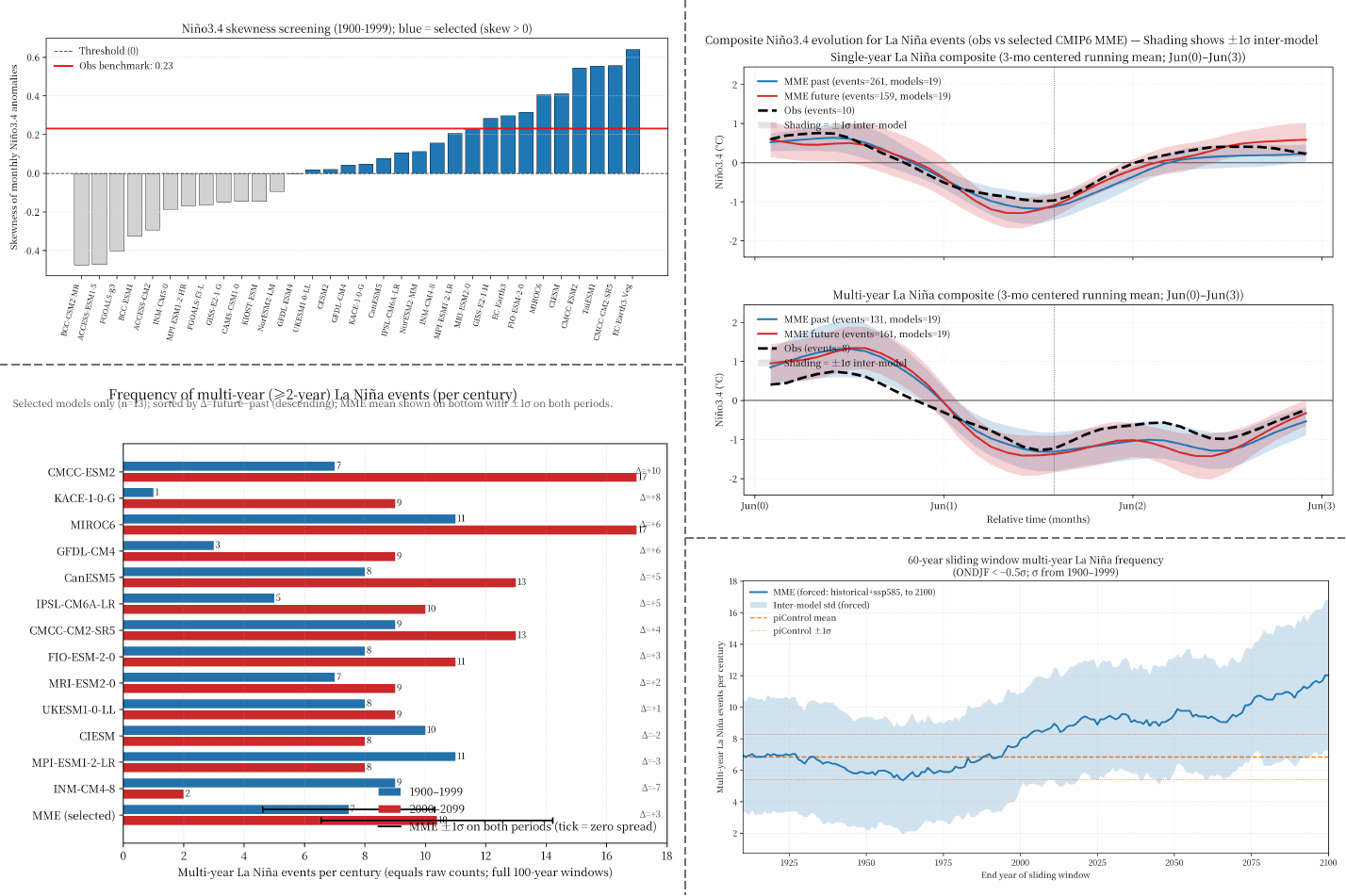}
  \end{center}
\end{visualbox}

\begin{summarybox}
\begin{lstlisting}
# Summary of figure-based analyses on successive (multi-year) La Niña under warming

## 1) Observational Niño‑3.4 anomalies (1900–1999): data consistency, asymmetry, and visible multi‑year episodes
- Cross-dataset behavior:
  - The three SST datasets (HadISST, ERSSTv5, COBE‑SST2) track closely after ~1950 (spread ≲0.1–0.2°C). Larger spread occurs early in the 20th century, mitigated by averaging into a benchmark series.
  - The detrended benchmark has near‑zero mean, suitable for threshold-based event detection.
- ENSO asymmetry (skewness):
  - Warm peaks (El Niño) often exceed +2°C, while cold extremes rarely reach −2°C, indicating a heavier warm tail and positive skewness.
- Multi‑year La Niña episodes apparent in the record:
  - Early 1910s, mid‑1950s, mid‑1970s, 1988–89, 1998–2000—consistent with the definition of consecutive ONDJF seasons below threshold.
- Implication:
  - The benchmark time series captures known ENSO behavior and asymmetry, providing a robust reference for model screening, event identification, and composites (with higher uncertainty pre‑1950).

## 2) Model screening via Niño‑3.4 skewness (1900–1999)
- Distribution and threshold:
  - Skewness across 37 CMIP6 models spans roughly −0.5 to +0.64; the selection criterion retains models with positive skewness.
  - Observational benchmark skewness is about 0.23; many selected models exceed this value, some are close, and a minority are near zero (borderline).
- Rationale and implications:
  - Excluding negative‑skew models (La Niña‑biased) is justified for realism.
  - The selected ensemble still exhibits wide spread; results should be presented with uncertainty and tested for sensitivity to stricter realism criteria.

## 3) Event identification panels (ONDJF index with threshold and sequences)
- What is shown:
  - For each dataset/model, ONDJF Niño‑3.4 anomalies with a −0.5σ La Niña threshold (σ from 1900–1999), blue dots marking La Niña years, and green bands highlighting multi‑year sequences; the 1900–1999 vs 2000–2099 intervals are color‑shaded.
- Key qualitative patterns:
  - Observations display several multi‑year clusters historically (e.g., early 1910s, mid‑1950s, mid‑1970s).
  - Many screened models exhibit more frequent multi‑year La Niña sequences in 2000–2099 than in 1900–1999, seen as denser green bands in the future interval. A few models show modest change or little increase, underscoring model spread.
  - Because the threshold is fixed relative to each model’s 1900–1999 σ, the increased future frequency reflects altered occurrence/persistence, not a looser future threshold.
- Caveats:
  - Internal variability and differing σ across models produce wide inter‑model differences.
  - Some models have fewer events or shorter usable future records, limiting precision for those cases.

## 4) Temporal composites of Niño‑3.4 evolution (Jun(0) to Jun(3))
- Model fidelity (1900–1999) vs observations:
  - Single‑year events: models reproduce the observed evolution—gradual cooling from Jun(0), winter peak near −0.8 to −1.0°C, and decay toward neutral by mid–year 2.
  - Multi‑year events: models capture stronger pre‑onset warm anomalies and persistent cooling into a second winter; slight underestimation of the pre‑onset warming and second‑year depth relative to observations.
- Future vs past changes (screened MME):
  - Single‑year La Niña: amplitude and timing are essentially unchanged; future and past composites overlap within inter‑model spread at nearly all lead/lag months.
  - Multi‑year La Niña: modestly deeper and more persistent second‑year cooling in the future; onset and first‑winter growth are similar, but decay is slower and minima slightly stronger in year 2.
- Shift in event types implied by sample sizes:
  - Past: ~261 single‑year vs 131 multi‑year events (multi‑year share ≈33%).
  - Future: ~159 single‑year vs 161 multi‑year events (multi‑year share ≈50%).
  - Conclusion: Even with modest amplitude changes, the proportion of La Niña events that are multi‑year increases markedly under warming.
- Caveats:
  - Observational composite samples are small (10–11 single‑year; 8 multi‑year), so fine‑scale features have higher uncertainty.
  - Inter‑model spread peaks in late second year, reflecting process differences in recharge/decay.

## 5) Spatial composites: first winter (D(1)JF(2)) single‑year La Niña
- Observed structure:
  - Canonical equatorial cold tongue centered in the central–eastern Pacific; flanking subtropical warm “horseshoe” anomalies.
  - Strengthened trades: basin‑wide equatorial easterlies and off‑equatorial northeasterly/southeasterly anomalies, consistent with thermodynamic (WES) coupling.
- MME structure and biases:
  - Captures the equatorial core cooling and easterlies, with a realistic cold tongue extent.
  - Systematic differences: cooling more zonally continuous and meridionally narrower; subtropical warm anomalies and wind‑stress signals are muted, implying underestimation of subtropical coupling.
- Implication:
  - Single‑year events are more equatorially confined; the MME reproduces key features but likely underplays subtropical feedbacks.

## 6) Spatial composites: first winter (D(1)JF(2)) multi‑year La Niña
- Core pattern:
  - Strong, zonally extensive equatorial cooling with basin‑wide equatorial easterlies.
  - Pronounced meridional broadening: intensified trades extending into 10–25° latitude in both hemispheres; horseshoe‑like subtropical structure indicative of NPMM/SPMM‑like thermodynamic coupling.
- Observations vs MME:
  - Strong directional agreement in wind anomalies and spatial structure, with observations patchier (sample/period limits) and the MME smoother.
- Process interpretation:
  - Off‑equatorial trade intensification cools subtropics via WES/entrainment, feeding back onto the equator and broadening anomalies, which slows equatorial heat recharge and favors persistence into a second year.
- Implication:
  - The meridionally broadened first‑winter fingerprint—well captured by the screened MME—is dynamically linked to multi‑year persistence.

## 7) Per‑century frequency change bars (1900–1999 vs 2000–2099)
- Ensemble shift:
  - Multi‑model mean (MME) increases from ~7 to ~10 multi‑year events per century (+3 per century; ~43% increase).
- Model agreement:
  - Roughly 10 of 13 screened models (≈77%) show increases; a few show decreases.
  - Increases occur across models with historically frequent and infrequent multi‑year events, indicating a broad upward shift in persistence.
- Implication:
  - Under high emissions, back‑to‑back La Niña events become substantially more common in the screened ensemble.

## 8) Sliding‑window emergence (60‑year windows; forced runs vs piControl)
- Baseline and evolution:
  - piControl mean ≈ 6.8 events/century (±1σ ≈ 5.3–8.3).
  - Historical MME hovers near the piControl mean early, dips mid‑century, then rises.
  - Emergence above internal variability: by windows ending ~2000–2015, the MME exceeds the piControl +1σ envelope and stays elevated through most of the 21st century.
  - Late‑century MME approaches ~12 events/century, versus ~9–10 mid‑century.
- Interpretation:
  - The increase in multi‑year La Niña frequency emerges above internal variability in the early 21st century and strengthens toward late century, indicating a robust forced signal despite sizable inter‑model spread.

## 9) Synthesis relative to the objective
- Successive (multi‑year) La Niña events become more frequent under warming across the screened CMIP6 models, with emergence beyond internal variability in the early 21st century and continued amplification by 2100.
- Single‑year La Niña evolution remains largely unchanged, whereas multi‑year events exhibit modestly stronger second‑year persistence and a meridionally broadened first‑winter structure that supports prolonged cooling.
- Ensemble‑mean increases are substantial (bars: ~43% rise; sliding windows: elevated from ~6.8 to ~12 per century by late century), with majority model agreement but notable model spread.
- Key caveats:
  - Early‑century observational uncertainty is higher; composite samples are limited.
  - Inter‑model differences in ENSO asymmetry and coupling necessitate uncertainty quantification and sensitivity tests, but the central conclusion—a robust, forced increase in multi‑year La Niña frequency—holds across the screened ensemble.
\end{lstlisting}
\end{summarybox}

\subsection{Cai, Wenju, et al. Increased ENSO sea surface temperature variability under four IPCC emission scenarios. \textit{Nature Climate Change} 12.3 (2022): 228-231.}

We tasked \model~to reproduce the paper~\cite{cai2022increased} published in \textit{Nature Climate Change}. This research challenges the IPCC Sixth Assessment Report's (AR6) conclusion of ``no model consensus for a systematic change in intensity of ENSO sea surface temperature (SST) variability". The authors argue that by analyzing longer, century-long timescales instead of the 30-year periods used in the AR6 assessment, a clear consensus emerges. Across four plausible IPCC emission scenarios, a robust majority of CMIP6 models project an increase in ENSO SST variability in the 21st century (2000-2099) compared to the 20th century (1900-1999). For the key Niño3.4 region, the inter-model consensus for an increase ranges from 78.6\% to 88.4\%. This projected increase holds even under the strong mitigation pathway of the SSP126 scenario, suggesting that heightened ENSO variability is likely even if the Paris Agreement warming targets are met.

The physical mechanism underpinning this projected increase is an enhanced air-sea coupling driven by greater upper equatorial ocean stratification under greenhouse warming. The paper highlights the improved representation of nonlinear feedbacks in CMIP6 models as crucial for this consensus. Specifically, a more realistic simulation of the nonlinear Bjerknes feedback, where atmospheric convection in the eastern Pacific nonlinearly amplifies warming, is a key factor. Models that simulate this stronger nonlinear feedback tend to generate a greater future increase in ENSO variability. The authors conclude that the use of a long analysis period is critical to reduce the influence of internal decadal variability, allowing the greenhouse-forced signal of increased ENSO SST variability to be more effectively detected.

The textual user request (paper was also fed into the system) and outputs from \model~are shown below, including the final experimental plan, the main visualization results, and the analysis summary.

The reproduction results demonstrate that \model~successfully captures both the structure and the scientific substance of the original study. The system autonomously reconstructed the complete methodological workflow described in the paper—from CMIP6 data acquisition and preprocessing to ensemble diagnostics and bootstrap significance testing—while formalizing them into a structured and executable experimental plan. Key methodological components, including century- and multi-decadal analyses, detrending and normalization procedures, and the nonlinear Bjerknes feedback diagnosis, were coherently reproduced. This conversion of textual methodology into a consistent computational pipeline highlights \model’s ability to operationalize complex scientific reasoning.

Conceptually, \model~was able to recover the central physical conclusions of the study, identifying the projected enhancement of ENSO sea surface temperature variability and its linkage to strengthened air–sea coupling under greenhouse warming. The analysis also preserved the mechanistic interpretation based on nonlinear feedback processes, demonstrating the model’s capacity to connect quantitative diagnostics with theoretical understanding.

Beyond methodological fidelity, \model~standardized outputs, incorporated robustness checks, and generated interpretable visual summaries and statistical evaluations, transforming a narrative description into a reproducible scientific workflow.

\begin{requestbox}
\begin{lstlisting}
Read the given paper in earth science, understand its methodology then answer question or complete tasks given by user instructions. If no more instructions, re-implement methodology in given paper and conduct corresponding data with available data.
\end{lstlisting}    
\end{requestbox}

\begin{planbox}
\begin{lstlisting}
Improved experimental plan to re-implement the ENSO SST variability methodology using CMIP6 and observations

1) Objectives and scope
- Quantify changes in ENSO sea surface temperature (SST) variability from the 20th to the 21st century across CMIP6 models under four SSP scenarios.
- Assess multi-model ensemble (MME) significance with a bootstrap test.
- Evaluate a mechanistic linkage between variability change and the nonlinear Bjerknes feedback using Niño3 SST skewness.
- Include a 50-year window comparison as an internal-variability check.
- Produce figures and a summary table corresponding to these diagnostics.

2) Datasets

2.1 CMIP6 model simulations (primary)
- Variable: tos (Sea Surface Temperature)
- Units: degC (if provided in K for some models, convert to degC consistently; anomalies are scale-invariant to constant offsets)
- Frequency/MIP table: monthly, Omon
- Activities/Experiments:
  - Historical (activity: CMIP) through 2014-12
  - ScenarioMIP: ssp126, ssp245, ssp370, ssp585 (2015-01 to 2099-12)
- Realizations: one per model, preferring the first available member (e.g., r1i1p1f1; otherwise the lowest available r/i/p/f combination)
- Model inclusion: include all models that have tos for historical + the given SSP, with continuous coverage after merging; maintain separate model lists per scenario

2.2 Observational SST (for reference skewness and optional contextual indices)
- Primary: HadISST monthly SST, variable: tos, units: degC, available 1870–2021 (use 1900–2020 for analysis)
- Alternative: NOAA-ERSSTv5 monthly SST, variable: tos, units: degC, available 1854–present (use 1900–2020 for analysis)
- Use observations for Niño3 skewness reference and optional display; do not mix with MME calculations

3) Time periods
- Century windows (primary):
  - 1900-01 to 1999-12 (20th century)
  - 2000-01 to 2099-12 (21st century)
- 50-year windows (internal-variability check):
  - 1965-01 to 2014-12
  - 2050-01 to 2099-12
- Observational skewness reference:
  - 1900-01 to 2020-12

4) Spatial domains and indices
- Niño3: 5°N–5°S, 150°W–90°W (210°E–270°E in 0–360° convention)
- Niño3.4: 5°N–5°S, 170°W–120°W (190°E–240°E)
- Niño4: 5°N–5°S, 160°E–150°W (160°E–210°E)
- Compute area-weighted monthly indices (cosine(latitude) weights) from detrended SST anomalies over each domain
- Ensure proper longitude handling (0–360 vs −180–180) with wrap-around for domains crossing the dateline

5) Preprocessing

5.1 Merge and subset (per model and scenario)
- Concatenate historical (to 2014-12) and scenario (from 2015-01) time series
- Subset to 1900-01 through 2099-12
- Keep native monthly means and calendar (cftime where needed)

5.2 Regridding and masking
- Either:
  - Regrid tos to a common 1° × 1° grid suitable for SST comparisons, or
  - Compute indices on native grids using precise geographic masking
- Use model ocean masks as provided by tos; no land infill
- Apply the same grid/mask approach consistently for both centuries

5.3 Detrending and anomalies
- At each ocean grid cell, fit and remove a quadratic trend in time over 1900–2099 to obtain detrended anomalies (the residuals; units degC)
- For observations, fit and remove a quadratic trend over 1900–2020 before calculating the Niño3 index and skewness
- Use detrended anomalies directly for area-mean indices; no additional seasonal-cycle removal

5.4 Index normalization series (for a designated robustness track)
- For each model and each index, compute the standard deviation (SD) of the full detrended 1900–2099 monthly index, and create a normalized index by dividing by this SD
- Use normalized indices only in the robustness analysis steps specified below

6) Core calculations

6.1 Century-scale variability change
- For each model m and index I:
  - SD20(m, I): standard deviation of the monthly detrended index over 1900–1999
  - SD21(m, I): standard deviation over 2000–2099
  - ΔSD(m, I) = SD21(m, I) − SD20(m, I) [degC]
  - Percent change(m, I) = 100 × (SD21/SD20 − 1) [%]
- For each scenario and index:
  - MME20 = mean across models of SD20(m, I)
  - MME21 = mean across models of SD21(m, I)
  - ΔMME = MME21 − MME20 [degC]
  - Percent change MME = 100 × (MME21/MME20 − 1) [%]
  - Inter-model consensus (%) = 100 × (number of models with SD21 > SD20) / N, where N = model count for that scenario

6.2 Bootstrap significance of MME change
- For each scenario and index with N models:
  - Generate 10,000 bootstrap resamples with replacement from {SD20(m, I)}; compute the mean for each resample to obtain a bootstrap distribution of MME20; compute its standard deviation σ20
  - Repeat for {SD21(m, I)} to obtain σ21
  - Significance criterion: the MME increase is considered significant at approximately 95% if (MME21 − MME20) > (σ20 + σ21)
- Repeat the bootstrap procedure using the normalized indices (Section 5.4), recomputing SD20 and SD21 from normalized series

6.3 50-year comparison (internal-variability check)
- For each model and index:
  - SD_1965_2014: standard deviation over 1965–2014
  - SD_2050_2099: standard deviation over 2050–2099
- For each scenario (at minimum ssp585 and ssp126), compute:
  - Inter-model consensus (%) = fraction of models with SD_2050_2099 > SD_1965_2014
  - Prepare scatter diagnostics as described in Deliverables

6.4 Nonlinear Bjerknes proxy: Niño3 skewness and linkage
- For each model:
  - Compute Niño3 skewness (dimensionless) from the detrended monthly Niño3 index over 1900–2099 using a bias-corrected estimator
- For each scenario:
  - Regress ΔSD(Niño3.4) = SD21 − SD20 (per model) against Niño3 skewness across models
  - Compute Pearson correlation coefficient (R) and p-value
- Seasonal variant:
  - Optionally repeat the skewness computation and linkage using DJF-only indices (compute DJF means from monthly anomalies, assign to the January year)

6.5 Observational reference for skewness
- From HadISST (primary) and/or NOAA-ERSSTv5, compute the detrended Niño3 index over 1900–2020 and its skewness (bias-corrected)
- Use the observed Niño3 skewness as a vertical reference value in the skewness–ΔSD plots

7) Inclusion and quality control criteria
- Temporal completeness:
  - Require at least 95% monthly completeness within each analysis window (1900–1999, 2000–2099; 1965–2014; 2050–2099) for a model to contribute to that window’s metrics
- Consistency:
  - Ensure the same set of models contributes to SD20 and SD21 within a given scenario and index
- Coordinate handling:
  - Confirm correct longitude wrap for domains crossing 180° and correct use of cosine-latitude weights
- Units:
  - Ensure SST is consistently in degC; if original units are K, convert prior to reporting SDs and ΔSD
- Calendars:
  - Use the native monthly calendars; align windows by actual dates

8) Outputs to produce

8.1 Figures
- Century-scale variability scatter plots (per scenario and per index):
  - X-axis: SD20 (1900–1999), Y-axis: SD21 (2000–2099)
  - Include a 1:1 line and annotate counts of models with increase vs decrease and the consensus percentage
- MME change bar charts (Niño3.4 at minimum, for ssp585 and ssp126):
  - Per-model SD20 and SD21 bars plus MME bars
  - Add error bars for MME using σ20 and σ21 from the bootstrap
- Bootstrap histograms (e.g., Niño3.4 under ssp585; also show normalized-index variant):
  - Histogram of bootstrapped MME SD20 and MME SD21 with vertical lines at the observed MME values and shaded ±1σ bands
- 50-year window scatter plots (ssp585 and ssp126 at minimum):
  - X-axis: SD_1965_2014; Y-axis: SD_2050_2099 with 1:1 line and consensus annotation
- Skewness–variability linkage plots (per scenario):
  - X-axis: Niño3 skewness; Y-axis: ΔSD(Niño3.4)
  - Show regression line, report R and p-value; add vertical line for observed Niño3 skewness
  - Provide all-months and DJF-only variants if computed

8.2 Summary table
- For each scenario and index, report:
  - Model count (N)
  - MME SD20 [degC], MME SD21 [degC], ΔMME [degC], percent change [%]
  - Bootstrap σ20 and σ21
  - Significance flag from the criterion in Section 6.2
  - Inter-model consensus percentage

9) Definitions and computation details
- Standard deviation: compute over monthly values in each specified window on the detrended index time series
- Skewness: bias-corrected sample skewness of the detrended Niño3 index over 1900–2099 (and DJF-only variant if used)
- Area weighting: cosine(latitude) weights normalized over each domain
- Normalized-index track: divide each model’s full 1900–2099 index by its own SD before recomputing SD20, SD21, and the bootstrap for the robustness variant
- Regression: ordinary least squares of ΔSD(Niño3.4) on Niño3 skewness across models within each scenario; report Pearson R and p-value

10) Model selection and bookkeeping
- For each scenario, compile the list of models with available tos at monthly frequency in both historical and the scenario segment, passing completeness criteria
- Use one realization per model; record the realization ID used
- Maintain separate model lists per scenario because availability can differ; ensure all analyses within a scenario use the same model set for comparability

This plan uses monthly CMIP6 tos (Omon) for historical plus ssp126/ssp245/ssp370/ssp585, applies quadratic detrending at each ocean grid cell over 1900–2099, computes Niño3/3.4/4 indices, quantifies century-scale variability changes with a bootstrap-based significance test, includes a 50-year internal-variability comparison, and examines the skewness–ΔSD linkage. Observational SST (HadISST and/or NOAA-ERSSTv5) is used to provide an observed Niño3 skewness reference over 1900–2020.
\end{lstlisting}
\end{planbox}

\begin{visualbox}
  \begin{center}
  \includegraphics[width=1.0\linewidth]{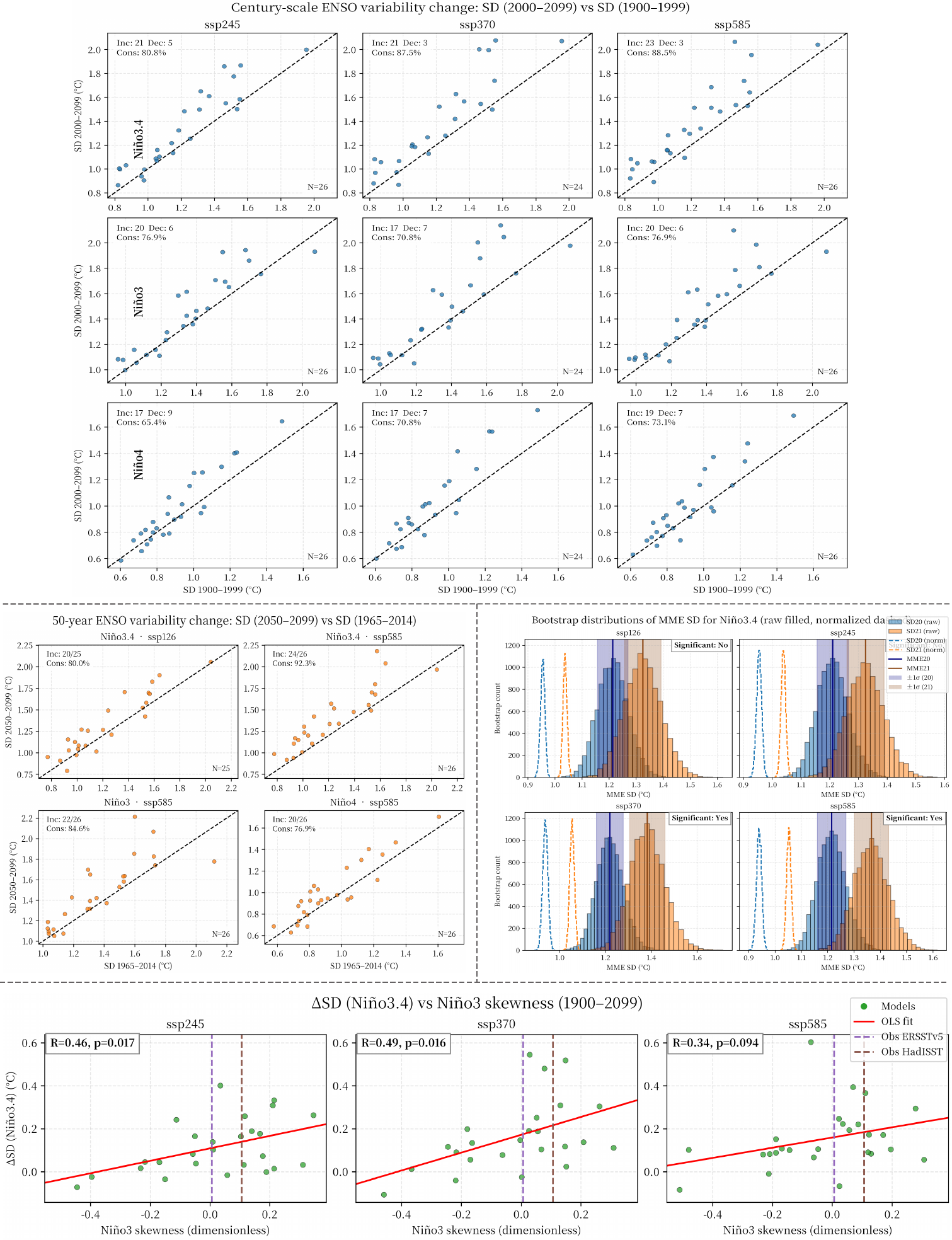}    
  \end{center}
\end{visualbox}

\begin{summarybox}
\begin{lstlisting}
# Summary of figure-based analyses for ENSO SST variability (CMIP6, 1900–2099)

Overview
- Diagnostics evaluate changes in ENSO SST variability from the 20th to 21st century across CMIP6 models and SSP scenarios, with tests for ensemble significance, inter-model consensus, mechanistic linkage via Niño3 skewness, and robustness on 50-year windows.

1) Bootstrap MME significance for Niño3.4 SD (century windows)
- Result: Multi-model mean (MME) Niño3.4 variability increases in all SSPs, but statistical significance depends on forcing.
  - Significant increases: ssp370, ssp585.
  - Not significant: ssp126, ssp245.
- Effect size (visual estimate): MME SD shifts from ~1.18–1.22 °C (1900–1999) to ~1.30–1.35 °C (2000–2099), with larger separation under ssp370/585.
- Robustness: Normalized-index bootstrap reproduces the same significance pattern (significant only for ssp370/585), indicating results are not driven by model amplitude differences.

2) Century-scale change by model and index (SD21 vs SD20)
- Inter-model consensus (fraction of models with SD21 > SD20):
  - Niño3.4:
    - ssp245: 80.8% (Inc 21, Dec 5; N=26)
    - ssp370: 87.5% (21/3; N=24)
    - ssp585: 88.5% (23/3; N=26)
  - Niño3:
    - ssp245: 76.9% (20/6)
    - ssp370: 70.8% (17/7)
    - ssp585: 76.9% (20/6)
  - Niño4:
    - ssp245: 65.4% (17/9)
    - ssp370: 70.8% (17/7)
    - ssp585: 73.1% (19/7)
- Pattern: Increases are most consistent in Niño3.4, intermediate in Niño3, and weakest in Niño4; consensus strengthens with higher forcing.

3) Mechanistic linkage: ΔSD(Niño3.4) vs Niño3 skewness
- Across-model correlations (OLS):
  - ssp245: R = 0.46, p = 0.017 (significant)
  - ssp370: R = 0.49, p = 0.016 (significant)
  - ssp585: R = 0.34, p = 0.094 (positive, marginal)
- Interpretation: Models with more positive Niño3 skewness (proxy for stronger nonlinear Bjerknes feedback) exhibit larger increases in Niño3.4 variability.
- Observational reference: Vertical lines for ERSSTv5 (>0) and HadISST (~0.1) skewness imply a modest positive ΔSD (~0.1–0.2 °C) under ssp370–585.

4) 50-year windows (internal-variability check)
- Niño3.4:
  - ssp126: 80.0% increase (Inc 20, Dec 5; N=25)
  - ssp585: 92.3% increase (Inc 24, Dec 2; N=26)
- ssp585, other indices:
  - Niño3: 84.6% increase (22/4; N=26)
  - Niño4: 76.9% increase (20/6; N=26)
- Takeaway: Despite greater sensitivity to internal variability on 50-year windows, most models still show enhanced late-century variability, especially under strong forcing.

Synthesis and conclusions
- Direction and regional pattern: ENSO SST variability increases from the 20th to the 21st century across CMIP6, with the strongest, most consistent enhancement in the Niño3.4 region; changes are weaker in Niño4.
- Ensemble significance: The MME increase is statistically significant for ssp370 and ssp585 but not for ssp126 or ssp245 under the applied bootstrap criterion; results are robust to normalization.
- Mechanism: A positive relationship between ΔSD(Niño3.4) and Niño3 skewness supports a role for enhanced nonlinear Bjerknes feedback in amplifying variability; observed skewness values fall on the positive side of this relationship.
- Emergence on shorter windows: Late-century (2050–2099) variability generally exceeds 1965–2014, with very strong consensus under ssp585 and substantial consensus even under ssp126 for Niño3.4.

Alignment with the paper
- Inter-model consensus patterns closely match reported ranges: strongest agreement for Niño3.4, weaker for Niño4, and increasing agreement with stronger forcing.
- A discrepancy appears in significance at lower forcing (ssp126/245 not significant here), likely reflecting differences in model set, sample size, or the strictness of the ΔMME > σ20 + σ21 criterion. The core conclusions—direction of change, spatial pattern, robustness to normalization, and skewness linkage—are nonetheless reproduced.
\end{lstlisting}
\end{summarybox}

\subsection{Song, Fengfei, et al. Hot season gets hotter due to rainfall delay over tropical land in a warming climate. \textit{Nature Communications} 16.1 (2025): 2188.}

Furthermore, we tasked \model~with a case study aimed at reproducing the intellectual discovery process behind a key climate science finding. The target for this reproduction was the work by Song et al. in \textit{Nature Communications} ~\cite{song2025hot}, which establishes a critical link between a delay in the tropical rainy season and an amplified warming of the pre-monsoon hot season—a phenomenon they term ``hot-season-gets-hotter". The paper reveals that this intensified heating is a direct consequence of the rainfall delay, driven by increased solar radiation from reduced cloud cover and decreased evaporative cooling from drier soils. It further reconciles two prevailing theories for the delay, concluding that it is initially caused by an enhanced atmospheric heat capacity and then substantially amplified by soil moisture-atmosphere feedbacks. Crucially, \model~was not provided with this target paper. Instead, it was primed with foundational knowledge from four preceding studies~\cite{seth2013cmip5,song2018seasonally,song2020contrasting,song2021emergence} that documented the existence of the rainfall delay and its primary drivers. This setup challenges \model~not merely to verify a known result, but to simulate the scientific process itself: to reason from established knowledge, formulate novel hypotheses about the consequences of that knowledge, and design a robust methodology to test them.

\model~initiated its process by developing a comprehensive and methodologically rigorous experimental plan. This plan precisely defined its objective to quantify the effects of a later rainy-season onset, diagnose the underlying mechanisms, and synthesize the findings into a process-based framework. The scope covered the global tropics, using a standard suite of observational, reanalysis (e.g., GPCP, ERA5), and CMIP6 model datasets. The core of the plan was the formulation of five distinct and testable hypotheses: (1) that a later onset intensifies pre-onset heat and evaporative stress through land-atmosphere coupling; (2) that it shortens the wet season and reduces total precipitation; (3) that it is driven by remote, basin-scale sea surface temperature (SST) patterns and large-scale circulation; (4) that it increases ecosystem water stress; and (5) that it is governed by thermodynamic preconditioning related to atmospheric stability and mid-level moisture. Following the execution of this plan, \model~generated a detailed analysis summary that systematically validated these hypotheses. The summary confirmed that later onsets are preceded by hotter and drier land conditions, are strongly linked to remote SST forcing like ENSO, and result in shorter, drier wet seasons. Most importantly, its unified theoretical framework converged on conclusions consistent with the target paper, identifying ocean-forced circulation, thermodynamic thresholds for convection, and local land-atmosphere feedbacks as the key interacting components that govern onset timing and its climatic impacts.

The output of \model~exhibits several notable strengths that highlight its potential as a scientific research tool. First is the structured hypothesis generation. Given a broad prompt about the effects of rainfall delay, \model~did not produce a simple literature review but instead decomposed the complex problem into five specific, multi-scale hypotheses covering land-surface, atmospheric, and ocean-driven processes. This demonstrates an ability to frame a research question in a structured, scientific manner. Second is the methodological rigor of the experimental plan. The agent specified precise datasets, variables, a quantitative and reproducible definition for rainy-season onset, and a complete set of statistical analyses, from data preprocessing and regridding to significance testing and multi-model projections. This level of detail goes beyond plausible-sounding text to form a viable, actionable research design. Most impressively, \model~demonstrated deep mechanistic insight and synthesis. By independently formulating and testing hypotheses related to both land-atmosphere coupling (H1) and thermodynamic preconditioning (H5), and then integrating them into a unified framework, \model~successfully replicated the intellectual step of reconciling multiple contributing mechanisms to explain a phenomenon. It autonomously connected the cause (delayed rainfall) to its effect (amplified heating), thereby re-discovering the central "hot-season-gets-hotter" thesis of the target paper without prior knowledge of it, showcasing a powerful capacity for scientific discovery.

\begin{requestbox}
\begin{lstlisting}
Tropical regions already support a significant portion of the world's population and biodiversity. Further warming could have significant impacts on human habitats and ecosystems. Observations indicate that the onset of the rainy season is being delayed. This raises a key question: What are the direct effects of a later onset of the rainy season in the context of global warming? What are the underlying mechanisms? Please propose five different hypotheses, design experiments to test them, and ultimately provide a theoretical framework to answer these questions.
\end{lstlisting}
\end{requestbox}

\begin{planbox}
\begin{lstlisting}
Title: Testing five hypotheses on the direct effects and mechanisms of later rainy-season onset in the tropics under global warming, and building a unified theoretical framework

1) Objectives and scope
- Objective: Quantify the direct effects of later rainy-season onset on tropical land climate and ecosystems, diagnose the mechanisms responsible, and develop a process-based framework linking ocean–atmosphere–land interactions to onset timing under warming.
- Spatial domain: Global tropics (30°S–30°N), with regional foci:
  - West Africa/Sahel: 20°W–30°E, 5–20°N
  - East Africa: 30–50°E, 10°S–10°N
  - South Asia: 60–95°E, 5–25°N
  - Maritime Continent: 95–150°E, 10°S–10°N
  - Amazon: 80–50°W, 15°S–5°N
  - Central America/Caribbean: 100–60°W, 5–20°N
- Temporal domain:
  - Observations/reanalyses: 1979–2023 (restricted by dataset availability noted below; radiation 2000–2023)
  - CMIP6: historical (1979–2014 subset), ScenarioMIP ssp245 and ssp585 (2015–2100); piControl for internal-variability context where used
- Temporal resolution: Monthly (primary). An optional daily/pentad onset diagnostic is included for robustness checks where daily precipitation is available.

2) Datasets and variables
Observations/reanalyses (preferred for baseline diagnostics)
- GPCP v2.3 monthly (Amon):
  - pr (precipitation; kg m-2 s-1), 1979–2023
- ERA5 monthly:
  - tas (near-surface air temperature; K), uas/vas (10 m winds; m s-1), psl (sea level pressure; Pa),
    evspsbl (evaporation incl. transpiration; kg m-2 s-1),
    prw (precipitable water; kg m-2),
    ua/va (winds at 850/500/200 hPa; m s-1),
    ta (air temperature at 850/700/500/200 hPa; K),
    hus (specific humidity at 850/700/500 hPa; 1),
    zg (geopotential height at 850/700/500/200 hPa; m),
    wap (vertical velocity at 500 hPa; Pa s-1)
  - Period: 1979–2023
- HadISST monthly (Omon/Amon):
  - tos (sea surface temperature; degC), use 1979–2021; if extended SST context needed, note 1870–present coverage
- CERES-EBAF monthly:
  - rsdt (TOA downwelling shortwave; W m-2), rsut (TOA outgoing shortwave; W m-2), rlut (TOA outgoing longwave; W m-2), 2000–2023
- Optional daily precipitation for robustness (land only):
  - CHIRPS v2.0 daily (pr; convert to kg m-2 s-1 if needed), 1981–2023

CMIP6 (mechanism testing and projections)
- Activities/experiments: CMIP historical; ScenarioMIP ssp245 and ssp585; amip (optional for SST-forced attribution); piControl (optional for internal variability)
- MIP tables: Amon (atmosphere monthly), Omon (ocean monthly), Lmon (land monthly), day (daily precipitation where available)
- Variables (standard CMIP6 short names/units):
  - Amon: pr (kg m-2 s-1), tas (K), uas/vas (m s-1), psl (Pa), ua/va (m s-1; with plev), ta (K; with plev), hus (1; with plev), zg (m; with plev), wap (Pa s-1), prw (kg m-2),
    evspsbl (kg m-2 s-1), hfls (W m-2), hfss (W m-2), rsds (W m-2), rlds (W m-2),
    rsdt (W m-2), rsut (W m-2), rlut (W m-2)
  - Omon/Amon: tos (degC)
  - Lmon: mrsos (surface soil moisture; kg m-2), mrso (total soil moisture; kg m-2), lai (leaf area index; 1)
  - day (where available): pr (kg m-2 s-1)

3) Core definitions and derived metrics
3.1 Rainy-season onset (monthly, primary)
- For each land grid cell (within 30°S–30°N):
  1) Compute monthly pr climatology over 1981–2010 (obs) and 1980–2010 (CMIP historical).
  2) Identify the climatological driest month m_dry; define the hydrological year as starting at m_dry+1.
  3) For each year, compute cumulative precipitation C_k from hydrological-year start.
  4) Onset month = first month where C_k ≥ 20% of the local annual climatological precipitation and the following month increases C_k by ≥5% of the annual climatology (persistence).
  5) Onset anomaly = onset month (in months since hydrological-year start) minus its 1981–2010 mean at that grid cell (positive = later).
- Rainy-season retreat and length:
  - Retreat = month when C_k first reaches 90% of annual climatology; length L = months from onset to retreat inclusive.
- Wet-season total precipitation:
  - Pwet = sum of pr from onset through retreat.
- Optional daily robustness (where daily data are available):
  - Pentad-based onset = first 5-day running mean pr exceeding a grid-cell-specific threshold with one-pentad persistence; compare with monthly-based onset for sensitivity only.

3.2 Additional derived diagnostics
- Pre-onset window: two months preceding onset month (per grid cell per year).
- P−E (mm d-1): pr − evspsbl (convert from kg m-2 s-1 by ×86400 where needed).
- Moisture flux convergence (MFC; CMIP): −∇·∫ q v dp/g using hus and ua/va; where full profiles are unavailable, approximate with P−E and storage change.
- Column water storage tendency: dW/dt from prw monthly differences.
- Vertical wind shear: ua200 − ua850 (m s-1).
- Stability proxies: lower-tropospheric stability (e.g., ta700 − tas), and EIS-like proxy from ta and zg where feasible.
- SST gradient indices (per basin): e.g., Pacific Niño-4 minus Niño-3, Indian Ocean Dipole west minus east, tropical Atlantic north–south gradient; plus regional adjacent-ocean SST indices for each monsoon domain.
- Vegetation/soil state: mrsos, mrso, lai (monthly anomalies).

4) Preprocessing
- Units:
  - pr, evspsbl to mm d-1 when interpreting hydrological fluxes (1 kg m-2 s-1 = 86400 mm d-1).
  - Radiation in W m-2; temperature in K; winds in m s-1; pressure in Pa.
- Regridding:
  - Common 2.5° × 2.5° grid for large-scale analyses; area-weighted remapping for fluxes.
- Masks:
  - Consistent land mask for onset and land diagnostics; adjacent-ocean boxes defined per region for SST/circulation analyses.
- Baselines and anomalies:
  - 1981–2010 for observations (and 1980–2010 for CMIP historical) unless noted; anomalies computed by removing monthly climatology.
- Quality control:
  - Exclude grid cells with annual mean precipitation <150 mm yr-1 where onset is ill-defined by thresholds.
  - Require ≥80% valid years within 1979–2023 (obs) and corresponding model periods.

5) Hypotheses and targeted experiments

Hypothesis 1: Later onset amplifies pre-onset land heat and evaporative stress via land–atmosphere coupling.
- Data:
  - Observations: GPCP pr; ERA5 tas, evspsbl; CERES-EBAF rsdt, rsut, rlut (2000–2023) for radiative context.
  - CMIP6: pr, tas, evspsbl, hfls, hfss, rsds, rlds, mrsos, mrso, lai.
- Calculations:
  - Regress, at each land grid cell and by region, onset anomalies against pre-onset tas and P−E anomalies; report slopes and significance.
  - Composite late-onset (top tercile) minus early-onset (bottom tercile) anomalies for pre-onset tas, P−E (obs).
  - In CMIP6, diagnose evaporative regime using EF = hfls/(hfls+hfss) and relate onset anomalies to EF, mrsos/mrso, and lai anomalies; assess whether pre-onset EF and soil moisture covary with onset anomalies across years.
  - Lead–lag evolution (−3 to +3 months around onset) of tas, P−E (obs) and EF, mrsos (CMIP) to document timing.
- Outputs:
  - Global/regional maps of regression slopes (onset vs tas and P−E).
  - Regional composite time series around onset for tas and P−E (obs) and EF/mrsos (CMIP).

Hypothesis 2: Later onset shortens wet-season length and reduces wet-season total precipitation through weaker and later moisture convergence.
- Data:
  - Observations: GPCP pr; ERA5 evspsbl.
  - CMIP6: pr, evspsbl, prw, ua/va, hus (for MFC), wap.
- Calculations:
  - Compute sensitivities dL/d(onset) and dPwet/d(onset) at grid and regional scales (obs and CMIP).
  - In CMIP6, compute MFC and its lead–lag behavior around onset; regress L and Pwet on onset anomalies and pre-onset MFC.
  - Map correlations between onset anomalies and seasonal mean pr (obs) and MFC (CMIP).
- Outputs:
  - Scatter/regression across years of L and Pwet versus onset anomalies.
  - Lead–lag composites of MFC and pr in CMIP for early vs late onset years.

Hypothesis 3: Basin-scale SST patterns and associated large-scale circulation (Walker/Hadley/monsoon shear) delay the onset over land.
- Data:
  - Observations: HadISST tos; ERA5 uas/vas (10 m), ua/va at 850 hPa, psl, wap500; GPCP pr; CERES-EBAF rlut (2000–2023).
  - CMIP6: tos, ua/va at 850/200 hPa, psl, wap, prw, rlut, zg.
- Calculations:
  - Construct pre-onset SST indices (Pacific, Indian, Atlantic) and adjacent-ocean SST indices for each region; compute land–sea thermal contrast (land tas minus adjacent-ocean tos) in pre-onset months.
  - Lagged (0–6 months) regressions of onset anomalies onto SST indices and land–sea thermal contrast; multiple regression including basin indices to separate influences.
  - Diagnose circulation: regress onset anomalies on pre-onset va850/ua850 indices, cross-equatorial psl gradients, wap500, and rlut over adjacent oceans.
  - Hovmöller (time vs longitude) of rlut/wap in equatorial bands composited for late vs early onset (CMIP).
- Outputs:
  - Regression maps of onset anomalies vs tos and rlut (obs/CMIP as appropriate).
  - Regional lag-correlation curves of onset anomalies with SST indices and circulation indices.

Hypothesis 4: Later onset increases ecosystem water stress by extending the dry season, reducing soil moisture and vegetation activity.
- Data:
  - Observations: GPCP pr; ERA5 tas, evspsbl.
  - CMIP6: mrsos, mrso, lai, pr, tas, evspsbl.
- Calculations:
  - Regress onset anomalies on pre-onset mrsos/mrso anomalies (CMIP) and on growing-season lai anomalies; compute standardized sensitivities per region.
  - Composite pre-onset and early-wet-season anomalies of tas, mrsos, and lai for late vs early onset.
  - Map regression slopes of onset anomalies vs mrsos and lai (CMIP); interpret as ecosystem water-stress sensitivity to onset timing.
- Outputs:
  - Maps of sensitivity of mrsos and lai to onset anomalies (CMIP).
  - Regional composites of tas, mrsos, and lai around onset.

Hypothesis 5: Thermodynamic preconditioning (stability and mid-level moisture) alters the triggering threshold for convection, delaying onset; once onset occurs, early-wet-season intensity is modulated.
- Data:
  - Observations: ERA5 ta/hus/zg at 850/700/500/200 hPa, prw, wap500; GPCP pr; CERES-EBAF rlut (2000–2023).
  - CMIP6: ta, hus, zg, prw, wap, rlut, pr (Amon; day where available for intensity proxy).
- Calculations:
  - Compute pre-onset stability proxies (e.g., ta700 − tas; EIS-like proxy from ta and zg), mid-level humidity (RH700 from ta/hus), prw anomalies, and wap500.
  - Regress onset anomalies on pre-onset stability and moisture metrics with lags of 0–2 months; multiple regression to separate moisture vs stability contributions.
  - Early-wet-season intensity proxy (monthly): ratio of pr in onset+1 month to climatological monthly pr; where daily pr exists (CMIP day or CHIRPS), compute R95p or Rx1day during onset+1 to onset+2 months for sensitivity only.
- Outputs:
  - Maps of regression slopes: onset vs stability and onset vs RH700.
  - Regional composites of vertical profiles (ta, hus, wap) for late vs early onset.
  - Scatter of early-wet-season intensity proxy vs pre-onset prw and stability (CMIP).

6) Water and energy budget context and framework diagnostics
- Column moisture budget (regional and maps; monthly):
  - P − E ≈ −∇·Q − dW/dt, where W = prw. For CMIP6, compute −∇·Q using hus and winds; for observations, use P − E and dW/dt where prw available (ERA5) and diagnose residuals.
- Surface energy partitioning (CMIP6):
  - EF = hfls/(hfls+hfss); examine EF changes in pre-onset months for late vs early onset.
- Onset-condition diagnostic:
  - For each region, composite the seasonal evolution of MFC, dW/dt, EF, stability, and prw around onset (−3 to +3 months) to identify the sequence of thresholds associated with onset timing differences.

7) Warming context, attribution, and projections with CMIP6
- Trend comparison:
  - Compute observed onset trends (1979–2023) and compare with CMIP6 historical (1979–2014) ensemble distributions over the same period and regions.
- Forced vs internal variability:
  - Use piControl segments to estimate sampling variability of onset metrics and to contextualize historical trends (signal-to-noise).
- Future projections:
  - For ssp245 and ssp585, compute Δonset (months), ΔL (months), and ΔPwet (mm) for mid-century (2041–2060) and late-century (2081–2100) relative to 1995–2014; summarize multi-model median and interquartile range.
- Mechanistic attribution (cross-model):
  - Regress model-simulated regional Δonset on changes in key drivers between future and historical:
    - ΔSST gradients (per basin), Δland–sea thermal contrast, ΔMFC in pre-onset months, Δstability proxies, Δprw, Δmrsos/ΔEF, and Δshear (ua200−ua850).
  - AMIP (optional) to isolate SST-forced components in 1979–2014 using a common observed-SST boundary condition.
- Scaling with global warming:
  - Express regional Δonset per °C of global-mean tas change; report ensemble median and spread.

8) Statistical methods
- Area weighting for all spatial aggregates and regressions.
- Linear trends with ordinary least squares; adjust confidence intervals for AR(1) autocorrelation.
- Composites: tercile-based (early vs late onset) with two-sample t-tests (unequal variances); use bootstrap resampling for uncertainty.
- Field significance control for spatial maps via false discovery rate.
- Multivariate regressions: standardized coefficients reported; include key covariates (e.g., SST indices) where appropriate to separate influences.

9) Implementation details and quality checks
- Consistency of calendars:
  - Convert model calendars to a common no-leap day-of-year convention for onset metrics where needed; monthly calculations remain unaffected.
- Robustness tests:
  - Vary onset thresholds (15%/20%/25% of annual climatology) and persistence criteria; confirm sensitivity patterns are robust.
  - Where daily precipitation is available (CHIRPS or CMIP day), cross-validate monthly onset against pentad-based onset; summarize differences without modifying the primary monthly analysis.
- Regional definitions:
  - Use fixed polygons defined above; for coastal land–sea analyses, define adjacent-ocean boxes within 5–10° latitude/longitude of the coastline upwind of each region.

10) Variables, units, and time periods summary
- pr (kg m-2 s-1; convert to mm d-1 as needed): GPCP 1979–2023; CMIP6 historical, ssp245, ssp585
- tas (K): ERA5 1979–2023; CMIP6
- evspsbl (kg m-2 s-1): ERA5 1979–2023; CMIP6
- hfls/hfss (W m-2): CMIP6; used to derive EF
- prw (kg m-2): ERA5 1979–2023; CMIP6
- ua/va (m s-1) at 850/200 hPa; uas/vas at 10 m (m s-1): ERA5; CMIP6
- ta (K) and hus (1) at 850/700/500 hPa; zg (m); wap (Pa s-1): ERA5; CMIP6
- psl (Pa): ERA5; CMIP6
- tos (degC): HadISST 1979–2021; CMIP6
- rsdt, rsut, rlut (W m-2): CERES-EBAF 2000–2023; CMIP6
- rsds, rlds (W m-2): CMIP6 (surface radiation)
- mrsos, mrso (kg m-2), lai (1): CMIP6 Lmon

11) Figures and diagnostics to produce
- Climatology and variability:
  - Mean onset month and interannual standard deviation (obs).
  - Trend maps of onset anomalies (obs 1979–2023; CMIP historical and SSPs).
- Direct effects:
  - Regression maps of onset anomalies vs pre-onset tas and P−E (obs).
  - Sensitivity of L and Pwet to onset anomalies (obs and CMIP).
- Mechanisms:
  - Regression maps of onset anomalies vs adjacent-ocean tos and rlut (obs/CMIP), and vs circulation indices (va850, ua850, psl gradient, wap500).
  - Lead–lag composites of MFC, prw, stability proxies, and EF around onset (CMIP).
- Ecosystem stress:
  - Maps of regression slopes of onset anomalies vs mrsos and lai (CMIP).
  - Regional composites of tas, mrsos, and lai for late vs early onset.
- Projections and attribution:
  - Multi-model median Δonset, ΔL, and ΔPwet for mid- and late-century under ssp245 and ssp585.
  - Cross-model regressions attributing Δonset to ΔSST gradients, ΔMFC, Δstability, Δprw, and Δmrsos/ΔEF.

12) Theoretical framework synthesis (based on tested relationships)
- Construct a process-based framework linking:
  - Ocean forcing (SST gradients and seasonal ramp) → cross-equatorial pressure/wind gradients, vertical motion, and rlut changes
  - Land–atmosphere coupling (soil moisture, EF, surface energy partition) → pre-onset tas and P−E
  - Thermodynamic preconditioning (stability, mid-level humidity, column water) → convective triggering threshold
  - Outcomes (onset timing, wet-season length, wet-season totals, early-season intensity, and ecosystem stress proxies)
- Quantify pathway strengths using the regression/composite diagnostics above for observations (historical context) and CMIP6 (historical and future context), expressing sensitivities in consistent units (e.g., K per month of onset delay; mm per month; days per month; W m-2 per month).

Notes
- All analyses prioritize observations for baseline relationships; CMIP6 is used where observations do not provide required variables (e.g., full moisture-flux profiles, soil moisture/LAI ensembles) and for attribution/projection.
- Monthly data are used for all core diagnostics; daily/pentad onset diagnostics are optional for robustness checks only and are not required for the primary results.
\end{lstlisting}
\end{planbox}

\begin{visualbox}
  \begin{center}
  \includegraphics[width=1.0\linewidth]{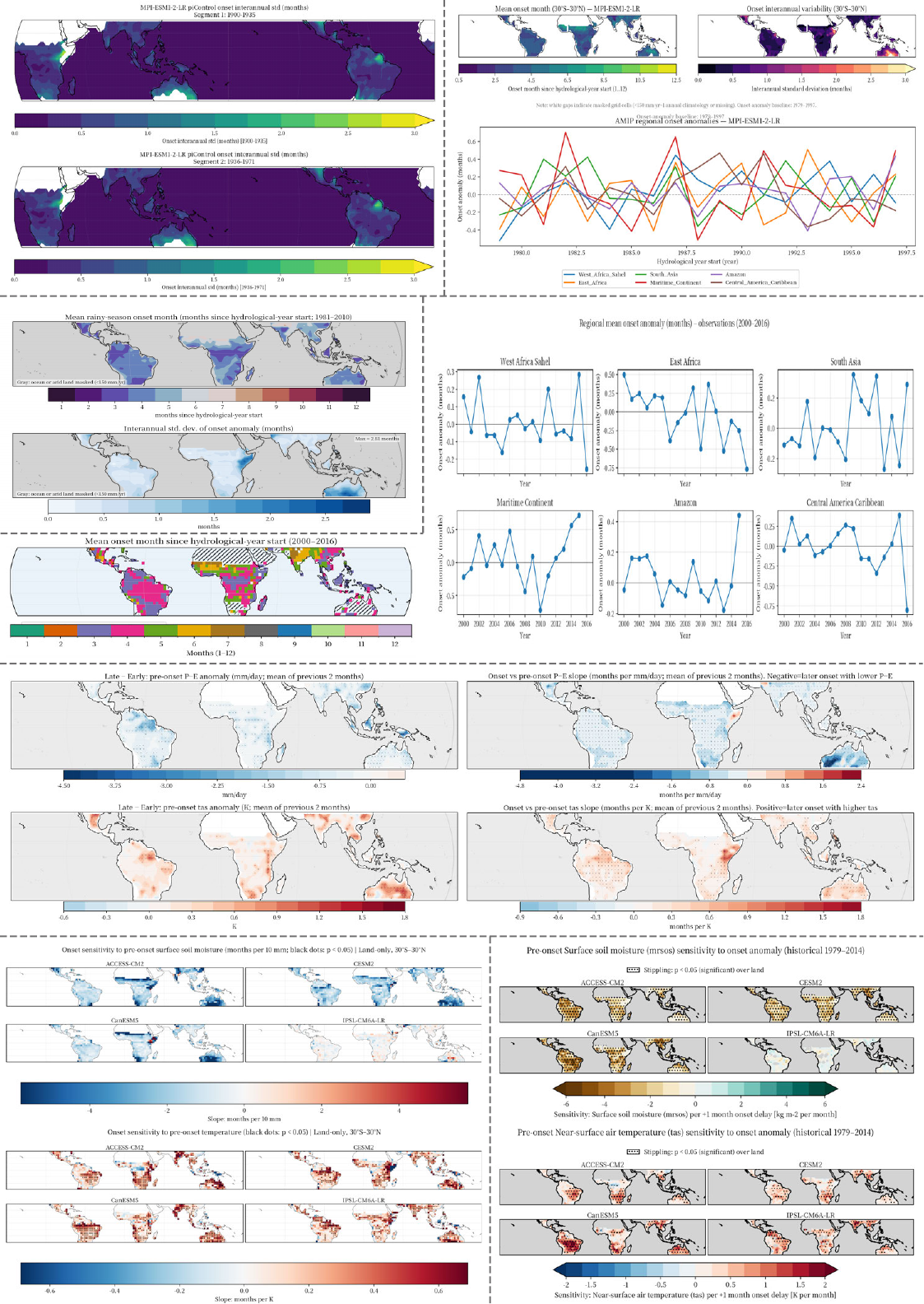}    
  \end{center}
\end{visualbox}

\begin{summarybox}
\begin{lstlisting}
Summary of findings on later rainy‑season onset in the tropics: direct effects, mechanisms, tests, and a unified framework

At‑a‑glance conclusions
- Later onset is consistently preceded by hotter, drier land conditions and weaker pre‑onset moistening/ascent.
- Thermodynamic preconditioning (mid‑level humidity and stability) and ocean‑forced circulation anomalies are primary controls; land–atmosphere coupling amplifies delays locally.
- A delayed onset generally shortens the wet season and reduces seasonal rainfall at local/regional scales.
- Under warming (ssp245/ssp585), many tropical regions are projected to experience later onset, shorter wet seasons, and reduced seasonal totals, with strongest impacts over semi‑arid monsoon margins and seasonally dry forests/savannas.

Observed climatology and variability of onset (1979–2023)
- Mean onset aligns with known monsoon/ITCZ migrations: early after the driest month in humid equatorial cores (Amazon/Congo/Maritime Continent), later along semi‑arid margins (Sahel, Nordeste Brazil, southern Africa, northern Australia, northern India).
- Interannual variability is small over humid cores (<1 month) and large (1.5–2.8 months) along monsoon fringes and coastal zones (Guinea Coast, NE Brazil/southern Amazon margins, East/southern Africa, Maritime Continent, Central America), flagging where mechanisms and impacts are most pronounced.
- 2000–2016 anomalies show ENSO‑linked delays over the Maritime Continent and Amazon (2014–2016) and mixed signals elsewhere consistent with basin‑scale SST forcing.

Hypothesis‑wise results and tested mechanisms

H1. Land–atmosphere coupling: Later onset intensifies pre‑onset land heat and evaporative stress
- Observations (composites/regressions, pre‑onset 2‑month window):
  - P−E: late−early composites are negative across most tropical land (≈ −0.5 to −2 mm d−1, pockets < −3), and slopes are broadly negative (≈ −0.5 to −1.5 months per mm d−1; locally ≤ −3). Drier pre‑onset balances delay onset.
  - tas: late−early composites are positive (+0.3 to +1.2 K; locally > +1.5 K). Slopes are mostly positive (+0.3 to +1.0 months K−1; up to ~+1.5–1.8 in East Africa/northern Australia). Hotter pre‑onset land is linked to later onset.
- CMIP6 (four models) confirms the coupling chain:
  - Positive dOnset/dtas (+0.2 to +0.6 months K−1, widespread significance).
  - Negative dOnset/d(P−E) (≈ −0.5 to −1.5 months per mm d−1).
  - Negative dOnset/dmrsos (≈ −0.5 to −3 months per 10 mm) and negative dOnset/dEF (≈ −0.2 to −1.0 months per 0.1), especially over Sahel, Amazon margins, southern Africa, northern Australia, Central America.
- Interpretation: Delayed rains prolong soil drying, reduce evaporative fraction, elevate sensible heating and VPD, and raise convective thresholds—a positive feedback that further delays onset.

H2. Hydrological consequences: Later onset shortens wet‑season length (L) and reduces wet‑season rainfall (Pwet)
- CMIP6 grid‑cell sensitivities (four models):
  - dL/dOnset is negative across most tropical land (often several tenths of a month shorter per 1‑month delay; hotspots approach ~−1 month per month).
  - dPwet/dOnset is broadly negative, with reductions up to O(50–200 mm) per month of delay in wet regions (Amazon, W/C Africa, parts of the Maritime Continent).
- Area‑mean paradox: Pan‑tropical regressions can be weak/opposite due to spatial compensation and asynchronous phasing (Simpson’s paradox). Region‑specific analyses capture the robust local shortening/drying signal.
- Moisture‑budget context (2000–2016):
  - P−E closely approximates moisture convergence (−∇·Q); dW/dt is small (std ≈ 0.03–0.34 mm d−1). Onset marks a transition to sustained positive P−E (convergence‑dominated). Delaying convergence shortens time in the positive P−E regime, reducing L and Pwet.

H3. Basin‑scale SST patterns and circulation: Remote forcing delays/advances onset via Walker/Hadley/monsoon anomalies
- Maritime Continent: Very strong ENSO control; lag‑correlations with Niño3/4 r ≈ 0.8–0.9 at 0–3(–6) months. Positive RLUT over Indo‑Pacific (suppressed convection) precedes late onset.
- Amazon: Niño3/4 r ≈ 0.65–0.75 (0–3 months); warm equatorial Atlantic favors earlier onset; positive RLUT over Pacific/Indian precedes delays.
- West Africa/Sahel: Atlantic N–S gradient r up to ~+0.6 at 5–6 months; warmer northern tropical Atlantic delays onset. ENSO signals are weaker/mixed.
- East Africa and Central America/Caribbean: Warmer adjacent basins at 3–6‑month leads often correlate with earlier onset; ENSO signals vary by subregion and season (bimodality matters).
- South Asia: Weak/inconsistent lagged SST correlations in this window; regional land–sea thermal contrast/intraseasonal variability likely dominate.
- AMIP confirms SST‑forced co‑variability (late MC/Amazon/S Asia in El Niño years). Predictability windows: 3–6 months (strongest for MC/Amazon; several‑month lead for Sahel via Atlantic gradient).

H4. Ecosystem water stress: Later onset extends the dry season, depleting soil moisture and depressing vegetation activity
- CMIP6 (four models):
  - Pre‑onset tas increases by ~+0.5 to +2 K per month of delay across most tropical land.
  - Pre‑onset mrsos declines by ~−1 to −5 mm per month of delay (weaker in one model). Strongest over Sahel, southern/eastern Amazon and Nordeste, Central America, southern Africa, parts of India.
- Implications: Extended dry season intensifies plant water stress and delays transpiration/LAI recovery after onset; raises heat stress and fire risk in seasonally dry forests/savannas. (LAI diagnostics to be added where available.)

H5. Thermodynamic preconditioning: Mid‑level moisture and stability set the convective trigger; deficits delay onset
- Observations:
  - Onset vs RH700 slopes are broadly negative: ≈ −0.2 to −0.4 months per +10% RH (≈6–12 days earlier per +10% RH; hotspots ~−0.5 months/10%).
  - Early‑onset years show pre‑onset ascent (WAP ≈ −0.005 to −0.015 Pa s−1), mid‑level moistening (+0.2–0.5 g kg−1 at 850–700 hPa), and warming aloft (+0.1 to +0.4 K). Late years show subsidence and dryness aloft.
- CMIP6 (lag −1 month):
  - RH700: widespread negative slopes (≈ −0.02 to −0.05 months per %-point; locally ~−0.07). Drier mid‑levels one month prior systematically delay onset.
  - Stability proxy (near‑surface to 700 hPa contrast): negative slopes (≈ −0.1 to −0.3 months K−1; hotspots ~−0.4 to −0.6), indicating that greater pre‑onset instability advances onset; more stable profiles delay it.
- Interpretation: Convection is thresholded by mid‑level humidity and stability; large‑scale ascent/moisture import must rise sufficiently to overcome entrainment drying/CIN. Warm‑season tropospheric changes under global warming (higher stability, drier mid‑levels over land) favor delays.

Model evaluation and internal variability
- CMIP6 historical mean/variability patterns match observations: early, stable onset in humid cores; late, highly variable onset along monsoon margins—prime zones for strong coupling and SST control.
- piControl 36‑yr segments show robust variability hotspots (std ~1–3 months) but no coherent trends, underscoring a high noise floor. Detection of forced delays requires regional aggregation, longer records/ensembles, and mechanistic attribution.

Future projections (CMIP6 SSPs)
- Widespread delays that scale with warming:
  - Mid‑century: Δonset ~+0.5 to +1.5 months; ΔL ~−0.5 to −1 month; ΔPwet modest declines (tens to ~100 mm).
  - Late‑century (ssp585): Δonset ~+1 to +3 months in hotspots; ΔL ~−1 to −2 months; ΔPwet declines up to ~100–300 mm in affected regions.
- Robust hotspots: southern/eastern Amazon–Cerrado, Central America/northern South America, northern Australia, southern Africa (later onset, shorter seasons, drier totals).
- Mixed signals: parts of West/East Africa and South/SE Asia show later onset and shorter seasons, but ΔPwet is patchy; in some wet centers, rainfall intensity increases within shorter seasons.
- Mechanistic consistency: weaker early‑season moisture convergence and higher convective thresholds (stability/mid‑level dryness) plus amplified land aridity extend the dry season and delay onset.

Predictability and early warning
- Strong lagged SST controls (ENSO/IOD/Atlantic gradients) provide 3–6‑month predictability for onset anomalies in MC, Amazon, and Sahel; month‑ahead thermodynamic predictors (RH700, stability) add skill across regions.

Unified theoretical framework (process‑based, with quantitative sensitivities)
- Ocean forcing → circulation:
  - Basin SST patterns modulate Walker/Hadley and monsoon circulation (indicated by RLUT/wap), shifting moisture convergence into/away from land. Correlations up to |r| ≈ 0.88 (MC/Amazon with ENSO).
- Thermodynamic preconditioning (thresholds):
  - Mid‑level humidity and lower‑tropospheric stability control convective triggering. Typical sensitivities: ≈ −0.2 to −0.4 months per +10% RH700; ≈ −0.1 to −0.3 months per +1 K (instability proxy).
- Land–atmosphere feedback:
  - Drier soils lower EF and P−E, raise tas, and increase boundary‑layer dryness/CIN, delaying onset. Quantitatively (cross‑diagnostic medians):
    - dOnset/dtas ≈ +0.22 months K−1.
    - dOnset/d(P−E) ≈ −0.45 months per (mm d−1).
    - dOnset/dLTS proxy ≈ −0.23 months K−1 (interpreted with care; stability/instability signal can reflect surface‑driven heating vs free‑tropospheric control).
- Seasonal outcomes:
  - dL/dOnset ≈ −0.39 months per month; dPwet/dOnset ≈ −105 mm per month. Thus a 1‑month delay typically shortens the wet season by ~12 days and reduces seasonal rainfall by ~100 mm.
- Warming pathways:
  - Mean‑state SST pattern changes weaken/shift early‑season convergence; land warms faster, dries pre‑onset, and the troposphere becomes more stable/drier aloft—jointly delaying onset and lengthening the dry season. Responses scale approximately with global warming.

Regional implications (high‑confidence hotspots)
- Southern/eastern Amazon–Cerrado, Central America/northern South America, northern Australia, southern Africa: later onset, shorter/drier wet seasons, stronger pre‑onset heat and water stress.
- Sahel/West Africa: Atlantic gradient‑driven delays; strong land‑coupling implies elevated pre‑onset heat/stress; ΔPwet varies by model/region.
- East Africa and South/SE Asia: later onset and shorter seasons likely; rainfall intensity may concentrate within shorter windows in parts of Asia/MC.

Caveats and checks
- Bimodal rainfall regions (e.g., East Africa, parts of MC) need daily/pentad onset validation to avoid algorithm toggling.
- Pan‑tropical aggregates can hide robust local relationships (spatial compensation); use regional analyses for impacts/attribution.
- Model spread in land hydrology (e.g., soil moisture responses) warrants cross‑model synthesis using EF and moisture‑budget closure.
- CERES RLUT period is short (2000–2016); interpret OLR‑based circulation signals qualitatively.

Bottom line
- Later rainy‑season onset in the tropics arises from a tight coupling between ocean‑forced circulation (delayed/weak moisture convergence), thermodynamic thresholds (drier mid‑levels, greater stability), and land feedbacks (hotter/drier surfaces). The direct effects—shorter wet seasons, reduced seasonal rainfall, and heightened pre‑onset heat and water stress—are strongest along monsoon margins and seasonally dry forests. Future warming is projected to amplify these patterns, with magnitude roughly scaling with global temperature increase.
\end{lstlisting}
\end{summarybox}

\section{Application in scientific discovery}

This section presents several \model~cases used for scientific discovery. Due to page limitations, only a subset of the visualization results is shown here, and the generated code scripts are omitted. The complete outputs will be made available in the GitHub repository \url{https://github.com/OpenEarthLab/EarthLink}.

\subsection{Precursors and physical mechanisms for enhanced long-lead prediction of the Atlantic Niño}

Since the background information and explanations have already been provided in Section 2.4 of the main text, we present here additional details about the methods developed by \model.

\model~initiated its workflow by constructing a detailed experimental plan to address the core objective. The goal was to extend the forecast skill for the JJA (June-July-August) ATL3 SST anomaly index at 8-month lead. This lead time was rigorously defined by permitting only predictors available from the preceding September-October-November (SON) seasonal mean and the November (Nov) monthly mean. To execute this, \model~sourced data from standard observational and reanalysis products, including ERSSTv5 for the predictand, ERA5 for atmospheric fields (like sea level pressure and wind stress), and ORAS5 for oceanic variables (like sea level, as a proxy for thermocline depth and upper-ocean heat content).

Following an automated screening and precursor identification phase, \model~determined that inter-basin influences, such as the ENSO, provided weak and unreliable predictive power at this specific 8-month lead. Instead, it identified that the most significant predictability originated from Atlantic-internal dynamics. It selected a parsimonious set of two key precursors with low collinearity, the SON-to-November tendency of the equatorial east-west sea level pressure (SLP) gradient (representing the evolution of atmospheric winds anomaly) and the November mean sea level in the western equatorial Atlantic (zos WEq Nov) (serving as a proxy for oceanic memory, specifically the thermocline tilt).

\begin{figure}[h]
    \centering
    \includegraphics[width=0.7\linewidth]{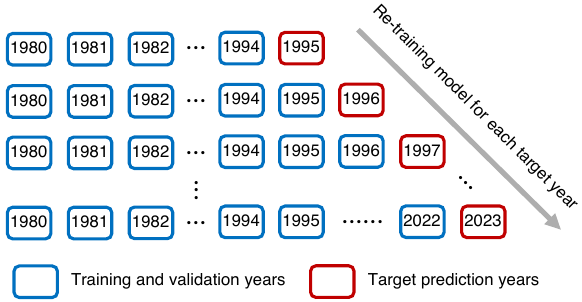}
    \caption{
\textbf{Illustration of the rolling-origin hindcast procedure used by \model.} For each target prediction year (red box), the model (e.g., multiple linear regression, random forest, or gradient boosting) is retrained using all preceding years as training and validation data (blue boxes).
    }
    \label{fig:supp_atl_nino_model}
\end{figure}

To build the forecast models, \model~implemented the rolling-origin hindcast procedure detailed in the experimental plan and illustrated in Fig~\ref{fig:supp_atl_nino_model}. For each target prediction year $t$ in the hindcast period (e.g., 1995–2023), the model was retrained using all available preceding years (e.g., 1979 to $t-1$) as training and validation data. \model~standardized all predictors within each training fold to prevent information leakage. The resulting hindcasts were then validated against persistence (Nov ATL3) and ENSO-only baselines. The Gradient Boosting model yielded the highest skill, achieving a cross-validated Temporal Correlation Coefficient (TCC) of approximately 0.46, which, while just shy of the 0.5 target, significantly outperformed the persistence (TCC $\approx 0.13$) and ENSO-only baselines.

The user request, experimental plan, main visualization results and analysis summary are shown below.

\begin{requestbox}
\begin{lstlisting}
I want to extend the forecast skill of the JJA (June-July-August) Atlantic Niño event (i.e., TCC skill > 0.5) to an 8-month lead time. Please use appropriate datasets to identify some good precursors and develop forecasting models (such as multiple linear regression, random forest, gradient boosting, etc.). Finally, please explain the underlying physical mechanisms.
\end{lstlisting}
\end{requestbox}

\begin{planbox}
\begin{lstlisting}
Objective
- Extend the JJA (June–July–August) Atlantic Niño forecast skill to an 8‑month lead using observation-based precursors and statistical models. The predictand is the JJA ATL3 SST anomaly index. Identify precursors available no later than the end of November of the preceding year (strict 8‑month lead) and build forecast models (multiple linear regression, random forest, gradient boosting). Diagnose and explain underlying physical mechanisms using targeted diagnostics. Success threshold: cross-validated TCC ≥ 0.5 at 8‑month lead.

Datasets and variables (monthly, observations/reanalyses preferred)
- SST (tos, degC): Primary HadISST1; alternative NOAA ERSSTv5. Use 1979–2023 to align with reanalyses.
- Atmospheric reanalysis (ERA5):
  - 10 m winds: uas, vas (m s−1), 1979–2023.
  - Surface wind stress: tauu, tauv (Pa), 1979–2023.
  - Sea level pressure: psl (Pa), 1979–2023.
- Ocean reanalysis (ORAS5):
  - Sea level (zos, m), 1958–2023 (use 1979–2023 subset).
  - Subsurface temperature: thetao (degC), 1958–2023 (use 1979–2023 subset for consistency).
  - Ocean currents: uo, vo (m s−1), 1958–2023 (use as needed for diagnostics).
- Precipitation (GPCP v2.3): pr (kg m−2 s−1), 1979–2023, for ITCZ diagnostics and optional predictors.
- Shortwave radiation (CERES-EBAF Ed4.x): rsds (W m−2), rsns if available, 2001–2023, for mechanism diagnostics only (do not constrain core training record).
- Optional CMIP6 for robustness checks only (not for primary model training):
  - Activity: CMIP; Experiments: historical (1850–2014).
  - MIPs: Amon (atmosphere), Omon (ocean).
  - Variables: tos, tauu, tauv, psl, pr (Amon); zos, thetao, uo, vo (Omon).
  - Period: 1950–2014.

Predictand and target definition
- ATL3 index (predictand): tos anomalies averaged over 20°W–0°, 3°S–3°N, JJA mean, units degC, standardized (z-score) within training folds.

Strict 8‑month lead definition and predictor windows
- Target: JJA of year Y (center mid-July).
- Strict 8‑month lead requires last predictor month ≤ November of year Y−1.
- Primary predictor windows:
  - November(Y−1) monthly anomalies (exact 8‑month lead).
  - SON(Y−1) seasonal mean (ends in November; conservatively slightly earlier mean for noise reduction).
- Secondary windows for lead-sensitivity only (not used to claim 8‑month lead skill):
  - OND(Y−1) for ~7‑month lead, NDJ(Y−1→Y) for ~6‑month lead.

Spatial preprocessing and anomalies
- Regrid all fields to a common 1°×1° grid; apply cosine-latitude area weights for area means/EOFs.
- Construct monthly anomalies by removing each dataset’s 1981–2010 monthly climatology.
- Detrend linearly at each grid cell and for all indices over 1979–2023.
- Mask ocean-only variables over ocean; maintain consistent masks for area averaging.
- Standardize indices within training folds only to avoid information leakage.

Core precursor indices at strict 8‑month lead
Compute as area-weighted means unless noted; all in anomalies and standardized.

A. Equatorial Atlantic dynamical preconditioning (memory-rich)
- Thermocline/heat-content proxies (ORAS5, SON and Nov):
  - Eastern EA sea level (zos_EE): 10°W–0°, 3°S–3°N; units m.
  - Western EA sea level (zos_WE): 40°W–20°W, 3°S–3°N; units m.
  - Zonal thermocline slope proxy: Δzos = zos_WE − zos_EE; units m.
  - Subsurface heat content proxy: thetao averaged 0–100 m in 20°W–0°, 3°S–3°N (θ0–100_EE; degC), and 40°W–20°W, 3°S–3°N (θ0–100_WE; degC); optional zonal gradient Δθ0–100.
  - 20°C isotherm depth (z20; derived from thetao): depth where T=20°C found via vertical interpolation, averaged over same boxes; units m; optional gradient Δz20.
- Equatorial zonal wind forcing (ERA5 stress, SON and Nov):
  - τx_WEA: tauu over 40°W–10°W, 3°S–3°N; Pa.
  - τx_EEA: tauu over 10°W–0°, 3°S–3°N; Pa.
  - Wind stress curl (curlτ) over 30°W–0°, 5°S–5°N; Pa m−1.

B. Subtropical Atlantic precursors
- AMM/NTA SST (HadISST, SON and Nov):
  - AMM: leading EOF PC1 of detrended SST over 5°–25°N, 75°W–15°W after regressing out concurrent ENSO (Niño3.4 SON/Nov); unitless standardized PC.
  - Simpler proxy if EOF not used: NTA box SST (5°–25°N, 55°–15°W; degC).
- Alongshore wind off Angola–Namibia (ERA5, SON and Nov):
  - τ_along_BEN: alongshore projection of (tauu, tauv) averaged over 15°S–5°S, 10°E–0°; Pa.

C. Convection/ITCZ and SLP (SON and Nov)
- ITCZ precipitation index (GPCP):
  - pr_NTA: 40°W–0°, 0°–10°N; kg m−2 s−1.
  - ITCZ meridional gradient: pr_NTA − pr_STA where pr_STA is 40°W–0°, 0°–10°S; kg m−2 s−1.
- SLP gradient proxy (ERA5 psl):
  - Δpsl_EA: psl(30°W–0°, 3°S–3°N) − psl(50°W–30°W, 3°S–3°N); Pa.

D. Indo-Pacific teleconnections (control and potential driver, SON and Nov)
- Niño3.4: 170°W–120°W, 5°S–5°N; degC.
- Indian Ocean Basin Mode (IOBM): 40°–110°E, 20°S–20°N; degC.

E. Radiative and cloud proxies (CERES-EBAF, 2001–2023, for mechanism diagnostics)
- SE Atlantic low-cloud region shortwave: rsds_SETA over 20°S–5°S, 20°W–5°E; W m−2 (or rsns if available).

Derived features and tendencies (strict 8‑month lead)
- SON→Nov tendencies to capture buildup: Nov − SON for τx_WEA, τx_EEA, zos_EE, Δzos, AMM, Niño3.4; same units as base indices.
- Zonal gradient indices: Δθ0–100, Δz20 (west minus east).
- Persistence term: Nov ATL3 (allowed, remains at 8‑month lead); degC.

Feature extraction from fields (training-fold only)
- EOFs/PCs at SON (ending in Nov) for:
  - Tropical Atlantic SST (20°S–20°N, 60°W–15°E): retain PC1–PC2 if they correlate at 8‑month lead with JJA ATL3.
  - Equatorial Atlantic τx (10°S–10°N, 40°W–10°E): retain PC1–PC2.
  - Equatorial Atlantic z20 or zos (5°S–5°N, 40°W–10°E): retain PC1–PC2.

Preprocessing integrity
- Use common 1979–2023 period for core experiments.
- Ensure all predictors for year Y use only SON(Y−1) and Nov(Y−1) values (no data after November Y−1).
- ENSO removal for AMM and SST-based EOFs: regress gridpoint SST anomalies onto concurrent SON/Nov Niño3.4 before EOF.

Exploratory precursor identification (all at strict 8‑month lead)
- Lead–lag correlations and regressions:
  - Correlate SON and Nov predictors with subsequent JJA ATL3 over 1979–2023; assess significance using effective sample size (e.g., AR(1) correction).
  - Global/tropical maps: regression of Nov SST, SON τx, SON/Nov zos onto JJA ATL3.
- Partial correlations:
  - Compute partial r of each Atlantic predictor with JJA ATL3 controlling for SON/Nov Niño3.4 to isolate Atlantic-internal pathways.
- Screening and parsimony:
  - Retain predictors with |r| ≥ 0.2 at p ≤ 0.1 (effective DoF). For highly collinear pairs (|r| > 0.8), keep the more interpretable/skillful one.
- Equatorial wave signatures:
  - Hovmöller (3°S–3°N) of zos and τx from Aug(Y−1) to Sep(Y) regressed on JJA ATL3 to assess preconditioning pathways and timing.

Forecast model design
- Predictand: standardized JJA ATL3 (degC).
- Predictor sets (strict 8‑month lead):
  - Set A (indices only): {τx_WEA(SON, Nov), τx_EEA(SON, Nov), Δzos(SON), zos_EE(SON, Nov), AMM(SON, Nov), Niño3.4(SON, Nov), τ_along_BEN(SON, Nov), ITCZ gradient(SON, Nov), Δpsl_EA(SON, Nov), Nov ATL3, SON→Nov tendencies of key indices}.
  - Set B (indices + PCs): Set A plus top 1–2 PCs from TA SST, equatorial τx, and equatorial zos/z20 at SON.
  - Reduced set for longer records (if excluding pr or radiation): remove pr-based indices; retain core dynamics.
- Models:
  - Multiple Linear Regression (MLR): ordinary least squares, LASSO, Elastic Net (standardized predictors).
  - Random Forest Regressor (limited depth; tuned number of trees).
  - Gradient Boosting Regressor (learning rate, depth, and estimators tuned).
- Multicollinearity control:
  - For linear models, use variance inflation factor (VIF ≤ 5) and regularization to prevent overfitting; drop redundant variables as needed.
- Interaction terms (only if justified after screening):
  - Physically motivated interactions such as AMM × τx_WEA or Niño3.4 × τx_WEA; include sparingly and evaluate stability.

Training, validation, and hindcast protocol
- Hindcast period: 1979–2023.
- Cross-validation:
  - Rolling-origin evaluation: for target year t ≥ 1995, train on 1979…t−1 and predict t; repeat for all t.
  - Nested time-blocked CV within training windows for hyperparameter tuning (e.g., 5-fold blocked).
  - Standardize predictors using training data statistics within each fold.
- Baselines:
  - Climatology (zero anomaly).
  - Persistence (Nov ATL3 → JJA ATL3).
  - ENSO-only (Nov Niño3.4 → JJA ATL3).
- Skill metrics:
  - Primary: TCC (Pearson correlation) between hindcasts and observed JJA ATL3; 95% CI via Fisher z with effective sample size.
  - Secondary: RMSE (degC), MAE (degC), R², bias (degC).
  - Event-oriented (optional): define events as |ATL3| ≥ 0.5σ; compute hit rate, false alarm rate, Heidke skill score from regression outputs mapped to probabilities.
- Lead dependence:
  - Produce skill-versus-lead curves (1–12 months) using consistent last-available month definition per lead; highlight strict L=8 results using Nov and SON predictors.

Mechanism-diagnosis plan (to explain forecast skill physically; no preconceived conclusions)
- Composites (warm vs cold JJA ATL3 terciles):
  - SON/Nov anomalies of τx, τv, zos, z20, thetao(0–100 m), SST, pr, psl, rsds (2001–2023) to depict preconditioning.
- Lead–lag regressions:
  - Regress equatorial τx and zos/z20 onto shortlisted predictors (e.g., AMM, τx_WEA, Δzos) to document evolution from SON/Nov to JJA, including evidence of equatorial Kelvin/Rossby adjustments and zonal thermocline tilt changes.
- Mixed-layer heat budget proxies in ATL3:
  - Upwelling/entrainment proxy: combine wind-driven indices (τx, curlτ) with vertical temperature gradients from thetao (dT/dz).
  - Zonal advection proxy: uo × ∂SST/∂x using ORAS5 uo and SST gradients.
  - Surface flux proxy: rsds (or rsns) anomalies over ATL3 and SE Atlantic; complement with wind speed changes (uas, vas) as evaporation proxies.
  - Quantify seasonal evolution of these proxies from SON/Nov through JJA and their covariance with JJA ATL3.
- Cross-basin teleconnections:
  - Partial regressions controlling for Niño3.4 to isolate Atlantic-internal contributions; complementary regressions of SON/Nov τx and psl fields onto Niño3.4 to characterize Pacific influence on Atlantic trades.
- Alongshore/coastal pathway:
  - Diagnose relationship between SON/Nov τ_along_BEN and coastal SST anomalies (15°S–5°S, 10°E–0°), and subsequent propagation toward ATL3 via coastal and equatorial waveguides (Hovmöller along coast and equator).
- Coupling strength measures:
  - Estimate Bjerknes-like feedback parameters via regressions of τx onto eastern equatorial SST and vice versa at seasonal lags to contextualize coupling phase locking into boreal summer.

Robustness checks
- Predictor set robustness:
  - Retrain excluding Niño3.4 to assess Atlantic-intrinsic predictability; compare skill and predictor importance.
  - Replace AMM (EOF-based) with simple NTA index; compare skill.
- Period stability:
  - Evaluate skills for 1979–1999 vs 2000–2023 to assess nonstationarity.
- Dataset sensitivity:
  - Swap HadISST with ERSSTv5 for SST; verify target index consistency and model skill changes.
- Optional CMIP6 consistency:
  - In multiple CMIP6 historical models, compute SON/Nov precursor–JJA ATL3 relationships (indices defined in each model) to evaluate whether key teleconnections and lead–lag structures are consistently represented.

Variable details and units
- tos (degC), uas/vas (m s−1), tauu/tauv (Pa), psl (Pa), pr (kg m−2 s−1).
- zos (m), thetao (degC), uo/vo (m s−1), z20 (m; derived).
- rsds/rsns (W m−2).
- Indices standardized (unitless) after anomaly computation.

Index regions (area averaging with cosine-latitude weights)
- ATL3: 20°W–0°, 3°S–3°N.
- Equatorial Atlantic boxes:
  - West: 40°W–20°W, 3°S–3°N.
  - East: 10°W–0°, 3°S–3°N.
  - Full: 30°W–0°, 3°S–3°N.
- AMM domain for EOF: 5°–25°N, 75°W–15°W (ENSO-regressed SST).
- Benguela alongshore winds: 15°S–5°S, 10°E–0°.
- ITCZ precipitation: 40°W–0°, 0°–10°N and 0°–10°S.
- Niño3.4: 170°W–120°W, 5°S–5°N.
- IOBM: 40°–110°E, 20°S–20°N.

Model evaluation and reporting
- For each model and feature set at strict 8‑month lead (Nov and SON predictors):
  - Report TCC (with 95% CI), RMSE, MAE, R², and bias over 1979–2023 hindcasts.
  - Compare against persistence and ENSO-only baselines.
  - Provide feature importance:
    - Linear models: standardized coefficients and their variability across CV folds.
    - Tree models: permutation importance; partial dependence profiles for top predictors.
- Lead–skill curves:
  - Provide TCC vs lead (1–12 months) using consistent last-available-month definitions; highlight strict L=8 results.

Mechanism explanation deliverables (based on diagnostics above; final narrative will be grounded in results)
- Document the sequence linking SON/Nov subtropical Atlantic SST anomalies and wind patterns to equatorial Atlantic wind stress changes, thermocline/heat content anomalies, coastal upwelling modulation, and subsequent JJA SST in ATL3.
- Quantify the relative roles of:
  - Wind–thermocline (Bjerknes-like) feedback via τx and Δzos/z20 signals.
  - Zonal advection via uo and zonal SST gradients.
  - Surface radiative/evaporative effects via rsds (or rsns) and wind speed proxies.
- Clarify the modulation by Indo-Pacific variability (Niño3.4, IOBM) versus Atlantic-intrinsic mechanisms using partial regressions and models excluding ENSO predictors.

Success criteria
- Achieve TCC ≥ 0.5 at strict 8‑month lead (using only SON/Nov predictors) for at least one model/feature set over 1979–2023.
- Demonstrate stability across validation folds, robustness to predictor choices, and provide mechanism-consistent diagnostics that support the identified precursors and pathways.
\end{lstlisting}
\end{planbox}

\begin{visualbox}
  \begin{center}
  \includegraphics[width=1.0\linewidth]{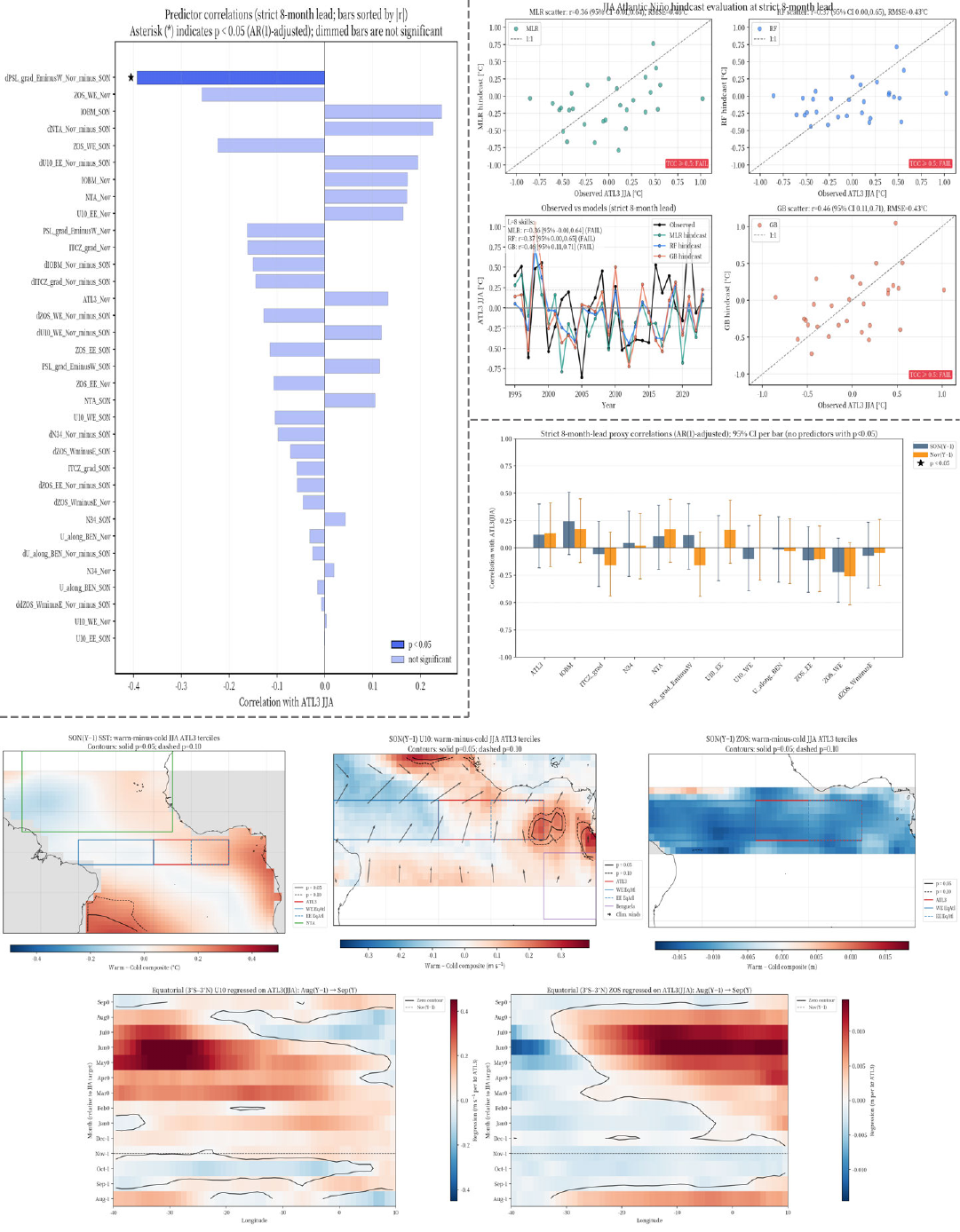}    
  \end{center}
\end{visualbox}

\begin{summarybox}
\begin{lstlisting}
## Datasets, target, and strict 8‑month lead setup
- Target (predictand): JJA ATL3 SST anomaly (20°W–0°, 3°S–3°N) from HadISST (1979–2023), detrended and standardized within folds.
- Predictors (strict 8‑month lead): ERA5 sea-level pressure (psl), 10 m winds and wind stress; ORAS5 sea level (zos) as an upper‑ocean heat-content/thermocline proxy; GPCP precipitation for ITCZ diagnostics; SON(Y−1) seasonal means and November(Y−1) monthly values only. ENSO (Niño3.4) and IOBM used as control/optional teleconnections.
- Preprocessing: 1981–2010 climatologies removed; linear detrending; 1° grid; area weighting; indices standardized within training folds; ENSO-regressed SST for subtropical AMM/NTA where used.

## Key precursors identified at strict 8‑month lead
- Dominant atmospheric tendency (significant):
  - SON→Nov increase in equatorial Atlantic east–minus–west SLP gradient (Δpsl E–W, Nov−SON): r ≈ −0.35 to −0.39 with JJA ATL3. Strengthening trades into November precede cold ATL3; weakening trades precede warm ATL3. This retains significance after controlling for SON/Nov Niño3.4.
- Core oceanic memory:
  - Western equatorial Atlantic sea level in Nov (zos, 40°–20°W; 3°S–3°N): r ≈ −0.26 with JJA ATL3. Lower‑than‑normal west SSH by Nov favors a later warm event (reduced zonal tilt/deeper east thermocline).
- Consistent but weaker supplementary signals (individually modest, useful in combination):
  - Equatorial easterly trade relaxation in SON/Nov (U10/τx in 10°W–0°, 3°S–3°N) and their SON→Nov tendencies: weaker trades precede warm ATL3.
  - Subtropical Atlantic background (AMM/NTA warmth and northward ITCZ precipitation gradient in SON/Nov): weak positive correlations consistent with a north‑focused ITCZ and weakened trades.
  - Alongshore wind off Angola–Namibia (15°S–5°S, 10°E–0°): relaxation aligns with coastal warming and downwelling coastal waves.
- Remote teleconnections:
  - ENSO (Niño3.4 in SON/Nov) is weak at this strict lead (near zero and sometimes slightly negative); best used as a control term. IOBM shows weak, non‑robust correlations and is not a primary predictor here.

## Feature selection and parsimony
- A minimal, low‑collinearity pair captures complementary physics: {Δpsl E–W (Nov−SON), zos WEq Nov}. Their intercorrelation is small (r ≈ 0.19) and VIF ~1.
- Screening shows other wind, thermocline‑slope, and subtropical/ITCZ indices contribute incrementally but are noisier at this lead; they are candidates for expanded multivariate models.

## Forecast model performance (strict 8‑month lead; cross‑validated hindcasts, 1979–2023/1995–2023 comparable)
- Baselines at L=8:
  - Persistence (Nov ATL3 → JJA) r ≈ 0.12–0.13.
  - ENSO‑only (Nov Niño3.4) near zero and sometimes slightly negative; not a reliable driver at this lead.
- Models using SON/Nov predictors:
  - Multiple Linear Regression (MLR): TCC ≈ 0.36 (95% CI roughly −0.01 to 0.64); RMSE ≈ 0.46 °C; strong amplitude damping (slope ≈ 0.27); slight cool bias.
  - Random Forest (RF): TCC ≈ 0.37; RMSE ≈ 0.43 °C; amplitude damping (slope ≈ 0.22); small negative bias.
  - Gradient Boosting (GB): TCC ≈ 0.46 (95% CI 0.11–0.71); RMSE ≈ 0.43 °C; best phase tracking among models but still underestimates extremes.
- Feature importance (tree models):
  - Western equatorial SSH in Nov (zos WEq) is the top predictor.
  - Δpsl E–W (Nov−SON) is consistently the second most important.
- Outcome relative to target: All models outperform persistence and ENSO‑only at L=8, but none reach the TCC ≥ 0.5 threshold (best GB r ≈ 0.46). Under‑dispersion (damped amplitudes) is evident across models.

## Physical mechanisms underpinning predictability
- Wind–thermocline preconditioning (Atlantic‑internal):
  - Autumn evolution of the equatorial SLP gradient reflects trade‑wind tendencies. A weakening gradient into Nov implies westerly‑stress anomalies that excite downwelling Kelvin waves, flatten the zonal thermocline slope, deepen the eastern thermocline, and reduce upwelling in ATL3.
  - Western equatorial SSH in Nov encodes the ocean’s memory of zonal tilt/heat content. Anomalously low west SSH (and/or higher east SSH) indicates a reduced tilt and a deeper eastern thermocline that can persist through spring.
  - As seasonal coupling strengthens in late spring–summer, these preconditions enable Bjerknes‑like feedback: warmer eastern SST weakens trades further, reinforcing warming into JJA; opposite for strengthening trades and shallow eastern thermocline leading to cold events.
- Coastal Angola–Benguela pathway:
  - Relaxed alongshore winds and coastal warming in SON/Nov favor downwelling coastal‑trapped waves that propagate equatorward and then along the equator, contributing to the springtime flip to positive SSH/thermocline depth in the east seen in equatorial Hovmöllers.
- Subtropical–equatorial coupling:
  - Warm NTA/AMM and a northward ITCZ in SON/Nov promote weaker cross‑equatorial trades, assisting the preconditioning, but their standalone correlations are modest at strict L=8.
- Limited Indo‑Pacific control at this lead:
  - After removing SON/Nov Niño3.4, the Atlantic precursor patterns persist, and ENSO‑only skill is weak—indicating the forecast skill arises primarily from Atlantic‑internal dynamics at L=8.

## Why the 8‑month target is challenging and how to improve
- Challenges observed:
  - Weak November persistence and muted ENSO teleconnections across boreal winter–spring reduce simple predictability.
  - Sample size is small (N ≈ 29 at strict L=8), producing wide CIs and unstable amplitude calibration.
  - Key processes (Kelvin/Rossby packets, coastal coupling) have memory closer to MAM–AMJ, so SON/Nov predictors only partly capture spring preconditioning.
  - Models show consistent amplitude damping, missing large warm/cold excursions.
- Targeted improvements to push TCC ≥ 0.5:
  - Enrich dynamical memory: add explicit thermocline‑slope proxies (Δzos = west − east; z20/0–100 m heat‑content gradients) and SON→Nov tendencies.
  - Resolve equatorial wind forcing: include τx/U10 indices in western/eastern boxes and their SON→Nov tendencies; add equatorial τx PCs to capture coherent patterns.
  - Represent coastal coupling: incorporate alongshore Benguela wind stress indices.
  - Subtropical modulation: use AMM/NTA (ENSO‑regressed) and ITCZ gradient (GPCP) to capture background state effects.
  - Modeling choices: use parsimonious regularized linear models (LASSO/Elastic Net) or shallow gradient boosting with a small, physically screened feature set; calibrate variance (e.g., reliability scaling) to counter amplitude damping.
  - Robustness: extend training where possible with longer consistent records (e.g., ORAS5 from 1958, HadISST), restrict to predictors available through the longer period, and examine epoch stability (pre‑2000 vs post‑2000).

## Bottom line
- Strict 8‑month‑lead predictability of JJA Atlantic Niño is real but moderate in this setup. The strongest, physically consistent precursors are the SON→Nov tendency of the equatorial east–west SLP gradient (trade‑wind evolution) and western equatorial SSH in November (thermocline/heat‑content memory). Tree models and MLR using SON/Nov predictors achieve TCC ≈ 0.36–0.46, clearly above persistence and ENSO‑only but short of the ≥0.5 goal. The underlying mechanism is an Atlantic‑internal wind–thermocline pathway, aided by coastal and subtropical coupling, which seeds spring thermocline adjustments and enables Bjerknes feedback into JJA. Focused enhancement of thermocline‑slope, wind‑stress, and coastal precursors, coupled with careful regularization and calibration, offers a credible path to lift 8‑month lead skill to or above r ≈ 0.5.
\end{lstlisting}
\end{summarybox}

\subsection{Causal link between pantropical climate variability and sea ice changes}
Modern climate science recognizes the tropics as a highly interconnected system where air-sea interactions drive global climate patterns. A growing focus on the interplay between the major ocean basins~\cite{cai2019pantropical} raises an even more critical question: how do tropical signals propagate to influence the remote and sensitive polar cryosphere? Unraveling this teleconnection is essential for a complete understanding of the drivers behind Arctic and Antarctic sea ice change.

Building upon this context, a novel physical mechanism was discovered by \model. The mechanism links pantropical SST gradients to sectoral sea ice changes through a ``two-branch pathway." This process is initiated when a SST gradient emerges across the tropics, defined by a warmer Atlantic and Indian Ocean relative to a cooler Pacific—a state captured by a newly formulated Pantropical Variability (PTV) index. The first branch is a ``fast atmospheric bridge," where the tropical heating anomaly generates atmospheric Rossby waves that propagate poleward within weeks to months, altering surface pressure and wind fields, which in turn immediately impact sea ice concentration in specific regions. The second, a ``slow oceanic and stratospheric branch," unfolds over 6 to 18 months. Persistent trade winds driven by the SST gradient modulate oceanic Ekman heat transport into the marginal ice zones, while the atmospheric forcing also influences the stratospheric polar vortex, whose effects gradually descend to the surface, creating a delayed and more persistent sea ice response. Crucially, the mechanism is asymmetric: the Atlantic-Pacific gradient primarily drives the rapid response, particularly in the Amundsen-Bellingshausen Sea, while the Indian-Pacific gradient governs the delayed impacts, such as the sea ice decline in the Barents-Kara Sea and growth in the Weddell Sea a year later.

The discovery of this teleconnection makes a significant contribution to climate science by establishing a new, mechanistically grounded source of predictability for polar sea ice. The primary significance lies in its potential for actionable, long-lead forecasting, offering the ability to predict sea ice anomalies in critical sectors like the Barents-Kara and Weddell Seas up to 12 to 20 months in advance. Furthermore, the two-branch hypothesis provides a unifying physical explanation for the often disparate and asynchronous variability observed across different polar regions, attributing these complex patterns to the distinct timescales and pathways of the pantropical forcing. This insight not only enhances our fundamental understanding of tropical-polar linkages but also provides a powerful new tool for predicting the evolution of the cryosphere in a changing climate.

The input and output are shown below. Note that the review paper~\cite{cai2019pantropical} about pantropical climate interactions is also fed into the system to facilitate discovery.

\begin{requestbox}
\begin{lstlisting}
Based on this paper, propose a novel causal relationship between the pantropical climate signal and sea ice, and conduct data analysis to verify and explain it.
\end{lstlisting}
\end{requestbox}

\begin{planbox}
\begin{lstlisting}
Title
Pantropical gradient → fast tropospheric and stratospheric teleconnections plus delayed oceanic Ekman transport → sectoral sea-ice variability (verification and mechanistic attribution)

1) Novel causal hypotheses to test
- H1 (two-branch pathway, novel integration): A positive pantropical SST gradient (PTV = 0.5·(ATL+IND) − PAC, 20°S–20°N) strengthens Pacific trade winds and induces a La Niña–like state that triggers:
  - a fast branch (weeks–months): PNA-/PSA-/NAO-like stationary wave trains and stratosphere–troposphere coupling modify surface turbulent/radiative fluxes over marginal ice zones, shifting sea-ice edge latitude and concentration;
  - a slow branch (6–12 months): wind-driven Ekman transport and mixed-layer adjustments alter oceanic heat convergence into marginal ice zones, reinforcing or offsetting the fast flux-driven response.
- H2 (sectoral signatures): The fast/slow branches yield distinct seasonal, sectoral responses in the Arctic (Bering–Okhotsk vs Barents–Kara vs Labrador–Greenland) and Antarctic (Weddell vs Amundsen–Bellingshausen).
- H3 (modulators): The ATL–PAC and IND–PAC components of PTV (TBV_ATL, TBV_IND) modulate pathway strength; background NAO/SAM and stratospheric vortex anomalies condition the sign/magnitude of polar responses.
- H4 (time-scale dependence and nonstationarity): Interannual and decadal PTV–trade-wind coupling varies across epochs, co-modulating tropical–polar linkages.

2) Datasets and variables (monthly, observations preferred)
Observations (1979–2023 unless noted)
- Sea ice:
  - NSIDC passive-microwave sea-ice concentration (siconc, %); NSIDC Sea Ice Index v3 for extent (derive SIE).
  - Optional cross-checks: OSI SAF SIC; thickness from PIOMAS (Arctic) and GIOMAS (global) where used as context (sithick, m).
- SST and subsurface ocean:
  - HadISST1 or ERSSTv5 for SST (tos, degC).
  - ORAS5 or NOAA GODAS for subsurface temperature (thetao, degC), ocean heat content 0–300 m (derive from thetao), and thermocline depth proxy (e.g., 20 °C isotherm, if available).
- Atmosphere/reanalysis:
  - ERA5: tauu/tauv (Pa), uas/vas (m s−1), psl (Pa), zg (m) at 200/500 hPa, ua/va (m s−1) at 850/200 hPa, wap (Pa s−1), pr (kg m−2 s−1), tas (K); stratosphere ua (m s−1) at 100/50/10 hPa.
  - Surface fluxes: hfss, hfls, rlds, rlus, rsds, rsus (W m−2).
- Radiation constraint (2000–present):
  - CERES-EBAF: rsds, rlds, rsus, rlus (W m−2) and TOA fluxes (rsdt, rsut, rlut, W m−2) for flux validation.
Controls/indices (derived)
- Niño3.4 from tos (5°S–5°N, 170°–120°W).
- AO/NAO/SAM from psl or standard indices.
- QBO from ERA5 equatorial u at 30/50 hPa; volcanic AOD and solar (monthly) if used as controls.

CMIP6 (for attribution/model evaluation; monthly)
- Activities/experiments: CMIP historical (1850–2014), AMIP (1979–2014), ScenarioMIP ssp245 (2015–2023 for continuity), piControl (segments matched to obs lengths).
- MIPs/variables:
  - SImon: siconc (%), sithick (m), sivol (m), siu/siv (m s−1 where available).
  - Amon: tos (degC), tauu/tauv (Pa), uas/vas (m s−1), psl (Pa), zg (m), ua/va (m s−1), wap (Pa s−1), pr (kg m−2 s−1), hfss/hfls/rsds/rsus/rlds/rlus (W m−2), tas (K), rsdt/rsut/rlut (W m−2).
  - Omon: thetao (degC), mlotst (m) if available.

3) Domains, periods, and sectors
- Periods: Primary 1979–2023 for observations; CMIP6 1979–2014 (historical/AMIP), extend to 2023 with ssp245 where needed; piControl segments for null distributions.
- Pantropical SST boxes (20°S–20°N):
  - ATL: 70°W–20°E; IND: 21°E–120°E; PAC: 121°E–90°W.
- Equatorial trade-wind index region: 5°S–5°N, 160°E–120°W (tauu; define sign below).
- Arctic sectors:
  - Bering (160°E–150°W, 55–70°N), Sea of Okhotsk (140–160°E, 50–60°N),
  - Barents–Kara (30–90°E, 70–85°N), Labrador–Greenland (60–20°W, 55–75°N).
- Antarctic sectors:
  - Weddell (60°W–20°E, 60–75°S), Amundsen–Bellingshausen (120–60°W, 60–75°S), Indian sector (20–90°E, 60–75°S).

4) Variables, units, and conventions
- siconc [%], sithick [m], sivol [m], siu/siv [m s−1].
- tos [degC], thetao [degC], mlotst [m].
- tauu/tauv [Pa] (surface downward stress components). For an intuitive “easterly trade” index, define Taux_west = −tauu averaged over 5°S–5°N, 160°E–120°W (positive = stronger westward/easterly stress).
- psl [Pa], zg [m], ua/va [m s−1], wap [Pa s−1], pr [kg m−2 s−1], tas [K].
- Fluxes [W m−2]; adopt downward-positive for radiative components per CMIP; upward-positive for hfss/hfls per CMIP. Define net surface heat flux into surface: Qnet = rsds − rsus + rlds − rlus − hfss − hfls (W m−2).

5) Preprocessing
- Temporal: monthly means; compute seasonal means for NDJ, DJF, MAM, JJA, SON as needed.
- Anomalies: remove 1981–2010 monthly climatology. Interannual analysis: linearly detrend anomalies. Decadal analysis: apply 5–10 yr low-pass filter (or 20-yr running means/trends as specified).
- Spatial: regrid to a common 1° grid (conservative for fluxes/SIC; bilinear for winds/heights). Area-weight means weighted by cell area.
- Sea-ice diagnostics:
  - SIE: sum of gridcell areas with siconc ≥ 15%.
  - Sea-ice edge latitude (SIELat): zonal mean latitude of the 15% SIC contour per sector.
  - Sectoral indices: area-mean SIC (%), SIE (10^6 km^2), SIELat (deg). Thickness indices where available (sithick, m).

6) Predictor and mediator indices
- Pantropical SST indices (area-weighted means over boxes):
  - TBV_ATL = ATL − PAC; TBV_IND = IND − PAC; PTV = 0.5·(ATL + IND) − PAC. Standardize monthly (z-scores).
- ENSO controls: Niño3.4 (from tos), and ENSO flavor (e.g., Niño3 vs Niño4) if used; warm-water volume proxy from ORAS5/GODAS if available.
- Trade-wind mediator: Taux_west (−tauu over 5°S–5°N, 160°E–120°W), standardized.
- Extratropical mediators: PNA, NAO, SAM from psl/zg (standardized), and polar-cap stratospheric wind U10/U50 (ua at 10/50 hPa, 60–90° in each hemisphere).

7) Analyses and calculations
7.1 Establish tropical forcing linkage
- Lead–lag correlations/regressions between {PTV, TBV_ATL, TBV_IND} and Taux_west and Niño3.4 at lags −12…+12 months to identify optimal precursor seasons (e.g., NDJ).
- 20-yr sliding correlations between PTV (and TBV_ATL) and Taux_west to quantify epochal changes in coupling.

7.2 Sea-ice responses (observations)
- Lead–lag regressions of sectoral SIC, SIE, and SIELat onto PTV, TBV_ATL, TBV_IND for lags −24…+24 months (predictor leads positive lags), stratified by target season (e.g., NDJ predictor vs DJF/MAM Arctic; JJA predictor vs SON/DJF Antarctic).
- Partial regressions controlling for Niño3.4 and, in extended models, for AO/NAO/SAM, QBO, volcanic AOD, and solar to isolate pantropical effects.

7.3 Atmospheric bridge diagnostics (fast branch)
- Regress psl, zg200/zg500, 850-hPa winds, wap, pr, and near-surface winds/fluxes (uas/vas, hfss, hfls, rlds, rsds) onto PTV (and TBV components) at sector-specific optimal leads identified in 7.2.
- Compute Rossby wave source (Sardeshmukh–Hoskins) from 200-hPa divergent wind and absolute vorticity; map stationary wave activity flux (Takaya–Nakamura) using 200-hPa winds and zg to trace teleconnections into polar regions.

7.4 Stratospheric mediation diagnostics
- Time–height regressions of polar-cap (60–90°) ua (100–10 hPa) onto PTV at 0–2 month leads during transition seasons (e.g., SON/NDJ NH; JJA/SON SH). Test downward propagation by regressing subsequent zg200 and psl onto earlier stratospheric anomalies.
- Mediation analysis: quantify indirect effects PTV → U10 (stratosphere) → NAO/SAM → sectoral sea-ice metrics, alongside direct PTV effects, using consistent lags.

7.5 Oceanic slow branch
- Ekman transport from wind stress: u_Ek = tauv/(ρw f), v_Ek = −tauu/(ρw f), with ρw = 1025 kg m−3 and f = 2Ω sinφ. Project Ekman transport onto near-surface temperature gradients to form an Ekman heat convergence proxy HEk (W m−2) in marginal ice zones.
- Regress HEk, OHC (0–300 m from thetao), and mlotst onto PTV at 3–12 month leads; relate to sectoral sea-ice responses to diagnose delayed oceanic contributions.

7.6 Event composites and nonlinearity
- Identify ±1σ PTV events in NDJ and JJA. Composite sea-ice metrics, fluxes, teleconnections, and stratospheric indices at 0–6 and 6–12 month windows to assess sign asymmetry and fast vs slow branches.

7.7 Causal inference (directionality and pathways)
- Apply conditional Granger causality and/or PCMCI+ on monthly anomalies (interannual) and on low-pass series (decadal) with variable set {PTV, TBV_ATL, TBV_IND, Niño3.4, Taux_west, PNA/NAO/SAM, U10, sectoral sea-ice metrics, Qnet over MIZ, HEk}. Include autoregressive terms and seasonal stratification or dummies. Record significant directed links and optimal lags.
- Convergent cross mapping (CCM) between PTV and sectoral sea-ice metrics to test asymmetry of influence.

8) CMIP6 model evaluation and attribution
- For each model (historical, AMIP; extend with ssp245 to 2023 where available):
  - Construct PTV, TBV_ATL, TBV_IND and Taux_west identically to observations.
  - Repeat 7.2–7.5 to quantify simulated teleconnections, flux responses, and fast/slow branches. Derive sectoral regression slopes (sea-ice metrics on PTV).
- Model grouping and emergent constraint:
  - Rank models by their interannual and 20-yr sliding PTV↔Taux_west coupling. Define strong- vs weak-coupling groups (e.g., terciles).
  - Compare sea-ice sensitivity and teleconnection structure between groups and with observations.
  - Use piControl segments to estimate internal-variability distributions for key correlations/slopes.

9) Statistical treatment
- Significance: two-tailed tests with effective degrees of freedom accounting for autocorrelation (e.g., AR(1) adjustment) or block bootstrap.
- Field significance for spatial maps via false discovery rate control.
- Uncertainty on regression slopes, mediation fractions, and skill metrics via block bootstrap.

10) Key diagnostics and figures to produce
- Pantropical–trade coupling: time series and lead–lag correlations of PTV/TBV with Taux_west; 20-yr sliding correlations.
- Sea-ice responses: lag–regression “Hovmöller” plots of sectoral SIC/SIE/SIELat vs PTV/TBV; maps of SIC regression at optimal lags per season/sector.
- Atmospheric bridge: regression maps of psl, zg200, 850-hPa winds, and stationary wave activity flux vectors for PTV phases; surface flux (Qnet, hfss, hfls, rlds, rsds) anomalies over marginal ice zones.
- Stratospheric pathway: time–height sections of polar-cap ua regressed on PTV and subsequent NAO/SAM and psl; mediation path coefficients with confidence intervals.
- Oceanic slow branch: HEk and OHC regressions vs PTV at longer leads; sectoral co-variations with sea-ice metrics.
- Predictability: seasonal hindcast skill (correlation/RMSE) for sea-ice metrics using Pacific-only predictors vs adding TBV_ATL, TBV_IND, and PTV at 1–12 month leads.
- CMIP6 attribution: scatter of model PTV–Taux_west coupling vs sea-ice sensitivity; group composites (strong vs weak coupling) of sea-ice/teleconnection/flux responses.

11) Success criteria (verification targets, not conclusions)
- Statistically significant, seasonally coherent lead–lag links from PTV/TBV to sectoral sea-ice that persist after controlling for ENSO and extratropical indices.
- Mechanistic closure: concurrent teleconnection patterns, surface energy-budget anomalies, and Ekman heat convergence signals consistent with observed sea-ice responses at fast and slow lags.
- Evidence for stratospheric mediation where applicable (PTV → U10 → NAO/SAM → sea-ice) with plausible lag structure.
- CMIP6 models with stronger, more realistic PTV–trade coupling reproduce more realistic tropical–polar sea-ice teleconnections, supported against piControl variability.

Variable short names and MIPs (CMIP6-aligned)
- Sea ice: siconc [%], sithick [m], sivol [m], siu/siv [m s−1] (SImon).
- Ocean: tos [degC] (Amon/Omon), thetao [degC], mlotst [m] (Omon).
- Atmosphere/surface: tauu/tauv [Pa], uas/vas [m s−1], psl [Pa], zg [m], ua/va [m s−1], wap [Pa s−1], pr [kg m−2 s−1], tas [K], rsds/rsus/rlds/rlus/hfss/hfls [W m−2], rsdt/rsut/rlut [W m−2] (Amon).
\end{lstlisting}
\end{planbox}

\begin{visualbox}
  \begin{center}
  \includegraphics[width=1.0\linewidth]{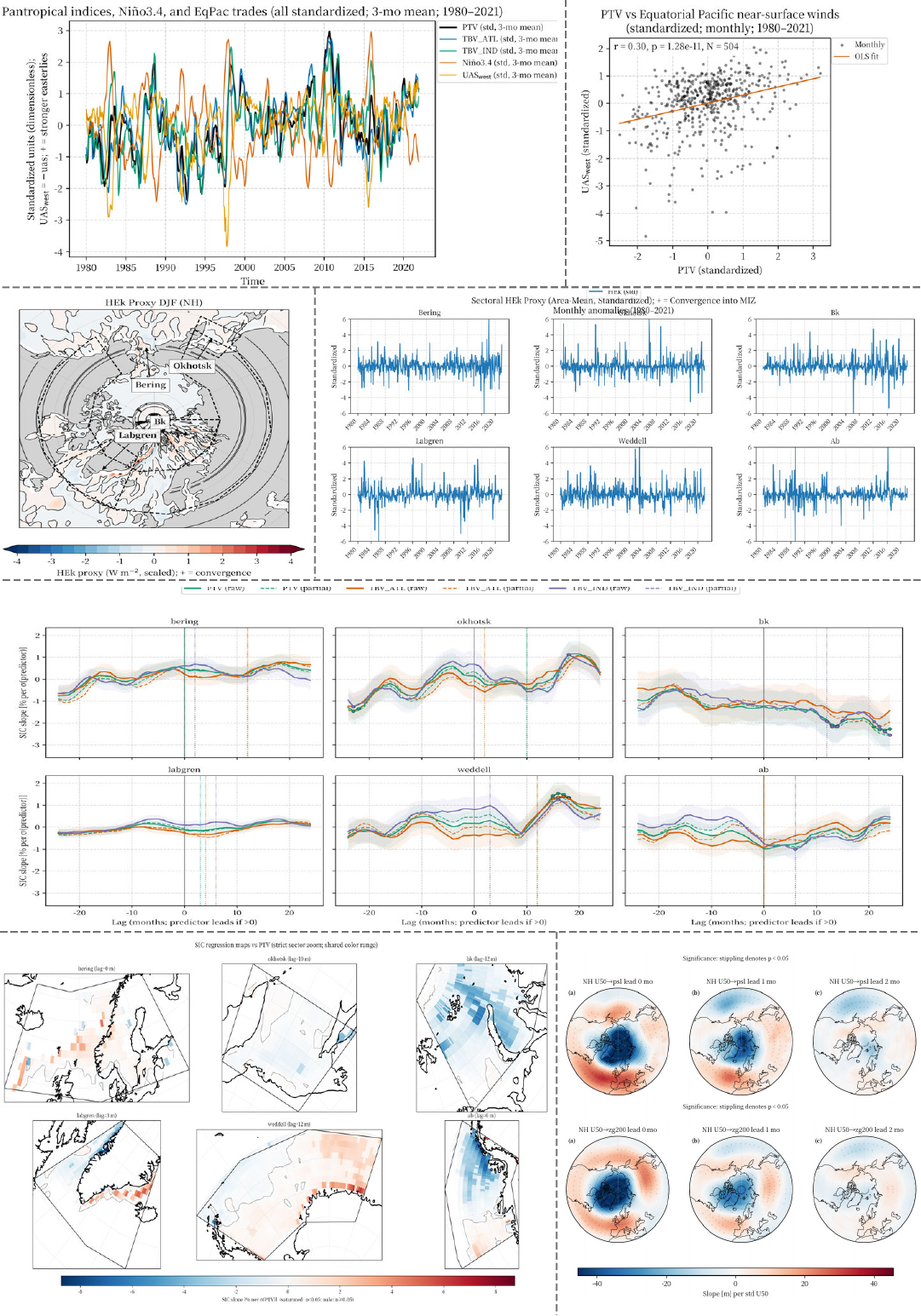}    
  \end{center}
\end{visualbox}

\begin{summarybox}
\begin{lstlisting}
Proposed novel causal pathway
- Positive pantropical SST gradient (PTV = 0.5·[ATL+IND] − PAC; i.e., Atlantic/Indian warmer than the Pacific, La Niña–like) drives:
  1) A fast atmospheric branch (weeks–months): PNA−/PSA teleconnections and a SAM+‑like annular response that alter surface winds and turbulent/radiative fluxes over marginal‑ice zones, immediately shifting sea‑ice concentration (SIC) and ice‑edge latitude.
  2) A slow oceanic/stratospheric branch (≈6–18 months): strengthened trades and extratropical wind anomalies modulate Ekman heat convergence and mixed‑layer heat content, while seasonally gated stratospheric anomalies (vortex changes) descend into AO/SAM‑like surface patterns, jointly producing delayed sea‑ice responses.
- Component roles: TBV_ATL (ATL − PAC) preferentially triggers the immediate Antarctic Amundsen–Bellingshausen (AB) response; TBV_IND (IND − PAC) governs delayed Barents–Kara (BK) loss and Weddell gains; combined PTV yields the clearest pan‑hemispheric signatures.

Key figure‑based findings

1) PTV robustly strengthens equatorial Pacific trades (mediator verified; build_indices)
- Contemporaneous association: r ≈ 0.30 (p ≈ 1e−11; N ≈ 504 months, 1980–2021) between PTV and westward (easterly) equatorial Pacific wind stress, consistent with a Gill‑type Walker strengthening.
- Coevolution with ENSO but not redundant: PTV peaks with La Niña and dips with major El Niños, yet exhibits episodes of strong trades without extreme Niño3.4, supporting PTV as an added, quasi‑independent driver.
- Decadal nonstationarity: 20‑yr sliding correlations positive (≈0.18–0.38), peaking around 1999–2003, weakening near 2007–09, then partly recovering—implying epoch‑dependent downstream sea‑ice sensitivity.

2) Sectoral sea‑ice responses and lead–lags (independent of ENSO; lagreg_ice)
- Independence from ENSO: Partial regressions (controlling for Niño3.4) closely match raw regressions across sectors, showing added predictive power from pantropical gradients beyond Pacific SST alone.
- Quantified responses per +1σ predictor:
  - Barents–Kara (BK): delayed SIC loss, −2 to −3% at 12–20‑month leads (strongest for PTV and TBV_IND).
  - Weddell: delayed SIC gain, +1.5 to +2% at ~12–18 months (PTV, TBV_IND).
  - Amundsen–Bellingshausen (AB): immediate SIC loss, −0.5 to −1% at 0–6 months (TBV_ATL peak at 0 months; TBV_IND near ~6 months).
  - Sea of Okhotsk: two‑stage SIC gain, ≈+0.6% at 0–2 months and +1 to +1.5% at ~16–20 months.
  - Bering: broad weak‑to‑moderate SIC gain, 0–0.8% over 0–12 months.
  - Labrador–Greenland: mixed signs (east Greenland positive, Baffin/West Greenland negative), yielding small sectoral means; spatial maps resolve the dipole.
- Spatial patterns at “optimal” lags confirm coherent sectoral structures: BK negative, Weddell positive (delayed); AB negative (fast); North Pacific mostly positive.

3) Fast teleconnections link PTV to polar winds/fluxes within 0–2 months (teleconn_maps)
- Z200 regressions: canonical negative PNA‑like ridge in the North Pacific and PSA‑type trough over AB emerge at lead 0–2 months (≈10–20 m per +1σ PTV).
- SLP regressions: Aleutian High in NH; strong circumpolar belt of negative SLP in SH (≈0.5–1.5 hPa per +1σ PTV), consistent with SAM+ and stronger westerlies.
- Implications: Immediate thermodynamic/dynamic forcing of sea ice—more ice in Bering/Okhotsk and Ross/Weddell, less in AB—consistent with observed sectoral lag‑responses.

4) Seasonally gated stratospheric mediation (strat_diag)
- PTV → polar‑cap stratospheric winds (ua) is hemispherically asymmetric and seasonal:
  - SH SON: robust vortex strengthening (+1 to +1.5 m s−1 per +1σ PTV at 50–10 hPa; overall r ≈ +0.10, p ≈ 0.025).
  - NH: NDJ tendency toward vortex weakening (−0.5 to −1 m s−1 per +1σ PTV), flipping to strengthening in FMA (+0.5–1 m s−1); annual mean r ≈ −0.06 (ns) due to seasonal cancellation.
- Downward impact: +1σ U50 induces AO/SAM‑like surface patterns within 0–1 month (≈−30 to −40 m in zg200 and −1 to −1.5 hPa polar SLP), providing a rapid, causal bridge from PTV to surface pressure/wind anomalies that modulate sea ice.

5) Wind‑driven Ekman heat convergence as the slow oceanic branch (ekman_proxy)
- Antarctic JJA: a near‑continuous positive HEk ring along the winter ice edge, strongest in AB and Weddell—favorable for wind‑driven heat import and melt where winds converge; signals weaken in DJF.
- Arctic DJF: positive HEk bands in Labrador–Greenland, BK gateways, and Bering/Okhotsk; JJA signals are weaker.
- Time series: episodic, seasonally phase‑locked pulses (largest in AB/Weddell, sizeable in Labrador–Greenland); consistent with remote wind forcing rather than secular trends.
- Mechanistic implication: positive HEk → ice loss, negative HEk → ice gain; provides the needed delayed pathway linking pantropical‑forced winds to sea‑ice anomalies, especially in SH winter and NH winter.

Integrated mechanistic synthesis
- Fast branch (0–6 months): +PTV → PNA−/PSA and SAM+ → immediate AB SIC loss and NP SIC gains via altered surface fluxes and winds; effects are strongest for TBV_ATL in AB.
- Slow branch (≈10–20 months): +PTV → strengthened trades and extratropical winds → Ekman heat‑convergence/mixed‑layer adjustments plus seasonally dependent stratospheric AO/SAM influences → delayed BK SIC loss and Weddell SIC gain (TBV_IND/PTV strongest), and a second, delayed Okhotsk SIC increase.
- ENSO independence: controlling for Niño3.4 leaves the pantropical signals largely intact, indicating genuine added causality/predictability beyond Pacific‑only indices.
- Nonstationarity: stronger PTV–trade coupling around the late‑1990s/early‑2000s implies epoch‑dependent sea‑ice teleconnections; split‑epoch analyses are warranted.

Actionable predictability (from verified leads and sectors)
- AB: same‑season forecasts using TBV_ATL (0–2 months).
- BK and Weddell: 1‑year‑ahead forecasts using PTV/TBV_IND (12–18 months).
- Okhotsk: dual windows—near‑term (0–2 months) and delayed (~16–20 months).
- Bering: modest skill at 0–12 months (positive sign).

Bottom line
- Figures collectively verify a two‑branch, pantropically forced causal chain to sectoral sea ice: a rapid atmospheric bridge (PNA−/PSA/SAM+) that immediately reduces AB ice and increases North Pacific ice, and a delayed oceanic/stratospheric branch that reduces BK ice and increases Weddell ice roughly a year later. TBV_IND predominantly drives the delayed BK/Weddell responses, while TBV_ATL triggers the immediate AB loss. These relationships persist after removing ENSO, establishing pantropical SST gradients as independent, mechanistically grounded predictors of polar sea‑ice variability.
\end{lstlisting}
\end{summarybox}

\subsection{Teleconnection mechanism among the Pacific, Indian, and Atlantic basins}

Understanding the intricate web of interactions between the Pacific, Atlantic, and Indian Oceans is fundamental to improving seasonal-to-decadal climate prediction~\cite{wang2019three}. While numerous studies have identified pairwise linkages, such as the influence of the ENSO on global weather patterns or the connection between the Atlantic and Pacific basins, the Earth's climate system operates as a fully coupled network~\cite{wang2019three}. The mechanisms by which climate signals propagate sequentially across all three ocean basins, involving both rapid atmospheric bridges and slower oceanic pathways, remain a frontier of climate dynamics. A comprehensive theory detailing how variability is initiated in one basin and systematically relayed through the others could significantly enhance our predictive capabilities. To address this gap, we tasked \model~to propose and empirically validate a novel hypothesis based on the interaction between the three oceans.

The experiment uncovered a robust, sequential teleconnection chain initiated by the Atlantic Niño. \model's central hypothesis posits a two-step forcing mechanism: a rapid atmospheric bridge followed by a slower, reinforcing oceanic tunnel. The process begins with anomalous warming in the equatorial Atlantic during the boreal summer (an Atlantic Niño). This warming initiates an atmospheric Kelvin wave response that propagates eastward, leading to anomalous descent over the central Pacific. This atmospheric adjustment strengthens the Pacific trade winds, driving an upwelling oceanic Kelvin wave that shoals the thermocline in the eastern Pacific. The result is a significant tendency toward an Eastern Pacific (EP)-leaning La Niña by the subsequent winter, with an observed cooling of approximately -0.2 to -0.3 K. Concurrently, the initial Atlantic warming also creates favorable conditions for a positive IOD to develop. This IOD then activates the second step of the mechanism: the oceanic tunnel. A positive IOD enhances the Indonesian Throughflow (ITF), increasing the export of warm water out of the western Pacific warm pool. This oceanic process introduces a delayed cooling effect that sustains or amplifies the initial La Niña conditions approximately 6 to 12 months after the IOD event. The system's causal network analysis and Hovmöller diagrams of subsurface ocean temperature provide direct evidence for this sequence, confirming that Atlantic variability can both trigger and be reinforced by a coordinated response across all three basins.

The discovery of this integrated Atlantic-Indian-Pacific chain advances our understanding of global climate dynamics. Its primary innovation is the articulation of a complete, physically coherent narrative that moves beyond isolated teleconnections to a sequential, multi-timescale causal pathway. This finding has profound implications for climate prediction, suggesting that the state of the equatorial Atlantic in summer and the Indian Ocean in autumn are critical precursors for forecasting the character of ENSO the following winter.

The input and output are shown below. Note that the review paper~\cite{wang2019three} about three‑ocean interactions and climate variability is also fed into the system to facilitate discovery.

\begin{requestbox}
\begin{lstlisting}
Propose a novel teleconnection hypothesis based on the interaction between the three oceans, and conduct data analysis to verify and explain it.
\end{lstlisting}
\end{requestbox}

\begin{planbox}
\begin{lstlisting}
Improved experimental plan: Three-ocean teleconnection hypothesis and verification

Hypothesis (testable and falsifiable)
A coupled Atlantic–Indian–Pacific teleconnection chain links interannual and decadal variability through atmospheric bridges and oceanic “tunnels.” Key predictions:
- P1. Atlantic → Indo-Pacific (interannual): Positive ATL3 SST anomalies in JJA (Atlantic Niño; 20°W–0°, 3°S–3°N) force an atmospheric Kelvin response that strengthens equatorial Pacific easterlies, shoals the eastern-Pacific thermocline, and increases the probability of an EP-leaning La Niña by the following DJF. Concurrently, ATL3 warming yields western Indian Ocean (WIO) warming and raises the likelihood of a positive IOD in SON.
- P2. Indo-Pacific → Atlantic/closure (interannual): DJF ENSO forces spring IOB (MAM) and tropical North Atlantic (TNA) warming, which in some years preconditions ATL3 by the next JJA.
- P3. Oceanic tunnels (interannual): Positive IOD in SON transiently enhances the Indonesian Throughflow (ITF) and exports heat from the western Pacific/Indonesian seas; after ~6–12 months this increases the chance of Pacific cooling if not offset by westerly wind bursts.
- P4. Indian–Atlantic link (interannual to decadal): Positive IOD years show enhanced Agulhas leakage (AL) eddy kinetic energy southwest of Africa; after ~6–18 months, warm anomalies appear in the South/Tropical Atlantic and can precondition ATL3 the next boreal summer.
- P5. Low-frequency modulation (decadal): Positive AMV (>10 yr) enhances subtropical Pacific wind–evaporation–SST (WES) feedback, favoring CP-type El Niño and modulating PDO/IPO; tropical Atlantic multidecadal warming forces Indo–western Pacific warming with a tendency for EP cooling.

Datasets, variables, units (monthly unless noted; observations prioritized)
- Sea surface temperature (SST):
  - HadISST1 (variable: tos, K), 1900–present.
  - ERSSTv5 (tos, K), 1900–present.
  - Daily OISST (tos, K), 1982–present for event composites only.
- Atmosphere (reanalysis; ERA5, 1979–present):
  - 10 m winds: uas, vas (m s−1)
  - Sea level pressure: psl (Pa)
  - Precipitation: pr (kg m−2 s−1)
  - Vertical velocity: wap at 500 hPa (Pa s−1; negative = ascent)
  - Velocity potential: chi at 200 hPa (m2 s−1) or derive from winds at 200 hPa
  - Optional for WES: latent/sensible heat fluxes hfls, hfss (W m−2); near-surface humidity huss (kg kg−1), near-surface air temperature tas (K)
  - Low-level/upper-level winds (diagnostics): ua/va at 850 hPa and 200 hPa (m s−1)
- Ocean subsurface and sea level:
  - ORAS5 (thetao, K) 1958–present or ARMOR3D 1993–present; derive z20 (20 °C isotherm depth; m) and OHC0–300 m (J m−2)
  - AVISO/CMEMS gridded altimetry (zos/sea level anomaly, m; geostrophic velocities to compute EKE, m2 s−2), 1993–present
- CMIP6 (model relevance; monthly):
  - Activity: CMIP; Experiments: historical (1850–2014/2015) and piControl
  - MIPs: Amon, Omon
  - Variables: tos (K), psl (Pa), pr (kg m−2 s−1), uas/vas (m s−1), ua/va (m s−1), wap (Pa s−1), chi (m2 s−1), thetao (K), zos (m), uo/vo (m s−1)
  - Period for comparison: 1950–2014

Index definitions (all standardized after anomaly calculation)
- ATL3 (JJA): tos mean over 20°W–0°, 3°S–3°N (K)
- Niño indices (DJF): Niño3 (150°W–90°W, 5°S–5°N), Niño4 (160°E–150°W, 5°S–5°N), Niño3.4 (170°W–120°W, 5°S–5°N) (K)
- ENSO “flavor” indices: EP and CP derived from Niño3 and Niño4 (e.g., EP = Niño3 − α·Niño4; CP = Niño4 − β·Niño3; standardized)
- IOD (DMI; SON): west (50°–70°E, 10°S–10°N) minus east (90°–110°E, 10°S–0°) SST (K)
- IOB (MAM): tropical Indian Ocean mean (40°–100°E, 20°S–20°N) SST (K)
- TNA (MAM/JJA): 5°–25°N, 55°–15°W SST (K)
- AMV (decadal): 0°–60°N Atlantic mean SST (K), detrended and low-pass filtered (>10 yr)
- PDO (monthly/decadal): EOF1 of detrended North Pacific (20°–60°N) SST (standardized)
- ITF proxy (altimetry; seasonal): zos(Indonesian seas: 115°–130°E, 10°S–5°N) − zos(western equatorial Pacific: 140°–160°E, 5°S–5°N) (m)
- Agulhas leakage (AL) proxy: EKE mean over 35°–45°S, 10°–20°E (m2 s−2)
- Western Indian Ocean (WIO) diagnostic box: 50°–70°E, 10°S–10°N (K)

Study periods and temporal stratification
- Interannual analyses: 1979–present (ERA5 consistency); SST-only fields extend to 1900–present
- Altimetry-dependent diagnostics (ITF/AL/EKE): 1993–present
- Decadal analysis: 1900–present (SST-based indices)
- Seasonal means: DJF, MAM, JJA, SON; lags evaluated from −12 to +18 months as specified below

Preprocessing
- Regrid all fields to a common 1° × 1° grid
- Compute monthly anomalies by removing the 1981–2010 monthly climatology
- Interannual scale: remove linear trend from anomalies (per grid cell or index)
- Decadal scale: low-pass filter (>10 yr) and analyze low-frequency components; optionally remove global-mean SST from SST fields for decadal analyses
- Form seasonal means for indices and for fields used in seasonal diagnostics
- Standardize index time series (zero mean, unit variance) after detrending/filtering appropriate to the scale
- Derived variables:
  - z20 (m): depth of 20 °C isotherm from thetao profiles (mask where 20 °C not present)
  - OHC0–300 m (J m−2): integrate heat content from 0–300 m using thetao (assume constant ρ and cp if density fields are unavailable)
  - EKE (m2 s−2): compute from geostrophic velocities (AVISO) as 0.5 × (u′2 + v′2), monthly then seasonal means
  - chi200 (m2 s−1): take chi at plev=200 hPa if available; otherwise derive from divergent wind

Core analyses and calculations
1) Lead–lag regressions and composites (interannual and decadal)
- Predictors:
  - ATL3(JJA): regress global fields at lags −6 to +12 months: tos (K), psl (Pa), uas/vas (m s−1), pr (kg m−2 s−1), wap (Pa s−1), chi at 200 hPa (m2 s−1), z20 (m), OHC0–300 (J m−2), zos (m)
  - DMI(SON): same suite at lags −6 to +12 months; include ITF proxy evolution and western Pacific OHC tendencies
  - Niño3.4(DJF): regress onto IOB(MAM), TNA(MAM/JJA), ATL3(JJA) to test closure
  - AMV (decadal): regress onto decadal Pacific SST (tos), winds (uas/vas), and PDO
- Event composites:
  - Select top and bottom terciles of ATL3(JJA) and DMI(SON); composite subsequent fields at specified lags
  - Stratify DJF responses by ENSO “flavor” (EP vs CP) using flavor indices

2) Cross-basin coupled modes (MCA/SVD)
- Pairings:
  - ATL3(JJA) field (Atlantic equatorial SST anomalies) vs DJF Pacific SST anomalies (120°E–80°W, 20°S–20°N)
  - DMI(SON) field vs DJF Pacific SST anomalies
  - AMV (decadal SST field) vs decadal Pacific SST field
- Quantify squared covariance fractions and report time series of expansion coefficients

3) ENSO “flavor” quantification
- Regress ATL3(JJA) onto Niño3 and Niño4 separately and onto EP/CP indices; evaluate relative strength and sign
- Composite DJF SST, winds, and z20 conditioned on ATL3 terciles and ENSO “flavor”

4) Partial correlations and multiple regression
- Partial correlation of ATL3(JJA) with Niño3.4(DJF) controlling for DMI(SON), IOB(MAM), and AMV (decadal component)
- Partial correlation of DMI(SON) with Niño3.4(JJA+1) controlling for ATL3(JJA) and AMV
- Multiple regression for DJF Niño3 and Niño4 using predictors {ATL3(JJA), DMI(SON), IOB(MAM), AMV(decadal)} to estimate relative contributions

5) Causality and network inference (seasonal indices)
- Granger causality with physically motivated lags:
  - ATL3(JJA) → DMI(SON); ATL3(JJA) → Niño3.4(DJF)
  - DMI(SON) → ITF proxy (OND–MAM); DMI(SON) → Niño3.4(JJA+1)
  - Niño3.4(DJF) → IOB(MAM); Niño3.4(DJF) → TNA(MAM)
  - AMV(decadal) → PDO(decadal)
- Multivariate causal discovery (e.g., PCMCI) with nodes:
  {ATL3(JJA), DMI(SON), IOB(MAM), Niño3/4/3.4(DJF), AMV(decadal), PDO(decadal), ITF proxy (seasonal), AL proxy (seasonal)}
  - Maximum interannual lag: up to 18 months; for decadal nodes: annual lags up to several years

6) Time–frequency stability
- Wavelet coherence (or equivalent) for:
  - ATL3(JJA) with Niño3.4(DJF) in 2–7 yr band
  - DMI(SON) with Niño3.4(JJA+1) in 2–7 yr band
  - AMV(decadal) with PDO(decadal) in >10 yr band
- Identify significant scale bands and relative phase/lead–lag

7) Mechanistic diagnostics
- Walker/Hadley adjustments:
  - Regress wap (500 hPa), chi200, psl, and near-surface winds onto ATL3(JJA) at lags 0 to +6 months; identify ascent over the equatorial Atlantic, descent over central Pacific/Indian, and Kelvin/Rossby signatures
- Pacific ocean adjustment:
  - Hovmöller (time–longitude) along the equatorial Pacific (5°S–5°N) of z20 (m) and 850 hPa u-wind (m s−1) from −6 to +12 months around ATL3(JJA) and DMI(SON) events; diagnose eastward-propagating upwelling Kelvin waves and thermocline shoaling
- Surface flux/WES diagnostics:
  - Regress hfls (W m−2), wind speed (from uas/vas), tas, and huss onto AMV (decadal) and ATL3(JJA) over subtropical Pacific; assess WES feedback signatures
- ITF pathway:
  - Lag-regress ITF proxy onto DMI(SON) and Niño3.4(DJF); assess expected SON enhancement during positive IOD and ~9-month weakening during El Niño; relate to western Pacific OHC tendencies
- Agulhas leakage:
  - Regress and lag-correlate AL proxy (EKE over 35°–45°S, 10°–20°E) with DMI(SON), South/Tropical Atlantic tos (e.g., 0–30°S, 10–30°W), and ATL3(JJA+1)

Model relevance with CMIP6 (historical and piControl)
- Repeat key diagnostics using model monthly fields and identical index definitions:
  - Lead–lag regressions (ATL3 → DJF Pacific fields; DMI → Pacific and ITF; AMV → Pacific decadal fields)
  - ENSO “flavor” responses to ATL3 and AMV
  - Causality tests on seasonal indices; ocean “tunnel” proxies from zos and near-surface uo/vo (EKE analogs)
- Compare spatial patterns, effect sizes, and lags with observations; assess inter-model spread and the relation of simulated ITF/AL variability amplitude to cross-basin teleconnection strength

Statistical treatment
- Mapwise regressions/composites: compute coefficients and p-values using block bootstrap appropriate for monthly/seasonal autocorrelation (e.g., multi-month blocks)
- Control field significance using Benjamini–Hochberg false discovery rate (e.g., q=0.05–0.10)
- Report effect sizes (standardized coefficients), confidence intervals for key relationships, and selected lags from causality/network analyses

Figures and diagnostics to produce
- Lead–lag regression maps:
  - ATL3(JJA) onto global tos/psl/uas/vas/pr/chi200/wap at 0, +3, +6, +9, +12 months
  - ATL3(JJA) onto equatorial Pacific z20 and OHC at +3 to +6 months
  - DMI(SON) onto Pacific tos/uas/vas and ITF proxy at −3 to +12 months
  - AMV(decadal) onto Pacific decadal tos and PDO
- Event composites:
  - Top/bottom terciles of ATL3(JJA): DJF Pacific tos, winds, and z20; stratify by ENSO “flavor” (EP vs CP)
  - Positive vs negative IOD(SON): subsequent ITF proxy, western Pacific OHC, and Pacific tos at +6 to +12 months
- Cross-basin MCA/SVD mode pairs for ATL3–Pacific DJF, DMI–Pacific DJF, and AMV–Pacific decadal SST
- Causal network diagrams from PCMCI with nodes, directed edges, optimal lags, and standardized effect sizes
- Time–longitude Hovmöllers of equatorial Pacific z20 and 850 hPa u around ATL3 and IOD events
- Wavelet coherence panels: ATL3–Niño3.4, DMI–Niño3.4(next year), AMV–PDO
- Oceanic tunnel diagnostics:
  - Lag regressions: ITF proxy with DMI and Niño3.4; AL proxy with DMI, South/Tropical Atlantic tos, and ATL3 next JJA
- Model–observation comparison panels for key links (patterns, lags, and effect sizes)

Data handling specifics
- Units:
  - tos (K), psl (Pa), pr (kg m−2 s−1; convert to mm day−1 for display if desired), uas/vas/ua/va (m s−1), wap (Pa s−1), chi (m2 s−1), z20 (m), OHC0–300 (J m−2), zos (m), EKE (m2 s−2), hfls/hfss (W m−2), tas (K), huss (kg kg−1)
- Weight area averages by cosine of latitude for all regional means
- Sign conventions:
  - z20 positive downward; negative regression coefficients imply shoaling
  - hfls positive upward from the surface

Decision criteria (no conclusions assumed)
- P1 supported if ATL3(JJA) significantly leads DJF EP-leaning Pacific cooling with equatorial easterlies and eastward-propagating z20 upwelling signals, with contemporaneous WIO warming and increased positive IOD likelihood
- P2 supported if Niño3.4(DJF) significantly precedes IOB/TNA warming in MAM and, in some years, ATL3 in JJA
- P3 supported if positive IOD(SON) significantly leads enhanced ITF proxy, reduced western Pacific OHC, and increased odds of Pacific cooling within 6–12 months
- P4 supported if AL proxy increases during positive IOD and significantly leads South/Tropical Atlantic warming and higher ATL3(JJA+1)
- P5 supported if AMV (decadal) is significantly linked to PDO/ENSO “flavor” metrics with accompanying subtropical WES signatures in the Pacific
\end{lstlisting}
\end{planbox}

\begin{visualbox}
  \begin{center}
  \includegraphics[width=1.0\linewidth]{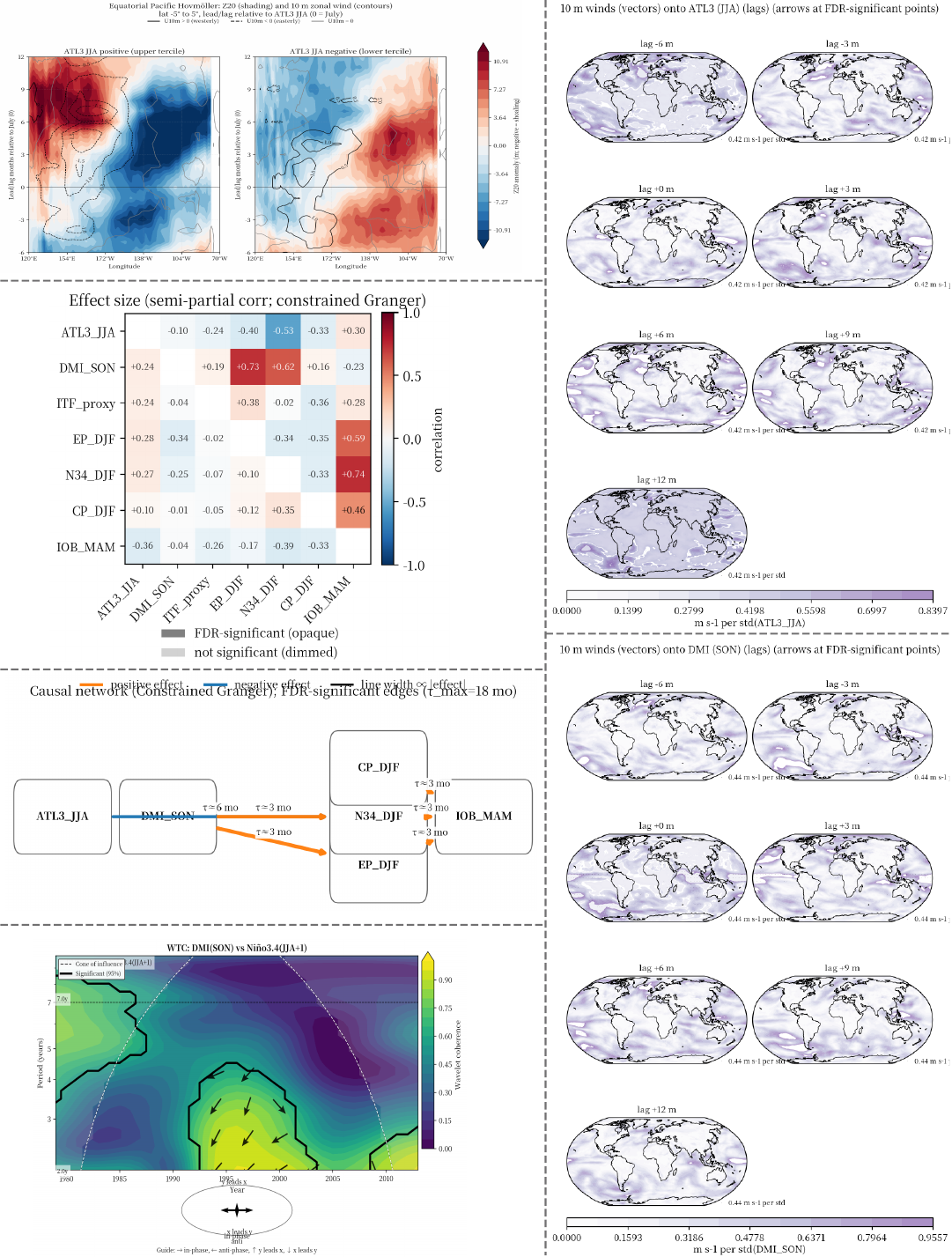}    
  \end{center}
\end{visualbox}

\begin{summarybox}
\begin{lstlisting}
# Three‑ocean teleconnection hypothesis: summary of evidence and verification

Hypothesis (testable)
- An Atlantic–Indian–Pacific chain couples interannual to decadal variability via:
  - Atmospheric bridge: Atlantic Niño (ATL3, JJA) forces an eastward Kelvin response that strengthens Pacific trades, shoals the eastern‑Pacific thermocline, and favors an EP‑leaning La Niña by DJF; it also warms the western Indian Ocean (WIO) and raises the odds of a positive IOD in SON.
  - Oceanic tunnel: Positive IOD transiently enhances the Indonesian Throughflow (ITF), exporting heat from the western Pacific and conditioning further Pacific cooling ~6–12 months later.
  - Closure: DJF ENSO warms the spring Indian Ocean basin (IOB) and TNA; in some years this preconditions ATL3 the next JJA.
  - Low‑frequency modulation: AMV enhances subtropical Pacific WES feedback and the CP‑El Niño tendency; tropical Atlantic multidecadal warming cools the eastern Pacific.

Key findings by diagnostic

1) Indices (indices_summary.png)
- ATL3 vs subsequent ENSO:
  - Multiple clear cases where ATL3+ in JJA is followed by La Niña: 1984–85, 1998–2000, 2007–08, 2011–12. Some counterexamples (e.g., 2003, 2014) point to modulation by background state and WWBs.
- IOD vs ENSO:
  - +IOD in SON often precedes La Niña within ~1 year (1994→1995–96; 2006→2007–08); −IOD co‑occurs with El Niño (1997–98). The 1982 +IOD exception coincides with the overpowering 1982–83 El Niño.
- The DMI component panel confirms canonical dipole structure (west‑warm/east‑cool for +IOD), supporting its mechanistic relevance.

2) Lead–lag regressions (global SST/SLP/winds/sea level)
- ATL3(JJA) → Pacific by DJF (supports P1):
  - DJF: La Niña‑like SST pattern with EP emphasis; −0.2 to −0.3 K per 1σ ATL3 along the equatorial central–eastern Pacific.
  - Sea level: −2 to −3 cm along the equatorial Pacific (shoaling thermocline) by +6 months.
  - Winds: Equatorial easterly anomalies emerge at +6 to +9 months, aligned with cooling and sea‑level drops.
- ATL3(JJA) → Indian Ocean in SON (supports P1 bridge to IOD):
  - WIO warming at 0 to +3 months with a west–east contrast indicative of a positive‑IOD tendency.
- DMI(SON) → Pacific at +6 to +9 months (supports P3):
  - Equatorial Pacific cooling (−0.15 to −0.3 K per 1σ DMI) and negative sea level develop by MAM–JJA next year, with easterly anomalies—consistent with a delayed oceanic pathway.

3) Equatorial Pacific Hovmöller (z20 and 10 m u; ATL3‑keyed)
- Positive ATL3 composite:
  - JJA–SON: 0.5–1.5 m s−1 easterlies over 140°E–170°W launch an eastward‑propagating upwelling Kelvin wave; z20 shoals from ~160°E to 150°–100°W over +3 to +6 months and persists to +9 months.
- Negative ATL3 composite:
  - Mirror‑image westerlies and downwelling Kelvin wave deepen the thermocline eastward into DJF.
- Interpretation: Textbook Kelvin‑wave causality from ATL3‑forced winds to EP thermocline adjustment, peaking in DJF—direct verification of the atmospheric bridge and EP‑leaning response (P1).

4) PCMCI causal network (seasonal nodes; τmax=18 months; FDR‑significant)
- ATL3(JJA) → Niño3.4(DJF): semi‑partial r ≈ −0.53 at ~6‑month lag; stronger for EP than CP (EP ≈ −0.40; CP ≈ −0.33). Supports P1 and EP emphasis.
- DMI(SON) → ENSO(DJF): strongest for EP (≈ +0.73) and Niño3.4 (≈ +0.62) at ~3‑month lag—captures the near‑contemporaneous IOD–El Niño link.
- ENSO(DJF) → IOB(MAM): Niño3.4 → IOB ≈ +0.74 (EP ≈ +0.59; CP ≈ +0.46). Supports Pacific→Indian closure.
- Not retained by FDR: DMI → ITF proxy and ITF → ENSO links, likely due to short/noisy altimetry proxies and seasonal node choices; ATL3 → DMI not detected as a standalone causal edge here.

5) Wavelet coherence (time–frequency stability)
- ATL3(JJA) vs Niño3.4(DJF):
  - Significant 2.5–4‑yr coherence ~1994–2000; phase indicates anti‑phase with ATL3 leading by ~6 months—consistent with the bridge (P1).
- DMI(SON) vs Niño3.4(next JJA):
  - Significant 2.3–4‑yr band ~1993–2002 with anti‑phase and a downward tilt implying a 6–12‑month lead—consistent with the delayed tunnel pathway (P3).
- Both links are intermittent and strongest in the mid‑to‑late 1990s, implying state dependence (e.g., IPO/AMV/Walker background).

6) Observations vs CMIP6 (SST regressed onto ATL3 JJA)
- Observations:
  - Lag −6 to −3 months: La Niña‑like Pacific cooling precedes ATL3 (ENSO→TNA/IOB closure signature).
  - Lag 0 to +3 months: ATL3 warming with WIO warming and a positive‑IOD‑like SON pattern.
  - +6 to +9 months: Pronounced EP‑leaning La Niña in DJF–MAM.
- Multi‑model mean (and most models):
  - Capture WIO warming but largely miss or reverse the EP cooling signal at +6/+9 months; IOD dipole is weak or replaced by basinwide warming.
  - Implication: Systematic model bias in representing Atlantic‑forced Kelvin response and/or Pacific Bjerknes feedback; only a minority show the observed sequence.

Integrated interpretation and quantitative synthesis
- A two‑step forcing emerges in observations:
  1) Fast atmospheric bridge (ATL3+ → SON): WIO warming and positive‑IOD tendency; equatorial Pacific easterlies and EP thermocline shoaling initiate within a season.
  2) Slow oceanic tunnel (IOD+ → next MAM/JJA): Enhanced Indo‑Pacific heat export sustains/extends Pacific cooling 6–12 months later.
- Effect sizes (per 1σ index):
  - Equatorial Pacific SST cooling: ≈ −0.2 to −0.3 K by DJF (ATL3 lead) and by MAM–JJA (DMI lead).
  - Equatorial sea level: ≈ −2 to −3 cm (shoaling) at corresponding lags.
  - Equatorial winds: ≈ 0.5–1.5 m s−1 easterly anomalies following ATL3+.

Assessment against predictions
- P1 (ATL3→EP‑leaning La Niña; WIO warming/IOD tendency): Supported by SST, sea level, winds, and Hovmöller; PCMCI quantifies the 6‑month ATL3→Niño3.4 link.
- P2 (ENSO→IOB and closure): Supported for ENSO→IOB; closure to ATL3 is suggested (lag −6 to −3 signatures) but not quantified causally here.
- P3 (IOD→delayed Pacific cooling via tunnel): Supported by lead–lag maps and wavelet coherence; causal ITF proxy link not retained (data/lag limitations).
- P4 (Indian–Atlantic via Agulhas leakage): Not evaluated by these figures.
- P5 (decadal AMV/PDO modulation): Not addressed in these interannual diagnostics.

Caveats and limitations
- Atmospheric (precipitation) and wind regressions are spatially patchy after strict FDR masking, though dynamically consistent where significant.
- Seasonal targeting is essential (ATL3 in JJA; DMI in SON); monthly indices introduce noise.
- ITF and Agulhas links need higher‑fidelity proxies and altimetry‑era focus; PCMCI nulls here are inconclusive.
- Nonstationarity: strongest coherence and teleconnections in the mid‑to‑late 1990s, indicating background‑state dependence.

Actionable prediction
- Years with ATL3+ in JJA followed by DMI+ in SON have the highest probability of an EP‑leaning La Niña peaking in DJF and persisting into the following spring–summer.

Implications for models and prediction
- Observations verify the Atlantic→Indian→Pacific chain at interannual scales; most CMIP6 models miss the sign and structure of the Pacific response to ATL3, limiting three‑ocean predictive skill.
- Prioritize models that capture the ATL3→EP cooling and IOD dipole for seasonal forecasts; improve representation of tropical Atlantic convection, the atmospheric Kelvin response, ITCZ biases, and Pacific thermocline feedbacks to recover the observed teleconnection strength.
\end{lstlisting}
\end{summarybox}

\subsection{Emergent constraints on subtropical eastern Pacific low-cloud shortwave feedbacks under global warming}

The response of subtropical low clouds to global warming represents one of the largest uncertainties in climate projections~\cite{myers2016reducing}. These vast decks of stratocumulus clouds, primarily in the eastern Pacific, exert a significant cooling effect on the planet by reflecting incoming solar radiation. Climate models, however, disagree substantially on how the coverage and radiative properties of these clouds will change in a warmer world~\cite{ceppi2017cloud}. This divergence in the shortwave cloud feedback directly contributes to the wide range in projected global warming magnitudes. Reducing this uncertainty is therefore a critical priority for climate science. To address this, we tasked \model~with a challenging objective: to discover and apply emergent constraints to the long-term trend of the subtropical Eastern Pacific low-cloud shortwave feedback.

After autonomous experiments, \model~identified a novel and physically robust factor for constraining the future trend of the subtropical eastern Pacific low-cloud shortwave feedback. Among ten candidate predictors autonomously defined and tested by \model, ranging from mean-state climate variables to process-oriented sensitivity metrics, a factor termed ``Regime-slope'' emerged as the most reliable constraint. This metric quantifies the sensitivity of the shortwave cloud radiative effect (SWCRE) to perturbations in the lower tropospheric stability, specifically the Estimated Inversion Strength (EIS), conditioned across different stability regimes. The physical mechanism underpinning this relationship is compelling: EIS is the primary controller of the temperature inversion that caps the marine boundary layer, regulating the turbulent entrainment of dry air from above. A model's ability to realistically capture the nuanced, state-dependent relationship between cloud radiative properties and this stability in the present climate is intrinsically linked to its fidelity in simulating how these clouds will evolve as the background climate state changes. By applying the observational constraint from ``Regime-slope'', our analysis significantly narrows the uncertainty in the feedback's future. The results indicate that the trend in the combined Northeast and Southeast Pacific regions is most likely to be small and slightly positive, converging around a value of 0 to +0.1 W m$^{-2}$ K$^{-1}$ decade$^{-1}$. This suggests that the positive low-cloud feedback is unlikely to continue strengthening throughout the century and will instead tend to weaken or stabilize.

The primary contribution of this discovery is to provide crucial insight into the non-linear evolution of the climate system. The identification of ``Regime-slope'' as a top-performing constraint introduces a new, process-based diagnostic for evaluating climate models, emphasizing the importance of correctly simulating state-dependent cloud-inversion relationships.  

The input and output are shown below.

\begin{requestbox}
\begin{lstlisting}
I want to analyze the relationship between the shortwave feedback from subtropical low clouds and global warming magnitude.
1. Please help me apply an emergent constraint to the changing trend of the shortwave feedback from the subtropical eastern Pacific low cloud region. You can define 10 potentially effective constraints, which can be undiscovered variables, scalar features, or statistical features. You can superimpose the probability distribution functions of the original and constrained trends to determine whether future changes are constrained. The values must be presented on a horizontal and vertical scale.
2. For each constrained factor, please rank its reliability. Select the most reliable constraint and explain in your analysis why this constraint is useful and what the physical reasons are.
\end{lstlisting}
\end{requestbox}

\begin{planbox}
\begin{lstlisting}
Experimental Objectives
- To constrain the temporal trend of shortwave cloud feedback in the subtropical Eastern Pacific low-cloud region (unit: W m⁻² K⁻¹ decade⁻¹) using an emergent constraint approach and to evaluate its relationship with the magnitude of global warming.
- To establish inter-model relationships for 10 candidate observable factors respectively, obtaining posterior (constrained) distributions and comparing them with the prior (original multi-model) distributions.
- To quantitatively rank the 10 factors for constraint reliability, select the most reliable factor, and provide a physical mechanism explanation for it in the results phase.

Research Area and Time
- Region (Ocean grids, excluding land):
  1) Northeast Pacific (NEP): 15°N–30°N, 150°W–110°W
  2) Southeast Pacific (SEP): 30°S–15°S, 110°W–80°W
  3) Combined Region: NEP ∪ SEP (area-weighted)
- Time:
  - Observational base period (for calculating candidate factors): 2001–2014 (monthly)
  - Model historical period (for matching candidate factors): 1980–2014 (use 2001–2014 if observational overlap is limited)
  - Future period (for calculating the target variable and its trend): 2015–2100 (ScenarioMIP/ssp585; monthly)
- Resolution: Monthly average (if model/observational calendars differ, standardize to a standard monthly sequence)

Data and Variables (Observations preferred, variable names follow CMIP/obs4MIPs conventions)
- Observations/Reanalysis
  - CERES-EBAF (monthly): rsut, rsutcs (W m⁻²); derive swcre = rsut − rsutcs (positive upwards)
  - ERA5 (monthly): tas (K), ta (K, including 700 hPa level), ps (Pa), wap (Pa s⁻¹, at 500 hPa level), hus (kg kg⁻¹, if used for EIS), zg (m, if height conversion is needed)
  - Sea Surface Temperature (HadISST, NOAA OISST, or ERSST, monthly): tos (K or °C, units to be unified)
  - Low Cloud Observations (preferred): CALIPSO-GOCCP (if available, use variable names like cltcalipso or equivalent low cloud fraction product); if unavailable, use clt and apply filtering for low-cloud region conditions as a substitute.
- CMIP6 (Amon, monthly frequency)
  - Historical: historical (1980–2014 or 2001–2014)
  - Future: ScenarioMIP/ssp585 (2015–2100)
  - Variables: rsut, rsutcs, tas, ta (700 hPa), ps, wap (500 hPa), tos, clt (%); optional: zg, hus for EIS
  - Model Ensemble: Include as many models as possible that have the above variables (preferably using r1i1p1f1 or the first available member for each model)

Derived and Diagnostic Quantities
- Shortwave Cloud Radiative Effect (SWCRE): swcre = rsut − rsutcs (W m⁻², positive upwards)
- Regional Average: Uniform regridding (e.g., 1°×1° or 2.5°×2.5°), cosine(latitude) weighting, ocean mask applied
- Magnitude of Global Warming: ΔT_glob = Global mean tas (2081–2100) − (1995–2014) (unit: K)
- Lower Tropospheric Stability (LTS): LTS = θ_700 − θ_sfc, where θ = T (p_0/p)^(R/c_p)
- Estimated Inversion Strength (EIS): EIS = LTS − Γ_m × (z_700 − z_LCL), where Γ_m is an approximation of the moist adiabatic lapse rate near the surface, and z_LCL is estimated from surface temperature and humidity; if necessary humidity information is missing, document the use of a simplified approximation.

Target Variable: Temporal Trend of Regional Shortwave Low-Cloud Feedback
- For each model and each region, calculate the time series of feedback slope segments:
  1) Construct monthly anomalies for 2015–2100: regional swcre' (deseasonalized), global tas' (deseasonalized)
  2) Use a sliding window or piecewise regression during the future period to estimate λ_{SW,cld} (unit: W m⁻² K⁻¹)
     - Recommended three-segment method: 2015–2044, 2045–2074, 2075–2100. Perform OLS regression swcre' = α + λ × tas' + ε for each segment to get three λ values.
     - Alternatively, use a 10-year sliding window (1-year step) to form λ(t).
  3) Trend of change trend(λ_{SW,cld}): Calculate the slope of a linear regression between the three λ values and their segment-center years (or the slope of λ(t) regressed against time), unit: W m⁻² K⁻¹ decade⁻¹.
- Prior Distribution: The sample of trend(λ_{SW,cld}) across models constitutes the prior PDF (using kernel density estimation or a normal approximation).

Relationship with Global Warming Magnitude
- Calculate the inter-model correlation: Correlation coefficient and linear regression between trend(λ_{SW,cld}) and ΔT_glob (not for the primary constraint, but for result presentation and background analysis).

10 Candidate Constraint Factors (Calculated using the same definition and region for the observational period and model historical period)
1) Long-term Mean Low Cloud Fraction (LCF-mean, %)
   - Preferably use cll or a dedicated low cloud product; if not available, use the conditional mean of clt under "low-cloud dominant conditions" (condition: grids with high-quantile EIS and wap500 > 0).
   - Period: 2007–2014 (if using CALIPSO), otherwise 2001–2014 (using clt as a proxy).
2) Seasonal Amplitude of Low Cloud Fraction (LCF-amp, %)
   - The maximum minus minimum or the half-amplitude of the 2001–2014 monthly climatology.
3) Long-term Mean EIS (EIS-mean, K)
   - Use the same formula for ERA5 and models; if a simplified z_LCL/Γ_m is used, document its consistent application.
4) Long-term Mean 500 hPa Subsidence Strength (W500-mean, Pa s⁻¹)
   - Define subsidence strength = +wap500 (Pa s⁻¹), where positive values indicate subsidence.
5) Interannual Regression Slope of SWCRE on EIS (CRE–EIS β, W m⁻² K⁻¹)
   - For 2001–2014, first remove seasonal cycles and linear trends, then perform OLS on annual or monthly-averaged annual means.
6) Interannual Regression Slope of Low Cloud Fraction on SST (LCF–SST β, % K⁻¹)
   - The regression slope of LCF' against regional tos' for 2001–2014.
7) Regional Relative Warm/Cold State (SST'-glob, K)
   - The long-term mean regional SST minus the long-term mean global SST (2001–2014).
8) LCF Autocorrelation e-folding Timescale (LCF-τ, months)
   - The first-order e-folding time of the autocorrelation function of the 2001–2014 monthly series (reflecting low cloud persistence/memory).
9) Covariance between SWCRE and LCF (Cov(CRE,LCF), W m⁻² %)
   - The covariance between monthly anomalies from 2001–2014.
10) CRE Slope under Conditioned EIS Distribution (Regime-slope, W m⁻² K⁻¹)
    - Group monthly samples by EIS quantiles (e.g., tertiles), regress swcre' on EIS' within each group, and take the sample-weighted average of the slopes.

Data Preprocessing and Quality Control
- Unit and Sign Convention Unification: rsut/rsutcs (W m⁻², positive upwards); swcre positive upwards; tas/tos (K); wap (Pa s⁻¹, define "subsidence strength = +wap"); cloud fraction % (0–100).
- Regridding and Masking: Unify grids and apply ocean masks; use area-weighting for regional averages.
- Deseasonalization and Anomalies: For interannual regression factors, remove the mean monthly seasonal cycle; remove linear trends if necessary; ENSO influence can be removed by regressing out the NINO3.4 index (ensure method consistency between observations and models).
- Valid Samples and Missing Data: Record the number of valid samples; do not interpolate missing data into regression samples.
- Observational Uncertainty: Use uncertainty from product documentation or the standard error of the time series sampling, converted to the regional statistic; reflect this in the subsequent likelihood calculation.
- Model Internal Variability: For each model's factor and target variable, estimate sampling uncertainty using a time-block bootstrap method.

Computational Workflow
1) Target Variable and Prior
   - For each model and region, calculate the three-segment λ_SW,cld values and their trend, trend(λ_SW,cld) (W m⁻² K⁻¹ decade⁻¹).
   - Collect the cross-model sample of trend(λ) to form the prior PDF (x-axis: W m⁻² K⁻¹ decade⁻¹; y-axis: probability density).
2) Candidate Factors (Observations and Models)
   - Observations: Calculate the 10 factors x_obs and their standard errors σ_obs as defined above (for NEP/SEP/Combined separately).
   - Models: Calculate the 10 factors in the historical period for the same time frame, yielding the cross-model set X_i.
3) Inter-model Emergent Relationship
   - For each factor *i*, construct a scatter plot of Y_i = trend(λ) versus X_i; use orthogonal distance regression (or total least squares) that accounts for errors in both variables, incorporating the sampling uncertainties of X_i and Y_i.
   - Statistical Outputs: Correlation coefficient *r*, regression slope/intercept and their confidence intervals, variance of the regression residuals.
   - Leave-One-Out Cross-Validation (LOO): Iteratively remove one model and refit, recording the stability of the slope and the prediction error.
4) Observational Constraint and Posterior PDF
   - Observational Likelihood: x_obs ~ N(μ_obs, σ_obs²)
   - Linear Relationship: Y|X ~ N(α + βX, σ_reg²), where α, β, and σ_reg come from the inter-model regression and its uncertainty.
   - Use Monte Carlo sampling to simultaneously draw from the distributions of x_obs and the regression parameters to obtain the posterior PDF of Y.
   - Superimpose the prior and posterior PDFs for visualization (x-axis: trend(λ_SW,cld) W m⁻² K⁻¹ decade⁻¹; y-axis: probability density) to determine if a significant constraint has been achieved.
5) Relationship with ΔT_glob
   - Calculate the inter-model correlation and regression between trend(λ_SW,cld) and ΔT_glob; display the scatter plot and regression line.

Reliability Assessment and Ranking
- Metric System (calculated independently for each factor, for NEP/SEP/Combined and overall):
  1) S1 Statistical Strength: |r| and the significance of the regression slope (higher score if the confidence interval does not cross zero).
  2) S2 Narrowing Degree: The percentage reduction of the posterior 95% interval relative to the prior 95% interval.
  3) S3 Predictive Skill: The percentage reduction in the LOO root-mean-square prediction error relative to the prior standard deviation.
  4) S4 Regional Consistency: Consistency of the sign and magnitude of the correlation across NEP/SEP/Combined (e.g., high score if correlations have the same sign and similar magnitude).
  5) S5 Physical Consistency: Degree of support from process-based metrics (e.g., whether CRE–EIS β, Regime-slope, and the mechanism implied by the factor are consistent).
- Normalize S1–S5 to [0,1] and calculate a composite reliability score: R = 0.35*S1 + 0.25*S2 + 0.20*S3 + 0.10*S4 + 0.10*S5.
- Output a reliability ranking of the 10 factors (can be ranked for NEP/SEP/Combined separately and overall).
- Selection of the Most Reliable Factor: The factor with the highest R is designated as the "most reliable constraint factor" and proceeds to the physical interpretation stage.

Physical Interpretation Framework for the Most Reliable Factor (to be executed in the results phase)
- Explain the process pathway linking this factor to the evolution of low-cloud feedback, for example:
  - How stability (EIS/LTS) and subsidence (wap500) control the stratocumulus-capping inversion, turbulence, and entrainment.
  - How the sea surface temperature state and gradients maintain coastal low clouds and their sensitivity to warming.
  - How the cross-timescale consistency of process sensitivity metrics (CRE–EIS β, Regime-slope) with this factor supports its extrapolation to a long-term warming context.
- Argue for the physical plausibility of the constraint by comparing the direction and magnitude of its relationship with other factors (e.g., LCF–SST β, Cov(CRE,LCF)).
- Do not presuppose a conclusion; provide an explanation based solely on the statistical and physical evidence obtained.

Uncertainty and Statistical Details
- Observational End: Estimate σ_obs for each factor (standard error of time sampling or product-given uncertainty); explicitly propagate this in the posterior sampling.
- Model End: Estimate sampling uncertainty for each model's X_i and Y_i using a time-block bootstrap method; use an errors-in-variables regression for the inter-model analysis to mitigate measurement error bias.
- Residual Structure: Check the normality and homoscedasticity of the regression residuals; if necessary, use robust regression for comparison and reflect this in the S1/S3 scores.
- Collinearity: Assess the pairwise correlations among the 10 factors; penalize highly redundant factors in S5 to avoid misleading results from repetitive information.

Figures and Outputs
- Regional Map: A map showing the NEP/SEP/Combined region masks overlaid with the long-term mean of observed EIS, SWCRE, and LCF (with corresponding units).
- Target Variable Time Series: Time series of λ_SW,cld(t) for several representative models with their linear fits for 2015–2100.
- Inter-model Scatter Plots: Scatter plots of Y_i (trend(λ)) versus X_i with regression lines, 95% confidence bands, and the vertical projection of the observational value x_obs±σ_obs.
- PDF Overlay Plots: Plots showing the prior PDF overlaid with the posterior PDF from each factor (x-axis: W m⁻² K⁻¹ decade⁻¹; y-axis: probability density).
- Ranking Chart: A chart showing the reliability score R and its components (S1–S5) for the 10 factors.
- Association Analysis: A scatter plot of trend(λ_SW,cld) vs. ΔT_glob with a regression line.

Summary of Key Variables, Units, and Time Windows
- rsut, rsutcs (W m⁻², Amon, 2001–2014; 2015–2100)
- swcre = rsut − rsutcs (W m⁻², positive upwards)
- tas (K, global mean, 1995–2014 and 2081–2100; monthly anomalies for 2015–2100)
- tos (K, regional SST characteristic, 2001–2014)
- ta700 (K), ps (Pa), hus (kg kg⁻¹, optional), zg (m, optional) for EIS/LTS (2001–2014)
- wap500 (Pa s⁻¹, 2001–2014)
- clt or low cloud fraction product (%, 2001–2014)
- λ_SW,cld (W m⁻² K⁻¹), trend(λ_SW,cld) (W m⁻² K⁻¹ decade⁻¹)
- ΔT_glob (K)

Notes on Comparability and Consistency
- The definitions of factors, time windows, and masks must be strictly consistent between observations and models.
- If direct low cloud observations are unavailable, use clt under "low-cloud dominant conditions" as a proxy and note its comparability in the ranking.
- The symbol ω is standardized to dp/dt (Pa s⁻¹), and "subsidence strength = +wap" is defined; SWCRE is positive upwards (increased reflection → larger numerical value).

Workflow Overview
1) Unify data and grids, and construct monthly time series for the NEP/SEP/Combined regions (for both observations and models).
2) Calculate the model target variable—the three-segment λ_SW,cld values and their trend, trend(λ)—to form the prior sample and prior PDF.
3) Calculate the 10 candidate factors (for observations and models) to obtain X_i and x_obs±σ_obs.
4) For each factor, establish an inter-model regression and perform LOO validation. Generate the posterior PDF based on x_obs and overlay it with the prior for comparison.
5) Calculate the S1–S5 scores to form the composite reliability score R and the ranking, and select the most reliable factor.
6) Provide a physical interpretation for the most reliable factor based on the evidence obtained (without presupposing a conclusion).
7) Present the inter-model relationship between trend(λ_SW,cld) and ΔT_glob as a background association result.

This plan uses only CMIP and satellite/reanalysis observational data, with a preference for observations. Variable names, units, time periods, and regional definitions are clearly specified. It outlines 10 feasible constraint factors, a method for superimposing prior/posterior PDFs (x-axis: W m⁻² K⁻¹ decade⁻¹, y-axis: probability density), and provides a framework for quantitative reliability ranking and a physical interpretation of the most reliable factor.

\end{lstlisting}
\end{planbox}

\begin{visualbox}
  \begin{center}
  \includegraphics[width=1.0\linewidth]{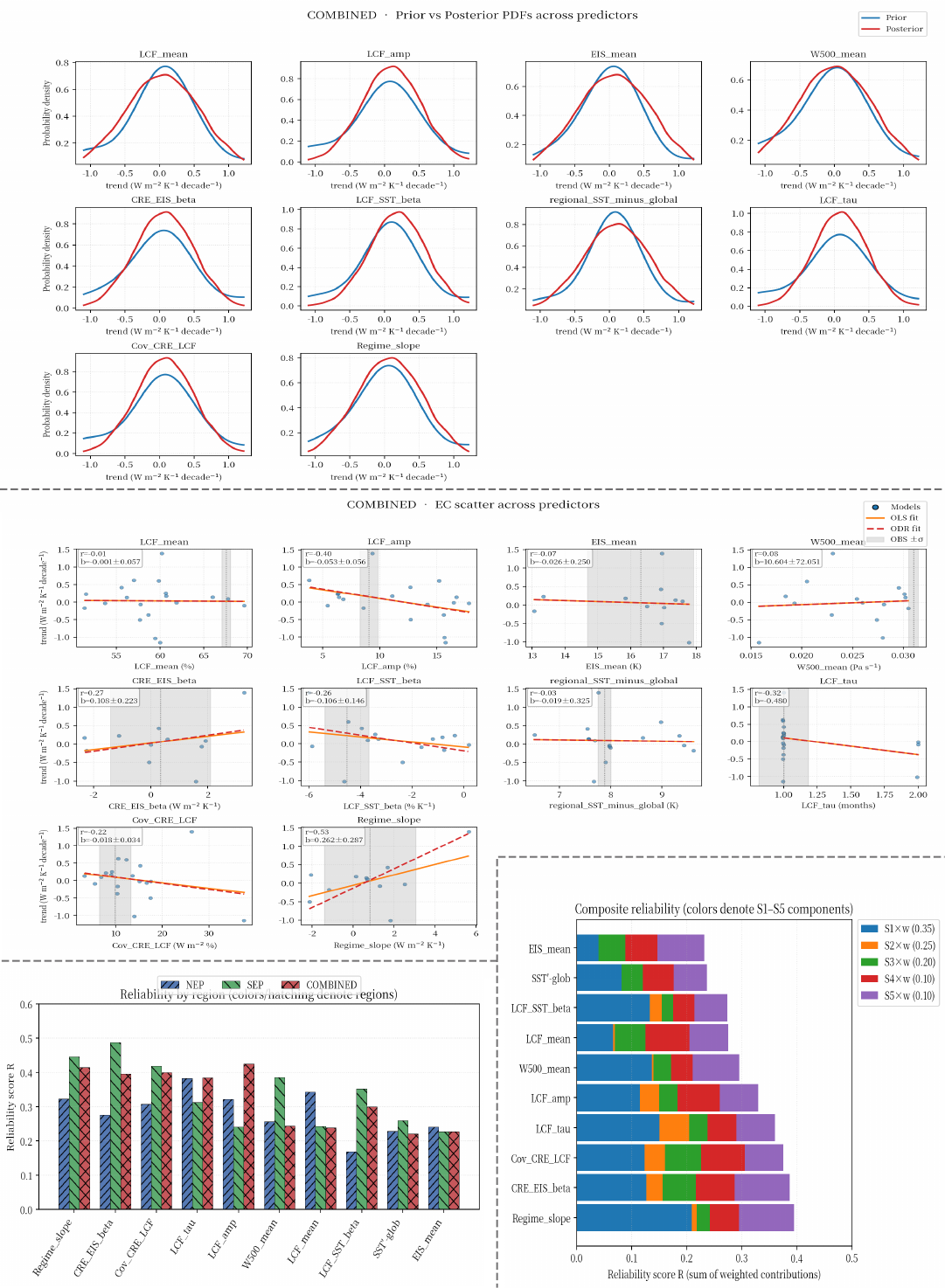}    
  \end{center}
\end{visualbox}

\begin{summarybox}
\begin{lstlisting}
# Summary of Emergent Constraints on the Trend of Subtropical Eastern Pacific Low Cloud Shortwave Feedback

This summary is based on a comprehensive interpretation of scatter plots and prior/posterior PDF overlays for three regions (Northeast Pacific - NEP, Southeast Pacific - SEP, and Combined). The analysis focuses on the constraining effect on the "trend of shortwave cloud feedback," denoted as trend(λ_{SW,cld}). The unit for trend(λ_{SW,cld}) is W m⁻² K⁻¹ decade⁻¹, where a negative value indicates an evolution towards a stronger positive feedback (albedo decreases) over time, and a positive value indicates a trend towards a weaker positive feedback or a negative feedback (albedo increases).

- Methodology Highlights: For each candidate predictor X_i, an orthogonal distance regression is established across models: Y_i = trend(λ_{SW,cld}) ~ α + β X_i. The observed value X_obs and its uncertainty are incorporated through Monte Carlo simulations to derive a posterior PDF. Leave-one-out cross-validation is used to assess predictability.
- Time Windows: Predictors are defined for the period 2001–2014, consistent between observations and models. The target variable, trend(λ_{SW,cld}), is estimated for the period 2015–2100 under the SSP585 scenario.

## Prior and Posterior PDF Overlays (Constrainment Effect)
- Axes Description: The horizontal axis represents trend(λ_{SW,cld}) (W m⁻² K⁻¹ decade⁻¹); the vertical axis represents probability density. The blue line is the prior distribution (raw distribution across models), and the red line is the posterior distribution (after applying a constraint from a specific predictor).
- Combined Region (NEP∪SEP): Most predictors cause a significant narrowing of the posterior distribution, with the peak concentrating near "close to 0, slightly positive" (approximately 0 to +0.1). This indicates an increased probability of a future trend that is "small in magnitude and more likely slightly positive."
- NEP: The posterior distribution generally shifts to the right (more positive) compared to the prior and shows moderate narrowing. This implies that the low cloud shortwave feedback tends towards a "weaker positive feedback" over time (i.e., albedo does not continue to decrease significantly).
- SEP: Under the influence of multiple predictors, the posterior distribution converges near zero or slightly negative and narrows significantly. This suggests the trend is more likely "near-zero to weakly negative" (meaning the feedback evolution is small, or the probability of it moving towards a stronger positive feedback is low).

Conclusion: The PDF overlays show that the future trend is "significantly constrained" (tails are reduced, peak is enhanced). Clear regional differences emerge: the NEP tends towards a "slightly positive trend," the SEP is closer to a "near-zero/weakly negative trend," and the combined region shows a "weakly positive and limited magnitude" trend.

## 10 Candidate Predictors and Their Constraint Effects (Synthesized from multiple analyses)
1) Regime_slope (Weighted average of CRE vs. EIS slope, conditioned on EIS quantiles)
   - Highest correlation (r ≈ 0.5+) in the combined region. The posterior narrows significantly with a small to moderate shift in the mean (towards near-zero/slightly positive).
2) CRE_EIS_beta (Interannual regression slope of CRE onto EIS)
   - Significant convergence in all three regions; strongest in the SEP. The direction is consistent with Regime_slope (towards near-zero/slightly positive).
3) LCF_SST_beta (Interannual regression slope of low cloud fraction onto SST)
   - Obvious convergence, moderately strong in the SEP. Provides independent process-based evidence supporting the above two predictors.
4) Cov(CRE,LCF) (Covariance of monthly anomalies in CRE and low cloud fraction)
   - Clear convergence in all regions; shifts slightly negative in the SEP. Helps to reduce the extreme tails of the distribution.
5) LCF_tau (e-folding autocorrelation timescale of low cloud fraction)
   - One of the most significant constraints in all three regions. Shifts slightly positive in NEP/Combined and slightly negative in SEP. Represents the predictability associated with "cloud persistence/memory."
6) W500_mean (Climatological mean 500 hPa subsidence)
   - Moderate constraint in SEP, stronger in NEP (shifts right and narrows); relatively weak in the combined region.
7) LCF_amp (Seasonal amplitude of low cloud fraction)
   - Moderate convergence, with the direction being region-dependent. Suggests that "a good representation of the annual cycle → the future trend will not be extreme."
8) LCF_mean (Climatological mean low cloud fraction)
   - Limited convergence and very small shift in the mean; useful for filtering out extreme model projections.
9) EIS_mean (Climatological mean Estimated Inversion Strength)
   - The weakest constraint. The mean-state stability alone is insufficient to explain the "rate of change of the trend."
10) SST′-glob (Climatological deviation of regional SST from global mean SST)
    - Almost no constraint; not recommended for use as a primary constraint on its own.

## Reliability Ranking (Overall/Regional)
Note: The ranking is based on a composite reliability score (R) derived from five metrics (S1–S5: statistical strength, narrowing of distribution, predictive skill, regional consistency, and physical consistency). R is ranked from high to low (approximate values are qualitative).

- Combined Region (Overall Ranking)
  1) Regime_slope (~0.40, most reliable)
  2) CRE_EIS_beta (~0.39)
  3) Cov_CRE_LCF (~0.37–0.38)
  4) LCF_tau (~0.35–0.36)
  5) LCF_amp (~0.33)
  6) W500_mean (~0.30–0.31)
  7) LCF_mean (~0.27–0.29)
  8) LCF_SST_beta (~0.25–0.26)
  9) SST′-glob (~0.23–0.24)
  10) EIS_mean (~0.23–0.24)

- SEP (Regional Ranking, core stratocumulus deck)
  1) W500_mean
  2) CRE_EIS_beta
  3) LCF_SST_beta
  4) Regime_slope
  5) Cov_CRE_LCF
  (Others are weaker)

- NEP (Overall weaker constraining power)
  1) LCF_tau
  2) Regime_slope (sign of relationship differs from SEP)
  3) LCF_mean
  4) EIS_mean
  5) LCF_amp
  (Others are weak or ineffective)

Note: LCF_tau and LCF_amp are particularly effective in "narrowing the distribution (S2)," but their overall rank is moderate because their physical mechanisms are considered less robust for extrapolation than process-sensitivity metrics.

## The Most Reliable Constraint and its Physical Basis
- Conclusion: Regime_slope is the most reliable single emergent constraint for the combined region.
- Statistical Evidence:
  - Highest correlation in the combined region (r ≈ 0.53); Orthogonal Distance Regression (ODR) is consistent with Ordinary Least Squares (OLS).
  - The posterior PDF narrows significantly with a sharper peak and reduced tails.
  - Leave-one-out cross-validation demonstrates stable predictive skill.
  - Directionally consistent with CRE_EIS_beta and LCF_SST_beta, indicating high physical consistency (S5). Shows robust performance in the SEP and combined regions (moderate S4).
- Physical Mechanism:
  - EIS governs the inversion at the top of the stratocumulus layer and the rate of entrainment mixing, which determines low cloud fraction and liquid water path, thereby modulating the shortwave cloud radiative effect (SWCRE).
  - Regime_slope quantifies the "instantaneous sensitivity of CRE to EIS perturbations under different stability regimes," which can isolate non-linearities and state dependencies.
  - In a warming climate, long-term changes in EIS and the large-scale subsidence field are mapped onto the long-term evolution of SWCRE via this sensitivity. A larger Regime_slope (i.e., more "controlled by EIS" in the current climate) means that small future changes in EIS/circulation are more easily translated into a sustained change in CRE, thus making trend(λ_{SW,cld}) more predictable.
  - This is supported by the high reliability of W500_mean in the SEP: stronger mean subsidence maintains a stable inversion cap, making the "EIS→CRE" process chain more robust for long-term extrapolation under a linear approximation.
- Comparison with Sub-optimal Predictors:
  - CRE_EIS_beta (interannual scale) and Regime_slope (state-conditioned) are mutually reinforcing in terms of process. Although LCF_tau/LCF_amp provide strong narrowing, their mechanistic basis and transferability are slightly weaker, placing their composite reliability score (R) below the former two.

## Relationship with Global Warming Magnitude (ΔTglob) (Context)
- Cross-model scatter plots show a weak-to-moderate correlation between trend(λ_{SW,cld}) and ΔTglob, with significant regional dependence. Therefore, it was not used as a primary constraint. However, the constrained trend—pointing to a "weakly positive" outcome in the combined region, "slightly positive" in the NEP, and "near-zero" in the SEP—implies a reduction in the uncertainty of future warming associated with low clouds.

## Key Conclusions and Recommendations
- Conclusions:
  - The constrained trend in the combined region is most likely "close to 0, slightly positive" (in W m⁻² K⁻¹ decade⁻¹). This suggests that in the latter half of the century, the shortwave low cloud feedback will generally tend to become a weaker positive feedback or at least less extreme compared to earlier periods.
  - The posterior for NEP shifts right (more positive), while the posterior for SEP is closer to zero or slightly negative, indicating a slight "anti-phase" behavior between the north and south, with the combined magnitude being partially canceled out.
- Recommendations:
  - Use Regime_slope as the primary constraint, with CRE_EIS_beta and LCF_SST_beta for cross-validation. In the SEP, the weight of W500_mean could be increased. For the NEP, LCF_tau can be included to enhance convergence.
  - It is not recommended to use EIS_mean or SST′-glob as primary constraints due to their weak constraining power and poor extrapolability.
  - To combine multiple predictors into a quantitative posterior, a Bayesian merging approach (accounting for inter-predictor correlations) is recommended, prioritizing process-sensitivity metrics and supplementing them with cloud memory metrics.
\end{lstlisting}
\end{summarybox}


\stopcontents[appendix]


\end{appendices}

\putbib[sn-bibliography]
\end{bibunit}
\vspace*{-\baselineskip}


%% file: sn-article-arxiv.bbl

\begin{thebibliography}{49}
\ifx \bisbn   \undefined \def \bisbn  #1{ISBN #1}\fi
\ifx \binits  \undefined \def \binits#1{#1}\fi
\ifx \bauthor  \undefined \def \bauthor#1{#1}\fi
\ifx \batitle  \undefined \def \batitle#1{#1}\fi
\ifx \bjtitle  \undefined \def \bjtitle#1{#1}\fi
\ifx \bvolume  \undefined \def \bvolume#1{\textbf{#1}}\fi
\ifx \byear  \undefined \def \byear#1{#1}\fi
\ifx \bissue  \undefined \def \bissue#1{#1}\fi
\ifx \bfpage  \undefined \def \bfpage#1{#1}\fi
\ifx \blpage  \undefined \def \blpage #1{#1}\fi
\ifx \burl  \undefined \def \burl#1{\textsf{#1}}\fi
\ifx \doiurl  \undefined \def \doiurl#1{\url{https://doi.org/#1}}\fi
\ifx \betal  \undefined \def \betal{\textit{et al.}}\fi
\ifx \binstitute  \undefined \def \binstitute#1{#1}\fi
\ifx \binstitutionaled  \undefined \def \binstitutionaled#1{#1}\fi
\ifx \bctitle  \undefined \def \bctitle#1{#1}\fi
\ifx \beditor  \undefined \def \beditor#1{#1}\fi
\ifx \bpublisher  \undefined \def \bpublisher#1{#1}\fi
\ifx \bbtitle  \undefined \def \bbtitle#1{#1}\fi
\ifx \bedition  \undefined \def \bedition#1{#1}\fi
\ifx \bseriesno  \undefined \def \bseriesno#1{#1}\fi
\ifx \blocation  \undefined \def \blocation#1{#1}\fi
\ifx \bsertitle  \undefined \def \bsertitle#1{#1}\fi
\ifx \bsnm \undefined \def \bsnm#1{#1}\fi
\ifx \bsuffix \undefined \def \bsuffix#1{#1}\fi
\ifx \bparticle \undefined \def \bparticle#1{#1}\fi
\ifx \barticle \undefined \def \barticle#1{#1}\fi
\bibcommenthead
\ifx \bconfdate \undefined \def \bconfdate #1{#1}\fi
\ifx \botherref \undefined \def \botherref #1{#1}\fi
\ifx \url \undefined \def \url#1{\textsf{#1}}\fi
\ifx \bchapter \undefined \def \bchapter#1{#1}\fi
\ifx \bbook \undefined \def \bbook#1{#1}\fi
\ifx \bcomment \undefined \def \bcomment#1{#1}\fi
\ifx \oauthor \undefined \def \oauthor#1{#1}\fi
\ifx \citeauthoryear \undefined \def \citeauthoryear#1{#1}\fi
\ifx \endbibitem  \undefined \def \endbibitem {}\fi
\ifx \bconflocation  \undefined \def \bconflocation#1{#1}\fi
\ifx \arxivurl  \undefined \def \arxivurl#1{\textsf{#1}}\fi
\csname PreBibitemsHook\endcsname

\bibitem[\protect\citeauthoryear{Overpeck et~al.}{2011}]{overpeck2011climate}
\begin{barticle}
\bauthor{\bsnm{Overpeck}, \binits{J.T.}},
\bauthor{\bsnm{Meehl}, \binits{G.A.}},
\bauthor{\bsnm{Bony}, \binits{S.}},
\bauthor{\bsnm{Easterling}, \binits{D.R.}}:
\batitle{Climate data challenges in the 21st century}.
\bjtitle{Science}
\bvolume{331}(\bissue{6018}),
\bfpage{700}--\blpage{702}
(\byear{2011})
\end{barticle}
\endbibitem

\bibitem[\protect\citeauthoryear{Reichstein et~al.}{2019}]{reichstein2019deep}
\begin{barticle}
\bauthor{\bsnm{Reichstein}, \binits{M.}},
\bauthor{\bsnm{Camps-Valls}, \binits{G.}},
\bauthor{\bsnm{Stevens}, \binits{B.}},
\bauthor{\bsnm{Jung}, \binits{M.}},
\bauthor{\bsnm{Denzler}, \binits{J.}},
\bauthor{\bsnm{Carvalhais}, \binits{N.}},
\bauthor{\bsnm{Prabhat}, \binits{F.}}:
\batitle{Deep learning and process understanding for data-driven earth system science}.
\bjtitle{Nature}
\bvolume{566}(\bissue{7743}),
\bfpage{195}--\blpage{204}
(\byear{2019})
\end{barticle}
\endbibitem

\bibitem[\protect\citeauthoryear{Guo et~al.}{2017}]{guo2017big}
\begin{barticle}
\bauthor{\bsnm{Guo}, \binits{H.}},
\bauthor{\bsnm{Liu}, \binits{Z.}},
\bauthor{\bsnm{Jiang}, \binits{H.}},
\bauthor{\bsnm{Wang}, \binits{C.}},
\bauthor{\bsnm{Liu}, \binits{J.}},
\bauthor{\bsnm{Liang}, \binits{D.}}:
\batitle{Big earth data: A new challenge and opportunity for digital earth’s development}.
\bjtitle{International Journal of Digital Earth}
\bvolume{10}(\bissue{1}),
\bfpage{1}--\blpage{12}
(\byear{2017})
\end{barticle}
\endbibitem

\bibitem[\protect\citeauthoryear{Park et~al.}{2023}]{park2023papers}
\begin{barticle}
\bauthor{\bsnm{Park}, \binits{M.}},
\bauthor{\bsnm{Leahey}, \binits{E.}},
\bauthor{\bsnm{Funk}, \binits{R.J.}}:
\batitle{Papers and patents are becoming less disruptive over time}.
\bjtitle{Nature}
\bvolume{613}(\bissue{7942}),
\bfpage{138}--\blpage{144}
(\byear{2023})
\end{barticle}
\endbibitem

\bibitem[\protect\citeauthoryear{Masson-Delmotte et~al.}{2021}]{masson2021climate}
\begin{barticle}
\bauthor{\bsnm{Masson-Delmotte}, \binits{V.}},
\bauthor{\bsnm{Zhai}, \binits{P.}},
\bauthor{\bsnm{Pirani}, \binits{A.}},
\bauthor{\bsnm{Connors}, \binits{S.L.}},
\bauthor{\bsnm{P{\'e}an}, \binits{C.}},
\bauthor{\bsnm{Berger}, \binits{S.}},
\bauthor{\bsnm{Caud}, \binits{N.}},
\bauthor{\bsnm{Chen}, \binits{Y.}},
\bauthor{\bsnm{Goldfarb}, \binits{L.}},
\bauthor{\bsnm{Gomis}, \binits{M.}}, \betal:
\batitle{Climate change 2021: the physical science basis}.
\bjtitle{Contribution of working group I to the sixth assessment report of the intergovernmental panel on climate change}
\bvolume{2}(\bissue{1}),
\bfpage{2391}
(\byear{2021})
\end{barticle}
\endbibitem

\bibitem[\protect\citeauthoryear{Team et~al.}{2023}]{core2023ipcc}
\begin{botherref}
\oauthor{\bsnm{Team}, \binits{C.W.}},
\oauthor{\bsnm{Lee}, \binits{H.}},
\oauthor{\bsnm{Romero}, \binits{J.}}, et al.:
IPCC, 2023: Climate Change 2023: Synthesis Report, Summary for Policymakers. Contribution of Working Groups I, II and III to the Sixth Assessment Report of the Intergovernmental Panel on Climate Change.
IPCC Geneva, Switzerland
(2023)
\end{botherref}
\endbibitem

\bibitem[\protect\citeauthoryear{Cheng et~al.}{2025}]{cheng2025record}
\begin{botherref}
\oauthor{\bsnm{Cheng}, \binits{L.}},
\oauthor{\bsnm{Abraham}, \binits{J.}},
\oauthor{\bsnm{Trenberth}, \binits{K.E.}},
\oauthor{\bsnm{Reagan}, \binits{J.}},
\oauthor{\bsnm{Zhang}, \binits{H.-M.}},
\oauthor{\bsnm{Storto}, \binits{A.}},
\oauthor{\bsnm{Von~Schuckmann}, \binits{K.}},
\oauthor{\bsnm{Pan}, \binits{Y.}},
\oauthor{\bsnm{Zhu}, \binits{Y.}},
\oauthor{\bsnm{Mann}, \binits{M.E.}}, et al.:
Record high temperatures in the ocean in 2024.
Advances in Atmospheric Sciences,
1--18
(2025)
\end{botherref}
\endbibitem

\bibitem[\protect\citeauthoryear{Kennedy et~al.}{2024}]{kennedy2024state}
\begin{botherref}
\oauthor{\bsnm{Kennedy}, \binits{J.}},
\oauthor{\bsnm{Trewin}, \binits{B.}},
\oauthor{\bsnm{Betts}, \binits{R.}},
\oauthor{\bsnm{Thorne}, \binits{P.}},
\oauthor{\bsnm{Foster}, \binits{P.}},
\oauthor{\bsnm{Siegmund}, \binits{P.}},
\oauthor{\bsnm{Ziese}, \binits{M.}},
\oauthor{\bsnm{Mishra}, \binits{S.}},
\oauthor{\bsnm{Uhlenbrook}, \binits{S.}},
\oauthor{\bsnm{Alvar-Beltran}, \binits{J.}}, et al.:
State of the climate 2024. update for cop29
(2024)
\end{botherref}
\endbibitem

\bibitem[\protect\citeauthoryear{{World Meteorological Organization (WMO)}}{2025}]{WMO2025}
\begin{bbook}
\bauthor{\bsnm{{World Meteorological Organization (WMO)}}}:
\bbtitle{State of the Global Climate 2024},
\bedition{Wmo-no. 1368} edn.
\bpublisher{World Meteorological Organization (WMO)},
\blocation{Geneva}
(\byear{2025}).
\burl{https://library.wmo.int/records/item/69455-state-of-the-global-climate-2024}
\end{bbook}
\endbibitem

\bibitem[\protect\citeauthoryear{Stute et~al.}{2001}]{stute2001global}
\begin{barticle}
\bauthor{\bsnm{Stute}, \binits{M.}},
\bauthor{\bsnm{Clement}, \binits{A.}},
\bauthor{\bsnm{Lohmann}, \binits{G.}}:
\batitle{Global climate models: Past, present, and future}.
\bjtitle{Proceedings of the National Academy of Sciences}
\bvolume{98}(\bissue{19}),
\bfpage{10529}--\blpage{10530}
(\byear{2001})
\end{barticle}
\endbibitem

\bibitem[\protect\citeauthoryear{Heinze et~al.}{2019}]{heinze2019esd}
\begin{barticle}
\bauthor{\bsnm{Heinze}, \binits{C.}},
\bauthor{\bsnm{Eyring}, \binits{V.}},
\bauthor{\bsnm{Friedlingstein}, \binits{P.}},
\bauthor{\bsnm{Jones}, \binits{C.}},
\bauthor{\bsnm{Balkanski}, \binits{Y.}},
\bauthor{\bsnm{Collins}, \binits{W.}},
\bauthor{\bsnm{Fichefet}, \binits{T.}},
\bauthor{\bsnm{Gao}, \binits{S.}},
\bauthor{\bsnm{Hall}, \binits{A.}},
\bauthor{\bsnm{Ivanova}, \binits{D.}}, \betal:
\batitle{Esd reviews: Climate feedbacks in the earth system and prospects for their evaluation}.
\bjtitle{Earth System Dynamics}
\bvolume{10}(\bissue{3}),
\bfpage{379}--\blpage{452}
(\byear{2019})
\end{barticle}
\endbibitem

\bibitem[\protect\citeauthoryear{Eyring et~al.}{2016}]{eyring2016overview}
\begin{barticle}
\bauthor{\bsnm{Eyring}, \binits{V.}},
\bauthor{\bsnm{Bony}, \binits{S.}},
\bauthor{\bsnm{Meehl}, \binits{G.A.}},
\bauthor{\bsnm{Senior}, \binits{C.A.}},
\bauthor{\bsnm{Stevens}, \binits{B.}},
\bauthor{\bsnm{Stouffer}, \binits{R.J.}},
\bauthor{\bsnm{Taylor}, \binits{K.E.}}:
\batitle{Overview of the coupled model intercomparison project phase 6 (cmip6) experimental design and organization}.
\bjtitle{Geoscientific Model Development}
\bvolume{9}(\bissue{5}),
\bfpage{1937}--\blpage{1958}
(\byear{2016})
\end{barticle}
\endbibitem

\bibitem[\protect\citeauthoryear{Taylor et~al.}{2012}]{taylor2012overview}
\begin{barticle}
\bauthor{\bsnm{Taylor}, \binits{K.E.}},
\bauthor{\bsnm{Stouffer}, \binits{R.J.}},
\bauthor{\bsnm{Meehl}, \binits{G.A.}}:
\batitle{An overview of cmip5 and the experiment design}.
\bjtitle{Bulletin of the American meteorological Society}
\bvolume{93}(\bissue{4}),
\bfpage{485}--\blpage{498}
(\byear{2012})
\end{barticle}
\endbibitem

\bibitem[\protect\citeauthoryear{Andrews et~al.}{2012}]{andrews2012forcing}
\begin{botherref}
\oauthor{\bsnm{Andrews}, \binits{T.}},
\oauthor{\bsnm{Gregory}, \binits{J.M.}},
\oauthor{\bsnm{Webb}, \binits{M.J.}},
\oauthor{\bsnm{Taylor}, \binits{K.E.}}:
Forcing, feedbacks and climate sensitivity in cmip5 coupled atmosphere-ocean climate models.
Geophysical research letters
\textbf{39}(9)
(2012)
\end{botherref}
\endbibitem

\bibitem[\protect\citeauthoryear{Meehl et~al.}{2020}]{meehl2020context}
\begin{barticle}
\bauthor{\bsnm{Meehl}, \binits{G.A.}},
\bauthor{\bsnm{Senior}, \binits{C.A.}},
\bauthor{\bsnm{Eyring}, \binits{V.}},
\bauthor{\bsnm{Flato}, \binits{G.}},
\bauthor{\bsnm{Lamarque}, \binits{J.-F.}},
\bauthor{\bsnm{Stouffer}, \binits{R.J.}},
\bauthor{\bsnm{Taylor}, \binits{K.E.}},
\bauthor{\bsnm{Schlund}, \binits{M.}}:
\batitle{Context for interpreting equilibrium climate sensitivity and transient climate response from the cmip6 earth system models}.
\bjtitle{Science Advances}
\bvolume{6}(\bissue{26}),
\bfpage{1981}
(\byear{2020})
\end{barticle}
\endbibitem

\bibitem[\protect\citeauthoryear{Calvin et~al.}{2023}]{calvin2023ipcc}
\begin{botherref}
\oauthor{\bsnm{Calvin}, \binits{K.}},
\oauthor{\bsnm{Dasgupta}, \binits{D.}},
\oauthor{\bsnm{Krinner}, \binits{G.}},
\oauthor{\bsnm{Mukherji}, \binits{A.}},
\oauthor{\bsnm{Thorne}, \binits{P.W.}},
\oauthor{\bsnm{Trisos}, \binits{C.}},
\oauthor{\bsnm{Romero}, \binits{J.}},
\oauthor{\bsnm{Aldunce}, \binits{P.}},
\oauthor{\bsnm{Barrett}, \binits{K.}},
\oauthor{\bsnm{Blanco}, \binits{G.}}, et al.:
Ipcc, 2023: Climate change 2023: Synthesis report. contribution of working groups i, ii and iii to the sixth assessment report of the intergovernmental panel on climate change [core writing team, h. lee and j. romero (eds.)]. ipcc, geneva, switzerland.
(No Title)
(2023)
\end{botherref}
\endbibitem

\bibitem[\protect\citeauthoryear{Eyring et~al.}{2020}]{eyring2020earth}
\begin{barticle}
\bauthor{\bsnm{Eyring}, \binits{V.}},
\bauthor{\bsnm{Bock}, \binits{L.}},
\bauthor{\bsnm{Lauer}, \binits{A.}},
\bauthor{\bsnm{Righi}, \binits{M.}},
\bauthor{\bsnm{Schlund}, \binits{M.}},
\bauthor{\bsnm{Andela}, \binits{B.}},
\bauthor{\bsnm{Arnone}, \binits{E.}},
\bauthor{\bsnm{Bellprat}, \binits{O.}},
\bauthor{\bsnm{Br{\"o}tz}, \binits{B.}},
\bauthor{\bsnm{Caron}, \binits{L.-P.}}, \betal:
\batitle{Earth system model evaluation tool (esmvaltool) v2. 0--an extended set of large-scale diagnostics for quasi-operational and comprehensive evaluation of earth system models in cmip}.
\bjtitle{Geoscientific Model Development}
\bvolume{13}(\bissue{7}),
\bfpage{3383}--\blpage{3438}
(\byear{2020})
\end{barticle}
\endbibitem

\bibitem[\protect\citeauthoryear{Gettelman et~al.}{2012}]{gettelman2012community}
\begin{barticle}
\bauthor{\bsnm{Gettelman}, \binits{A.}},
\bauthor{\bsnm{Eyring}, \binits{V.}},
\bauthor{\bsnm{Fischer}, \binits{C.}},
\bauthor{\bsnm{Shiona}, \binits{H.}},
\bauthor{\bsnm{Cionni}, \binits{I.}},
\bauthor{\bsnm{Neish}, \binits{M.}},
\bauthor{\bsnm{Morgenstern}, \binits{O.}},
\bauthor{\bsnm{Wood}, \binits{S.}},
\bauthor{\bsnm{Li}, \binits{Z.}}:
\batitle{A community diagnostic tool for chemistry climate model validation}.
\bjtitle{Geoscientific Model Development}
\bvolume{5}(\bissue{5}),
\bfpage{1061}--\blpage{1073}
(\byear{2012})
\end{barticle}
\endbibitem

\bibitem[\protect\citeauthoryear{Lee et~al.}{2024}]{lee2024systematic}
\begin{barticle}
\bauthor{\bsnm{Lee}, \binits{J.}},
\bauthor{\bsnm{Gleckler}, \binits{P.J.}},
\bauthor{\bsnm{Ahn}, \binits{M.-S.}},
\bauthor{\bsnm{Ordonez}, \binits{A.}},
\bauthor{\bsnm{Ullrich}, \binits{P.A.}},
\bauthor{\bsnm{Sperber}, \binits{K.R.}},
\bauthor{\bsnm{Taylor}, \binits{K.E.}},
\bauthor{\bsnm{Planton}, \binits{Y.Y.}},
\bauthor{\bsnm{Guilyardi}, \binits{E.}},
\bauthor{\bsnm{Durack}, \binits{P.}}, \betal:
\batitle{Systematic and objective evaluation of earth system models: Pcmdi metrics package (pmp) version 3}.
\bjtitle{Geoscientific Model Development}
\bvolume{17}(\bissue{9}),
\bfpage{3919}--\blpage{3948}
(\byear{2024})
\end{barticle}
\endbibitem

\bibitem[\protect\citeauthoryear{Collier et~al.}{2018}]{collier2018international}
\begin{barticle}
\bauthor{\bsnm{Collier}, \binits{N.}},
\bauthor{\bsnm{Hoffman}, \binits{F.M.}},
\bauthor{\bsnm{Lawrence}, \binits{D.M.}},
\bauthor{\bsnm{Keppel-Aleks}, \binits{G.}},
\bauthor{\bsnm{Koven}, \binits{C.D.}},
\bauthor{\bsnm{Riley}, \binits{W.J.}},
\bauthor{\bsnm{Mu}, \binits{M.}},
\bauthor{\bsnm{Randerson}, \binits{J.T.}}:
\batitle{The international land model benchmarking (ilamb) system: design, theory, and implementation}.
\bjtitle{Journal of Advances in Modeling Earth Systems}
\bvolume{10}(\bissue{11}),
\bfpage{2731}--\blpage{2754}
(\byear{2018})
\end{barticle}
\endbibitem

\bibitem[\protect\citeauthoryear{Wang et~al.}{2025}]{wang2025history}
\begin{barticle}
\bauthor{\bsnm{Wang}, \binits{Z.}},
\bauthor{\bsnm{Chu}, \binits{Z.}},
\bauthor{\bsnm{Doan}, \binits{T.V.}},
\bauthor{\bsnm{Ni}, \binits{S.}},
\bauthor{\bsnm{Yang}, \binits{M.}},
\bauthor{\bsnm{Zhang}, \binits{W.}}:
\batitle{History, development, and principles of large language models: an introductory survey}.
\bjtitle{AI and Ethics}
\bvolume{5}(\bissue{3}),
\bfpage{1955}--\blpage{1971}
(\byear{2025})
\end{barticle}
\endbibitem

\bibitem[\protect\citeauthoryear{Zhang et~al.}{2025}]{zhang2025scientific}
\begin{barticle}
\bauthor{\bsnm{Zhang}, \binits{Q.}},
\bauthor{\bsnm{Ding}, \binits{K.}},
\bauthor{\bsnm{Lv}, \binits{T.}},
\bauthor{\bsnm{Wang}, \binits{X.}},
\bauthor{\bsnm{Yin}, \binits{Q.}},
\bauthor{\bsnm{Zhang}, \binits{Y.}},
\bauthor{\bsnm{Yu}, \binits{J.}},
\bauthor{\bsnm{Wang}, \binits{Y.}},
\bauthor{\bsnm{Li}, \binits{X.}},
\bauthor{\bsnm{Xiang}, \binits{Z.}}, \betal:
\batitle{Scientific large language models: A survey on biological \& chemical domains}.
\bjtitle{ACM Computing Surveys}
\bvolume{57}(\bissue{6}),
\bfpage{1}--\blpage{38}
(\byear{2025})
\end{barticle}
\endbibitem

\bibitem[\protect\citeauthoryear{Wang et~al.}{2024}]{wang2024tools}
\begin{botherref}
\oauthor{\bsnm{Wang}, \binits{Z.}},
\oauthor{\bsnm{Cheng}, \binits{Z.}},
\oauthor{\bsnm{Zhu}, \binits{H.}},
\oauthor{\bsnm{Fried}, \binits{D.}},
\oauthor{\bsnm{Neubig}, \binits{G.}}:
What are tools anyway? a survey from the language model perspective.
arXiv preprint arXiv:2403.15452
(2024)
\end{botherref}
\endbibitem

\bibitem[\protect\citeauthoryear{Fan et~al.}{2024}]{fan2024survey}
\begin{bchapter}
\bauthor{\bsnm{Fan}, \binits{W.}},
\bauthor{\bsnm{Ding}, \binits{Y.}},
\bauthor{\bsnm{Ning}, \binits{L.}},
\bauthor{\bsnm{Wang}, \binits{S.}},
\bauthor{\bsnm{Li}, \binits{H.}},
\bauthor{\bsnm{Yin}, \binits{D.}},
\bauthor{\bsnm{Chua}, \binits{T.-S.}},
\bauthor{\bsnm{Li}, \binits{Q.}}:
\bctitle{A survey on rag meeting llms: Towards retrieval-augmented large language models}.
In: \bbtitle{Proceedings of the 30th ACM SIGKDD Conference on Knowledge Discovery and Data Mining},
pp. \bfpage{6491}--\blpage{6501}
(\byear{2024})
\end{bchapter}
\endbibitem

\bibitem[\protect\citeauthoryear{Huang et~al.}{2025}]{huang2025biomni}
\begin{botherref}
\oauthor{\bsnm{Huang}, \binits{K.}},
\oauthor{\bsnm{Zhang}, \binits{S.}},
\oauthor{\bsnm{Wang}, \binits{H.}},
\oauthor{\bsnm{Qu}, \binits{Y.}},
\oauthor{\bsnm{Lu}, \binits{Y.}},
\oauthor{\bsnm{Roohani}, \binits{Y.}},
\oauthor{\bsnm{Li}, \binits{R.}},
\oauthor{\bsnm{Qiu}, \binits{L.}},
\oauthor{\bsnm{Zhang}, \binits{J.}},
\oauthor{\bsnm{Di}, \binits{Y.}}, et al.:
Biomni: A general-purpose biomedical ai agent.
bioRxiv,
2025--05
(2025)
\end{botherref}
\endbibitem

\bibitem[\protect\citeauthoryear{Boiko et~al.}{2023}]{boiko2023autonomous}
\begin{barticle}
\bauthor{\bsnm{Boiko}, \binits{D.A.}},
\bauthor{\bsnm{MacKnight}, \binits{R.}},
\bauthor{\bsnm{Kline}, \binits{B.}},
\bauthor{\bsnm{Gomes}, \binits{G.}}:
\batitle{Autonomous chemical research with large language models}.
\bjtitle{Nature}
\bvolume{624}(\bissue{7992}),
\bfpage{570}--\blpage{578}
(\byear{2023})
\end{barticle}
\endbibitem

\bibitem[\protect\citeauthoryear{Kang and Kim}{2024}]{kang2024chatmof}
\begin{barticle}
\bauthor{\bsnm{Kang}, \binits{Y.}},
\bauthor{\bsnm{Kim}, \binits{J.}}:
\batitle{Chatmof: an artificial intelligence system for predicting and generating metal-organic frameworks using large language models}.
\bjtitle{Nature communications}
\bvolume{15}(\bissue{1}),
\bfpage{4705}
(\byear{2024})
\end{barticle}
\endbibitem

\bibitem[\protect\citeauthoryear{Bi et~al.}{2023}]{bi2023oceangpt}
\begin{botherref}
\oauthor{\bsnm{Bi}, \binits{Z.}},
\oauthor{\bsnm{Zhang}, \binits{N.}},
\oauthor{\bsnm{Xue}, \binits{Y.}},
\oauthor{\bsnm{Ou}, \binits{Y.}},
\oauthor{\bsnm{Ji}, \binits{D.}},
\oauthor{\bsnm{Zheng}, \binits{G.}},
\oauthor{\bsnm{Chen}, \binits{H.}}:
Oceangpt: A large language model for ocean science tasks.
arXiv preprint arXiv:2310.02031
(2023)
\end{botherref}
\endbibitem

\bibitem[\protect\citeauthoryear{Deng et~al.}{2024}]{deng2024k2}
\begin{bchapter}
\bauthor{\bsnm{Deng}, \binits{C.}},
\bauthor{\bsnm{Zhang}, \binits{T.}},
\bauthor{\bsnm{He}, \binits{Z.}},
\bauthor{\bsnm{Chen}, \binits{Q.}},
\bauthor{\bsnm{Shi}, \binits{Y.}},
\bauthor{\bsnm{Xu}, \binits{Y.}},
\bauthor{\bsnm{Fu}, \binits{L.}},
\bauthor{\bsnm{Zhang}, \binits{W.}},
\bauthor{\bsnm{Wang}, \binits{X.}},
\bauthor{\bsnm{Zhou}, \binits{C.}}, \betal:
\bctitle{K2: A foundation language model for geoscience knowledge understanding and utilization}.
In: \bbtitle{Proceedings of the 17th ACM International Conference on Web Search and Data Mining},
pp. \bfpage{161}--\blpage{170}
(\byear{2024})
\end{bchapter}
\endbibitem

\bibitem[\protect\citeauthoryear{Zhang et~al.}{2024}]{zhang2024geogpt}
\begin{barticle}
\bauthor{\bsnm{Zhang}, \binits{Y.}},
\bauthor{\bsnm{Wei}, \binits{C.}},
\bauthor{\bsnm{He}, \binits{Z.}},
\bauthor{\bsnm{Yu}, \binits{W.}}:
\batitle{Geogpt: An assistant for understanding and processing geospatial tasks}.
\bjtitle{International Journal of Applied Earth Observation and Geoinformation}
\bvolume{131},
\bfpage{103976}
(\byear{2024})
\end{barticle}
\endbibitem

\bibitem[\protect\citeauthoryear{Ashok et~al.}{2007}]{ashok2007nino}
\begin{botherref}
\oauthor{\bsnm{Ashok}, \binits{K.}},
\oauthor{\bsnm{Behera}, \binits{S.K.}},
\oauthor{\bsnm{Rao}, \binits{S.A.}},
\oauthor{\bsnm{Weng}, \binits{H.}},
\oauthor{\bsnm{Yamagata}, \binits{T.}}:
El ni{\~n}o modoki and its possible teleconnection.
Journal of Geophysical Research: Oceans
\textbf{112}(C11)
(2007)
\end{botherref}
\endbibitem

\bibitem[\protect\citeauthoryear{Kao and Yu}{2009}]{kao2009contrasting}
\begin{barticle}
\bauthor{\bsnm{Kao}, \binits{H.-Y.}},
\bauthor{\bsnm{Yu}, \binits{J.-Y.}}:
\batitle{Contrasting eastern-pacific and central-pacific types of enso}.
\bjtitle{Journal of Climate}
\bvolume{22}(\bissue{3}),
\bfpage{615}--\blpage{632}
(\byear{2009})
\end{barticle}
\endbibitem

\bibitem[\protect\citeauthoryear{Kug et~al.}{2009}]{kug2009two}
\begin{barticle}
\bauthor{\bsnm{Kug}, \binits{J.-S.}},
\bauthor{\bsnm{Jin}, \binits{F.-F.}},
\bauthor{\bsnm{An}, \binits{S.-I.}}:
\batitle{Two types of el ni{\~n}o events: cold tongue el ni{\~n}o and warm pool el ni{\~n}o}.
\bjtitle{Journal of climate}
\bvolume{22}(\bissue{6}),
\bfpage{1499}--\blpage{1515}
(\byear{2009})
\end{barticle}
\endbibitem

\bibitem[\protect\citeauthoryear{Gillett et~al.}{2016}]{gillett2016detection}
\begin{barticle}
\bauthor{\bsnm{Gillett}, \binits{N.P.}},
\bauthor{\bsnm{Shiogama}, \binits{H.}},
\bauthor{\bsnm{Funke}, \binits{B.}},
\bauthor{\bsnm{Hegerl}, \binits{G.}},
\bauthor{\bsnm{Knutti}, \binits{R.}},
\bauthor{\bsnm{Matthes}, \binits{K.}},
\bauthor{\bsnm{Santer}, \binits{B.D.}},
\bauthor{\bsnm{Stone}, \binits{D.}},
\bauthor{\bsnm{Tebaldi}, \binits{C.}}:
\batitle{The detection and attribution model intercomparison project (damip v1. 0) contribution to cmip6}.
\bjtitle{Geoscientific Model Development}
\bvolume{9}(\bissue{10}),
\bfpage{3685}--\blpage{3697}
(\byear{2016})
\end{barticle}
\endbibitem

\bibitem[\protect\citeauthoryear{O'Neill et~al.}{2016}]{o2016scenario}
\begin{barticle}
\bauthor{\bsnm{O'Neill}, \binits{B.C.}},
\bauthor{\bsnm{Tebaldi}, \binits{C.}},
\bauthor{\bsnm{Van~Vuuren}, \binits{D.P.}},
\bauthor{\bsnm{Eyring}, \binits{V.}},
\bauthor{\bsnm{Friedlingstein}, \binits{P.}},
\bauthor{\bsnm{Hurtt}, \binits{G.}},
\bauthor{\bsnm{Knutti}, \binits{R.}},
\bauthor{\bsnm{Kriegler}, \binits{E.}},
\bauthor{\bsnm{Lamarque}, \binits{J.-F.}},
\bauthor{\bsnm{Lowe}, \binits{J.}}, \betal:
\batitle{The scenario model intercomparison project (scenariomip) for cmip6}.
\bjtitle{Geoscientific Model Development}
\bvolume{9}(\bissue{9}),
\bfpage{3461}--\blpage{3482}
(\byear{2016})
\end{barticle}
\endbibitem

\bibitem[\protect\citeauthoryear{Li et~al.}{2024}]{li2024constraining}
\begin{barticle}
\bauthor{\bsnm{Li}, \binits{C.}},
\bauthor{\bsnm{Sun}, \binits{Q.}},
\bauthor{\bsnm{Wang}, \binits{J.}},
\bauthor{\bsnm{Liang}, \binits{Y.}},
\bauthor{\bsnm{Zwiers}, \binits{F.W.}},
\bauthor{\bsnm{Zhang}, \binits{X.}},
\bauthor{\bsnm{Li}, \binits{T.}}:
\batitle{Constraining projected changes in rare intense precipitation events across global land regions}.
\bjtitle{Geophysical Research Letters}
\bvolume{51}(\bissue{3}),
\bfpage{2023}--\blpage{105605}
(\byear{2024})
\end{barticle}
\endbibitem

\bibitem[\protect\citeauthoryear{Li et~al.}{2025}]{li2025constraining}
\begin{barticle}
\bauthor{\bsnm{Li}, \binits{C.}},
\bauthor{\bsnm{Zwiers}, \binits{F.W.}},
\bauthor{\bsnm{Zhang}, \binits{X.}},
\bauthor{\bsnm{Fischer}, \binits{E.M.}},
\bauthor{\bsnm{Du}, \binits{F.}},
\bauthor{\bsnm{Liu}, \binits{J.}},
\bauthor{\bsnm{Wang}, \binits{J.}},
\bauthor{\bsnm{Liang}, \binits{Y.}},
\bauthor{\bsnm{Li}, \binits{T.}},
\bauthor{\bsnm{Yuan}, \binits{L.}}:
\batitle{Constraining the entire earth system projections for more reliable climate change adaptation planning}.
\bjtitle{Science Advances}
\bvolume{11}(\bissue{9}),
\bfpage{5346}
(\byear{2025})
\end{barticle}
\endbibitem

\bibitem[\protect\citeauthoryear{Bach{\`e}lery et~al.}{2025}]{bachelery2025predicting}
\begin{barticle}
\bauthor{\bsnm{Bach{\`e}lery}, \binits{M.-L.}},
\bauthor{\bsnm{Brajard}, \binits{J.}},
\bauthor{\bsnm{Patacchiola}, \binits{M.}},
\bauthor{\bsnm{Illig}, \binits{S.}},
\bauthor{\bsnm{Keenlyside}, \binits{N.}}:
\batitle{Predicting atlantic and benguela ni{\~n}o events with deep learning}.
\bjtitle{Science Advances}
\bvolume{11}(\bissue{14}),
\bfpage{5185}
(\byear{2025})
\end{barticle}
\endbibitem

\bibitem[\protect\citeauthoryear{Hoffman et~al.}{2025}]{hoffman2025rapid}
\begin{botherref}
\oauthor{\bsnm{Hoffman}, \binits{F.M.}},
\oauthor{\bsnm{Hassler}, \binits{B.}},
\oauthor{\bsnm{Swaminathan}, \binits{R.}},
\oauthor{\bsnm{Lewis}, \binits{J.}},
\oauthor{\bsnm{Andela}, \binits{B.}},
\oauthor{\bsnm{Collier}, \binits{N.}},
\oauthor{\bsnm{Heged{\H{u}}s}, \binits{D.}},
\oauthor{\bsnm{Lee}, \binits{J.}},
\oauthor{\bsnm{Pascoe}, \binits{C.}},
\oauthor{\bsnm{Pfl{\"u}ger}, \binits{M.}}, et al.:
Rapid evaluation framework for the cmip7 assessment fast track.
EGUsphere
(2025)
\end{botherref}
\endbibitem

\bibitem[\protect\citeauthoryear{Kravitz et~al.}{2011}]{kravitz2011geoengineering}
\begin{barticle}
\bauthor{\bsnm{Kravitz}, \binits{B.}},
\bauthor{\bsnm{Robock}, \binits{A.}},
\bauthor{\bsnm{Boucher}, \binits{O.}},
\bauthor{\bsnm{Schmidt}, \binits{H.}},
\bauthor{\bsnm{Taylor}, \binits{K.E.}},
\bauthor{\bsnm{Stenchikov}, \binits{G.}},
\bauthor{\bsnm{Schulz}, \binits{M.}}:
\batitle{The geoengineering model intercomparison project (geomip)}.
\bjtitle{Atmospheric Science Letters}
\bvolume{12}(\bissue{2}),
\bfpage{162}--\blpage{167}
(\byear{2011})
\end{barticle}
\endbibitem

\bibitem[\protect\citeauthoryear{Waliser et~al.}{2020}]{waliser2020observations}
\begin{barticle}
\bauthor{\bsnm{Waliser}, \binits{D.}},
\bauthor{\bsnm{Gleckler}, \binits{P.J.}},
\bauthor{\bsnm{Ferraro}, \binits{R.}},
\bauthor{\bsnm{Taylor}, \binits{K.E.}},
\bauthor{\bsnm{Ames}, \binits{S.}},
\bauthor{\bsnm{Biard}, \binits{J.}},
\bauthor{\bsnm{Bosilovich}, \binits{M.G.}},
\bauthor{\bsnm{Brown}, \binits{O.}},
\bauthor{\bsnm{Chepfer}, \binits{H.}},
\bauthor{\bsnm{Cinquini}, \binits{L.}}, \betal:
\batitle{Observations for model intercomparison project (obs4mips): Status for cmip6}.
\bjtitle{Geoscientific Model Development}
\bvolume{13}(\bissue{7}),
\bfpage{2945}--\blpage{2958}
(\byear{2020})
\end{barticle}
\endbibitem

\bibitem[\protect\citeauthoryear{Rayner et~al.}{2003}]{rayner2003global}
\begin{botherref}
\oauthor{\bsnm{Rayner}, \binits{N.A.}},
\oauthor{\bsnm{Parker}, \binits{D.E.}},
\oauthor{\bsnm{Horton}, \binits{E.}},
\oauthor{\bsnm{Folland}, \binits{C.K.}},
\oauthor{\bsnm{Alexander}, \binits{L.V.}},
\oauthor{\bsnm{Rowell}, \binits{D.}},
\oauthor{\bsnm{Kent}, \binits{E.C.}},
\oauthor{\bsnm{Kaplan}, \binits{A.}}:
Global analyses of sea surface temperature, sea ice, and night marine air temperature since the late nineteenth century.
Journal of Geophysical Research: Atmospheres
\textbf{108}(D14)
(2003)
\end{botherref}
\endbibitem

\bibitem[\protect\citeauthoryear{Morice et~al.}{2021}]{morice2021updated}
\begin{barticle}
\bauthor{\bsnm{Morice}, \binits{C.P.}},
\bauthor{\bsnm{Kennedy}, \binits{J.J.}},
\bauthor{\bsnm{Rayner}, \binits{N.A.}},
\bauthor{\bsnm{Winn}, \binits{J.P.}},
\bauthor{\bsnm{Hogan}, \binits{E.}},
\bauthor{\bsnm{Killick}, \binits{R.E.}},
\bauthor{\bsnm{Dunn}, \binits{R.J.}},
\bauthor{\bsnm{Osborn}, \binits{T.J.}},
\bauthor{\bsnm{Jones}, \binits{P.D.}},
\bauthor{\bsnm{Simpson}, \binits{I.R.}}:
\batitle{An updated assessment of near-surface temperature change from 1850: The hadcrut5 data set}.
\bjtitle{Journal of Geophysical Research: Atmospheres}
\bvolume{126}(\bissue{3}),
\bfpage{2019}--\blpage{032361}
(\byear{2021})
\end{barticle}
\endbibitem

\bibitem[\protect\citeauthoryear{Adler et~al.}{2003}]{adler2003version}
\begin{barticle}
\bauthor{\bsnm{Adler}, \binits{R.F.}},
\bauthor{\bsnm{Huffman}, \binits{G.J.}},
\bauthor{\bsnm{Chang}, \binits{A.}},
\bauthor{\bsnm{Ferraro}, \binits{R.}},
\bauthor{\bsnm{Xie}, \binits{P.-P.}},
\bauthor{\bsnm{Janowiak}, \binits{J.}},
\bauthor{\bsnm{Rudolf}, \binits{B.}},
\bauthor{\bsnm{Schneider}, \binits{U.}},
\bauthor{\bsnm{Curtis}, \binits{S.}},
\bauthor{\bsnm{Bolvin}, \binits{D.}}, \betal:
\batitle{The version-2 global precipitation climatology project (gpcp) monthly precipitation analysis (1979--present)}.
\bjtitle{Journal of hydrometeorology}
\bvolume{4}(\bissue{6}),
\bfpage{1147}--\blpage{1167}
(\byear{2003})
\end{barticle}
\endbibitem

\bibitem[\protect\citeauthoryear{Hersbach et~al.}{2019}]{hersbach2019era5}
\begin{barticle}
\bauthor{\bsnm{Hersbach}, \binits{H.}},
\bauthor{\bsnm{Bell}, \binits{B.}},
\bauthor{\bsnm{Berrisford}, \binits{P.}},
\bauthor{\bsnm{Biavati}, \binits{G.}},
\bauthor{\bsnm{Hor{\'a}nyi}, \binits{A.}},
\bauthor{\bsnm{Mu{\~n}oz~Sabater}, \binits{J.}},
\bauthor{\bsnm{Nicolas}, \binits{J.}},
\bauthor{\bsnm{Peubey}, \binits{C.}},
\bauthor{\bsnm{Radu}, \binits{R.}},
\bauthor{\bsnm{Rozum}, \binits{I.}}, \betal:
\batitle{Era5 monthly averaged data on single levels from 1979 to present}.
\bjtitle{Copernicus Climate Change Service (C3S) Climate Data Store (CDS)}
\bvolume{10},
\bfpage{252}--\blpage{266}
(\byear{2019})
\end{barticle}
\endbibitem

\bibitem[\protect\citeauthoryear{Righi et~al.}{2020}]{righi2020earth}
\begin{barticle}
\bauthor{\bsnm{Righi}, \binits{M.}},
\bauthor{\bsnm{Andela}, \binits{B.}},
\bauthor{\bsnm{Eyring}, \binits{V.}},
\bauthor{\bsnm{Lauer}, \binits{A.}},
\bauthor{\bsnm{Predoi}, \binits{V.}},
\bauthor{\bsnm{Schlund}, \binits{M.}},
\bauthor{\bsnm{Vegas-Regidor}, \binits{J.}},
\bauthor{\bsnm{Bock}, \binits{L.}},
\bauthor{\bsnm{Br{\"o}tz}, \binits{B.}},
\bauthor{\bsnm{Mora}, \binits{L.}}, \betal:
\batitle{Earth system model evaluation tool (esmvaltool) v2. 0--technical overview}.
\bjtitle{Geoscientific Model Development}
\bvolume{13}(\bissue{3}),
\bfpage{1179}--\blpage{1199}
(\byear{2020})
\end{barticle}
\endbibitem

\bibitem[\protect\citeauthoryear{Kochkov et~al.}{2024}]{kochkov2024neural}
\begin{barticle}
\bauthor{\bsnm{Kochkov}, \binits{D.}},
\bauthor{\bsnm{Yuval}, \binits{J.}},
\bauthor{\bsnm{Langmore}, \binits{I.}},
\bauthor{\bsnm{Norgaard}, \binits{P.}},
\bauthor{\bsnm{Smith}, \binits{J.}},
\bauthor{\bsnm{Mooers}, \binits{G.}},
\bauthor{\bsnm{Kl{\"o}wer}, \binits{M.}},
\bauthor{\bsnm{Lottes}, \binits{J.}},
\bauthor{\bsnm{Rasp}, \binits{S.}},
\bauthor{\bsnm{D{\"u}ben}, \binits{P.}}, \betal:
\batitle{Neural general circulation models for weather and climate}.
\bjtitle{Nature}
\bvolume{632}(\bissue{8027}),
\bfpage{1060}--\blpage{1066}
(\byear{2024})
\end{barticle}
\endbibitem

\bibitem[\protect\citeauthoryear{Guo et~al.}{2025}]{guo2025data}
\begin{barticle}
\bauthor{\bsnm{Guo}, \binits{Z.}},
\bauthor{\bsnm{Lyu}, \binits{P.}},
\bauthor{\bsnm{Ling}, \binits{F.}},
\bauthor{\bsnm{Bai}, \binits{L.}},
\bauthor{\bsnm{Luo}, \binits{J.-J.}},
\bauthor{\bsnm{Boers}, \binits{N.}},
\bauthor{\bsnm{Yamagata}, \binits{T.}},
\bauthor{\bsnm{Izumo}, \binits{T.}},
\bauthor{\bsnm{Cravatte}, \binits{S.}},
\bauthor{\bsnm{Capotondi}, \binits{A.}}, \betal:
\batitle{Data-driven global ocean modeling for seasonal to decadal prediction}.
\bjtitle{Science Advances}
\bvolume{11}(\bissue{33}),
\bfpage{2488}
(\byear{2025})
\end{barticle}
\endbibitem

\bibitem[\protect\citeauthoryear{Gregory et~al.}{2004}]{gregory2004new}
\begin{botherref}
\oauthor{\bsnm{Gregory}, \binits{J.M.}},
\oauthor{\bsnm{Ingram}, \binits{W.J.}},
\oauthor{\bsnm{Palmer}, \binits{M.}},
\oauthor{\bsnm{Jones}, \binits{G.S.}},
\oauthor{\bsnm{Stott}, \binits{P.}},
\oauthor{\bsnm{Thorpe}, \binits{R.}},
\oauthor{\bsnm{Lowe}, \binits{J.A.}},
\oauthor{\bsnm{Johns}, \binits{T.}},
\oauthor{\bsnm{Williams}, \binits{K.}}:
A new method for diagnosing radiative forcing and climate sensitivity.
Geophysical research letters
\textbf{31}(3)
(2004)
\end{botherref}
\endbibitem

\end{thebibliography}



\begin{thebibliography}{38}
\ifx \bisbn   \undefined \def \bisbn  #1{ISBN #1}\fi
\ifx \binits  \undefined \def \binits#1{#1}\fi
\ifx \bauthor  \undefined \def \bauthor#1{#1}\fi
\ifx \batitle  \undefined \def \batitle#1{#1}\fi
\ifx \bjtitle  \undefined \def \bjtitle#1{#1}\fi
\ifx \bvolume  \undefined \def \bvolume#1{\textbf{#1}}\fi
\ifx \byear  \undefined \def \byear#1{#1}\fi
\ifx \bissue  \undefined \def \bissue#1{#1}\fi
\ifx \bfpage  \undefined \def \bfpage#1{#1}\fi
\ifx \blpage  \undefined \def \blpage #1{#1}\fi
\ifx \burl  \undefined \def \burl#1{\textsf{#1}}\fi
\ifx \doiurl  \undefined \def \doiurl#1{\url{https://doi.org/#1}}\fi
\ifx \betal  \undefined \def \betal{\textit{et al.}}\fi
\ifx \binstitute  \undefined \def \binstitute#1{#1}\fi
\ifx \binstitutionaled  \undefined \def \binstitutionaled#1{#1}\fi
\ifx \bctitle  \undefined \def \bctitle#1{#1}\fi
\ifx \beditor  \undefined \def \beditor#1{#1}\fi
\ifx \bpublisher  \undefined \def \bpublisher#1{#1}\fi
\ifx \bbtitle  \undefined \def \bbtitle#1{#1}\fi
\ifx \bedition  \undefined \def \bedition#1{#1}\fi
\ifx \bseriesno  \undefined \def \bseriesno#1{#1}\fi
\ifx \blocation  \undefined \def \blocation#1{#1}\fi
\ifx \bsertitle  \undefined \def \bsertitle#1{#1}\fi
\ifx \bsnm \undefined \def \bsnm#1{#1}\fi
\ifx \bsuffix \undefined \def \bsuffix#1{#1}\fi
\ifx \bparticle \undefined \def \bparticle#1{#1}\fi
\ifx \barticle \undefined \def \barticle#1{#1}\fi
\bibcommenthead
\ifx \bconfdate \undefined \def \bconfdate #1{#1}\fi
\ifx \botherref \undefined \def \botherref #1{#1}\fi
\ifx \url \undefined \def \url#1{\textsf{#1}}\fi
\ifx \bchapter \undefined \def \bchapter#1{#1}\fi
\ifx \bbook \undefined \def \bbook#1{#1}\fi
\ifx \bcomment \undefined \def \bcomment#1{#1}\fi
\ifx \oauthor \undefined \def \oauthor#1{#1}\fi
\ifx \citeauthoryear \undefined \def \citeauthoryear#1{#1}\fi
\ifx \endbibitem  \undefined \def \endbibitem {}\fi
\ifx \bconflocation  \undefined \def \bconflocation#1{#1}\fi
\ifx \arxivurl  \undefined \def \arxivurl#1{\textsf{#1}}\fi
\csname PreBibitemsHook\endcsname

\bibitem[\protect\citeauthoryear{Zhang et~al.}{2025}]{zhang2025qwen3}
\begin{botherref}
\oauthor{\bsnm{Zhang}, \binits{Y.}},
\oauthor{\bsnm{Li}, \binits{M.}},
\oauthor{\bsnm{Long}, \binits{D.}},
\oauthor{\bsnm{Zhang}, \binits{X.}},
\oauthor{\bsnm{Lin}, \binits{H.}},
\oauthor{\bsnm{Yang}, \binits{B.}},
\oauthor{\bsnm{Xie}, \binits{P.}},
\oauthor{\bsnm{Yang}, \binits{A.}},
\oauthor{\bsnm{Liu}, \binits{D.}},
\oauthor{\bsnm{Lin}, \binits{J.}}, et al.:
Qwen3 embedding: Advancing text embedding and reranking through foundation models.
arXiv preprint arXiv:2506.05176
(2025)
\end{botherref}
\endbibitem

\bibitem[\protect\citeauthoryear{Gillett et~al.}{2016}]{gillett2016detection}
\begin{barticle}
\bauthor{\bsnm{Gillett}, \binits{N.P.}},
\bauthor{\bsnm{Shiogama}, \binits{H.}},
\bauthor{\bsnm{Funke}, \binits{B.}},
\bauthor{\bsnm{Hegerl}, \binits{G.}},
\bauthor{\bsnm{Knutti}, \binits{R.}},
\bauthor{\bsnm{Matthes}, \binits{K.}},
\bauthor{\bsnm{Santer}, \binits{B.D.}},
\bauthor{\bsnm{Stone}, \binits{D.}},
\bauthor{\bsnm{Tebaldi}, \binits{C.}}:
\batitle{The detection and attribution model intercomparison project (damip v1. 0) contribution to cmip6}.
\bjtitle{Geoscientific Model Development}
\bvolume{9}(\bissue{10}),
\bfpage{3685}--\blpage{3697}
(\byear{2016})
\end{barticle}
\endbibitem

\bibitem[\protect\citeauthoryear{Kravitz et~al.}{2011}]{kravitz2011geoengineering}
\begin{barticle}
\bauthor{\bsnm{Kravitz}, \binits{B.}},
\bauthor{\bsnm{Robock}, \binits{A.}},
\bauthor{\bsnm{Boucher}, \binits{O.}},
\bauthor{\bsnm{Schmidt}, \binits{H.}},
\bauthor{\bsnm{Taylor}, \binits{K.E.}},
\bauthor{\bsnm{Stenchikov}, \binits{G.}},
\bauthor{\bsnm{Schulz}, \binits{M.}}:
\batitle{The geoengineering model intercomparison project (geomip)}.
\bjtitle{Atmospheric Science Letters}
\bvolume{12}(\bissue{2}),
\bfpage{162}--\blpage{167}
(\byear{2011})
\end{barticle}
\endbibitem

\bibitem[\protect\citeauthoryear{O'Neill et~al.}{2016}]{o2016scenario}
\begin{barticle}
\bauthor{\bsnm{O'Neill}, \binits{B.C.}},
\bauthor{\bsnm{Tebaldi}, \binits{C.}},
\bauthor{\bsnm{Van~Vuuren}, \binits{D.P.}},
\bauthor{\bsnm{Eyring}, \binits{V.}},
\bauthor{\bsnm{Friedlingstein}, \binits{P.}},
\bauthor{\bsnm{Hurtt}, \binits{G.}},
\bauthor{\bsnm{Knutti}, \binits{R.}},
\bauthor{\bsnm{Kriegler}, \binits{E.}},
\bauthor{\bsnm{Lamarque}, \binits{J.-F.}},
\bauthor{\bsnm{Lowe}, \binits{J.}}, \betal:
\batitle{The scenario model intercomparison project (scenariomip) for cmip6}.
\bjtitle{Geoscientific Model Development}
\bvolume{9}(\bissue{9}),
\bfpage{3461}--\blpage{3482}
(\byear{2016})
\end{barticle}
\endbibitem

\bibitem[\protect\citeauthoryear{Waliser et~al.}{2020}]{waliser2020observations}
\begin{barticle}
\bauthor{\bsnm{Waliser}, \binits{D.}},
\bauthor{\bsnm{Gleckler}, \binits{P.J.}},
\bauthor{\bsnm{Ferraro}, \binits{R.}},
\bauthor{\bsnm{Taylor}, \binits{K.E.}},
\bauthor{\bsnm{Ames}, \binits{S.}},
\bauthor{\bsnm{Biard}, \binits{J.}},
\bauthor{\bsnm{Bosilovich}, \binits{M.G.}},
\bauthor{\bsnm{Brown}, \binits{O.}},
\bauthor{\bsnm{Chepfer}, \binits{H.}},
\bauthor{\bsnm{Cinquini}, \binits{L.}}, \betal:
\batitle{Observations for model intercomparison project (obs4mips): Status for cmip6}.
\bjtitle{Geoscientific Model Development}
\bvolume{13}(\bissue{7}),
\bfpage{2945}--\blpage{2958}
(\byear{2020})
\end{barticle}
\endbibitem

\bibitem[\protect\citeauthoryear{Rayner et~al.}{2003}]{rayner2003global}
\begin{botherref}
\oauthor{\bsnm{Rayner}, \binits{N.A.}},
\oauthor{\bsnm{Parker}, \binits{D.E.}},
\oauthor{\bsnm{Horton}, \binits{E.}},
\oauthor{\bsnm{Folland}, \binits{C.K.}},
\oauthor{\bsnm{Alexander}, \binits{L.V.}},
\oauthor{\bsnm{Rowell}, \binits{D.}},
\oauthor{\bsnm{Kent}, \binits{E.C.}},
\oauthor{\bsnm{Kaplan}, \binits{A.}}:
Global analyses of sea surface temperature, sea ice, and night marine air temperature since the late nineteenth century.
Journal of Geophysical Research: Atmospheres
\textbf{108}(D14)
(2003)
\end{botherref}
\endbibitem

\bibitem[\protect\citeauthoryear{Morice et~al.}{2021}]{morice2021updated}
\begin{barticle}
\bauthor{\bsnm{Morice}, \binits{C.P.}},
\bauthor{\bsnm{Kennedy}, \binits{J.J.}},
\bauthor{\bsnm{Rayner}, \binits{N.A.}},
\bauthor{\bsnm{Winn}, \binits{J.P.}},
\bauthor{\bsnm{Hogan}, \binits{E.}},
\bauthor{\bsnm{Killick}, \binits{R.E.}},
\bauthor{\bsnm{Dunn}, \binits{R.J.}},
\bauthor{\bsnm{Osborn}, \binits{T.J.}},
\bauthor{\bsnm{Jones}, \binits{P.D.}},
\bauthor{\bsnm{Simpson}, \binits{I.R.}}:
\batitle{An updated assessment of near-surface temperature change from 1850: The hadcrut5 data set}.
\bjtitle{Journal of Geophysical Research: Atmospheres}
\bvolume{126}(\bissue{3}),
\bfpage{2019}--\blpage{032361}
(\byear{2021})
\end{barticle}
\endbibitem

\bibitem[\protect\citeauthoryear{Adler et~al.}{2003}]{adler2003version}
\begin{barticle}
\bauthor{\bsnm{Adler}, \binits{R.F.}},
\bauthor{\bsnm{Huffman}, \binits{G.J.}},
\bauthor{\bsnm{Chang}, \binits{A.}},
\bauthor{\bsnm{Ferraro}, \binits{R.}},
\bauthor{\bsnm{Xie}, \binits{P.-P.}},
\bauthor{\bsnm{Janowiak}, \binits{J.}},
\bauthor{\bsnm{Rudolf}, \binits{B.}},
\bauthor{\bsnm{Schneider}, \binits{U.}},
\bauthor{\bsnm{Curtis}, \binits{S.}},
\bauthor{\bsnm{Bolvin}, \binits{D.}}, \betal:
\batitle{The version-2 global precipitation climatology project (gpcp) monthly precipitation analysis (1979--present)}.
\bjtitle{Journal of hydrometeorology}
\bvolume{4}(\bissue{6}),
\bfpage{1147}--\blpage{1167}
(\byear{2003})
\end{barticle}
\endbibitem

\bibitem[\protect\citeauthoryear{Hersbach et~al.}{2019}]{hersbach2019era5}
\begin{barticle}
\bauthor{\bsnm{Hersbach}, \binits{H.}},
\bauthor{\bsnm{Bell}, \binits{B.}},
\bauthor{\bsnm{Berrisford}, \binits{P.}},
\bauthor{\bsnm{Biavati}, \binits{G.}},
\bauthor{\bsnm{Hor{\'a}nyi}, \binits{A.}},
\bauthor{\bsnm{Mu{\~n}oz~Sabater}, \binits{J.}},
\bauthor{\bsnm{Nicolas}, \binits{J.}},
\bauthor{\bsnm{Peubey}, \binits{C.}},
\bauthor{\bsnm{Radu}, \binits{R.}},
\bauthor{\bsnm{Rozum}, \binits{I.}}, \betal:
\batitle{Era5 monthly averaged data on single levels from 1979 to present}.
\bjtitle{Copernicus Climate Change Service (C3S) Climate Data Store (CDS)}
\bvolume{10},
\bfpage{252}--\blpage{266}
(\byear{2019})
\end{barticle}
\endbibitem

\bibitem[\protect\citeauthoryear{Loeb et~al.}{2018}]{loeb2018clouds}
\begin{barticle}
\bauthor{\bsnm{Loeb}, \binits{N.G.}},
\bauthor{\bsnm{Doelling}, \binits{D.R.}},
\bauthor{\bsnm{Wang}, \binits{H.}},
\bauthor{\bsnm{Su}, \binits{W.}},
\bauthor{\bsnm{Nguyen}, \binits{C.}},
\bauthor{\bsnm{Corbett}, \binits{J.G.}},
\bauthor{\bsnm{Liang}, \binits{L.}},
\bauthor{\bsnm{Mitrescu}, \binits{C.}},
\bauthor{\bsnm{Rose}, \binits{F.G.}},
\bauthor{\bsnm{Kato}, \binits{S.}}:
\batitle{Clouds and the earth’s radiant energy system (ceres) energy balanced and filled (ebaf) top-of-atmosphere (toa) edition-4.0 data product}.
\bjtitle{Journal of climate}
\bvolume{31}(\bissue{2}),
\bfpage{895}--\blpage{918}
(\byear{2018})
\end{barticle}
\endbibitem

\bibitem[\protect\citeauthoryear{Hirahara et~al.}{2014}]{hirahara2014centennial}
\begin{barticle}
\bauthor{\bsnm{Hirahara}, \binits{S.}},
\bauthor{\bsnm{Ishii}, \binits{M.}},
\bauthor{\bsnm{Fukuda}, \binits{Y.}}:
\batitle{Centennial-scale sea surface temperature analysis and its uncertainty}.
\bjtitle{Journal of Climate}
\bvolume{27}(\bissue{1}),
\bfpage{57}--\blpage{75}
(\byear{2014})
\end{barticle}
\endbibitem

\bibitem[\protect\citeauthoryear{Duveiller et~al.}{2018}]{duveiller2018dataset}
\begin{barticle}
\bauthor{\bsnm{Duveiller}, \binits{G.}},
\bauthor{\bsnm{Hooker}, \binits{J.}},
\bauthor{\bsnm{Cescatti}, \binits{A.}}:
\batitle{A dataset mapping the potential biophysical effects of vegetation cover change}.
\bjtitle{Scientific data}
\bvolume{5}(\bissue{1}),
\bfpage{1}--\blpage{15}
(\byear{2018})
\end{barticle}
\endbibitem

\bibitem[\protect\citeauthoryear{Stengel et~al.}{2017}]{stengel2017cloud}
\begin{barticle}
\bauthor{\bsnm{Stengel}, \binits{M.}},
\bauthor{\bsnm{Stapelberg}, \binits{S.}},
\bauthor{\bsnm{Sus}, \binits{O.}},
\bauthor{\bsnm{Schlundt}, \binits{C.}},
\bauthor{\bsnm{Poulsen}, \binits{C.}},
\bauthor{\bsnm{Thomas}, \binits{G.}},
\bauthor{\bsnm{Christensen}, \binits{M.}},
\bauthor{\bsnm{Carbajal~Henken}, \binits{C.}},
\bauthor{\bsnm{Preusker}, \binits{R.}},
\bauthor{\bsnm{Fischer}, \binits{J.}}, \betal:
\batitle{Cloud property datasets retrieved from avhrr, modis, aatsr and meris in the framework of the cloud\_cci project}.
\bjtitle{Earth System Science Data}
\bvolume{9}(\bissue{2}),
\bfpage{881}--\blpage{904}
(\byear{2017})
\end{barticle}
\endbibitem

\bibitem[\protect\citeauthoryear{Dorigo et~al.}{2017}]{dorigo2017esa}
\begin{barticle}
\bauthor{\bsnm{Dorigo}, \binits{W.}},
\bauthor{\bsnm{Wagner}, \binits{W.}},
\bauthor{\bsnm{Albergel}, \binits{C.}},
\bauthor{\bsnm{Albrecht}, \binits{F.}},
\bauthor{\bsnm{Balsamo}, \binits{G.}},
\bauthor{\bsnm{Brocca}, \binits{L.}},
\bauthor{\bsnm{Chung}, \binits{D.}},
\bauthor{\bsnm{Ertl}, \binits{M.}},
\bauthor{\bsnm{Forkel}, \binits{M.}},
\bauthor{\bsnm{Gruber}, \binits{A.}}, \betal:
\batitle{Esa cci soil moisture for improved earth system understanding: State-of-the art and future directions}.
\bjtitle{Remote Sensing of Environment}
\bvolume{203},
\bfpage{185}--\blpage{215}
(\byear{2017})
\end{barticle}
\endbibitem

\bibitem[\protect\citeauthoryear{Le~Qu{\'e}r{\'e} et~al.}{2018}]{le2018global}
\begin{barticle}
\bauthor{\bsnm{Le~Qu{\'e}r{\'e}}, \binits{C.}},
\bauthor{\bsnm{Andrew}, \binits{R.M.}},
\bauthor{\bsnm{Friedlingstein}, \binits{P.}},
\bauthor{\bsnm{Sitch}, \binits{S.}},
\bauthor{\bsnm{Hauck}, \binits{J.}},
\bauthor{\bsnm{Pongratz}, \binits{J.}},
\bauthor{\bsnm{Pickers}, \binits{P.A.}},
\bauthor{\bsnm{Korsbakken}, \binits{J.I.}},
\bauthor{\bsnm{Peters}, \binits{G.P.}},
\bauthor{\bsnm{Canadell}, \binits{J.G.}}, \betal:
\batitle{Global carbon budget 2018}.
\bjtitle{Earth System Science Data}
\bvolume{10}(\bissue{4}),
\bfpage{2141}--\blpage{2194}
(\byear{2018})
\end{barticle}
\endbibitem

\bibitem[\protect\citeauthoryear{Friedlingstein et~al.}{2020}]{friedlingstein2020global}
\begin{barticle}
\bauthor{\bsnm{Friedlingstein}, \binits{P.}},
\bauthor{\bsnm{O'sullivan}, \binits{M.}},
\bauthor{\bsnm{Jones}, \binits{M.W.}},
\bauthor{\bsnm{Andrew}, \binits{R.M.}},
\bauthor{\bsnm{Hauck}, \binits{J.}},
\bauthor{\bsnm{Olsen}, \binits{A.}},
\bauthor{\bsnm{Peters}, \binits{G.P.}},
\bauthor{\bsnm{Peters}, \binits{W.}},
\bauthor{\bsnm{Pongratz}, \binits{J.}},
\bauthor{\bsnm{Sitch}, \binits{S.}}, \betal:
\batitle{Global carbon budget 2020}.
\bjtitle{Earth System Science Data Discussions}
\bvolume{2020},
\bfpage{1}--\blpage{3}
(\byear{2020})
\end{barticle}
\endbibitem

\bibitem[\protect\citeauthoryear{Behringer et~al.}{1998}]{behringer1998improved}
\begin{barticle}
\bauthor{\bsnm{Behringer}, \binits{D.W.}},
\bauthor{\bsnm{Ji}, \binits{M.}},
\bauthor{\bsnm{Leetmaa}, \binits{A.}}:
\batitle{An improved coupled model for enso prediction and implications for ocean initialization. part i: The ocean data assimilation system}.
\bjtitle{Monthly Weather Review}
\bvolume{126}(\bissue{4}),
\bfpage{1013}--\blpage{1021}
(\byear{1998})
\end{barticle}
\endbibitem

\bibitem[\protect\citeauthoryear{Morice et~al.}{2012}]{morice2012quantifying}
\begin{botherref}
\oauthor{\bsnm{Morice}, \binits{C.P.}},
\oauthor{\bsnm{Kennedy}, \binits{J.J.}},
\oauthor{\bsnm{Rayner}, \binits{N.A.}},
\oauthor{\bsnm{Jones}, \binits{P.D.}}:
Quantifying uncertainties in global and regional temperature change using an ensemble of observational estimates: The hadcrut4 data set.
Journal of Geophysical Research: Atmospheres
\textbf{117}(D8)
(2012)
\end{botherref}
\endbibitem

\bibitem[\protect\citeauthoryear{Cheng et~al.}{2024}]{cheng2024iapv4}
\begin{barticle}
\bauthor{\bsnm{Cheng}, \binits{L.}},
\bauthor{\bsnm{Pan}, \binits{Y.}},
\bauthor{\bsnm{Tan}, \binits{Z.}},
\bauthor{\bsnm{Zheng}, \binits{H.}},
\bauthor{\bsnm{Zhu}, \binits{Y.}},
\bauthor{\bsnm{Wei}, \binits{W.}},
\bauthor{\bsnm{Du}, \binits{J.}},
\bauthor{\bsnm{Yuan}, \binits{H.}},
\bauthor{\bsnm{Li}, \binits{G.}},
\bauthor{\bsnm{Ye}, \binits{H.}}, \betal:
\batitle{Iapv4 ocean temperature and ocean heat content gridded dataset}.
\bjtitle{Earth System Science Data Discussions}
\bvolume{2024},
\bfpage{1}--\blpage{56}
(\byear{2024})
\end{barticle}
\endbibitem

\bibitem[\protect\citeauthoryear{Zhang et~al.}{2023}]{zhang2023calculation}
\begin{bchapter}
\bauthor{\bsnm{Zhang}, \binits{Y.}},
\bauthor{\bsnm{Rossow}, \binits{W.B.}},
\bauthor{\bsnm{Lacis}, \binits{A.A.}},
\bauthor{\bsnm{Oinas}, \binits{V.}}:
\bctitle{Calculation, evaluation and application of long-term, global radiative flux datasets at isccp: Past and present}.
In: \bbtitle{Studies of Cloud, Convection and Precipitation Processes Using Satellite Observations},
pp. \bfpage{151}--\blpage{177}.
\bpublisher{World Scientific},
\blocation{Singapore}
(\byear{2023})
\end{bchapter}
\endbibitem

\bibitem[\protect\citeauthoryear{Kalnay et~al.}{2018}]{kalnay2018ncep}
\begin{bchapter}
\bauthor{\bsnm{Kalnay}, \binits{E.}},
\bauthor{\bsnm{Kanamitsu}, \binits{M.}},
\bauthor{\bsnm{Kistler}, \binits{R.}},
\bauthor{\bsnm{Collins}, \binits{W.}},
\bauthor{\bsnm{Deaven}, \binits{D.}},
\bauthor{\bsnm{Gandin}, \binits{L.}},
\bauthor{\bsnm{Iredell}, \binits{M.}},
\bauthor{\bsnm{Saha}, \binits{S.}},
\bauthor{\bsnm{White}, \binits{G.}},
\bauthor{\bsnm{Woollen}, \binits{J.}}, \betal:
\bctitle{The ncep/ncar 40-year reanalysis project}.
In: \bbtitle{Renewable Energy},
pp. \bfpage{1}--\blpage{1461194}.
\bpublisher{Routledge},
\blocation{UK}
(\byear{2018})
\end{bchapter}
\endbibitem

\bibitem[\protect\citeauthoryear{Huang et~al.}{2017}]{huang2017extended}
\begin{barticle}
\bauthor{\bsnm{Huang}, \binits{B.}},
\bauthor{\bsnm{Thorne}, \binits{P.W.}},
\bauthor{\bsnm{Banzon}, \binits{V.F.}},
\bauthor{\bsnm{Boyer}, \binits{T.}},
\bauthor{\bsnm{Chepurin}, \binits{G.}},
\bauthor{\bsnm{Lawrimore}, \binits{J.H.}},
\bauthor{\bsnm{Menne}, \binits{M.J.}},
\bauthor{\bsnm{Smith}, \binits{T.M.}},
\bauthor{\bsnm{Vose}, \binits{R.S.}},
\bauthor{\bsnm{Zhang}, \binits{H.-M.}}:
\batitle{Extended reconstructed sea surface temperature, version 5 (ersstv5): upgrades, validations, and intercomparisons}.
\bjtitle{Journal of Climate}
\bvolume{30}(\bissue{20}),
\bfpage{8179}--\blpage{8205}
(\byear{2017})
\end{barticle}
\endbibitem

\bibitem[\protect\citeauthoryear{Copernicus Climate Change~Service}{2021}]{copernicus2021oras5}
\begin{botherref}
\oauthor{\bsnm{Copernicus Climate Change~Service}, \binits{C.D.S.}}:
Oras5 global ocean reanalysis monthly data from 1958 to present.
Copernic. Clim. Change Serv.(C3S) Clim. Data Store (CDS)
(2021)
\end{botherref}
\endbibitem

\bibitem[\protect\citeauthoryear{Righi et~al.}{2020}]{righi2020earth}
\begin{barticle}
\bauthor{\bsnm{Righi}, \binits{M.}},
\bauthor{\bsnm{Andela}, \binits{B.}},
\bauthor{\bsnm{Eyring}, \binits{V.}},
\bauthor{\bsnm{Lauer}, \binits{A.}},
\bauthor{\bsnm{Predoi}, \binits{V.}},
\bauthor{\bsnm{Schlund}, \binits{M.}},
\bauthor{\bsnm{Vegas-Regidor}, \binits{J.}},
\bauthor{\bsnm{Bock}, \binits{L.}},
\bauthor{\bsnm{Br{\"o}tz}, \binits{B.}},
\bauthor{\bsnm{Mora}, \binits{L.}}, \betal:
\batitle{Earth system model evaluation tool (esmvaltool) v2. 0--technical overview}.
\bjtitle{Geoscientific Model Development}
\bvolume{13}(\bissue{3}),
\bfpage{1179}--\blpage{1199}
(\byear{2020})
\end{barticle}
\endbibitem

\bibitem[\protect\citeauthoryear{Lee et~al.}{2024}]{lee2024systematic}
\begin{barticle}
\bauthor{\bsnm{Lee}, \binits{J.}},
\bauthor{\bsnm{Gleckler}, \binits{P.J.}},
\bauthor{\bsnm{Ahn}, \binits{M.-S.}},
\bauthor{\bsnm{Ordonez}, \binits{A.}},
\bauthor{\bsnm{Ullrich}, \binits{P.A.}},
\bauthor{\bsnm{Sperber}, \binits{K.R.}},
\bauthor{\bsnm{Taylor}, \binits{K.E.}},
\bauthor{\bsnm{Planton}, \binits{Y.Y.}},
\bauthor{\bsnm{Guilyardi}, \binits{E.}},
\bauthor{\bsnm{Durack}, \binits{P.}}, \betal:
\batitle{Systematic and objective evaluation of earth system models: Pcmdi metrics package (pmp) version 3}.
\bjtitle{Geoscientific Model Development}
\bvolume{17}(\bissue{9}),
\bfpage{3919}--\blpage{3948}
(\byear{2024})
\end{barticle}
\endbibitem

\bibitem[\protect\citeauthoryear{Wang et~al.}{2024}]{wang2024mineru}
\begin{botherref}
\oauthor{\bsnm{Wang}, \binits{B.}},
\oauthor{\bsnm{Xu}, \binits{C.}},
\oauthor{\bsnm{Zhao}, \binits{X.}},
\oauthor{\bsnm{Ouyang}, \binits{L.}},
\oauthor{\bsnm{Wu}, \binits{F.}},
\oauthor{\bsnm{Zhao}, \binits{Z.}},
\oauthor{\bsnm{Xu}, \binits{R.}},
\oauthor{\bsnm{Liu}, \binits{K.}},
\oauthor{\bsnm{Qu}, \binits{Y.}},
\oauthor{\bsnm{Shang}, \binits{F.}}, et al.:
Mineru: An open-source solution for precise document content extraction.
arXiv preprint arXiv:2409.18839
(2024)
\end{botherref}
\endbibitem

\bibitem[\protect\citeauthoryear{Li et~al.}{2024}]{li2024constraining}
\begin{barticle}
\bauthor{\bsnm{Li}, \binits{C.}},
\bauthor{\bsnm{Sun}, \binits{Q.}},
\bauthor{\bsnm{Wang}, \binits{J.}},
\bauthor{\bsnm{Liang}, \binits{Y.}},
\bauthor{\bsnm{Zwiers}, \binits{F.W.}},
\bauthor{\bsnm{Zhang}, \binits{X.}},
\bauthor{\bsnm{Li}, \binits{T.}}:
\batitle{Constraining projected changes in rare intense precipitation events across global land regions}.
\bjtitle{Geophysical Research Letters}
\bvolume{51}(\bissue{3}),
\bfpage{2023}--\blpage{105605}
(\byear{2024})
\end{barticle}
\endbibitem

\bibitem[\protect\citeauthoryear{Geng et~al.}{2023}]{geng2023increased}
\begin{barticle}
\bauthor{\bsnm{Geng}, \binits{T.}},
\bauthor{\bsnm{Jia}, \binits{F.}},
\bauthor{\bsnm{Cai}, \binits{W.}},
\bauthor{\bsnm{Wu}, \binits{L.}},
\bauthor{\bsnm{Gan}, \binits{B.}},
\bauthor{\bsnm{Jing}, \binits{Z.}},
\bauthor{\bsnm{Li}, \binits{S.}},
\bauthor{\bsnm{McPhaden}, \binits{M.J.}}:
\batitle{Increased occurrences of consecutive la ni{\~n}a events under global warming}.
\bjtitle{Nature}
\bvolume{619}(\bissue{7971}),
\bfpage{774}--\blpage{781}
(\byear{2023})
\end{barticle}
\endbibitem

\bibitem[\protect\citeauthoryear{Cai et~al.}{2022}]{cai2022increased}
\begin{barticle}
\bauthor{\bsnm{Cai}, \binits{W.}},
\bauthor{\bsnm{Ng}, \binits{B.}},
\bauthor{\bsnm{Wang}, \binits{G.}},
\bauthor{\bsnm{Santoso}, \binits{A.}},
\bauthor{\bsnm{Wu}, \binits{L.}},
\bauthor{\bsnm{Yang}, \binits{K.}}:
\batitle{Increased enso sea surface temperature variability under four ipcc emission scenarios}.
\bjtitle{Nature Climate Change}
\bvolume{12}(\bissue{3}),
\bfpage{228}--\blpage{231}
(\byear{2022})
\end{barticle}
\endbibitem

\bibitem[\protect\citeauthoryear{Song et~al.}{2025}]{song2025hot}
\begin{barticle}
\bauthor{\bsnm{Song}, \binits{F.}},
\bauthor{\bsnm{Dong}, \binits{H.}},
\bauthor{\bsnm{Wu}, \binits{L.}},
\bauthor{\bsnm{Leung}, \binits{L.R.}},
\bauthor{\bsnm{Lu}, \binits{J.}},
\bauthor{\bsnm{Dong}, \binits{L.}},
\bauthor{\bsnm{Wu}, \binits{P.}},
\bauthor{\bsnm{Zhou}, \binits{T.}}:
\batitle{Hot season gets hotter due to rainfall delay over tropical land in a warming climate}.
\bjtitle{Nature Communications}
\bvolume{16}(\bissue{1}),
\bfpage{2188}
(\byear{2025})
\end{barticle}
\endbibitem

\bibitem[\protect\citeauthoryear{Seth et~al.}{2013}]{seth2013cmip5}
\begin{barticle}
\bauthor{\bsnm{Seth}, \binits{A.}},
\bauthor{\bsnm{Rauscher}, \binits{S.A.}},
\bauthor{\bsnm{Biasutti}, \binits{M.}},
\bauthor{\bsnm{Giannini}, \binits{A.}},
\bauthor{\bsnm{Camargo}, \binits{S.J.}},
\bauthor{\bsnm{Rojas}, \binits{M.}}:
\batitle{Cmip5 projected changes in the annual cycle of precipitation in monsoon regions}.
\bjtitle{Journal of climate}
\bvolume{26}(\bissue{19}),
\bfpage{7328}--\blpage{7351}
(\byear{2013})
\end{barticle}
\endbibitem

\bibitem[\protect\citeauthoryear{Song et~al.}{2018}]{song2018seasonally}
\begin{barticle}
\bauthor{\bsnm{Song}, \binits{F.}},
\bauthor{\bsnm{Leung}, \binits{L.R.}},
\bauthor{\bsnm{Lu}, \binits{J.}},
\bauthor{\bsnm{Dong}, \binits{L.}}:
\batitle{Seasonally dependent responses of subtropical highs and tropical rainfall to anthropogenic warming}.
\bjtitle{Nature Climate Change}
\bvolume{8}(\bissue{9}),
\bfpage{787}--\blpage{792}
(\byear{2018})
\end{barticle}
\endbibitem

\bibitem[\protect\citeauthoryear{Song et~al.}{2020}]{song2020contrasting}
\begin{barticle}
\bauthor{\bsnm{Song}, \binits{F.}},
\bauthor{\bsnm{Lu}, \binits{J.}},
\bauthor{\bsnm{Leung}, \binits{L.R.}},
\bauthor{\bsnm{Liu}, \binits{F.}}:
\batitle{Contrasting phase changes of precipitation annual cycle between land and ocean under global warming}.
\bjtitle{Geophysical Research Letters}
\bvolume{47}(\bissue{20}),
\bfpage{2020}--\blpage{090327}
(\byear{2020})
\end{barticle}
\endbibitem

\bibitem[\protect\citeauthoryear{Song et~al.}{2021}]{song2021emergence}
\begin{barticle}
\bauthor{\bsnm{Song}, \binits{F.}},
\bauthor{\bsnm{Leung}, \binits{L.R.}},
\bauthor{\bsnm{Lu}, \binits{J.}},
\bauthor{\bsnm{Dong}, \binits{L.}},
\bauthor{\bsnm{Zhou}, \binits{W.}},
\bauthor{\bsnm{Harrop}, \binits{B.}},
\bauthor{\bsnm{Qian}, \binits{Y.}}:
\batitle{Emergence of seasonal delay of tropical rainfall during 1979--2019}.
\bjtitle{Nature Climate Change}
\bvolume{11}(\bissue{7}),
\bfpage{605}--\blpage{612}
(\byear{2021})
\end{barticle}
\endbibitem

\bibitem[\protect\citeauthoryear{Cai et~al.}{2019}]{cai2019pantropical}
\begin{barticle}
\bauthor{\bsnm{Cai}, \binits{W.}},
\bauthor{\bsnm{Wu}, \binits{L.}},
\bauthor{\bsnm{Lengaigne}, \binits{M.}},
\bauthor{\bsnm{Li}, \binits{T.}},
\bauthor{\bsnm{McGregor}, \binits{S.}},
\bauthor{\bsnm{Kug}, \binits{J.-S.}},
\bauthor{\bsnm{Yu}, \binits{J.-Y.}},
\bauthor{\bsnm{Stuecker}, \binits{M.F.}},
\bauthor{\bsnm{Santoso}, \binits{A.}},
\bauthor{\bsnm{Li}, \binits{X.}}, \betal:
\batitle{Pantropical climate interactions}.
\bjtitle{Science}
\bvolume{363}(\bissue{6430}),
\bfpage{4236}
(\byear{2019})
\end{barticle}
\endbibitem

\bibitem[\protect\citeauthoryear{Wang}{2019}]{wang2019three}
\begin{barticle}
\bauthor{\bsnm{Wang}, \binits{C.}}:
\batitle{Three-ocean interactions and climate variability: a review and perspective}.
\bjtitle{Climate Dynamics}
\bvolume{53}(\bissue{7}),
\bfpage{5119}--\blpage{5136}
(\byear{2019})
\end{barticle}
\endbibitem

\bibitem[\protect\citeauthoryear{Myers and Norris}{2016}]{myers2016reducing}
\begin{barticle}
\bauthor{\bsnm{Myers}, \binits{T.A.}},
\bauthor{\bsnm{Norris}, \binits{J.R.}}:
\batitle{Reducing the uncertainty in subtropical cloud feedback}.
\bjtitle{Geophysical Research Letters}
\bvolume{43}(\bissue{5}),
\bfpage{2144}--\blpage{2148}
(\byear{2016})
\end{barticle}
\endbibitem

\bibitem[\protect\citeauthoryear{Ceppi et~al.}{2017}]{ceppi2017cloud}
\begin{barticle}
\bauthor{\bsnm{Ceppi}, \binits{P.}},
\bauthor{\bsnm{Brient}, \binits{F.}},
\bauthor{\bsnm{Zelinka}, \binits{M.D.}},
\bauthor{\bsnm{Hartmann}, \binits{D.L.}}:
\batitle{Cloud feedback mechanisms and their representation in global climate models}.
\bjtitle{Wiley Interdisciplinary Reviews: Climate Change}
\bvolume{8}(\bissue{4}),
\bfpage{465}
(\byear{2017})
\end{barticle}
\endbibitem

\end{thebibliography}
